\documentclass{article}
\usepackage[final]{neurips_2025}

\usepackage[utf8]{inputenc}
\usepackage[LAE,LFE,T1]{fontenc} 
\usepackage{CJKutf8}             

\usepackage{arabtex}
\usepackage{utf8}                
\setcode{utf8}                   

\makeatletter
\protected\def\begin#1{%
  \UseHook{env/#1/before}%
  \@ifundefined{#1}%
    {\def\reserved@a{\@latex@error{Environment #1 undefined}\@eha}}%
    {\def\reserved@a{\def\@currenvir{#1}%
        \edef\@currenvline{\on@line}%
        \@execute@begin@hook{#1}%
        \csname #1\endcsname}}%
  \@ignorefalse
  \begingroup
  \let\end\a@l@end  
  \@endpefalse\reserved@a}
\makeatother

\newenvironment{UrduText}{\begin{arabtext}}{\end{arabtext}}

\makeatletter
\renewcommand{\@noticestring}{}%
\makeatother

\makeatletter
\let\vlu@orig@quote\quote
\let\vlu@orig@endquote\endquote
\renewenvironment{quote}%
  {\list{}{\rightmargin\leftmargin}\item\relax}
  {\endlist}
\makeatother

\makeatletter
\renewenvironment{abstract}{%
  \par
  \begingroup
    \centering \bfseries Abstract\par
  \endgroup
  \vspace{-0.6em}%
  \list{}{\leftmargin\parindent \rightmargin\parindent}%
  \item\relax
}{%
  \endlist
  \par
}
\makeatother

\usepackage{url}
\usepackage{booktabs}
\usepackage{amsfonts}
\usepackage{amsmath}
\usepackage{nicefrac}
\usepackage{microtype}
\usepackage{xcolor}
\usepackage{multirow}
\usepackage{float}
\usepackage{colortbl}
\usepackage{natbib}
\usepackage{graphicx}            
\usepackage{caption}
\usepackage{subcaption}
\usepackage{wrapfig}
\usepackage{adjustbox}
\usepackage{pifont}
\usepackage{listings}
\usepackage{tikz}
\usetikzlibrary{patterns}
\usepackage{pgfplots}
\pgfplotsset{compat=1.18}
\usepgfplotslibrary{colormaps}
\usepgfplotslibrary{polar}
\usepackage{tcolorbox}
\tcbuselibrary{skins, breakable}

\definecolor{bblue}{HTML}{4F81BD}
\definecolor{rred}{HTML}{C0504D}
\definecolor{ggreen}{HTML}{9BBB59}
\definecolor{ppurple}{HTML}{9F4C7C}
\definecolor{baseOlive}{HTML}{808000}
\definecolor{darkOlive}{HTML}{404000}
\definecolor{lightOlive}{HTML}{CCCC99}
\definecolor{darkCyan}{HTML}{008B8B}
\definecolor{lightCyan}{HTML}{E0FFFF}
\definecolor{darkGreen}{HTML}{006400}
\definecolor{lightGreen}{HTML}{90EE90}
\definecolor{darkLime}{HTML}{238F23}
\definecolor{Honeydew}{HTML}{F0FFF0}
\definecolor{darkOrange}{HTML}{D35400}
\definecolor{lightOrange}{HTML}{FFEBCD}
\definecolor{promptboxTitleBg}{HTML}{595959}
\definecolor{promptboxFrame}{HTML}{666666}
\definecolor{promptboxContentBg}{HTML}{F2F2F2}
\definecolor{apricot}{rgb}{0.98, 0.81, 0.69}
\definecolor{modelBrown}{HTML}{8B4513}
\definecolor{modelSlateGray}{HTML}{708090}

\newcommand{\cmark}{\textcolor{green}{\ding{52}}}
\newcommand{\xmark}{\textcolor{red}{\ding{56}}}

\pgfplotscreateplotcyclelist{mybarlist}{%
  {fill=bblue,   draw=bblue!60!black},
  {fill=rred,    draw=rred!60!black},
  {fill=ggreen,  draw=ggreen!60!black},
  {fill=ppurple, draw=ppurple!60!black},
}

\DeclareUnicodeCharacter{0394}{\ensuremath{\Delta}}
\DeclareUnicodeCharacter{2192}{\ensuremath{\to}}

\newcommand{\textjapanese}[1]{\begin{CJK*}{UTF8}{min}#1\end{CJK*}}

\newcommand{\jptext}[1]{\textjapanese{#1}}
\newcommand{\jp}[1]{\jptext{#1}}

\newcommand{\textswahili}[1]{#1}
\newcommand{\swtext}[1]{\textswahili{#1}}
\newcommand{\sw}[1]{\swtext{#1}}

\newcommand{\urduinline}[1]{\begin{arabtext}#1\end{arabtext}}

\date{}

\title{
\emph{\textbf{VLURes}}: Benchmarking VLM Visual and Linguistic Understanding in Low-Resource Languages
}

\author{%
  Jesse Atuhurra$^{1, * , \dagger}$ \quad
  Iqra Ali$^{2, \dagger}$ \quad
  Tomoya Iwakura$^{3}$ \quad
  Hidetaka Kamigaito$^{1}$ \\
  \textbf{Tatsuya Hiraoka}$^{1,4}$ \\[2ex] %
  \small $^{1}$NAIST \quad
  $^{2}$QMUL \quad
  $^{3}$Fujitsu Ltd.  \quad
  $^{4}$MBZUAI \\[1ex] %
  \small\texttt{atuhurra.jesse.ag2@naist.ac.jp, kamigaito.h@naist.ac.jp, iqra.ali@qmul.ac.uk} \\
  \small\texttt{iwakura.tomoya@fujitsu.com, tatsuya.hiraoka@mbzuai.ac.ae}
  \thanks{Corresponding author: atuhurra.jesse.ag2@naist.ac.jp.}
  \thanks{Work based on an internship at Fujitsu AI Lab.}
}

\begin{document}

\maketitle

\begin{abstract}
{\footnotesize
Vision Language Models (VLMs) are pivotal for advancing perception in intelligent agents. Yet, evaluation of VLMs remains limited to predominantly English-centric benchmarks in which the image-text pairs comprise short texts.
To evaluate VLM fine-grained abilities, in \textbf{four languages} under long-text settings, we introduce a novel multilingual benchmark \textbf{\emph{VLURes}} featuring \textbf{eight} vision-and-language tasks, and a pioneering \textit{unrelatedness} task, to probe the fine-grained \underline{\textbf{V}}isual and \underline{\textbf{L}}inguistic \underline{\textbf{U}}nderstanding capabilities of VLMs across English, Japanese, and low \underline{\textbf{Res}}ource languages, Swahili, and Urdu.
Our datasets, curated from web resources in the target language, encompass ten diverse \textit{image} categories and rich \textit{textual} context, introducing valuable vision-language resources for Swahili and Urdu.
By prompting VLMs to generate \textit{responses} and \textit{rationales}, evaluated automatically and by native speakers, we uncover performance disparities across languages and tasks critical to intelligent agents, such as object recognition, scene understanding, and relationship understanding.
We conducted evaluations of ten VLMs with \textbf{\emph{VLURes}}. The best performing model, GPT-4o, achieves an overall accuracy of 90.8\% and lags human performance by 6.7\%, though the gap is larger for open-source models. The gap highlights \textbf{\emph{VLURes}}’ critical role in developing intelligent agents to tackle multi-modal visual reasoning.
}
\end{abstract}

\begin{figure}[H]
\centering
\includegraphics[width=0.600\linewidth]{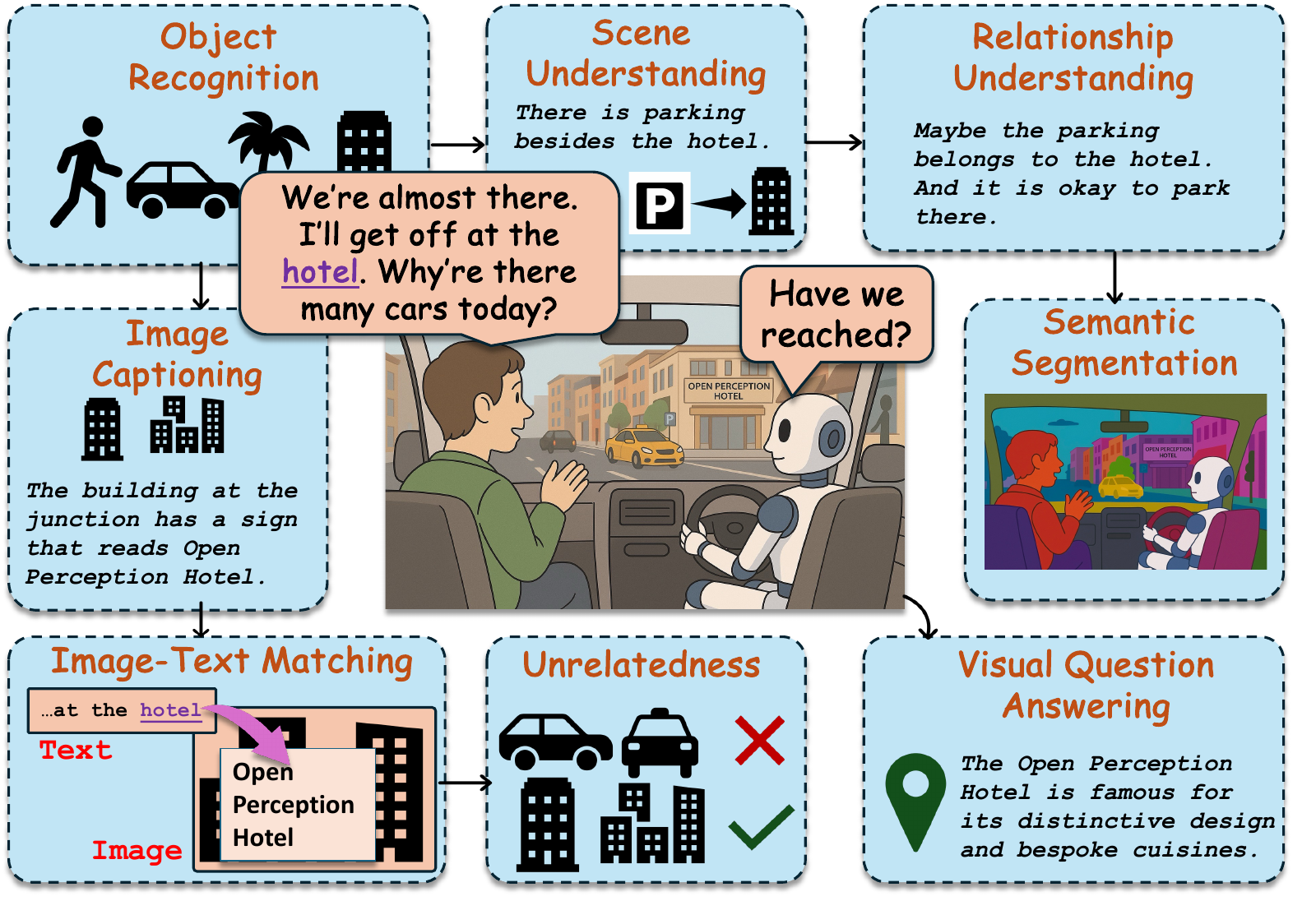} 
\caption{
\textbf{ \emph{VLURes}} consists of \textbf{\textcolor{brown}{eight tasks}} shown above. 
The tasks enable the intelligent agent to understand its surroundings. 
We evaluate Vision Language Models' performance on  \textbf{\emph{VLURes}}.
It offers \textit{image-text} pairs in \textbf{En, Jp, Sw, Ur}, and each \textit{image is embedded in article-length prose}.
}
\label{fig:IntroFigure}
\end{figure}

\begin{table}[H]
    \centering
    \tiny
    \caption{ \textbf{\emph{VLURes}} includes eight \colorbox{red!10}{tasks}, and \textcolor{violet}{prompts} specific to each task needed to generate the \textcolor{teal}{responses} from the VLM. The dots (\textcolor{teal}{......}) indicate truncated parts of the VLM response. }
    \resizebox{\textwidth}{!}{
    \begin{tabular}{p{3.5cm} p{4.5cm} p{8cm}}
        \toprule
        \textbf{Task} & \textbf{Task Question (Prompt) } & \textbf{Response from Vision Language Model} \\
        \hline
        \hline
        \multirow{1}{*}{\textbf{   \colorbox{red!10}{1. Object Recognition (OR)} }} & 
        \textcolor{violet}{Analyze this image and list all objects present.} Categorize each object into groups such as furniture, electronic devices, clothing, etc. Be thorough and specific in your identification.
        & 
        \textcolor{teal}{Objects Present: 1. Infrastructure: - Bridge - Railings - Streetlights - Road markings 2. Buildings: - High-rise buildings - Residential or commercial structures 3. People:  - Individuals walking or cycling 4. Transportation: - Bicycles. Categories: Infrastructure: Bridge, railings, streetlights, road markings. Buildings: High-rise buildings. People: Individuals. Transportation: Bicycles.} \\ 
        \hline
        \multirow{1}{*}{\textbf{   \colorbox{red!10}{2. Scene Understanding (SU)} }} & 
        \textcolor{violet}{Describe the overall scene in this image.} What is the setting, and what activities or events are taking place? Provide a comprehensive overview of the environment and any actions occurring.
        & 
        \textcolor{teal}{The scene depicts an urban setting with a modern bridge leading towards a cityscape of tall buildings. The bridge has a dedicated pathway for pedestrians and cyclists, as indicated by the markings on the ground. There are a few people walking and cycling along this path, suggesting it is a popular route for commuting or leisure activities......}
        \\ 
        \hline
        \multirow{1}{*}{\textbf{   \colorbox{red!10}{3. Relationship Understanding (RU)}  }} & 
        \textcolor{violet}{Identify any interactions or relationships between objects or entities in this image.} How are they related or interacting with each other? Explain any spatial, functional, or social connections you observe.
        & 
        \textcolor{teal}{Interactions and Relationships. 1. Spatial Relationships: - The image shows a bridge or elevated pathway leading towards a cityscape with several tall buildings in the background. The pathway is divided into lanes, likely for pedestrians and cyclists. - The bridge is flanked by railings on both sides, providing safety and separation from the surrounding environment. 2. Functional Connections: - The bridge serves...... 
        3. Social Interactions: - There are people walking...... 
        }
        \\ 
        \hline
        \multirow{1}{*}{\textbf{   \colorbox{red!10}{4. Semantic Segmentation (SS)}  }} & 
        \textcolor{violet}{Divide this image into different semantic regions.} Label each region (e.g., sky, buildings, people, street) and briefly describe its contents. Provide a clear breakdown of the image's composition.
        & 
        \textcolor{teal}{Semantic Regions. 1. Sky: - Location: Top portion of the image. - Description: The sky is mostly clear with some scattered clouds, providing a bright backdrop to the scene. 
        2. Buildings: - Location: Middle section of the image, in the background.  - Description: A series of modern high-rise buildings with glass facades. They vary in height and design, contributing to an urban skyline. 
        3. Bridge/Pathway: ......  
        4. People: ...... 
        5. Street Lamps: ...... 
        6. Water: ...... 
        This breakdown highlights the urban setting and infrastructure, emphasizing the interaction between architecture and public spaces. 
        }
        \\ 
        \hline
        \multirow{1}{*}{\textbf{   \colorbox{red!10}{5. Image Captioning (IC)} }} & 
        \textcolor{violet}{Provide a detailed, natural language description of what is happening in this image.} Narrate the scene as if you were explaining it to someone who cannot see it, including all relevant details and actions.
        & 
        \textcolor{teal}{The image depicts a modern urban scene featuring a pedestrian and cycling path on a bridge. The path is divided by a dashed white line, indicating separate lanes for cyclists and pedestrians. The bridge is flanked by metal railings on both sides, providing safety for those using the path. In the background, a skyline of tall, contemporary buildings is visible,...... } 
        \\ 
        \hline
        \multirow{1}{*}{\textbf{   \colorbox{red!10}{6. Image-Text Matching (ITM)}  }} & 
        \textcolor{violet}{Extract and list the specific parts of the text that closely match or directly reference entities, objects, or scenes depicted in the image.} Be precise in identifying these connections and explain the visual evidence that supports each textual reference.
        & 
         \textcolor{teal}{The image depicts a modern urban area with high-rise buildings, which aligns with the description of Wentworth Point in the text. Here are the specific connections between the text and the image: 
        1. Wentworth Point: The text mentions that the incident occurred in Wentworth Point, a suburb with high-rise buildings. The image shows a skyline with several tall buildings, which is characteristic of urban areas like Wentworth Point. 
        2. Apartment Buildings:......
        These elements in the image visually support textual references to locations and settings of events described. 
        }
        \\ 
        \hline
        \multirow{1}{*}{\textbf{    \colorbox{red!10}{7. Unrelatedness (U)}  }} & 
        \textcolor{violet}{Identify which parts of the text are not relevant to or not represented in the image.} Explain why these elements are unrelated by describing what is missing in the image that would be needed to illustrate these textual elements.
        & 
        \textcolor{teal}{The image shows a view of a bridge or walkway leading towards a cluster of modern buildings, likely in an urban area. 
        Here's how the text relates to the image: 
        1. Location Context:.....
        2. Unrelated Text Elements: 
        - Specific Events: The text describes a specific incident involving a killing, a text message, and interactions between individuals. None of these events or interactions are depicted in the image.....
        3. Missing Elements for Illustration: - To illustrate the textual elements, the image would need to show the specific individuals involved, the Toyota Kluger, the garage, or the interior of the building mentioned in the text. 
        Overall, while the image provides a general sense of the location, it does not depict the specific events or details described in the text. 
        }
        \\ 
        \hline
        \multirow{1}{*}{\textbf{    \colorbox{red!10}{8. Visual Question Answering (VQA)}  }} & 
        \textcolor{violet}{What places are mentioned in the text or shown in the image?} For each place identified, indicate whether it appears in the text, the image, or both. If any of these places are famous or well-known locations, explain why they are significant.
        & 
        \textcolor{teal}{The text mentions the following places: 
        1. Wentworth Point:..... 
        2. Granville:..... 
        Significance:..... 
        - Wentworth Point: Known for its residential developments and proximity to Sydney Olympic Park, making it a notable suburb in Sydney. 
        - Granville: Known for its diverse community and historical significance in Sydney. 
        In summary, Wentworth Point and Granville are mentioned in the text, while the image could potentially depict a similar urban environment but does not explicitly show any identifiable location from the text. 
        }
        \\ 
        \bottomrule
    \end{tabular}
    \label{Table:VLMoutputsIntro}
    }
\end{table}

\section{Introduction}
\label{sec:Introduction}
Vision Language Models (VLMs) have emerged as transformative tools in artificial intelligence. They integrate visual and textual understanding to advance perception in intelligent agents, such as robots.
Agents must interpret complex scenes, recognize objects, and comprehend contextual relationships to navigate and interact effectively in diverse environments.
VLMs make up the perception component of modern intelligent agents.
Yet, to measure the technological progress made in intelligent agents' perception, the state of the art is to benchmark agents against task-specific datasets. 
However, existing datasets, such as MSCOCO \cite{zaidan-etal-2007-using} and Visual Genome \cite{krishna2017visualgenome} mainly focus on high-resource languages like English. This restricts the assessment of agents' perception in low-resource languages.\footnote{The languages like Swahili and Urdu, though low-resource, have 200 million and 230 million speakers, respectively, and would benefit from the availability of multimodal evaluation data.}
A further limitation of existing datasets, such as MaRVL \cite{liu-etal-2021-visually} (comprising image-text pairs in Indonesian, Chinese, Swahili, Tamil, and Turkish), is that they offer short \textit{textual descriptions} of the images, hampering efforts to assess the visual and linguistic abilities of agents when thorough textual descriptions exist\footnote{Rich textual information supports the agent to perform VL tasks by providing the semantics and context about the environment and the task \cite{zhang2024visionlanguagemodelsvisiontasks}.}, for a wide range of vision-language (VL) tasks. 

We address the above limitations, to (i) unravel the \underline{\textbf{V}}isual and \underline{\textbf{L}}inguistic \underline{\textbf{U}}nderstanding abilities of VLMs in low \underline{\textbf{Res}}ource languages, (ii) facilitate a thorough analysis of abilities of VLMs' multimodal especially when rich textual information accompanies the image, (iii) and comprehensively evaluate of VLMs at fine-grained VL tasks. 

Therefore, we first construct a multilingual benchmark\footnote{We extend our analysis in \cite{atuhurra2024c} to include comprehensive cross-lingual and multi-LLM evaluations.}, \textbf{\emph{VLURes}}, containing image-text pairs in \textit{English, Japanese, Swahili, and Urdu}.
Second, we embed each image with detailed context derived from article-length texts.
Third, we introduce  eight VL tasks, illustrated in Figure \ref{fig:IntroFigure}.  

Given an image-text pair, the agent must decide whether to use information from the image, text, or both to accomplish the task.
The tasks in \emph{VLURes} are suitable to evaluate agent abilities like identifying objects (OR), interpreting scenes (SU), and inferring object interactions (RU), which are essential for agent navigation and manipulation. 
Our novel \emph{unrelatedness} task challenges VLMs to discard irrelevant text about the image and retain relevant texts, enhancing their robustness in noisy, real-world settings.

Then, given an image-text pair as input, we leverage VLMs to generate responses (shown in Table \ref{Table:VLMoutputsIntro}) and \textbf{rationales}, which are evaluated automatically by LLM judges, and manually by native speakers. 
Spanning ten image categories, our benchmark provides rich, \textit{article-length descriptions} drawn from diverse sources, such as websites, blogs, news articles, Wikinews and Wikipedia, contrasting with the short captions in prior benchmarks (see Table \ref{Table:VLMBenchmarksCompared}).

This work marks several key contributions. 
\textbf{First,} we introduce multi-task, multimodal VL datasets for Swahili and Urdu, addressing the evaluation gap in both languages \cite{thapliyal-etal-2022-crossmodal}. 
\textbf{Second,} we introduce article-length prose in every image-text pair in \emph{VLURes}, providing rich context to the image. 
\textbf{Third,} the inclusion of rationales offers transparency into VLM reasoning, aiding evaluation by both native speakers and the LLM-judges. 
\textbf{Fourth,} our work introduces the novel \textit{unrelatedness} task to evaluate the model's ability to discard irrelevant information. 
\textbf{Finally}, we validate our benchmark's effectiveness by fine-tuning open-source VLMs, demonstrating their potential for developing multilingual intelligent agents.

\section{Related Work}
\label{sec:RelatedWork}

\begin{table*}[!t]
\centering
\caption{A comparison between our dataset and existing datasets.
In this table, \cmark~means ``availabe'' while \xmark~means ``unavailable''. Yet * means the original paper did not explicitly mention the number of images or questions. Compared to all these datasets, only our dataset introduces rationales.}   
\resizebox{0.85\textwidth}{!}{
\begin{tabular}{lllllllrrr}
\toprule
\textbf{Dataset} & \textbf{Task}  & \textbf{\#Tasks} & \textbf{Multilingual} & \textbf{Language} & \textbf{\#Languages} & \textbf{Rationales} & \textbf{\#Images} & \textbf{\#Questions} & \textbf{Article-level Prose} \\
\midrule
VQAv2~\cite{goyal2017making}        & VQA  & 1 &  \xmark & En & 1 & \xmark & 265K & 1.1M  & \xmark \\ %
OK-VQA~\cite{okvqa}       & VQA  & 1 & \xmark & En & 1  & \xmark & 14K & 14K  & \xmark \\ %
OCR-VQA~\cite{mishraICDAR19}   & VQA & 1 & \xmark & En & 1  & \xmark & 207K & 1M & \xmark \\ %
GQA~\cite{hudson2019gqa}   & VQA & 1 & \xmark & En & 1  & \xmark & 113K & 22M & \xmark \\ %
Visual Genome \cite{Krishna2016VisualGC} & VQA & 1 & \xmark & En & 1  & \xmark & 108K & 1.7M & \xmark \\ %
VizWizQA \cite{gurari2018vizwiz}  & VQA & 1 & \xmark & En & 1  & \xmark & * & 31.1K & \xmark \\ %
TextVQA \cite{singh2019towards}   & VQA & 1 & \xmark & En  & 1 & \xmark & 28K & 45.3K & \xmark \\ %
\midrule
LAION 5B~\cite{schuhmann2022laion5b}     & IC & 1 & \cmark & En, Zh, .. & many  & \xmark &  5.85B & * & \xmark\\ %
LAION-COCO & IC & 1 & \xmark & En & 1  & \xmark & 600M & * & \xmark \\ %
Visual Genome \cite{Krishna2016VisualGC}  & IC & 1 & \xmark & En & 1  & \xmark & 108KM & 1.77M & \xmark \\
MSCOCO~\cite{lin2015microsoft}  & IC & 1  & \xmark & En & 1  & \xmark & 328K & * & \xmark \\
Flickr30k~\cite{plummer2016flickr30k} & IC & 1 & \xmark & En & 1  & \xmark & 30K & * & \xmark \\
Crossmodal-3600~\cite{thapliyal-etal-2022-crossmodal}  & IC & 1 & \cmark & En, Jp, Sw, .. & 36 & \xmark & 3.6K & * & \xmark\\
\midrule
RefClef \cite{kazemzadeh-etal-2014-referitgame} & REG & 1 & \xmark & En & 1 & \xmark & 19.9K & * & \xmark \\
RefCOCO \cite{kazemzadeh-etal-2014-referitgame}  & REG & 1 & \xmark & En & 1 & \xmark & 19.9K & *& \xmark \\
RefCOCO+ \cite{kazemzadeh-etal-2014-referitgame}  & REG & 1 & \xmark & En & 1  & \xmark & 19.9K & *& \xmark \\
RefCOCOg \cite{mao2016generation}  & REG & 1 & \xmark & En & 1 & \xmark & 25.7K & * & \xmark \\
\midrule
MMMU \cite{yue2023mmmu} & VQA & -- &  \xmark & En & 1 & \xmark & 11K & 11.5K & \xmark \\ 
MME$+$ \cite{fu2023mme} & OR, OCR, .. & 7 &  \xmark & En & 1 & \xmark & 1K & 2K & \xmark \\ 
MMBench~\cite{liu2023mmbench}& OCR, .. & 20 & \cmark & En, Zh & 2 & \xmark & 2.9K & 2.9K & \xmark \\ 
SEED-Bench~\cite{li2023seedbench} & OCR,..& 12 &  \xmark & En & 1  & \xmark & 19K & 19K & \xmark \\ 
MathVista \cite{lu2024mathvista}   & Maths & 12 & \xmark & En & 1  & \xmark & 6.1K & 6.1K & \xmark \\ 
MM-Vet \cite{yu2023mmvet} & OCR,OR,.. & 6 &  \xmark & En & 1  & \xmark & 200 & 218 & \xmark \\ 
Q-Bench \cite{wu2024qbench}  & VQA,.. & 3 & \xmark & En & 1 & \xmark & 3.4K & 2.9K & \xmark \\ 
\midrule
MaRVL \cite{liu-etal-2021-visually} &  Reasoning & 1 & \cmark & Id, Sw, Zh, Ta, Tr & 5  & \xmark & 5K & * & \xmark \\ %
\midrule
\textbf{\emph{VLURes} (Ours)} &  OR,IC,VQA,.. & 8 & \cmark  & En, Jp, Sw, Ur & 4 & \cmark & 4K & 8K & \cmark \\
\bottomrule
\end{tabular} }
\label{Table:VLMBenchmarksCompared}
\end{table*}

\textbf{Multimodal Benchmarks.} 
Many benchmarks have been developed to evaluate VLMs' visual/multimodal understanding abilities. 
The benchmarks include text-and-image pairs, for instance, MMMU \cite{yue2023mmmu}, TextVQA \cite{singh2019towards}, MathVista \cite{lu2024mathvista}, MME~\cite{fu2023mme}, MMBench~\cite{liu2023mmbench}, SEED-Bench~\cite{li2023seedbench}; video, such as Perception Test \cite{patraucean2023perception}, \textit{inter alia}. 
Table~\ref{Table:VLMBenchmarksCompared} shows prior benchmarks, and compares our benchmark to prior datasets. 
As shown in the table, IC, REG, VQA datasets and benchmarks above did not introduce ``rationale'' information about the \textit{caption, reference expression, or answer} generated/provided by the model or human. 
Moreover, the benchmarks above did not provide thorough textual context for images in the image-text pairs. These drawbacks motivated us to construct a new VL benchmark. 
In contrast, we develop a multi-lingual, multi-modal, multi-task benchmark containing image-text pairs mainly from web resources in four languages. (We describe more details about this decision later in \S \ref{sec:Data source}.) 
We have included further details about related datasets in \S \ref{App:MoreRelatedWorks}.
Additionally, recent works, such as \cite{fu2023mme, ye2023mplugowl, liu2023mmbench} and more,  which leverage a small yet well-curated sample size of images to give instructions \cite{xu-etal-2023-multiinstruct, fu2023mme, yue2023mmmu, yu2023mmvet} to VLMs to perform VL tasks on those images, inspire our work. For example, \cite{fu2023mme} assembled a set of 1,077 \textit{images} and corresponding \textit{instruction-answer pairs} to perform eleven perception and cognition tasks in the MME benchmark. \\
\textbf{Rationales in LLMs.} 
Rationales \cite{zaidan-etal-2007-using} serve as a gateway towards understanding the thought process of LLMs, by explaining how LLMs arrived at the final answer \cite{ling-etal-2017-program, hsieh-etal-2023-distilling}. 
Moreover, rationales break down problems into manageable steps, improving accuracy in tasks like math and reasoning \cite{yao2023tree, zhou2023leasttomost, zhang2024multimodal, hao2023reasoning, dhuliawala-etal-2024-chain, wang-etal-2024-absinstruct, jiang-etal-2024-rora}. 
In this study, we prompted VLMs to generate rationales and hypothesized that the rationales are valuable for human and automatic evaluation of VLM responses.

In summary, the benchmarks above are largely limited to English or Chinese. Prior Swahili or Urdu corpora contained $\leq 100$ images or only one-sentence captions, which limits comprehensive evaluation. By contrast, \emph{VLURes} provides sizable VL resources for Swahili and Urdu with 1{,}000 image–text pairs per language; supports long-text grounding by embedding each image in article-level prose, enabling discourse-level analysis; and explicitly tests a model’s ability to discard text that is unrelated to the picture via the \textit{unrelatedness} task.

\begin{figure*}[!t]
\centering
\includegraphics[width=0.6\textwidth]{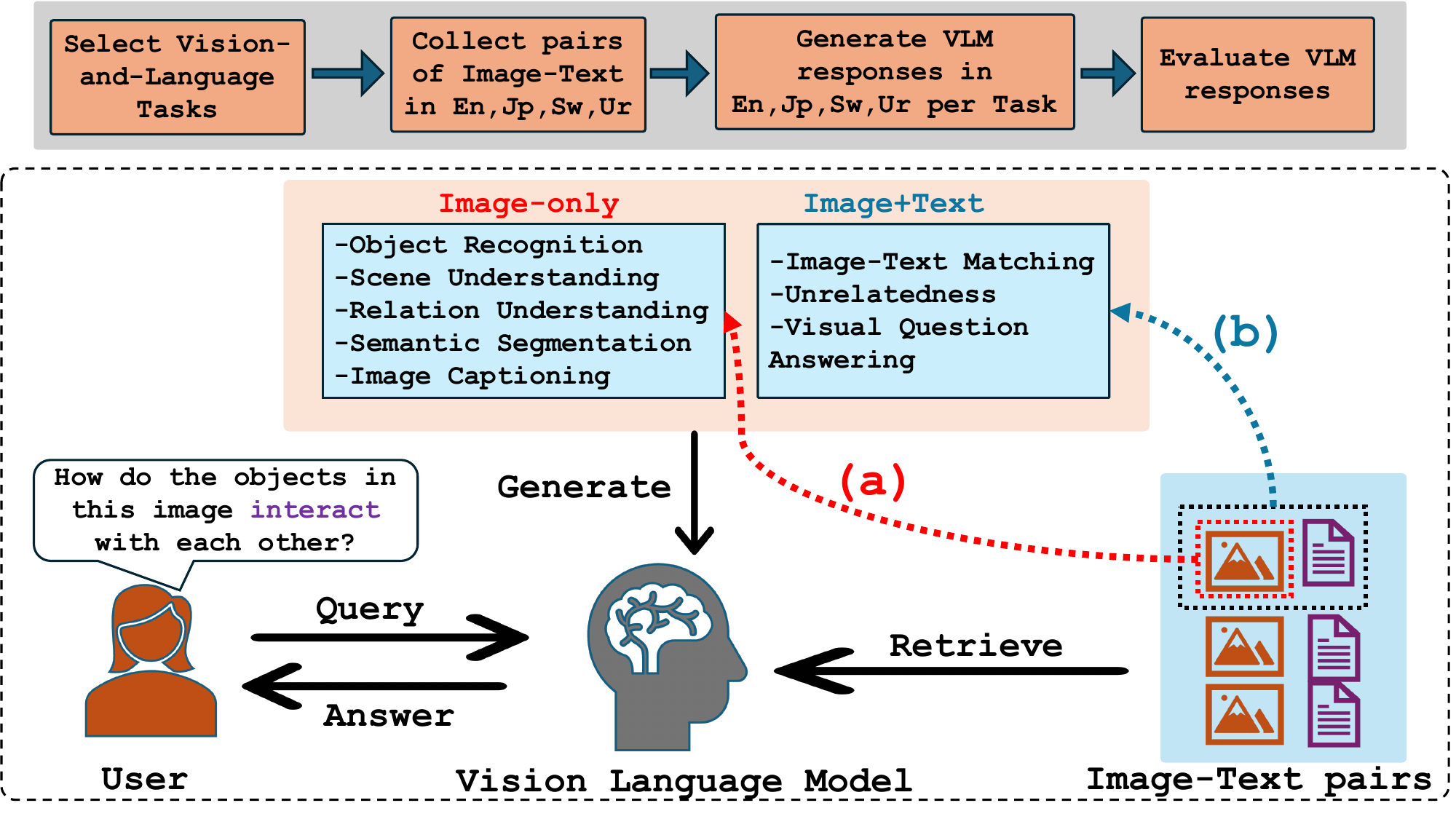} 
\caption{ \emph{VLURes} development process. \textbf{Top:} We selected the VL tasks and constructed datasets for each task and language. We leverage the image-text pairs as input to the VLMs and prompt the VLMs to generate responses for each task. 
Lastly, we deploy both automatic evaluation using \textit{LLM-as-a-Judge} approach and hire native speakers to judge VLM-response quality manually. 
\textbf{Bottom:} We chose the tasks in \emph{VLURes} to compare two kinds of visual reasoning for each image-text pair in the input: \textit{image-only} reasoning (shown in \textcolor{red}{(a)})  and \textit{image+text} reasoning (shown in \textcolor{blue}{(b)}). }
\label{fig:PaperSummary}
\end{figure*}

\section{ \emph{VLURes} Benchmark}
\label{sec:Datasets}
Our work is motivated by the need to train intelligent agents to choose the correct modality required to accomplish a VL task, in the presence of both image and textual information. Therefore, the input is always an image text pair, but the agent must apply the appropriate \texttt{image} or \texttt{image+text} visual reasoning to accomplish the VL task at hand. 

\subsection{Data Curation Principles}
\label{subsec:dataPrinciples}
\emph{VLURes} is constructed from web pages because they frequently contain multimedia content like images, making it a suitable basis for developing a multimodal benchmark. \emph{VLURes} aims to address the significant gap in current benchmarks that primarily assess visual and linguistic understanding in English. Hence, \emph{VLURes} follows these collection principles: 
(1) encompass multiple tasks and various topics to reflect real-world agent applications; 
(2) offer \textit{image-text} pairs in \textit{En, Jp, Sw, Ur}, and the data is native to each language, facilitating multilingual comparison; 
(3) varying difficulty levels of \textit{image-only} or \textit{image+text} reasoning, to identify limitations in existing models effectively; 
(4) suitability for the \textit{multimodal input} processing of modern VLMs, and 
(5) collect lengthy texts, to provide thorough context to the image and to facilitate the evaluation on a wide range of VL tasks.

The taxonomy for this work is introduced as follows: we identify two types of visual reasoning (Figure \ref{fig:PaperSummary}): \textit{image-only reasoning} and \textit{image+text reasoning}. \textit{Image-only reasoning} includes five primary tasks: \textit{Object Recognition (OR), Scene Understanding (SU), Relationship Understanding (RU), Semantic Segmentation (SS), and Image Captioning (IC)}. Moreover, \textit{Image+Text reasoning} includes three primary tasks: \textit{Image-Text Matching (ITM), Unrelatedness (U), and Visual Question Answering (VQA)}. All eight tasks are formally defined in \S \ref{subsec:DefinitionsVLTasks}.
In addition, we account for diverse visual contexts, including archival artifacts, photographic images, data visualizations, maps, infographics, and more, from diverse web resources.
Lastly, recognizing the lack of multi-task image-text datasets in Sw and Ur, we aimed at evaluating VLMs across many VL tasks.

\begin{figure}[!t]
\centering
\begin{minipage}{0.24\columnwidth}
    \centering
    \includegraphics[width=\linewidth]{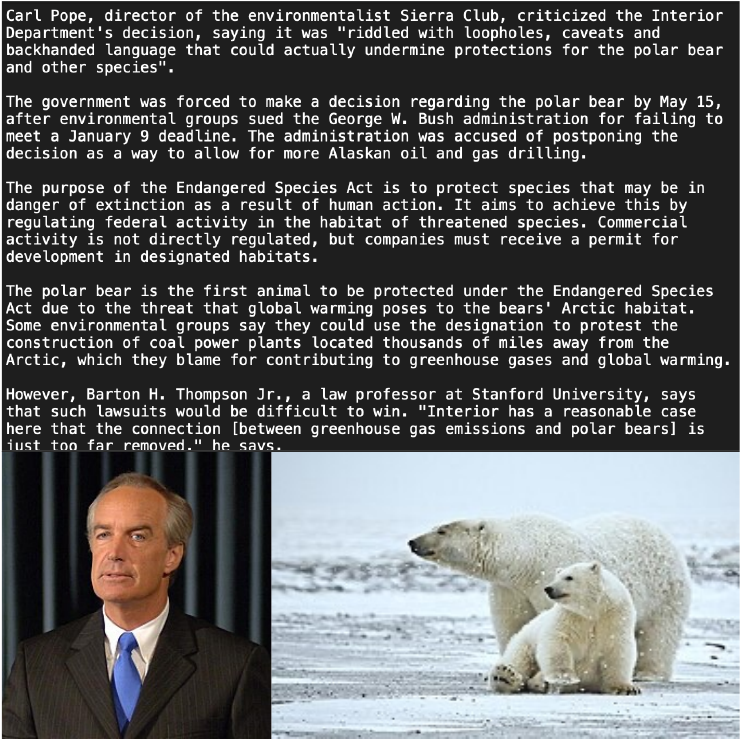}
    \caption*{(a) \textbf{English Wikinews} article, in \texttt{animals}.}
\end{minipage}
\hfill
\begin{minipage}{0.24\columnwidth}
    \centering
    \includegraphics[width=\linewidth]{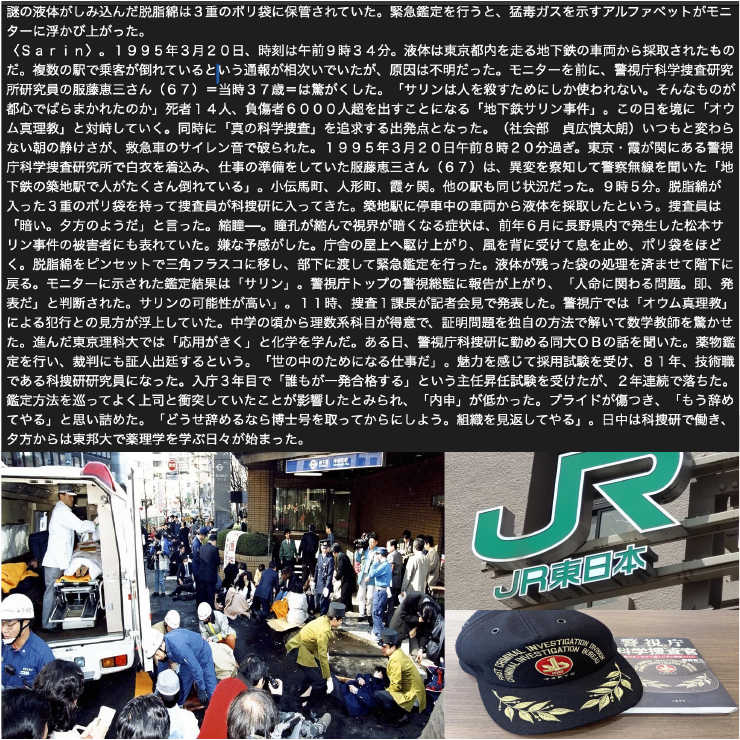}
    \caption*{(b) \textbf{Japanese news} article, in \texttt{events}.}
\end{minipage}
\hfill
\begin{minipage}{0.24\columnwidth}
    \centering
    \includegraphics[width=\linewidth]{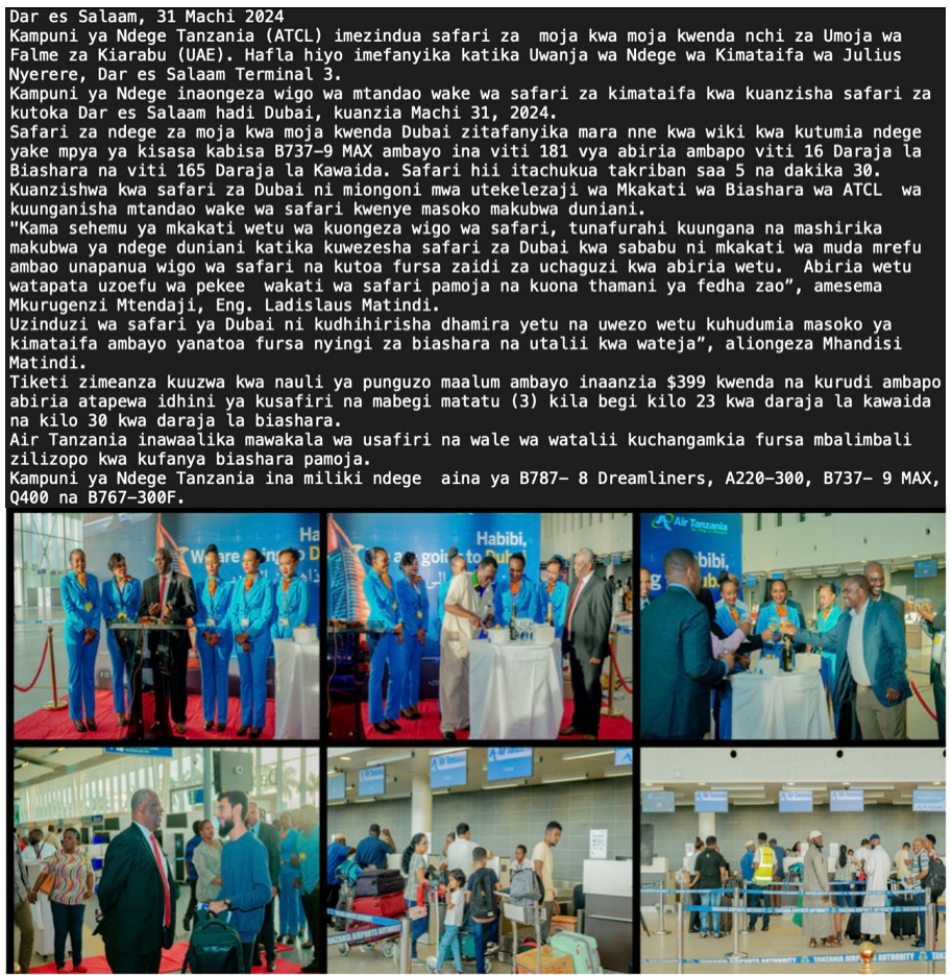}
    \caption*{(c) \textbf{Swahili blog} post, in \texttt{events}.}
\end{minipage}
\hfill
\begin{minipage}{0.26\columnwidth}
    \centering
    \includegraphics[width=\linewidth]{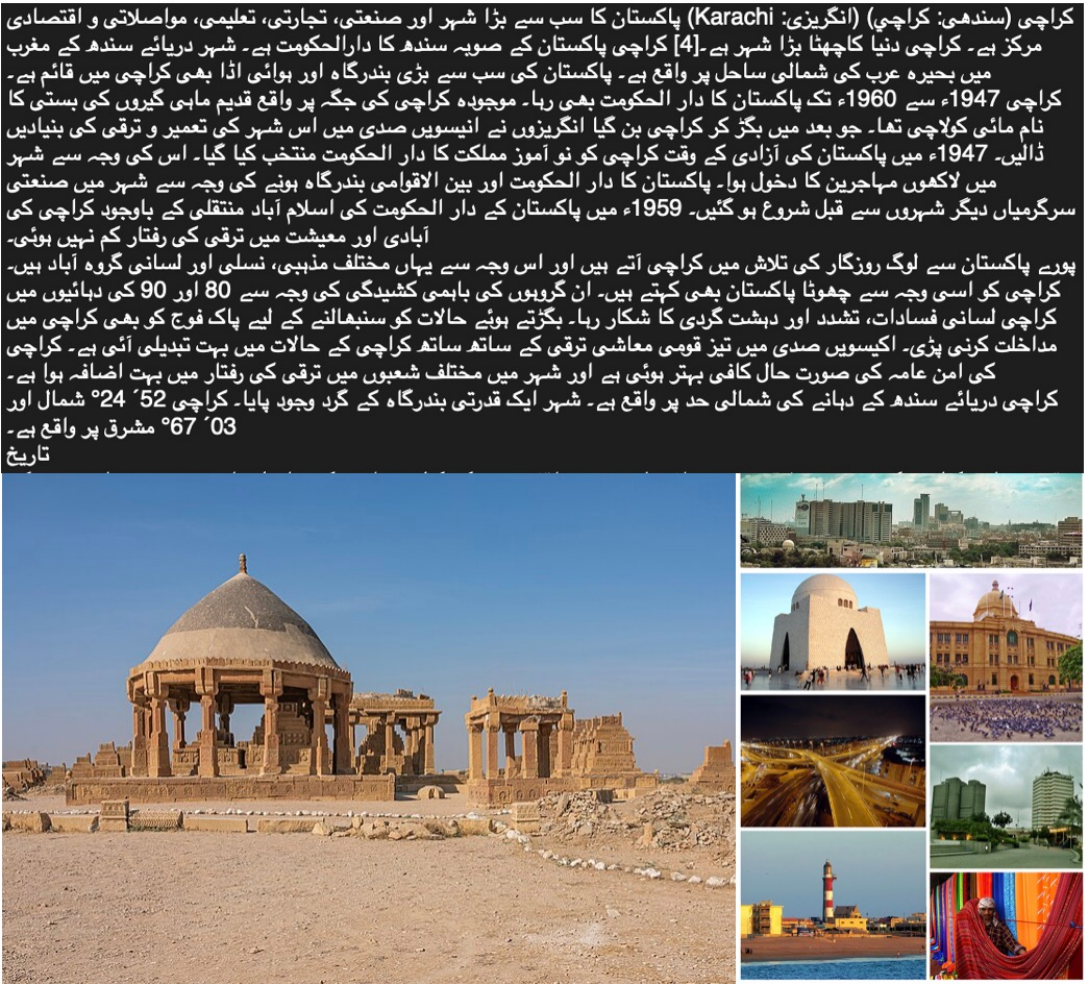}
    \caption*{(d) \textbf{Urdu Wikipedia} article, in \texttt{locations}.}
\end{minipage}
\caption{We collected all images and texts from each article, then used CLIP scores to select the most relevant image. 
\textit{We embed each chosen image with article-level prose.}
The chosen images and texts constitute the \textit{image-text} pairs in our benchmark. Texts shown above are truncated for brevity.}
\label{fig:CollectImageTextPairs}
\end{figure}

\subsection{Language Selection}
\label{subsc:LanguageSelection}
We chose languages in \emph{VLURes} under this criteria: (1) languages from distinct families (En is Indo-European, Jp is Japonic, Sw is Niger-Congo, Ur is Indo-European), (2) geographically diverse under the \cite{dryer2013wals} distinction (En from Eurasia, Jp from Eurasia, Sw from Africa, Ur from Eurasia)\footnote{Continent-wise: En, Jp, Sw, Ur originate from Europe, Asia, Africa, and Asia, respectively.}, (3) comprise different writing scripts (En uses Latin, Jp uses Kanji, Sw uses Latin, Ur uses Nastaliq), and (4) includes low-resource languages (Sw, Ur) \cite{nigatu-etal-2024-zenos}, alongside high resource languages (En, Jp).    

\subsection{Data Sources and Data Collection}
\label{sec:Data source}
Our dataset comprises En, Jp, Sw, and Ur image-text pairs. However, unlike image captioning with short texts \cite{lin2015microsoft, plummer2016flickr30k}, we provide ``rich context'' for each image. An article accompanies every image.

\subsubsection{Text+Image Collection}
We collected articles from Wikinews and Wikipedia due to their permissive license and from news websites, travelogues, restaurant review forums, and blogs. We kept all the texts and images available in each article. 
Examples are shown in Figure \ref{fig:CollectImageTextPairs}.\footnote{This is a link to Wikinews articles in English \url{https://en.wikinews.org/wiki/Main_Page}.}\footnote{Data dumps from Wikipedia are available at \url{https://dumps.wikimedia.org/enwiki/}.}
To diversify our dataset, we gathered image-text pairs for ten categories: \emph{animals, products, buildings, locations, events, food, drinks, hobbies, works of art, and organization}, because articles about these categories are useful to elicit cultural nuances per language, and the articles are readily available on web resources, such as Wikinews, Wikipedia, web forums, and the like. 
We collected over 1,000 image-text pairs per language and filtered them in the subsequent clean-up process.

Each `text' in \emph{VLURes} consists of the texts gathered from one URL.  
During the collection of images from URLs, we restrict the image extension to only \texttt{ png/jpeg/jpg}, and exclude image URLs that contain the following tokens: \texttt{logo, button, icon, plugin, widget}. 
Moreover, we exclude any articles that contain no valid, downloadable images. We manually checked all images and confirmed no NSFW images in our datasets. 

We conducted several data-cleaning steps, for example, removing articles whose main language is not the target language, and URLs containing zero words inside. 
Moreover, none of the collected documents contained any word on the `List of Dirty, Naughty, Obscene or Otherwise Bad Words'\footnote{https://github.com/LDNOOBW/List-of-Dirty-Naughty-Obscene-and-Otherwise-Bad-Words}. 

\subsubsection{Aligning Images with Texts} 
To assign an image, from among many images, to the texts contained in one article, we treat every article as a bipartite assignment problem \cite{hessel-etal-2019-unsupervised, kuhn1955hungarian}. 
Then, we deploy CLIP \cite{radford2021learning} ViT-L/14 to calculate the pairwise similarity between all text in the article and every image. 
Images that do not achieve a CLIP cosine similarity of at least 0.15 with any sentence are discarded. 
Finally, we sorted the CLIP similarity and selected only the image with the highest similarity score to create the \textit{image-text} pair from that article. 

\begin{wrapfigure}{r}{0.48\textwidth} 
    \vspace{-\intextsep} 
    \centering 
    \tiny
    \captionof{table}{Statistics of \emph{VLURes}.} 
    \label{tab:vlu-stats} 
    \resizebox{\linewidth}{!}{
      \begin{tabular}{l|cccc} 
        \toprule
        \textbf{Metric} & \textbf{En} & \textbf{Jp} & \textbf{Sw} & \textbf{Ur} \\
        \midrule
        \#Images              & 1000   & 1000   & 1130   & 996 \\
        \#Texts      & 1000   & 1000   & 1130   & 996 \\
        \midrule
        \multicolumn{5}{l}{\textbf{Text Lengths}} \\ %
        \midrule
        Min. Length     & 12     & 46     & 14     & 10 \\
        Max. Length     & 1716   & 3993   & 7766   & 3712 \\
        Median Length   & 242    & 381    & 335    & 231 \\
        Avg. Length     & 270    & 447    & 392    & 373 \\
        \bottomrule
      \end{tabular}
    } 
    \vspace{\medskipamount} 
    \includegraphics[width=\linewidth]{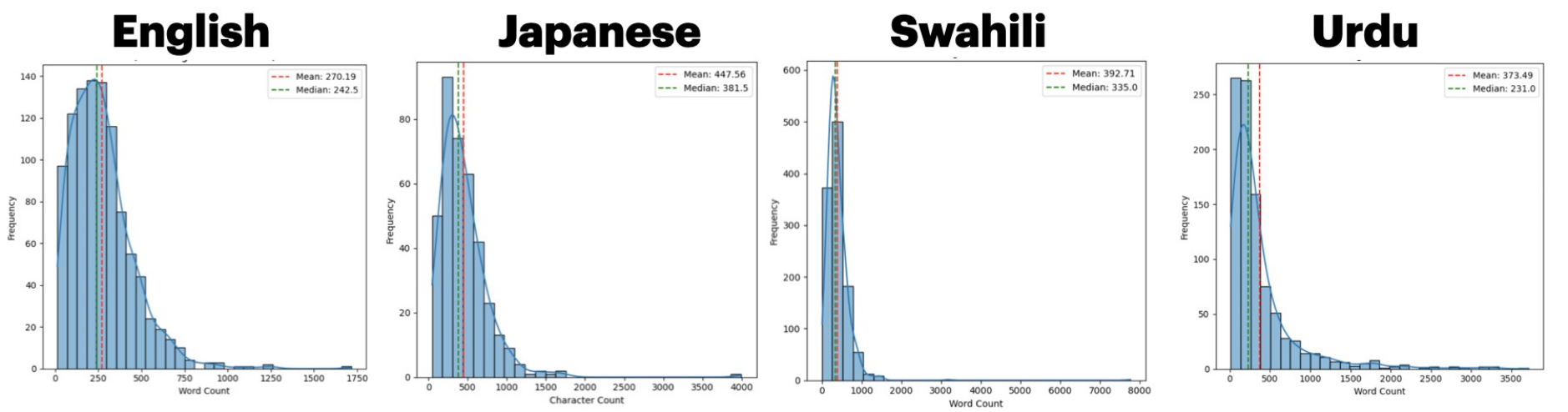}
    \captionof{figure}{Distribution of text lengths in \emph{VLURes} indicated by the number of words, for En, Sw, Ur; and the number of characters, for Jp.}
    \label{fig:vlu-lengths}
    \vspace{\medskipamount} 
    \includegraphics[width=\linewidth]{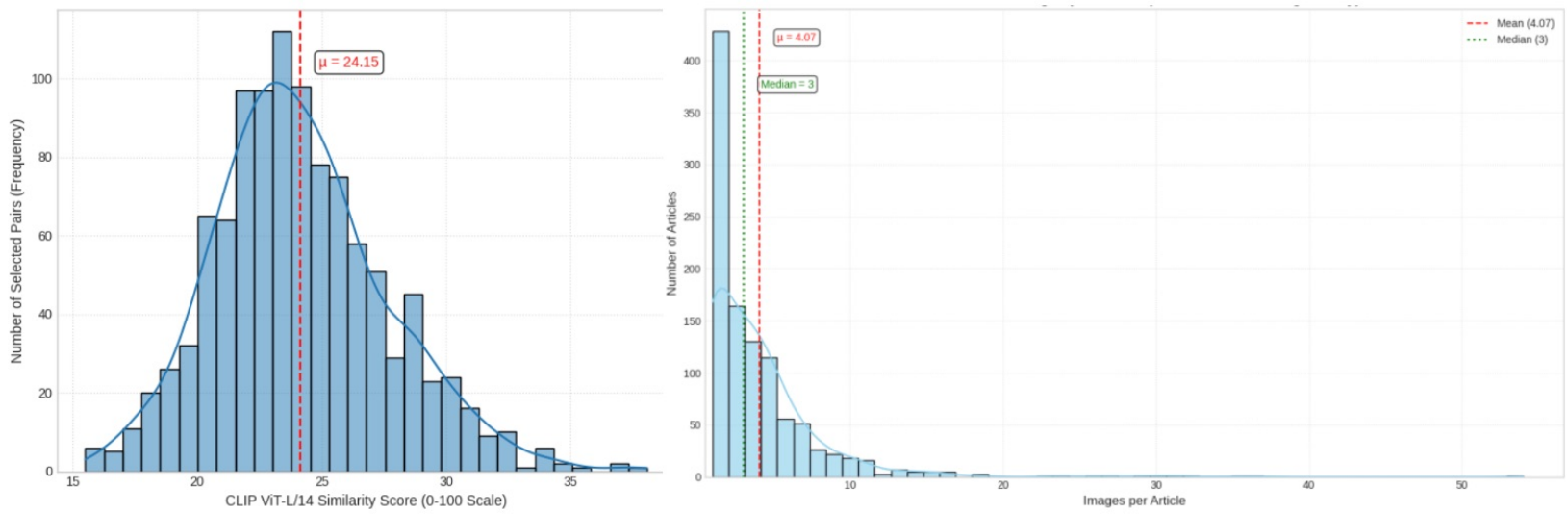} 
    \captionof{figure}{Example distribution of CLIP scores for Sw image-text pairs (left), and number of images in each Sw article (right).} 
    \label{fig:swCLIPdsbn} 
    \vspace{-35pt} 
\end{wrapfigure}

\subsection{Data Statistics}
\label{subsec:DataStatistics}
We show detailed statistics about data created in this study in Table~\ref{tab:vlu-stats} and Figures~\ref{fig:vlu-lengths}, \ref{fig:swCLIPdsbn}. 
We ensured an even distribution of image-text pairs (1K) across languages. On average, the texts contained in \emph{VLURes} are long enough: 270 words for En, 392 words for Sw, 373 words for Ur, and 447 characters for Jp. 
These long texts address the previous datasets' limited context per image-text pair. 
In Figure \ref{fig:vlu-lengths}, we visualize the length distributions of the texts across En, Jp, Sw, and Ur.  
Moreover, Figure \ref{fig:swCLIPdsbn} shows the distribution of CLIP scores for Sw image-text pairs, and the number of images in each Sw article.

\subsection{Benchmark Difficulty and Robustness}
\label{sec:formal_difficulty_robustness}
Beyond basic statistics, we formalize aspects of \emph{VLURes'} difficulty and its potential to evaluate model robustness, particularly concerning its multilingual nature and the inclusion of complex reasoning tasks with long-form text.\\
\textbf{Task and Language Difficulty.}
While inherent task difficulty is complex to define absolutely, we quantify the \textit{empirical difficulty} for a given VLM on a specific task $\tau_i$ and language $\ell \in \mathcal{L} = \{\text{En, Jp, Sw, Ur}\}$. Using the primary accuracy metric reported in our experiments (\S \ref{sec:Experiments}), we define a difficulty score:
\begin{equation}
    \text{Diff}(\tau_i, \ell | \text{VLM}) = 1 - \text{Acc}_{\text{VLM}}(\tau_i, \ell)
    \label{eq:difficulty_score}
\end{equation}
where $\text{Acc}_{\text{VLM}}(\tau_i, \ell)$ is the accuracy achieved by the specific VLM on task $\tau_i$ in language $\ell$. A higher $\text{Diff}$ score indicates greater difficulty for that model on that specific task-language pair. We hypothesize that factors, such as the (1)  low-resource nature of Swahili and Urdu, (2) requirement for grounding long textual contexts (\S \ref{subsec:dataPrinciples}), (3) and complexity of tasks like RU ($\tau_{\text{RU}}$) and U ($\tau_{\text{U}}$) contribute to higher average difficulty scores on \emph{VLURes} compared to benchmarks focused solely on high-resource languages or short captions. Analyzing $\text{Diff}(\tau_i, \ell | \text{VLM})$ across VLMs (using results in \S \ref{sec:Results}) enables a quantitative comparison of model capabilities on these challenging dimensions.
Overall, English is easier for VLMs but Urdu, and Swahili are most difficult (more details in \S \ref{subsec:Robustness}). 

\textbf{Cross-Lingual Robustness.}
A key goal of \emph{VLURes} is to assess how robustly VLMs perform across different languages. We can quantify the robustness of a VLM on a specific task $\tau_i$ by measuring the consistency of its performance across the languages $\mathcal{L}$. A simple measure uses the variance of the accuracy scores:
\begin{equation}
    \text{Robustness}(\tau_i | \text{VLM}) = 1 - \text{Var}_{\ell \in \mathcal{L}} \left( \text{Acc}_{\text{VLM}}(\tau_i, \ell) \right)
    \label{eq:robustness_variance}
\end{equation}
where $\text{Var}_{\ell \in \mathcal{L}}(\cdot)$ calculates the sample variance across the four languages. A robustness score closer to 1 indicates stable performance across languages for task $\tau_i$. In contrast, a lower score signifies significant performance disparities, potentially highlighting weaknesses in the model's multilingual capabilities or biases towards high-resource languages like English. \emph{VLURes} is designed to elicit lower robustness scores compared to English-centric benchmarks, thereby providing a clearer signal on the need for improved language-inclusive perception.

Alternatively, the distribution of performance across languages can be captured using the entropy of the normalized accuracy scores. Let $p(\ell | \tau_i, \text{VLM}) = \frac{\text{Acc}_{\text{VLM}}(\tau_i, \ell)}{\sum_{\ell' \in \mathcal{L}} \text{Acc}_{\text{VLM}}(\tau_i, \ell')}$ be the normalized performance in language $\ell$. The performance entropy for task $\tau_i$ is:
\begin{equation}
    H(\tau_i | \text{VLM}) = - \sum_{\ell \in \mathcal{L}} p(\ell | \tau_i, \text{VLM}) \log p(\ell | \tau_i, \text{VLM})
    \label{eq:performance_entropy}
\end{equation}
Higher entropy suggests more evenly distributed performance across languages (higher effective robustness), while lower entropy indicates performance concentrated in only a few languages.

By applying these metrics (Eq. \ref{eq:difficulty_score}, \ref{eq:robustness_variance}, \ref{eq:performance_entropy}) to the results obtained in \S \ref{sec:Results}, we provide a rigorous quantitative assessment of model performance profiles, highlighting that \emph{VLURes} is suitable to evaluate the VLMs because it is sufficiently complex and generalizable. 
Notably, robustness scores could only be computed for models demonstrating broad multilingual support (GPT-4o, GPT-4o-mini, Gemini 2.0 Flash Lite, Gemini 1.5 Flash 8B), as other models lacked comprehensive evaluation data across all four benchmark languages (En, Jp, Sw, Ur) primarily due to missing Swahili results. GPT-4o is the most robust VLM in our evaluation (details in \S \ref{subsec:Robustness}). 

\section{Experiments}
\label{sec:Experiments}
\subsection{Evaluation Criteria.}
To evaluate the quality of responses generated by VLMs, researchers may use human-curated rules or deploy template-matching rules \cite{lu2022learn}. 
However, we take inspiration from \cite{lu2024mathvista} and follow their evaluation steps, that is, (1) prompt VLMs to generate responses for a task, (2) extract the answer relevant to the task from the model's output, and (3) assign scores on a scale of 1 to 100 to quantify the quality of VLM response. 
We deploy \textit{accuracy} as the primary metric. \\
\textbf{Automatic Evaluation.}
For automatic/quantitative evaluation, we deploy \textit{Gemini 1.5 Pro} via the Gemini API as the \textit{LLM-judge}\footnote{We made this decision due to: the need to minimize API costs, the model's ability to understand both image and text inputs, large context window, and the high token/sec throughput of Gemini models, and the need to avoid overlap between the evaluation models and target models \cite{duan2024vlmevalkit, zheng2023judging, artificial_analysis_leaderboard}.}. 
To assign scores for the model's response to any task, the input to the LLM-judge includes: the image, texts, response from VLM, and prompt describing the evaluation criteria. Then, the LLM-judge outputs a score, on a scale of one to one hundred, such that one is the lowest and one hundred is the highest. \\
\textbf{Human Evaluation.} 
We conducted human evaluations to incorporate a human perspective in evaluating cultural nuance and multilingual understanding of VLMs and to validate rationales. 
We recruited eight evaluators, two native speakers per language (En, Jp, Sw, Ur), and they possess sufficient visual-linguistic knowledge. 
Each data sample (consisting of an image-text pair, task instructions, and the VLM's response/rationale) was judged independently by the two native speakers assigned to that language.  
These  native speakers assign a score between one and one hundred, aligning with the accuracy metric reported in our experiments.
Full evaluation guidelines in \S \ref{HumanEvalCriteria}. 
\subsection{Model Selection.} We chose ten VLMs that have shown competitive results in recent studies. We categorize VLMs as: (1) \textit{proprietary VLMs}, including GPT-4o \cite{openai2024gpt4o}, GPT-4o-mini \cite{openai2024gpt4omini}, Gemini 2.0 Flash Lite \cite{google2025gemini2flashlite}, Gemini 1.5 Flash 8B \cite{google2024gemini15flash8b}, and (2) \textit{open-source VLMs}, including LlaVa-NeXT Mistral 7B \cite{liu2024llavanext},  Qwen2VL 7B \cite{wang2024qwen2vlenhancingvisionlanguagemodels}, PALO 7B \cite{Rasheed_2025_WACV}, MAYA 8B \cite{alam2024maya}, LlaVa-NeXT Mistral 13B \cite{liu2024llavanext}, and PALO 13B \cite{Rasheed_2025_WACV}. 
\subsection{Experiment Setup.} 
We evaluate the models under \textit{zero-shot, one-shot} settings, before and after \textit{fine-tuning}. Note that we only fine-tuned the open-source models. We access all proprietary models via the relevant API and open-source models via the Hugging Face Transformers \cite{wolf-etal-2020-transformers} library. 
In all three settings, we further compare the results \textit{without rationales} against \textit{with rationales}.  
\begin{table*}[t]
    \centering
    \caption{Performance of VLMs on \textbf{eight VL tasks} under \colorbox{violet!20}{zero-shot} and \colorbox{teal!20}{one-shot} settings, measured by Accuracy (\%). 
            \textbf{Input:} English Texts + Images; \textbf{Output:} En, Jp, Sw, Ur responses.
            \colorbox{orange!20}{Shaded columns represent \{En\} in input and \{En\} in output VLM results.}
            }
    \resizebox{\textwidth}{!}{ 
    \begin{tabular}{r|| >{\columncolor{apricot}}c c c c | >{\columncolor{apricot}}c c c c | >{\columncolor{apricot}}c c c c | >{\columncolor{apricot}}c c c c | >{\columncolor{apricot}}c c c c || >{\columncolor{apricot}}c c c c | >{\columncolor{apricot}}c c c c | >{\columncolor{apricot}}c c c c }
        \toprule
        \multirow{2}{*}{\textbf{Model}} & \multicolumn{4}{c|}{\textbf{Object Recognition}} & \multicolumn{4}{c|}{\textbf{Scene Understanding}} & \multicolumn{4}{c|}{\textbf{Relation Understanding}} & \multicolumn{4}{c|}{\textbf{Semantic Segmentation}} & \multicolumn{4}{c|}{\textbf{Image Captioning}} & \multicolumn{4}{c||}{\textbf{Image-Text Matching}} & \multicolumn{4}{c|}{\textbf{Unrelatedness}} & \multicolumn{4}{c}{\textbf{Visual Question Answering}} \\
        \cmidrule(lr){2-5} \cmidrule(lr){6-9} \cmidrule(lr){10-13} \cmidrule(lr){14-17} \cmidrule(lr){18-21} \cmidrule(lr){22-25} \cmidrule(lr){26-29} \cmidrule(lr){30-33}
        & En & Jp & Sw & Ur & En & Jp & Sw & Ur & En & Jp & Sw & Ur & En & Jp & Sw & Ur & En & Jp & Sw & Ur & En & Jp & Sw & Ur & En & Jp & Sw & Ur & En & Jp & Sw & Ur \\
        \midrule
        \rowcolor{gray!20}
        \multicolumn{33}{c}{\textcolor{violet}{\textbf{Zero-shot, \textit{Without Rationales}}}} \\
        \midrule
        GPT-4o                 & 89.8 & 88.5 & 86.0 & 87.8 & 89.0 & 88.0 & 88.1 & 87.4 & 88.9 & 83.9 & 85.7 & 82.5 & 84.9 & 80.7 & 82.9 & 80.9 & 85.1 & 79.0 & 83.9 & 78.5 & 89.8 & \textbf{89.0} & 88.8 & 86.3 & 91.4 & 90.8 & 91.0 & 90.2 & 89.4 & 89.2 & 86.8 & 87.5 \\
        GPT-4o-mini            & 80.0 & 82.8 & 78.4 & 80.2 & 80.1 & 79.8 & 78.5 & 78.9 & 78.8 & 75.2 & 76.1 & 75.8 & 77.3 & 72.8 & 73.7 & 74.5 & 78.3 & 71.8 & 73.4 & 72.5 & 78.0 & 77.3 & 80.5 & 79.7 & 83.0 & 81.9 & 81.6 & 82.0 & 81.0 & 81.7 & 76.8 & 77.4 \\
        Gemini 2.0 Flash Lite  & 83.7 & 86.5 & 82.1 & 83.9 & 83.8 & 83.5 & 82.2 & 82.6 & 82.5 & 78.9 & 79.8 & 79.5 & 81.0 & 76.5 & 77.4 & 78.2 & 81.9 & 75.5 & 77.1 & 76.1 & 81.7 & 81.0 & 84.2 & 83.4 & 86.7 & 85.6 & 85.3 & 85.7 & 84.7 & 85.0 & 80.5 & 81.1 \\
        Gemini 1.5 Flash 8B    & 78.3 & 81.1 & 76.7 & 78.5 & 78.4 & 78.1 & 76.8 & 77.2 & 77.1 & 73.5 & 74.4 & 74.1 & 75.6 & 71.1 & 72.0 & 72.8 & 76.6 & 70.1 & 71.7 & 70.8 & 76.3 & 75.6 & 78.8 & 78.0 & 81.3 & 80.2 & 79.9 & 80.3 & 79.3 & 79.6 & 75.1 & 75.7 \\
        LlaVa Mistral 7B       & 38.6 & 41.4 & 0.0 & 0.0 & 38.6 & 37.8 & 0.0 & 0.0 & 37.3 & 33.7 & 0.0 & 0.0 & 36.2 & 30.9 & 0.0 & 0.0 & 36.6 & 30.2 & 0.0 & 0.0 & 37.2 & 36.4 & 0.0 & 0.0 & 41.5 & 40.4 & 0.0 & 0.0 & 40.2 & 39.5 & 0.0 & 0.0 \\ 
        Qwen2VL 7B             & 60.0 & 62.7 & 0.0 & 0.0 & 60.9 & 59.0 & 0.0 & 0.0 & 58.0 & 55.0 &  0.0 &  0.0 & 57.2 & 52.3 &  0.0 &  0.0 & 58.4 & 51.0 &  0.0 &  0.0 & 58.3 & 57.4 &  0.0 &  0.0 & 63.0 & 61.7 &  0.0 &  0.0 & 61.4 & 61.4 &  0.0 &  0.0 \\
        PALO 7B                & 39.5 & 40.5 & 0.0   & 50.7 & 39.6 & 37.8 & 0.0   & 48.9 & 38.3 & 33.2 & 0.0   & 45.9 & 36.8 & 30.8 & 0.0   & 44.8 & 37.8 & 29.6 & 0.0   & 42.5 & 37.5 & 35.3 & 0.0   & 49.7 & 42.5 & 39.9 & 0.0   & 52.0 & 40.5 & 39.3 & 0.0   & 47.4 \\
        MAYA 8B                & 40.8 & 46.6 & 0.0   & 49.8 & 40.9 & 43.6 & 0.0   & 48.5 & 39.6 & 39.0 & 0.0   & 45.5 & 38.1 & 36.6 & 0.0   & 44.1 & 39.3 & 35.0 & 0.0   & 42.1 & 38.8 & 41.1 & 0.0   & 49.3 & 43.8 & 45.7 & 0.0   & 51.6 & 41.8 & 45.1 & 0.0   & 47.0 \\
        LlaVa Mistral 13B      & 46.9 & 49.8 & 0.0 & 0.0 & 46.9 & 46.4 & 0.0 & 0.0 & 45.5 & 42.3 & 0.0 & 0.0 & 44.2 & 39.5 & 0.0 & 0.0 & 45.4 & 38.6 & 0.0 & 0.0 & 45.0 & 44.3 & 0.0 & 0.0 & 50.1 & 48.7 & 0.0 & 0.0 & 47.9 & 48.4 & 0.0 & 0.0 \\ 
        PALO 13B               & 41.1 & 43.4 & 0.0   & 57.2 & 41.1 & 40.4 & 0.0   & 55.0 & 39.8 & 36.0 & 0.0   & 52.9 & 38.6 & 33.9 & 0.0   & 51.0 & 39.0 & 32.4 & 0.0   & 49.0 & 39.2 & 38.1 & 0.0   & 56.5 & 44.3 & 42.9 & 0.0   & 59.2 & 42.5 & 42.3 & 0.0   & 54.0 \\
        \midrule
        \rowcolor{gray!20} 
        \multicolumn{33}{c}{\textcolor{violet}{\textbf{Zero-shot, \textit{With Rationales}}}} \\
        \midrule
        GPT-4o                & 89.1 & 86.5 & 88.2 & 87.5 & 89.7 & 88.3 & 87.3 & 86.2 & 88.0 & 85.9 & 86.3 & 85.2 & 88.3 & 86.2 & 86.0 & 84.5 & 86.5 & 82.2 & 83.3 & 82.3 & 89.5 & 86.7 & 87.0 & 86.0 & 91.3 & 90.7 & 90.8 & 90.3 & 88.2 & 86.7 & 86.0 & 86.1 \\
        GPT-4o-mini           & 80.0 & 83.0 & 78.3 & 81.9 & 79.8 & 79.9 & 78.5 & 78.4 & 79.6 & 79.8 & 77.7 & 76.4 & 80.0 & 80.7 & 76.8 & 75.6 & 78.2 & 75.4 & 71.6 & 72.7 & 79.4 & 78.0 & 75.5 & 76.2 & 83.1 & 82.1 & 80.9 & 82.4 & 76.0 & 76.6 & 76.6 & 78.3 \\
        Gemini 2.0 Flash Lite & 83.7 & 86.7 & 81.8 & 85.6 & 83.5 & 83.6 & 82.2 & 82.1 & 83.3 & 83.6 & 81.4 & 80.0 & 83.7 & 84.4 & 80.5 & 79.3 & 81.9 & 79.0 & 75.3 & 76.4 & 83.0 & 81.7 & 79.2 & 79.9 & 86.8 & 85.8 & 84.6 & 86.1 & 79.7 & 80.3 & 80.3 & 81.9 \\
        Gemini 1.5 Flash 8B   & 78.3 & 81.3 & 76.4 & 80.2 & 78.1 & 78.2 & 76.8 & 76.7 & 77.9 & 78.4 & 76.0 & 74.7 & 78.3 & 79.0 & 75.1 & 73.9 & 76.5 & 73.7 & 69.9 & 71.0 & 77.7 & 76.3 & 73.8 & 74.5 & 81.4 & 80.4 & 79.2 & 80.7 & 74.3 & 74.9 & 74.9 & 76.6 \\
        LlaVa Mistral 7B      & 39.4 & 41.9 & 0.0 & 0.0 & 38.3 & 38.4 & 0.0 & 0.0 & 38.1 & 38.1 & 0.0 & 0.0 & 39.2 & 39.3 & 0.0 & 0.0 & 37.0 & 34.4 & 0.0 & 0.0 & 38.0 & 36.7 & 0.0 & 0.0 & 42.0 & 41.3 & 0.0 & 0.0 & 34.7 & 35.1 & 0.0 & 0.0 \\ 
        Qwen2VL 7B            & 60.0 & 63.1 &  0.0 &  0.0 & 59.0 & 59.9 &  0.0 &  0.0 & 59.7 & 59.7 &  0.0 &  0.0 & 60.3 & 60.4 &  0.0 &  0.0 & 58.7 & 55.8 &  0.0 &  0.0 & 59.3 & 58.5 &  0.0 &  0.0 & 63.0 & 62.4 &  0.0 &  0.0 & 56.6 & 56.9 &  0.0 &  0.0 \\
        PALO 7B               & 39.5 & 41.5 & 0.0   & 51.9 & 39.3 & 37.2 & 0.0   & 48.0 & 39.1 & 37.9 & 0.0   & 46.4 & 39.5 & 38.7 & 0.0   & 45.6 & 37.8 & 33.3 & 0.0   & 42.7 & 38.9 & 36.0 & 0.0   & 46.2 & 42.6 & 40.1 & 0.0   & 52.4 & 35.5 & 34.6 & 0.0   & 48.3 \\
        MAYA 8B               & 40.8 & 46.8 & 0.0   & 51.6 & 40.6 & 43.7 & 0.0   & 48.1 & 40.4 & 43.5 & 0.0   & 46.0 & 40.8 & 44.5 & 0.0   & 45.0 & 39.0 & 39.2 & 0.0   & 42.3 & 40.2 & 41.8 & 0.0   & 45.8 & 43.9 & 45.9 & 0.0   & 52.1 & 36.8 & 40.4 & 0.0   & 47.9 \\
        LlaVa Mistral 13B     & 47.2 & 50.1 & 0.0 & 0.0 & 45.9 & 46.0 & 0.0 & 0.0 & 46.6 & 46.6 & 0.0 & 0.0 & 47.2 & 47.5 & 0.0 & 0.0 & 45.3 & 41.9 & 0.0 & 0.0 & 46.4 & 45.0 & 0.0 & 0.0 & 49.9 & 49.2 & 0.0 & 0.0 & 43.0 & 43.1 & 0.0 & 0.0 \\ 
        PALO 13B              & 41.0 & 44.0 & 0.0   & 58.8 & 40.8 & 40.7 & 0.0   & 55.4 & 40.6 & 40.7 & 0.0   & 53.3 & 41.4 & 41.5 & 0.0   & 52.4 & 39.3 & 36.4 & 0.0   & 49.4 & 40.4 & 39.5 & 0.0   & 53.2 & 44.3 & 43.6 & 0.0   & 59.7 & 37.3 & 37.7 & 0.0   & 55.3 \\
        \midrule
        \rowcolor{gray!20}
        \multicolumn{33}{c}{\textcolor{teal}{\textbf{One-shot, \textit{Without Rationales}}}} \\
        \midrule
        GPT-4o                & 90.5 & 87.2 & 87.3 & 86.6 & 90.5 & 88.1 & 88.4 & 87.6 & 89.4 & 87.4 & 87.3 & 84.4 & 89.1 & 85.3 & 86.0 & 84.9 & 88.0 & 78.9 & 85.3 & 79.7 & 89.1 & 86.8 & 87.2 & 86.1 & 91.0 & 90.5 & 91.0 & 90.6 & 89.2 & 87.0 & 88.1 & 86.2 \\
        GPT-4o-mini           & 82.0 & 79.2 & 79.0 & 78.2 & 82.1 & 79.2 & 78.9 & 78.6 & 79.6 & 78.1 & 76.9 & 75.8 & 81.5 & 77.9 & 77.4 & 75.7 & 80.0 & 73.0 & 74.4 & 73.3 & 80.2 & 76.9 & 77.9 & 75.2 & 83.0 & 82.5 & 82.5 & 82.2 & 81.7 & 78.9 & 77.6 & 77.1 \\
        Gemini 2.0 Flash Lite & 85.7 & 82.9 & 82.7 & 81.9 & 85.8 & 82.9 & 82.6 & 82.3 & 83.3 & 81.8 & 80.6 & 79.5 & 85.2 & 81.6 & 81.1 & 79.4 & 83.7 & 76.7 & 78.1 & 77.0 & 83.9 & 80.6 & 81.6 & 78.9 & 86.7 & 86.2 & 86.2 & 85.9 & 85.4 & 82.6 & 81.3 & 80.8 \\
        Gemini 1.5 Flash 8B   & 80.3 & 77.5 & 77.3 & 76.5 & 80.4 & 77.5 & 77.2 & 76.9 & 77.9 & 76.5 & 75.2 & 74.1 & 79.8 & 76.2 & 75.7 & 74.0 & 78.3 & 71.3 & 72.7 & 71.6 & 78.5 & 75.2 & 76.2 & 73.5 & 81.3 & 80.8 & 80.8 & 80.5 & 80.0 & 77.2 & 75.9 & 75.4 \\
        LlaVa Mistral 7B      & 40.6 & 37.8 & 0.0 & 0.0 & 41.2 & 37.9 & 0.0 & 0.0 & 38.0 & 36.5 & 0.0 & 0.0 & 40.0 & 36.0 & 0.0 & 0.0 & 39.3 & 32.3 & 0.0 & 0.0 & 39.0 & 35.4 & 0.0 & 0.0 & 42.0 & 41.2 & 0.0 & 0.0 & 40.0 & 36.7 & 0.0 & 0.0 \\ 
        Qwen2VL 7B            & 62.0 & 59.1 &  0.0 &  0.0 & 62.1 & 59.4 & 0.0 &  0.0 & 59.6 & 58.1 &  0.0 &  0.0 & 61.7 & 57.9 &  0.0 &  0.0 & 60.3 & 53.0 &  0.0 &  0.0 & 60.3 & 56.5 &  0.0 &  0.0 & 63.7 & 62.5 &  0.0 &  0.0 & 61.1 & 58.5 &  0.0 &  0.0 \\
        PALO 7B               & 41.5 & 37.7 & 0.0   & 48.4 & 41.6 & 37.2 & 0.0   & 48.6 & 39.1 & 36.0 & 0.0   & 45.4 & 41.0 & 35.7 & 0.0   & 45.9 & 39.5 & 31.4 & 0.0   & 43.3 & 39.7 & 34.8 & 0.0   & 45.2 & 42.5 & 40.5 & 0.0   & 52.2 & 41.2 & 36.9 & 0.0   & 47.1 \\
        MAYA 8B               & 42.8 & 43.0 & 0.0   & 47.8 & 42.9 & 43.0 & 0.0   & 48.2 & 40.4 & 41.8 & 0.0   & 45.3 & 42.3 & 41.8 & 0.0   & 45.3 & 40.8 & 36.8 & 0.0   & 42.9 & 41.0 & 40.7 & 0.0   & 44.8 & 43.8 & 46.3 & 0.0   & 51.8 & 42.5 & 42.7 & 0.0   & 46.7 \\
        LlaVa Mistral 13B     & 49.3 & 46.4 & 0.0 & 0.0 & 49.2 & 46.3 & 0.0 & 0.0 & 46.3 & 45.4 & 0.0 & 0.0 & 48.3 & 44.6 & 0.0 & 0.0 & 47.2 & 40.3 & 0.0 & 0.0 & 47.4 & 43.4 & 0.0 & 0.0 & 50.3 & 49.3 & 0.0 & 0.0 & 47.9 & 45.4 & 0.0 & 0.0 \\ 
        PALO 13B              & 43.7 & 40.2 & 0.0   & 55.2 & 43.3 & 40.2 & 0.0   & 55.0 & 40.6 & 39.1 & 0.0   & 52.6 & 42.4 & 38.5 & 0.0   & 52.5 & 41.1 & 34.0 & 0.0   & 50.4 & 41.2 & 37.6 & 0.0   & 52.4 & 44.6 & 43.7 & 0.0   & 59.2 & 42.4 & 39.7 & 0.0   & 54.1 \\
        \midrule
        \rowcolor{gray!20} 
        \multicolumn{33}{c}{\textcolor{teal}{\textbf{One-shot, \textit{With Rationales}}}} \\
        \midrule
        GPT-4o                & \textbf{90.8} & \textbf{89.8} & \textbf{89.9} & \textbf{88.1} & \textbf{90.7} & \textbf{88.8} & \textbf{88.8} & \textbf{87.9} & \textbf{90.4} & \textbf{89.1} & \textbf{89.2} & \textbf{87.7} & \textbf{90.8} & \textbf{89.1} & \textbf{89.3} & \textbf{86.4} & \textbf{88.5} & \textbf{84.9} & \textbf{85.5} & \textbf{84.0} & \textbf{90.3} & 88.3 & \textbf{88.9} & \textbf{87.8} & \textbf{91.7} & \textbf{91.3} & \textbf{91.6} & \textbf{91.0} & \textbf{90.8} & \textbf{90.7} & \textbf{88.5} & \textbf{88.4} \\
        GPT-4o-mini           & 82.5 & 83.3 & 80.1 & 80.4 & 82.3 & 80.2 & 78.4 & 78.2 & 81.6 & 81.8 & 78.9 & 78.5 & 82.6 & 80.5 & 79.7 & 78.2 & 80.3 & 76.6 & 76.0 & 74.7 & 81.9 & 76.8 & 78.6 & 79.3 & 83.3 & 81.1 & 82.0 & 82.7 & 82.6 & 79.8 & 78.7 & 79.3 \\
        Gemini 2.0 Flash Lite & 86.2 & 87.0 & 83.8 & 84.1 & 85.9 & 83.8 & 82.1 & 81.8 & 85.3 & 85.5 & 82.6 & 82.2 & 86.3 & 84.2 & 83.4 & 81.9 & 84.0 & 80.3 & 79.7 & 78.4 & 85.6 & 80.5 & 82.3 & 83.0 & 87.0 & 84.8 & 85.7 & 86.4 & 86.3 & 83.5 & 82.4 & 83.0 \\
        Gemini 1.5 Flash 8B   & 80.8 & 81.6 & 78.6 & 78.7 & 80.5 & 78.5 & 76.7 & 76.5 & 79.9 & 80.3 & 77.4 & 76.8 & 80.9 & 78.8 & 78.0 & 76.5 & 78.6 & 74.9 & 74.3 & 73.0 & 80.2 & 75.1 & 76.9 & 77.6 & 81.6 & 79.4 & 80.3 & 81.0 & 80.9 & 78.1 & 77.0 & 77.6 \\
        LlaVa Mistral 7B      & 40.6 & 42.4 & 0.0 & 0.0 & 40.7 & 38.9 & 0.0 & 0.0 & 39.5 & 40.2 & 0.0 & 0.0 & 40.5 & 39.2 & 0.0 & 0.0 & 39.4 & 34.5 & 0.0 & 0.0 & 40.2 & 35.0 & 0.0 & 0.0 & 41.5 & 40.0 & 0.0 & 0.0 & 41.1 & 38.1 & 0.0 & 0.0 \\ 
        Qwen2VL 7B            & 62.5 & 63.6 & 60.5 & 60.5 & 62.2 & 60.3 & 58.0 & 58.4 & 61.0 & 61.6 & 58.7 & 58.6 & 62.6 & 60.3 & 59.6 & 58.3 & 60.7 & 56.4 & 56.5 & 54.8 & 61.3 & 56.1 & 58.5 & 59.3 & 63.5 & 61.3 & 62.2 & 62.8 & 62.5 & 59.2 & 58.3 & 59.3 \\
        PALO 7B               & 42.2 & 41.8 & 0.0   & 50.4 & 41.7 & 38.1 & 0.0   & 48.1 & 41.1 & 39.2 & 0.0   & 48.3 & 42.1 & 38.5 & 0.0   & 48.6 & 39.8 & 34.6 & 0.0   & 44.0 & 41.4 & 34.8 & 0.0   & 49.3 & 42.8 & 39.1 & 0.0   & 52.3 & 42.1 & 37.8 & 0.0   & 49.5 \\
        MAYA 8B               & 43.3 & 47.7 & 0.0   & 50.0 & 43.0 & 44.0 & 0.0   & 47.8 & 42.4 & 45.6 & 0.0   & 48.1 & 43.4 & 44.3 & 0.0   & 47.8 & 41.1 & 40.4 & 0.0   & 44.3 & 42.7 & 40.5 & 0.0   & 48.9 & 44.1 & 44.9 & 0.0   & 52.3 & 43.4 & 43.6 & 0.0   & 48.6 \\
        LlaVa Mistral 13B     & 49.5 & 50.4 & 0.0 & 0.0 & 49.4 & 47.4 & 0.0 & 0.0 & 48.6 & 48.8 & 0.0 & 0.0 & 49.6 & 47.8 & 0.0 & 0.0 & 47.4 & 43.6 & 0.0 & 0.0 & 48.5 & 43.5 & 0.0 & 0.0 & 50.4 & 48.3 & 0.0 & 0.0 & 49.2 & 46.0 & 0.0 & 0.0 \\ 
        PALO 13B              & 43.5 & 44.1 & 0.0   & 57.8 & 43.3 & 41.2 & 0.0   & 55.4 & 42.5 & 42.7 & 0.0   & 55.5 & 43.6 & 41.3 & 0.0   & 55.0 & 41.3 & 37.5 & 0.0   & 51.3 & 42.3 & 37.0 & 0.0   & 56.0 & 44.0 & 42.9 & 0.0   & 59.7 & 43.0 & 40.8 & 0.0   & 56.0 \\
        \midrule
        \midrule
        \textbf{Human Performance}    & 97.5 & -- & -- & -- & 98.8 & -- & -- & --  & 99.0 & -- & -- & -- & 97.5 & -- & -- & -- & 98.7 & -- & -- & -- & 94.7 & -- & -- & -- & 93.8 & -- & -- & -- & 81.9 & -- & -- & -- \\ 
        \bottomrule
    \end{tabular}
    }
    \label{tab:ZerooneshotAccuracyEn}   
\end{table*}
\section{Results}
\label{sec:Results}
We compare the performance of several models on \emph{VLURes}. 
We set the \textit{zero-shot no rationale} setting as our baseline VLM assessment. In addition, we establish a baseline for human performance in En, Jp, Sw, and Ur by recruiting eight native speakers. All human evaluators are graduate students, and they possess sufficient vision-language knowledge.  

In Tables \ref{tab:ZerooneshotAccuracyEn}, \ref{tab:FinetuneAccuracyEn}, we show the performance for all VLMs when the input contains image-text pairs and those texts are in En. Then, VLMs generate responses in En, Jp, Sw, Ur. 
This setting results in \textit{En-En} as well as \textit{En-Jp, En-Sw}, and \textit{En-Ur} outputs, making it possible to evaluate VLM performance when the language of the VLM response is different from the language in the input texts.\footnote{We conduct this evaluation because VLMs are known to perform better with English prompts due to training data bias than with prompts in other languages \cite{geigle-etal-2024-babel, chen2023pali}. We check for any VLM performance drops.}

Under all settings, the \textit{1-shot with rationale} setting results in the best performance, such that GPT-4o achieves slightly above 90\% accuracy across all eight tasks. 
GPT-4o achieves the best accuracy compared to other VLMs but still lags human performance in almost all tasks.
For example, in OR, GPT-4o scores a 90.8\% accuracy but falls 6.7\% short of human performance. 
The gap highlights the room for improvement on \emph{VLURes}. 
In contrast, open-source models achieve weaker performance on \emph{VLURes} than proprietary models. 
In OR, the best open-source model, Qwen2VL, achieves 62.5\%, lagging GPT-4o by 28.3 points.  
The performance deficit may be attributed to factors including limited access to high-quality data during training, a lower level of sophistication in the model architecture, the absence of advanced multimodal fusion techniques needed for seamless interaction between modalities, and the lack of resources or data necessary for extensive fine-tuning.
After fine-tuning, Qwen2VL accuracy increased from 62.5\% to 71.3\%, promising additional benefits in accuracy with additional fine-tuning. 

Tables for additional results, given the same language in both input and output, that is \textbf{Jp-Jp, Sw-Sw}, and \textbf{Ur-Ur}, are detailed in \S \ref{Appendix_ExperimentsJapaneseData}, \ref{Appendix_ExperimentsSwahiliData}, \ref{Appendix_ExperimentsUrduData}.   
\begin{table*}[t]
    \centering
    \caption{ Performance of VLMs on \textbf{eight VL tasks} under \colorbox{blue!20}{finetuning} settings, measured by Accuracy (\%). 
                \textbf{Input:} English Texts + Images; \textbf{Output:} En, Jp, Sw, Ur responses.
                \colorbox{orange!20}{Shaded columns represent \{En\} in input and \{En\} in output VLM results.}
            }
    \resizebox{\textwidth}{!}{
    \begin{tabular}{r|| >{\columncolor{apricot}}c c c c | >{\columncolor{apricot}}c c c c | >{\columncolor{apricot}}c c c c | >{\columncolor{apricot}}c c c c | >{\columncolor{apricot}}c c c c || >{\columncolor{apricot}}c c c c | >{\columncolor{apricot}}c c c c | >{\columncolor{apricot}}c c c c }
        \toprule
        \multirow{2}{*}{\textbf{Model}} & \multicolumn{4}{c|}{\textbf{Object Recognition}} & \multicolumn{4}{c|}{\textbf{Scene Understanding}} & \multicolumn{4}{c|}{\textbf{Relation Understanding}} & \multicolumn{4}{c|}{\textbf{Semantic Segmentation}} & \multicolumn{4}{c|}{\textbf{Image Captioning}} & \multicolumn{4}{c||}{\textbf{Image-Text Matching}} & \multicolumn{4}{c|}{\textbf{Unrelatedness}} & \multicolumn{4}{c}{\textbf{Visual Question Answering}} \\
        \cmidrule(lr){2-5} \cmidrule(lr){6-9} \cmidrule(lr){10-13} \cmidrule(lr){14-17} \cmidrule(lr){18-21} \cmidrule(lr){22-25} \cmidrule(lr){26-29} \cmidrule(lr){30-33}
        & En & Jp & Sw & Ur & En & Jp & Sw & Ur & En & Jp & Sw & Ur & En & Jp & Sw & Ur & En & Jp & Sw & Ur & En & Jp & Sw & Ur & En & Jp & Sw & Ur & En & Jp & Sw & Ur \\
        \midrule
        \rowcolor{blue!20}
        \multicolumn{33}{c}{\textcolor{violet}{\textbf{Zero-shot, \textit{Without Rationales}}}} \\
        \midrule
        LlaVa Mistral 7B       & 48.9 & 51.7 & 0.0 & 0.0 & 48.9 & 48.1 & 0.0 & 0.0 & 47.6 & 44.0 & 0.0 & 0.0 & 46.5 & 41.2 & 0.0 & 0.0 & 46.9 & 40.5 & 0.0 & 0.0 & 47.5 & 46.7 & 0.0 & 0.0 & 51.8 & 50.7 & 0.0 & 0.0 & 50.5 & 49.8 & 0.0 & 0.0 \\
        Qwen2VL 7B             & 68.8 & 71.5 &  0.0  &  0.0  & 69.7 & 67.8 &  0.0  &  0.0  & 66.8 & 63.8 &  0.0  &  0.0  & 66.0 & 61.1 &  0.0  &  0.0  & 67.2 & 59.8 &  0.0  &  0.0  & 67.1 & 66.2 &  0.0  &  0.0 & 71.8 & 70.5 &  0.0 &  0.0   & 70.2 & 70.2 & 0.0   &  0.0  \\
        PALO 7B                & 48.3 & 49.3 & 0.0   & 59.5 & 48.4 & 46.6 & 0.0   & 57.7 & 47.1 & 42.0 & 0.0   & 54.7 & 45.6 & 39.6 & 0.0   & 53.6 & 46.6 & 38.4 & 0.0   & 51.3 & 46.3 & 44.1 & 0.0   & 58.5 & 51.3 & 48.7 & 0.0   & 60.8 & 49.3 & 48.1 & 0.0   & 56.2 \\
        MAYA 8B                & 49.6 & 55.4 & 0.0   & 58.6 & 49.7 & 52.4 & 0.0   & 57.3 & 48.4 & 47.8 & 0.0   & 54.3 & 46.9 & 45.4 & 0.0   & 52.9 & 48.1 & 43.8 & 0.0   & 50.9 & 47.6 & 49.9 & 0.0   & 58.1 & 52.6 & 54.5 & 0.0   & 60.4 & 50.6 & 53.9 & 0.0   & 55.8 \\
        LlaVa Mistral 13B      & 54.8 & 57.7 & 0.0 & 0.0 & 54.8 & 54.3 & 0.0 & 0.0 & 53.4 & 50.2 & 0.0 & 0.0 & 52.1 & 47.4 & 0.0 & 0.0 & 53.3 & 46.5 & 0.0 & 0.0 & 52.9 & 52.2 & 0.0 & 0.0 & 58.0 & 56.6 & 0.0 & 0.0 & 55.8 & 56.3 & 0.0 & 0.0 \\
        PALO 13B               & 49.9 & 52.2 & 0.0   & 66.0 & 49.9 & 49.2 & 0.0   & 63.8 & 48.6 & 44.8 & 0.0   & 61.7 & 47.4 & 42.7 & 0.0   & 59.8 & 47.8 & 41.2 & 0.0   & 57.8 & 48.0 & 46.9 & 0.0   & 65.3 & 53.1 & 51.7 & 0.0   & 68.0 & 51.3 & 51.1 & 0.0   & 62.8 \\
        \midrule
        \rowcolor{blue!20}
        \multicolumn{33}{c}{\textcolor{violet}{\textbf{Zero-shot, \textit{With Rationales}}}} \\
        \midrule
        LlaVa Mistral 7B      & 49.7 & 52.2 & 0.0 & 0.0 & 48.6 & 48.7 & 0.0 & 0.0 & 48.4 & 48.7 & 0.0 & 0.0 & 49.5 & 49.6 & 0.0 & 0.0 & 47.3 & 44.7 & 0.0 & 0.0 & 48.3 & 47.0 & 0.0 & 0.0 & 52.3 & 51.6 & 0.0 & 0.0 & 45.0 & 45.4 & 0.0 & 0.0 \\
        Qwen2VL 7B            & 68.8 & 71.9 & 0.0 & 0.0 & 67.8 & 68.7 & 0.0 & 0.0 & 68.5 & 68.5 & 0.0 & 0.0 & 69.1 & 69.2 & 0.0 & 0.0 & 67.5 & 64.6 & 0.0 & 0.0 & 68.1 & 67.3 & 0.0 & 0.0 & 71.8 & 71.2 & 0.0 & 0.0 & 65.4 & 65.7 & 0.0 & 0.0 \\
        PALO 7B               & 48.3 & 50.3 & 0.0   & 60.7 & 48.1 & 46.0 & 0.0   & 56.8 & 47.9 & 46.7 & 0.0   & 55.2 & 48.3 & 47.5 & 0.0   & 54.4 & 46.6 & 42.1 & 0.0   & 51.5 & 47.7 & 44.8 & 0.0   & 55.0 & 51.4 & 48.9 & 0.0   & 61.2 & 44.3 & 43.4 & 0.0   & 57.1 \\
        MAYA 8B               & 49.6 & 55.6 & 0.0   & 60.4 & 49.4 & 52.5 & 0.0   & 56.9 & 49.2 & 52.3 & 0.0   & 54.8 & 49.6 & 53.3 & 0.0   & 53.8 & 47.8 & 48.0 & 0.0   & 51.1 & 49.0 & 50.6 & 0.0   & 54.6 & 52.7 & 54.7 & 0.0   & 60.9 & 45.6 & 49.2 & 0.0   & 56.7 \\
        LlaVa Mistral 13B     & 55.1 & 58.0 & 0.0 & 0.0 & 53.8 & 53.9 & 0.0 & 0.0 & 54.6 & 54.6 & 0.0 & 0.0 & 55.1 & 55.4 & 0.0 & 0.0 & 53.3 & 49.9 & 0.0 & 0.0 & 54.4 & 53.0 & 0.0 & 0.0 & 57.9 & 57.2 & 0.0 & 0.0 & 50.9 & 51.0 & 0.0 & 0.0 \\
        PALO 13B              & 49.8 & 52.8 & 0.0   & 67.6 & 49.6 & 49.5 & 0.0   & 64.2 & 49.4 & 49.5 & 0.0   & 62.1 & 50.2 & 50.3 & 0.0   & 61.2 & 48.1 & 45.2 & 0.0   & 58.2 & 49.2 & 48.3 & 0.0   & 62.0 & 53.1 & 52.4 & 0.0   & 68.5 & 46.1 & 46.5 & 0.0   & 64.1 \\
        \midrule
        \rowcolor{blue!20}
        \multicolumn{33}{c}{\textcolor{teal}{\textbf{One-shot, \textit{Without Rationales}}}} \\
        \midrule
        LlaVa Mistral 7B      & 50.9 & 48.1 & 0.0 & 0.0 & 51.5 & 48.2 & 0.0 & 0.0 & 48.3 & 46.8 & 0.0 & 0.0 & 50.3 & 46.3 & 0.0 & 0.0 & 49.6 & 42.6 & 0.0 & 0.0 & 49.3 & 45.7 & 0.0 & 0.0 & 52.3 & 51.5 & 0.0 & 0.0 & 50.3 & 47.0 & 0.0 & 0.0 \\
        Qwen2VL 7B            & 70.8 & 67.9 & 0.0 & 0.0 & 70.9 & 68.2 & 0.0 & 0.0 & 68.4 & 66.9 & 0.0 & 0.0 & 70.5 & 66.7 & 0.0 & 0.0 & 69.1 & 61.8 & 0.0 & 0.0 & 69.1 & 65.3 & 0.0 & 0.0 & 72.5 & 71.3 & 0.0 & 0.0 & 69.9 & 67.3 & 0.0  & 0.0 \\
        PALO 7B               & 50.3 & 46.5 & 0.0   & 57.2 & 50.4 & 46.0 & 0.0   & 57.4 & 47.9 & 44.8 & 0.0   & 54.2 & 49.8 & 44.5 & 0.0   & 54.7 & 48.3 & 40.2 & 0.0   & 52.1 & 48.5 & 43.6 & 0.0   & 54.0 & 51.3 & 49.3 & 0.0   & 61.0 & 50.0 & 45.7 & 0.0   & 55.9 \\
        MAYA 8B               & 51.6 & 51.8 & 0.0   & 56.6 & 51.7 & 51.8 & 0.0   & 57.0 & 49.2 & 50.6 & 0.0   & 54.1 & 51.1 & 50.6 & 0.0   & 54.1 & 49.6 & 45.6 & 0.0   & 51.7 & 49.8 & 49.5 & 0.0   & 53.6 & 52.6 & 55.1 & 0.0   & 60.6 & 51.3 & 51.5 & 0.0   & 55.5 \\
        LlaVa Mistral 13B     & 57.2 & 54.3 & 0.0 & 0.0 & 57.1 & 54.2 & 0.0 & 0.0 & 54.2 & 53.3 & 0.0 & 0.0 & 56.2 & 52.5 & 0.0 & 0.0 & 55.1 & 48.2 & 0.0 & 0.0 & 55.3 & 51.3 & 0.0 & 0.0 & 58.2 & 57.2 & 0.0 & 0.0 & 55.8 & 53.3 & 0.0 & 0.0 \\
        PALO 13B              & 52.5 & 49.0 & 0.0   & 64.0 & 52.1 & 49.0 & 0.0   & 63.8 & 49.4 & 47.9 & 0.0   & 61.4 & 51.2 & 47.3 & 0.0   & 61.3 & 49.9 & 42.8 & 0.0   & 59.2 & 50.0 & 46.4 & 0.0   & 61.2 & 53.4 & 52.5 & 0.0   & 68.0 & 51.2 & 48.5 & 0.0   & 62.9 \\
        \midrule
        \rowcolor{blue!20}
        \multicolumn{33}{c}{\textcolor{teal}{\textbf{One-shot, \textit{With Rationales}}}} \\
        \midrule
         LlaVa Mistral 7B     & 50.9 & 52.7 & 0.0 & 0.0 & 51.0 & 49.2 & 0.0 & 0.0 & 49.8 & 50.5 & 0.0 & 0.0 & 50.8 & 49.5 & 0.0 & 0.0 & 49.7 & 44.8 & 0.0 & 0.0 & 50.5 & 45.3 & 0.0 & 0.0 & 51.8 & 50.3 & 0.0 & 0.0 & 51.4 & 48.4 & 0.0 & 0.0 \\
        Qwen2VL 7B            & \textbf{71.3} & \textbf{72.4} & 0.0 & 0.0 & \textbf{70.9} & \textbf{69.1} & 0.0 & 0.0 & \textbf{69.8} & \textbf{70.4} & 0.0 & 0.0 & \textbf{71.4} & \textbf{69.1} & 0.0 & 0.0 & \textbf{65.2} & \textbf{65.3} & 0.0 & 0.0 & \textbf{64.9} & \textbf{67.3} & 0.0 & 0.0 & \textbf{70.1} & \textbf{71.0} & 0.0 & 0.0 & \textbf{68.0} & \textbf{67.1} & 0.0  & 0.0 \\
        PALO 7B               & 51.0 & 50.6 & 0.0   & 59.2 & 50.5 & 46.9 & 0.0   & 56.9 & 49.9 & 48.0 & 0.0   & 57.1 & 50.9 & 47.3 & 0.0   & 57.4 & 48.6 & 43.4 & 0.0   & 52.8 & 50.2 & 43.6 & 0.0   & 58.1 & 51.6 & 47.9 & 0.0   & 61.1 & 50.9 & 46.6 & 0.0   & 58.3 \\
        MAYA 8B               & 52.1 & 56.5 & 0.0   & 58.8 & 51.8 & 52.8 & 0.0   & 56.6 & 51.2 & 54.4 & 0.0   & 56.9 & 52.2 & 53.1 & 0.0   & 56.6 & 49.9 & 49.2 & 0.0   & 53.1 & 51.5 & 49.3 & 0.0   & 57.7 & 52.9 & 53.7 & 0.0   & 61.1 & 52.2 & 52.4 & 0.0   & 57.4 \\
        LlaVa Mistral 13B     & 57.4 & 58.3 & 0.0 & 0.0 & 57.3 & 55.3 & 0.0 & 0.0 & 56.5 & 56.7 & 0.0 & 0.0 & 57.5 & 55.7 & 0.0 & 0.0 & 55.3 & 51.5 & 0.0 & 0.0 & 56.4 & 51.4 & 0.0 & 0.0 & 58.3 & 56.2 & 0.0 & 0.0 & 57.1 & 53.9 & 0.0 & 0.0 \\
        PALO 13B              & 52.3 & 52.9 & 0.0   & 66.6 & 52.1 & 50.0 & 0.0   & 64.2 & 51.3 & 51.5 & 0.0   & 64.3 & 52.4 & 50.1 & 0.0   & 63.8 & 50.1 & 46.3 & 0.0   & 60.1 & 51.1 & 45.8 & 0.0   & 64.8 & 52.8 & 51.7 & 0.0   & 68.5 & 51.8 & 49.6 & 0.0   & 64.8 \\
        \bottomrule
    \end{tabular}
    }
    \label{tab:FinetuneAccuracyEn} 
\end{table*}
\paragraph{Comparison with MaRVL.}
\label{sec:MaRVL}
\cite{liu-etal-2021-visually} introduced the MaRVL dataset, mentioned in \S \ref{sec:Introduction}. 
Moreover, \cite{bugliarello-etal-2022-iglue} created IGLUE, a benchmark comprising 20 languages, including Sw, and four tasks: natural language inference, question answering, reasoning, and cross-modal retrieval. 
However, the Sw portion of IGLUE uses data from MaRVL. 
Hence, we compare VLM performance on Sw data in our benchmark to Sw data in MaRVL.
The results in Table \ref{tab:MaRVLsummarized} indicate that GPT-4o achieves better accuracy on MaRVL than on \emph{VLURes}, mainly due to the short captions which are found in MaRVL. 
GPT-4o achieves more significant gains on \textit{U, VQA}, that is 10.7\% and 7.8\%, respectively, because the model comprehends the short captions in MaRVL much better. 
Detailed results in \S \ref {subsec:MaRVL}.
\begin{table*}[t]
    \centering
    \caption{Performance of VLMs on \textbf{eight VL tasks}, measured by Accuracy (\%). 
            \textbf{Input:} Swahili Texts + Images; \textbf{Output:} Sw responses. All images and texts are from the \textbf{MaRVL} dataset.
            We report the best results under \textit{one-shot with rationales} setting, from GPT-4o.}
    \resizebox{\textwidth}{!}{ 
    \begin{tabular}{r|| >{\columncolor{apricot}}c c | >{\columncolor{apricot}}c c | >{\columncolor{apricot}}c c | >{\columncolor{apricot}}c c | >{\columncolor{apricot}}c c | >{\columncolor{apricot}}c c || >{\columncolor{apricot}}c c | >{\columncolor{apricot}}c c }
        \toprule
        \multirow{2}{*}{\textbf{Model}} & \multicolumn{2}{c|}{\textbf{Object Recognition}} & \multicolumn{2}{c|}{\textbf{Scene Understanding}} & \multicolumn{2}{c|}{\textbf{Relation Understanding}} & \multicolumn{2}{c|}{\textbf{Semantic Segmentation}} & \multicolumn{2}{c|}{\textbf{Image Captioning}} & \multicolumn{2}{c||}{\textbf{Image-Text Matching}} & \multicolumn{2}{c|}{\textbf{Unrelatedness}} & \multicolumn{2}{c}{\textbf{Visual Question Answering}} \\ 
        \cmidrule(lr){2-3} \cmidrule(lr){4-5} \cmidrule(lr){6-7} \cmidrule(lr){8-9} \cmidrule(lr){10-11} \cmidrule(lr){12-13} \cmidrule(lr){14-15} \cmidrule(lr){16-17}
        & Sw & $\Delta$Acc. & Sw & $\Delta$Acc. & Sw & $\Delta$Acc. & Sw & $\Delta$Acc. & Sw & $\Delta$Acc. & Sw & $\Delta$Acc. & Sw & $\Delta$Acc. & Sw & $\Delta$Acc. \\
        \midrule
        GPT-4o & 88.3 & \textcolor{blue}{+2.7} & 87.6 & \textcolor{blue}{+2.9} & 87.8 & \textcolor{blue}{+2.0} & 85.4 & \textcolor{blue}{+0.5} & 88.0 & \textcolor{blue}{+7.3} & 89.2 & \textcolor{blue}{+5.1} & 97.8 & \textcolor{blue}{10.7} & 94.3 & \textcolor{blue}{+7.8} \\
        \bottomrule
    \end{tabular}
    }
    \label{tab:MaRVLsummarized} 
\end{table*}

\section{Discussion}
\label{sec:Discussion}
The results presented in \S \ref{sec:Results} demonstrate that VLMs still lag humans on \emph{VLURes} by 6.7\%. 
Hence, VLMs can be improved, and our benchmark can be used to further develop them. 
Moreover, whereas VLMs struggle with tasks in \emph{VLURes}, humans efficiently perform such tasks, highlighting the gap between humans and VLMs. 
In addition, it is evident that the addition of rationales into VLM reasoning improves performance on our benchmark, in line with previous findings.
In Table \ref{tab:ZerooneshotAccuracyEn}, the human performance is quite low at 81.9\%, compared to the other tasks where human accuracy is near perfect. We do not have explicit reasons but this probably happened due to fatigue or human error during the evaluation. 
We observed that VLMs performed best when the input instructions or prompts were given in English. This is a previously reported behavior and probably caused by data bias at the time of model pretraining \cite{geigle-etal-2024-babel, chen2023pali}. Moreover, only a few models were robust across tasks and languages, that is, GPT-4o, GPT-4o-mini, Gemini 2.0 Flash Lite, and Gemini 1.5 Flash 8B. For the other VLMs, a low robustness is due to the limited ability to understand Swahili and Urdu. For example, despite the impressive performance of Qwen2 VL 7B on English and Japanese data, the model generated unintelligible responses when Swahili or Urdu data are fed as input to this model. 
Despite the gap between proprietary and open-source models, fine-tuning of the models, such as Qwen2VL, improved the performance on all tasks (see Table \ref{tab:FinetuneAccuracyEn}). 

\textbf{Relevance of Rationales.} During human evaluation, evaluators emphasized that rationales provided more evidence necessary for the evaluators to make an informed decision about VLM performance.

\section{Conclusion}
\label{sec:Conclusion}
Our work introduces \emph{VLURes}, a benchmark designed to analyze VLM visual reasoning capabilities across four languages systematically. 
We evaluated ten vision language models, and our findings indicate that the disparity between humans and the best performing model (GPT-4o) is less than that of the open-source models evaluated. 
We also find that eliciting rationales is vital to improve VLM performance, and this finding is consistent across all tasks, languages, and models. 
This work further introduced a significant resource for VLM evaluation, especially for Swahili and Urdu. 
Due to the disparity between human performance and models, we hope that this benchmark will be helpful in the future development of models.

\clearpage
\bibliography{main}
\bibliographystyle{plain}

\clearpage
\section{Appendix}
\label{sec:Appedndix}


\subsection{Dataset Examples}
\label{App:DatasetExamples}
In this section, we show image–text pairs and the associated CLIP scores. In Tables \ref{tab:ClipSimExapleUr}, \ref{tab:ClipSimExapleSw}, we have shown examples of Urdu and Swahili, respectively, the two low-resource languages in this study.  

\begin{table*}[h]
  \centering
  \scriptsize
  \caption{Data instances retrieved from the \textbf{\emph{VLURes}} benchmark comprising the \textit{texts} and associated \textit{images}, together with the CLIP ViT-B/32 image–text \textit{similarity scores}. The language is \textbf{Urdu}.}
  \resizebox{0.9\textwidth}{!}{
    \begin{tabular}{>{\raggedright\arraybackslash}m{0.68\textwidth}
                    >{\centering\arraybackslash}m{0.17\textwidth}
                    >{\centering\arraybackslash}m{0.14\textwidth}}
      \toprule
      \textbf{Text} & \textbf{Image} & \textbf{CLIP Sim.} \\
      \midrule
      \begin{UrduText}شاورما ایک شامی عربی گوشت کی تیاری ہے جس میں بھیڑ، مرغی، ٹرکی، گائے، اونٹ یا مخلوط گوشت کو ایک سلاخ پر پرو کر پکایا جاتا ہے۔ شاورما عرب ممالک کے علاوہ ہندوستان اور پاکستان میں بھی لوگوں کی مرغوب غذا ہے۔\end{UrduText} \textcolor{teal}{\textbf{ [Shawarma]}} &
      \includegraphics[width=\linewidth,height=2.1cm]{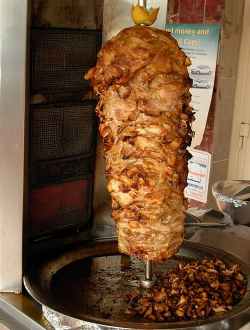} & 0.3648 \\ \midrule

      \begin{UrduText}طغرا عثمانی سلاطین کی تمام سرکاری دستاویزات اور خط کتابت پر ثبت کی جانے والی مہر یا دستخط ہے۔\end{UrduText} \textcolor{teal}{\textbf{ [Tughra]}} &
      \includegraphics[width=\linewidth]{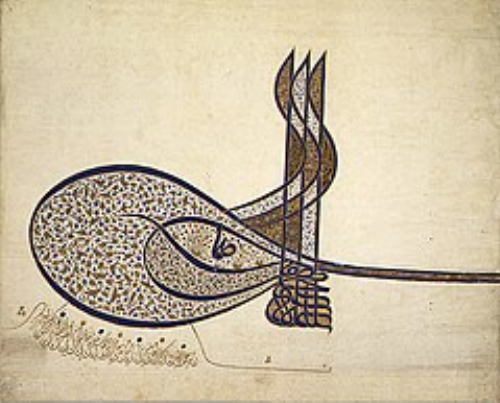} & 0.3444 \\ \midrule

      \begin{UrduText}لا ٹوماٹینا ہسپانیہ میں تقریباً 70 سالوں سے منایا جا رہا ہے، یہ تہوار یہاں کے لوگوں کے درمیان بہت مقبول ہے، نیز یہ تہوار غیر ملکی سیاحوں کی توجہ کا مرکز بھی ہے اور ہزاروں کی تعداد میں سیاح اس میں شریک ہوتے ہیں۔ اس تہوار کے سبب آغاز میں اختلاف ہے، بہت سے لوگ اسے 61 سال پرانا مانتے ہیں، تہوار کے حوالہ سے اہم کہانی یہ ہے کہ سالوں پہلے کچھ دوستوں کی ایک ٹیم نے ایک دوسرے پر ٹماٹر پھینکنا شروع کیا جو بعد میں ایک سالانہ تہوار کے طور پر منایا جانے لگا۔\end{UrduText} \textcolor{teal}{\textbf{ [La Tomatina]}} &
      \includegraphics[width=\linewidth,height=2.2cm]{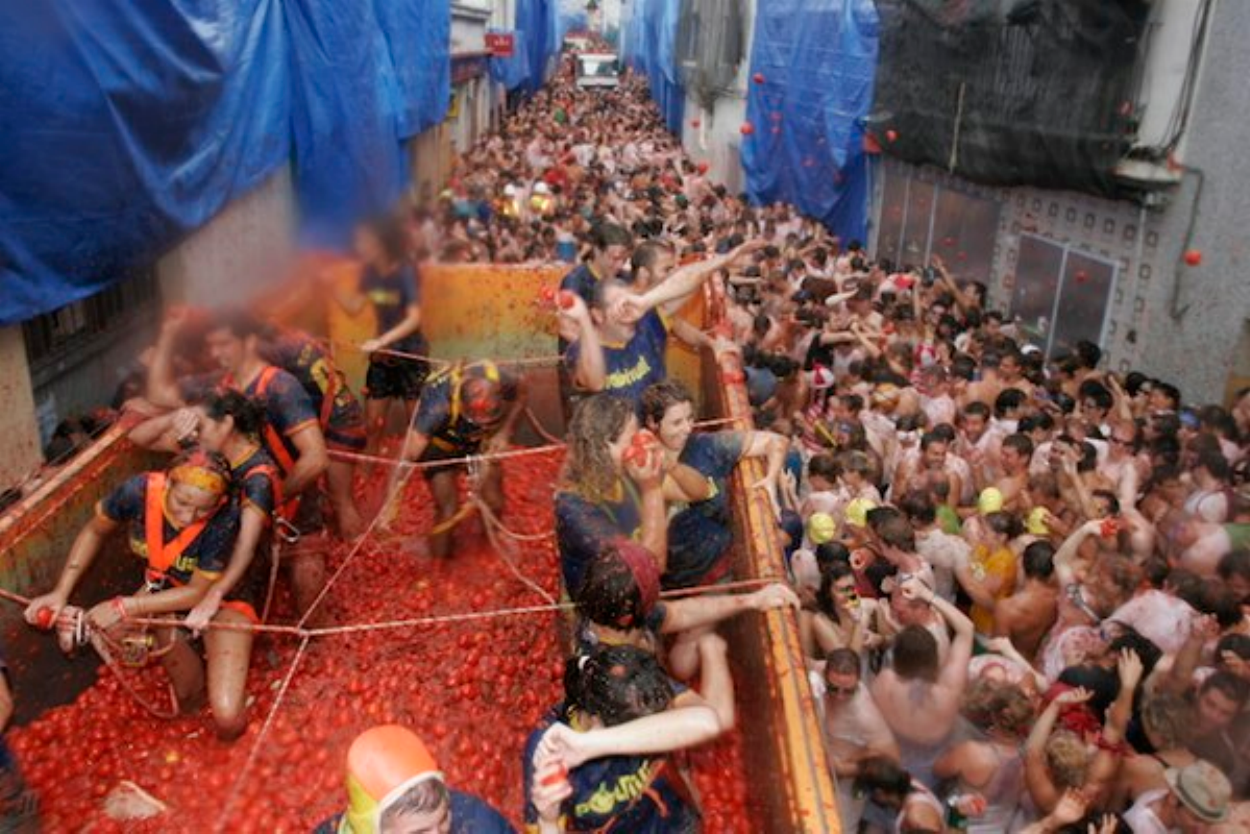} & 0.3539 \\ \midrule

      \begin{UrduText}پاکستانی روپیہ 1948ء سے پاکستان کا سرکاری زرمبادلہ ہے۔ سکے اور نوٹ مرکزی بینک یعنی بینک دولت پاکستان کے ذریعے جاری کیے جاتے ہیں اور انضباط میں رکھے جاتے ہیں۔\end{UrduText} \textcolor{teal}{\textbf{ [Pakistani Rupee]}} &
      \includegraphics[width=\linewidth]{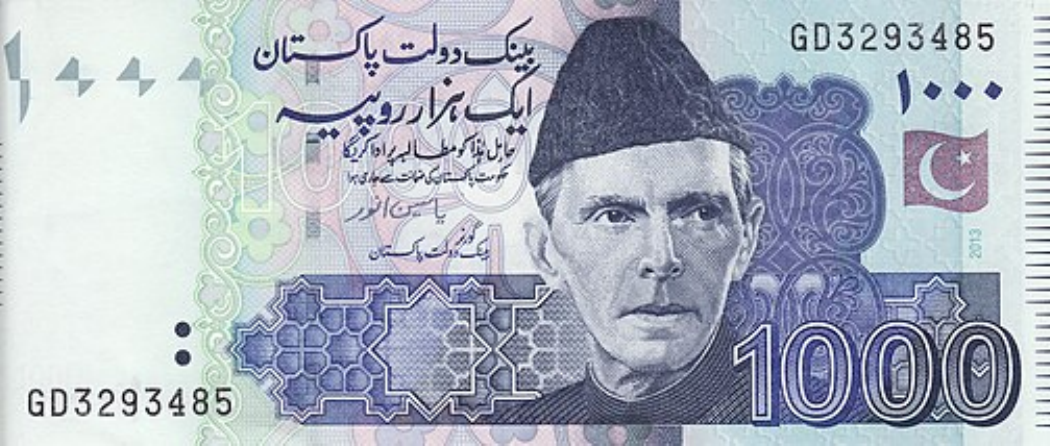} & 0.3344 \\ \midrule

      \begin{UrduText}شارق برصغیر میں پایا جانے والا ایک عام پرندہ ہے، جو اپنے بھورے جسم، سیاہ سر، آنکھوں کے گرد پیلے حلقوں اور پیلی ٹانگوں کی وجہ سے آسانی سے پہچانا جا سکتا ہے۔ شارق کی لمبائی 23 سینٹی میٹر ہوتی ہے۔ جبکہ نر کا اوسط وزن 109.8 گرام اور مادہ کا اوسط وزن 120–138 گرام ہوتا ہے۔ شارق کا اصل مسکن ایشیا ہے، لیکن یہ دیگر ممالک میں بھی پایا جاتا ہے۔ ایشیا میں پاکستان، بھارت، سری لنکا، نیپال، بھوٹان، بنگلہ دیش، ترکمانستان، افغانستان، قازقستان، میانمار، ملائیشیا اور تھائی لینڈ میں پایا جاتا ہے۔ شارق درختوں، درخت کی کھوہ میں، عمارتوں اور روشندانوں میں گھونسلا بنانا پسند کرتا ہے، اور مادہ 4 سے 6 انڈے دیتی ہے۔ اس کا انڈا ہلکے نیلے رنگ کا ہوتا ہے۔ شارق بنیادی طور پر ایک جارحانہ مزاج کا حامل پرندہ ہے، جو کسی اور پرندے کو اپنے گھونسلے کے قریب پھٹکنے بھی نہیں دیتا۔\end{UrduText} \textcolor{teal}{\textbf{ [Common myna]}} &
      \includegraphics[width=\linewidth]{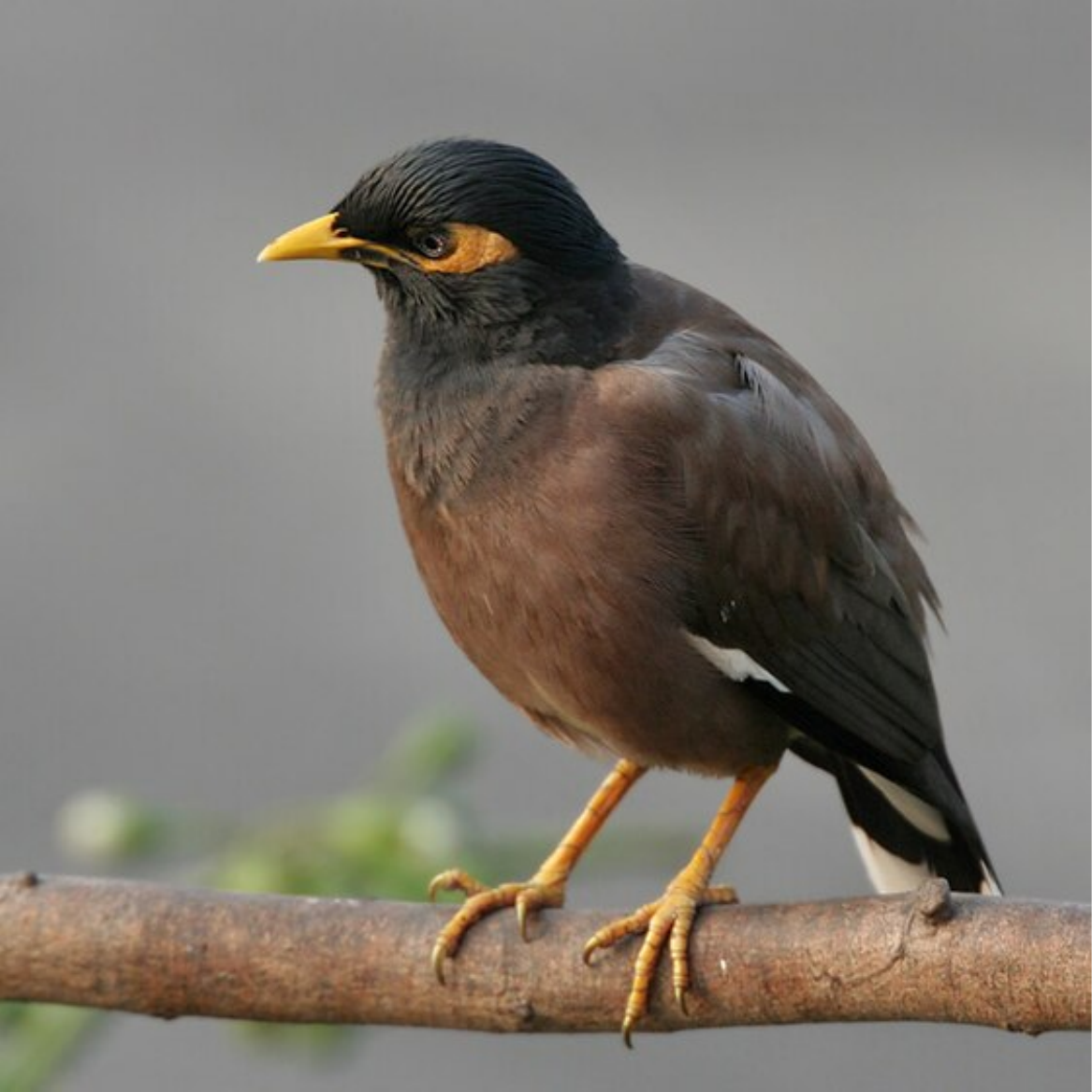} & 0.3746 \\
      \bottomrule
    \end{tabular}}
  \label{tab:ClipSimExapleUr}
\end{table*}

\clearpage
\begin{table*}[h]
    \centering
    \tiny
    \caption{A data instance retrieved from the \textbf{\emph{VLURes}} benchmark comprising the \textit{texts} and associated \textit{images}, together with the CLIP ViT-B/32 image–text \textit{similarity scores}.
    The language is \textbf{Swahili}.}
    \begin{tabular}{>{\raggedright\arraybackslash}m{0.68\textwidth} >{\centering\arraybackslash}m{0.17\textwidth} >{\centering\arraybackslash}m{0.08\textwidth}}
    \toprule
    \textbf{Text} & \textbf{Image} & \textbf{CLIP Similarity} \\
    \midrule
    \texttt{KANDA YA ZIWA KUONJA RADHA YA ROYAL BOXING TOUR. Na.Khadija Seif, Michuzi blog.
        KATIKa kuunga mkono juhudi za Rais Dk. Samia Suluhu Hassan za kutangaza na kukuza utalii wa Tanzania, kampuni ya HB Sacs Boxing inatarajia kuandaa pambano la 'Royal Tour Boxing Kanda ya Ziwa litakalofanyika Septemba 30 mwaka huu jijini Mwanza katika Ukumbi wa Rock City Mall.
        Akizungumza na Wanahabari Promota wa pambano hilo, Hassan Kumbucha amesema lengo la kuandaa mtanange huo ni kutangaza utalii, kuhamasisha mchezo huo Kanda ya Ziwa na kuleta uzalendo.
        "Kampuni yetu imekuja kitofauti sana tunahitaji kutoa fursa za utalii, pia kuleta hamasa Kwa mabondia wa kanda ya Ziwa na kuunga mkono Rais Dk. Samia, "alisema Kambucha.
        Promota huyo ameweka wazi kuwa  pambano kuu litakuwa kati ya bondia mtanzania, Fadhili Majiha 'Stopper' dhidi ya Renz Rosia raia wa Ufilipino kuwania mkanda wa Ubingwa wa ABU.
        "Kabla ya pambano la Majiha na mfilipino kutakuwa na mapambano ya vijana ambayo yatafanyika Karagwe Buringi Septemba 25 mwaka huu baada hapo wataenda kuangalia utalii Chato Septemba 26.
        Bondia, Fadhili Majiha amesema amefurahi kupata nafasi hiyo ya kucheza katika pambano kubwa ambalo linatangaza utalii wa ndani.
        Ambapo mara nyingi amekuwa akikosa nafasi ya kucheza, ila kupitia kampuni hiyo atapambana na kuonyesha uwezo wake.
        "Pongezi kwa kampuni ya HB Sacs Boxing Kwa kuandaa pambano hili na kuniamini, nimejiandaa vema na bado naendelea na mazoezi chini ya kocha wangu, Kwa me Hamisi watanzania waniombee dua mkanda ubaki nyumbani." Makamu Mwenyekiti wa Kamisheni ya Ngumi za Kulipwa Tanzania (TPBRC), Nassoro Chuma ameeleza kuwa  jambo nzuri ni kuona wadau wa michezo wanamuunga mkono Rais Dk. Samia.
        Aidha mabondia ambao watapanda ulingoni siku hiyo wanatakiwa kutoa burudani na kucheza Kwa viwango vikubwa.
        Pamoja na Majiha na Mfilipino mapambano mengine ambayo yatasindikiza pambano hilo ni Saleh Kassim dhidi ya Freddy Sayuni wakati Stumai Muki atacheza na Engine Kayange huku Shomari Milundi na Abdullah Rashid na Ahmed Pelembela ataonyesha ubabe na Issa Maneva. Wengine Luckman Ramadhani atapigana na Khalid Karama wakati John Chua dhidi ya Ramadhani Kumbele huku Said Mkola akiwa na kibarua kizito na Francis Miyeyusho.}            & \includegraphics[width=\linewidth]{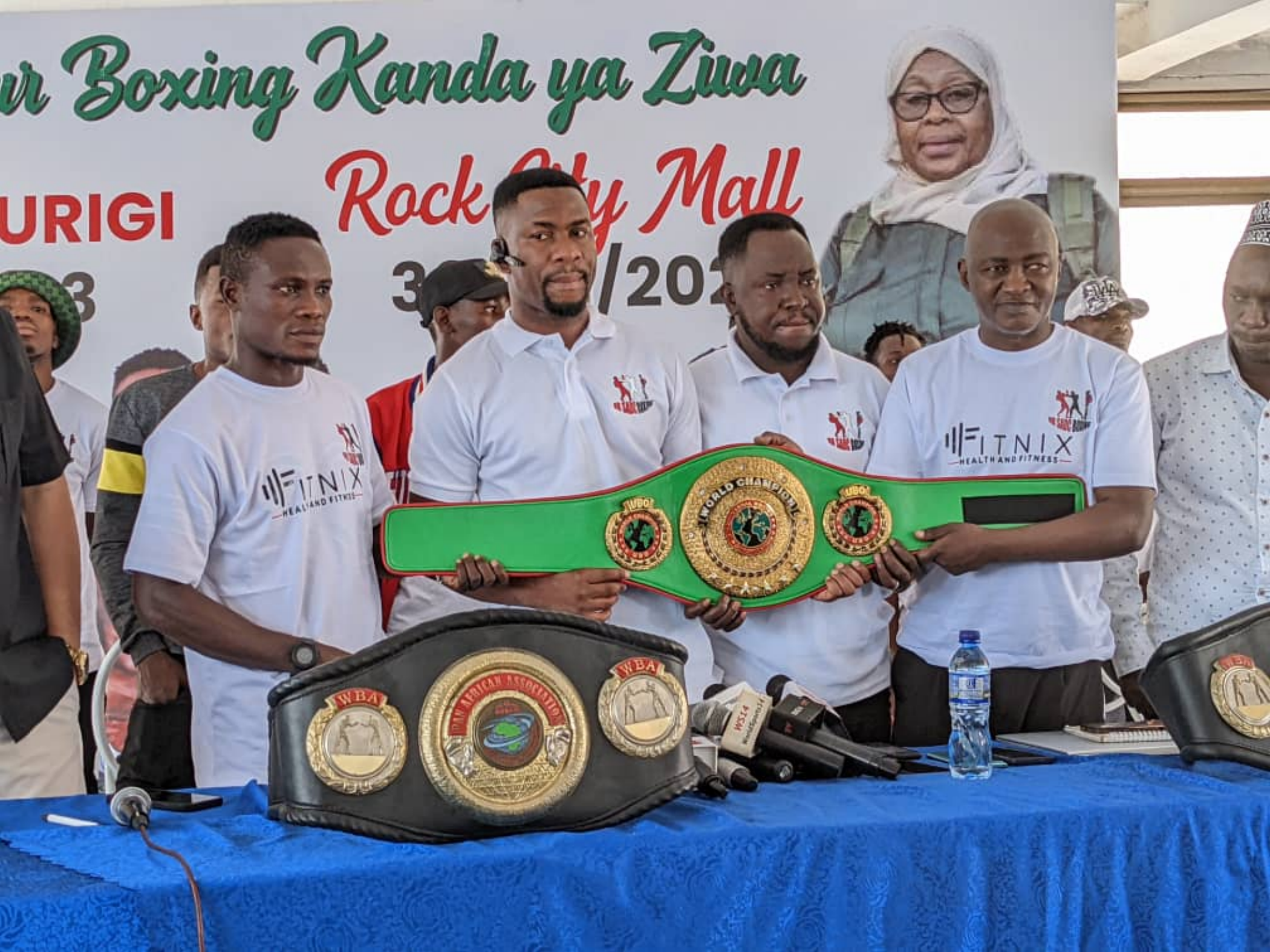}   & 0.3806 \\
        \midrule
        \texttt{MAJALIWA AZUNGUMZA NA WAZIRIi MKUU WA JAPAN SHINZO ABE. Waziri Mkuu, Kassim Majaliwa akiteta na Waziri Mkuu wa Japan, Shinzo Abe baada ya mazungumzo yao yaliyofanyika kwenye hoteli ya The New Otani iliyopo Tokyo nchini Japan Waziri Mkuu, Kassim Majaliwa na ujumbe wake wakiwa katika picha ya pamoja na Waziri Mkuu wa Japan, Shinzo Abe baada ya mazungumzo yao yaliyofanyika kwenye hoteli ya The New Otani ya Tokyo nchini Japan, Agosti 31, 2019. Kutoka kushoto ni Afisa Msimamizi wa Fedha na Mipango wa Zanzibar, Sharif Bakari, Afisa Msimamizi wa Fedha, Wizara ya Fedha Mipango, Athumani Msabila, Naibu Katibu Mkuu, Wizara ya Fedha na Mipango Zanzibar, Iddi Makame Haji, Kaimu Balozi wa Tanzania Nchini Japan, John Kambona, Naibu Katibu Mkuu Wizara ya Fedha na Mipango Zanzibar, Amina Khamis Shaaban, Kaimu Mkurugenzi, Idara ya Asia na Australia, Wizara ya Mambo ya Nje na Ushirikiano wa Afrika ya Mashariki, Size Waitara, Mganga Mkuu wa Serikali, Wizara ya Afya, Maendelo ya Jamii, Jinsia, Wazee na Watoto, Profesa Bakari Kambi na kulia ni Kamishina wa Fedha za Nje, Wizara ya Fedha na Mipango, John Lubuga.
        (Picha na Ofisi ya Waziri Mkuu). }  & \includegraphics[width=\linewidth]{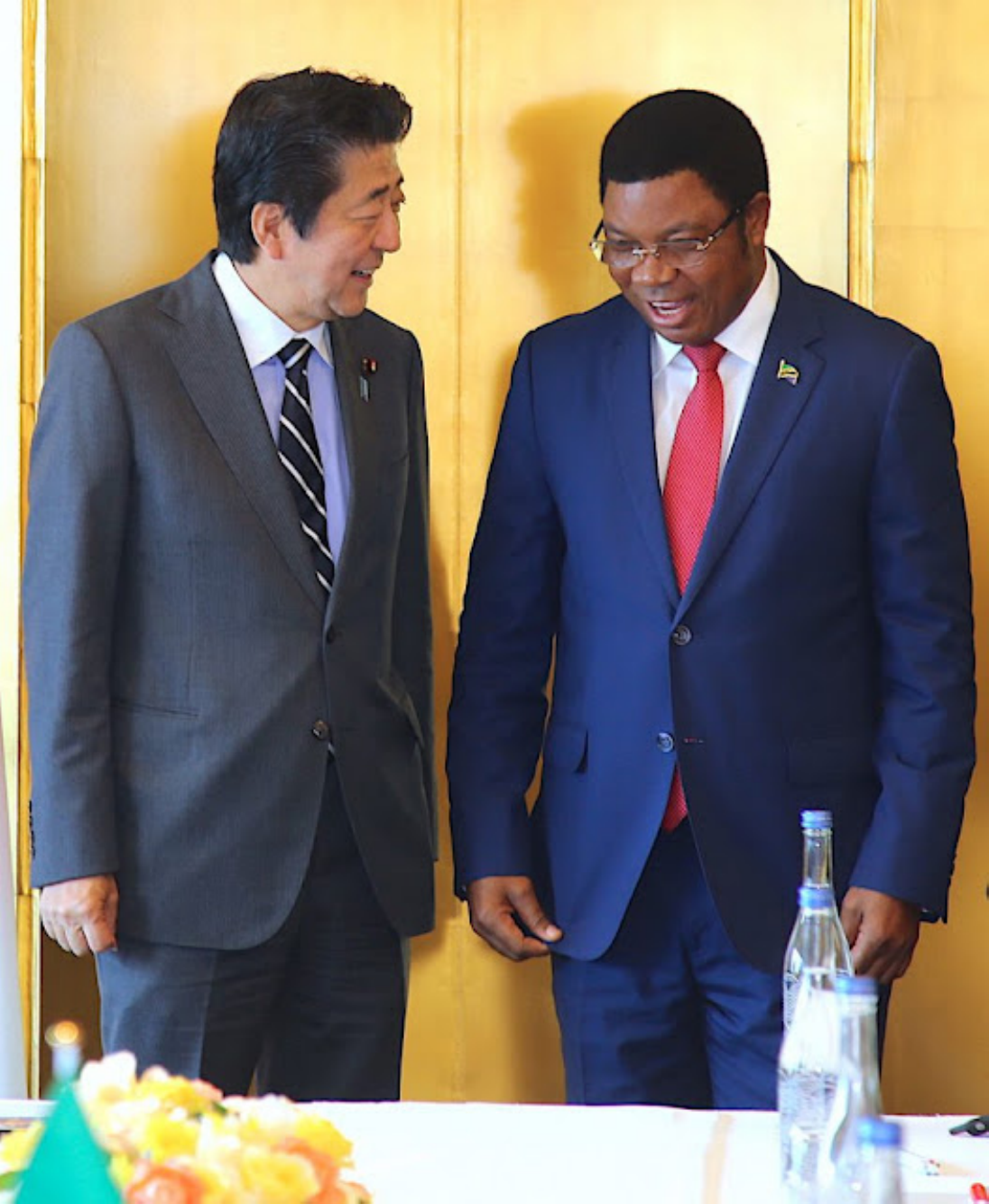}   & 0.3397 \\
        \midrule
        \texttt{CASTLE LAGER KUWAPELEKA WATANZANIA ‘WORLD CUP’. Meneja wa Bia ya Castle Lager, Pamela. Kikuli, akionyesha moja ya kipeperushi kinachoelezea mashindano maalumu ya soka yanayofahamika kama ‘Castle Lager Africa 5s’ (5 – Aside) yaliyozinduliwa leo katika viwanja vya Leaders, Kinondoni jijini Dar es salaam. Kulia ni Meneja Masoko, Udhamini na Promosheni za Wateja George Kavishe na kushoto ni Mkufunzi wa Timu ya Castle Lager Tanzania, Ivo Mapunda. Kampuni ya Bia Tanzania (TBL) kupitia kinywaji chake cha Bia ya Castle Lager, imezindua rasmi mashindano maalumu ya soka yanayofahamika kama ‘Castle Lager Africa 5s’ (5 – Aside). Mashindano haya yenye hadhi ya kimataifa yanashirikisha timu za wachezaji watano (5) kila upande, na kocha mmoja. Mechi zake zitakuwa zikichezwa kwa dakika saba (7) ili kata mshindi. Bingwa wa michuano hiyo ataiwakilisha Tanzania huko Zambia katika michuano ya kimataifa dhidi ya nchi nyingine 5 kutoka Afrika ambazo ni: Afrika Kusini, Zambia, Zimbabwe, Swaziland, Lesotho. Baada ya hapo mshindi atapata fursa ya kwenda Urusi kushuhudia michuano ya Kombe la Dunia baadaye mwaka huu, akiambatana na mashabiki wawili ambao watapatikana katika mchakato malumu pamoja na mchezaji mmoja mkongwe kutoka nchi ambayo ni bingwa. Kama Tanzania itafanikiwa kushinda, timu pamoja na mashabiki hao wawili wataongozana na mkongwe katika soka Ivo Mapunda aliyekua kipa wa Taifa Stars na Simba Sports Club. Akizungumza na waandishi wa habari jijini Dar es Salaam, Meneja wa Bia ya Castle Lager, Pamela Kikuli, alisema kuwa, kuna michakato maalumu ya kupata timu shiriki kupitia katika baa mbalimbali jijini Dar es Salaam. “Utaratibu wa kupata timu utawahusisha moja kwa moja mashabiki wa baa husika kupendekeza timu zao, kabla ya hatua nyingine za kuzipigia kura timu zinazotakiwa kushiriki. Mteja wetu atapata maelekezo maalumu kutoka kwa watu wetu ambao watakuwa wakipita katika baa hizo kwa nyakati tofauti. Sifa kubwa ya mteja wetu kushiriki katika mchakato huo ni pamoja na kununua bia ya Castle Lager,” alisema Kikuli. Akitolea maelezo za ziada jinsi mchakato huo wa Castle Lager Africa 5s utakavyokua, Meneja Masoko, Udhamini na Promosheni za Wateja George Kavishe amesema kua, michuano hiyo inashirikisha baa 160, zitakazogawanywa katika makundi 10, huku kila kundi likiwa na baa 16. Kila kundi litatakiwa kuwa na kiwanja chake katika michuano ya awali ya timu 16 na kupata timu 8. Wilaya zote za Mkoa wa Dar es Salaam zitashiriki katika michuano hii. Hapa tukimaanisha Kinondoni, Ilala, Temeke, Ubungo na Kigamboni. Bonanza za kuanza kuchagua timu wakilishi za wilaya zitaanza tarehe 17 Machi hadi 21 Aprili kisha Bonanza kuu itakayoleta timu 10 za fainali itakua tarehe 28 Aprili 2018. “Katika hatua ya pili ya bonanza tutapata timu moja kutoka katika yale makundi 10 ya awali an kufikisha idadi ya timu 10 zitakazocheza katika bonanza kubwa pale Viwanja vya Leaders jijini Dar es Salaam, kupata timu tano (5) zitakazocheza kwa mtindo wa ligi na hatimaye kupata bingwa.” Alisema Meneja Kavishe. Michuano hii inafanyika kwa miezi miwili, ambapo washiriki wanatakiwa kuwa na umri wa miaka kati ya 24 na 34. Lengo la mashindano haya ni kuwaleta pamoja marafiki na mashabiki wa soka. Wachezaji walioko katika ligi mbalimbali nchini hawatoruhusiwa kushiriki katika michuano hii, hata kama wana kigezo cha umri uliotajwa hapo juu.}    & \includegraphics[width=\linewidth, height=2.5cm]{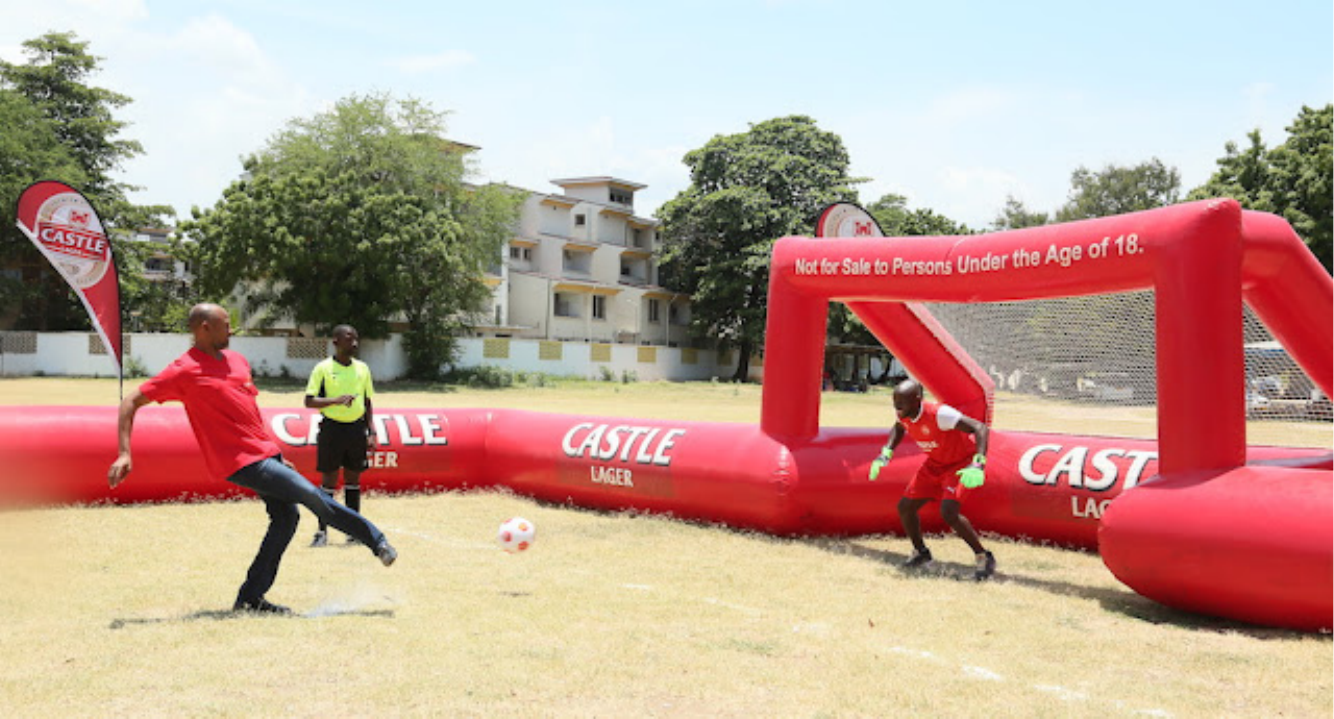}   & 0.3436 \\
    \bottomrule
    \end{tabular}
\label{tab:ClipSimExapleSw}
\end{table*}

\clearpage
\subsection{Radar plots for VLM performance across En, Jp, Sw, and Ur.}
We visualize the performance of VLMs under one-shot settings with rationales because this setting resulted in the best performance, from our experiments. Figure \ref{fig:radar_vlm_en_jp_final} is a radar plot containing results for En and Jp, while Figure \ref{fig:radar_vlm_sw_ur} shows the Sw and Ur results. Across En, JP, Sw, Ur, the GPT-4o model demonstrated superior performance compared to other models. 
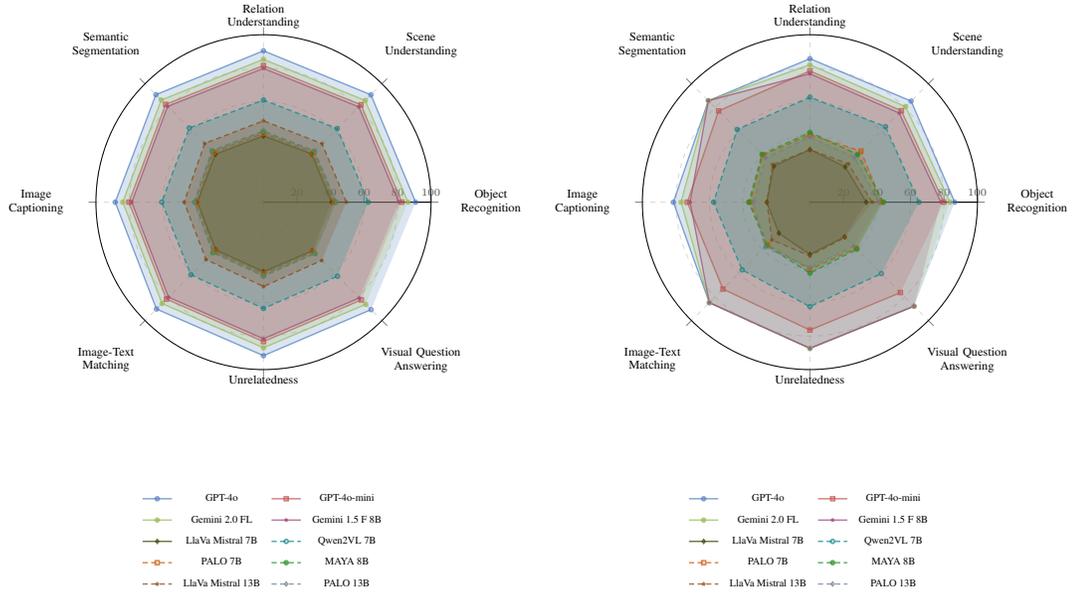
\begin{figure*}[h]
    \centering
    \begin{minipage}[t]{0.48\textwidth}
        \centering
        \begin{tikzpicture}[scale=0.65]
            \begin{polaraxis}[
                title style={font=\small, yshift=-2ex},
                ymin=0, ymax=100,
                ytick={20,40,60,80,100},
                yticklabel style={font=\tiny, color=black!70},
                yticklabel pos=right,
                xtick={0,45,90,135,180,225,270,315},
                xticklabels={
                    Object\\Recognition,
                    Scene\\Understanding,
                    Relation\\Understanding,
                    Semantic\\Segmentation,
                    Image\\Captioning,
                    Image-Text\\Matching,
                    Unrelatedness,
                    Visual Question\\Answering
                },
                xticklabel style={font=\scriptsize, text width=2.2cm, align=center},
                ymajorgrids=true,
                xmajorgrids=true,
                grid style={dashed, gray!30},
                every axis plot/.append style={
                    thick,
                    mark=*,
                    mark size=1.2pt,
                    fill opacity=0.2,
                    draw opacity=0.75
                },
                legend style={
                    at={(0.5, -0.35)},
                    anchor=north,
                    font=\tiny,
                    draw=none,
                    column sep=0.15cm,
                    row sep=0.1cm
                },
                legend columns=2
            ]

            \addplot+[color=bblue, fill=bblue] coordinates {
                (0, 90.8) (45, 90.7) (90, 90.4) (135, 90.8) (180, 88.5) (225, 90.3) (270, 91.7) (315, 90.8)
            }; \addlegendentry{GPT-4o}

            \addplot+[color=rred, fill=rred] coordinates {
                (0, 82.5) (45, 82.3) (90, 81.6) (135, 82.6) (180, 80.3) (225, 81.9) (270, 83.3) (315, 82.6)
            }; \addlegendentry{GPT-4o-mini}

            \addplot+[color=ggreen, fill=ggreen] coordinates {
                (0, 86.2) (45, 85.9) (90, 85.3) (135, 86.3) (180, 84.0) (225, 85.6) (270, 87.0) (315, 86.3)
            }; \addlegendentry{Gemini 2.0 FL}

            \addplot+[color=ppurple, fill=ppurple] coordinates {
                (0, 80.8) (45, 80.5) (90, 79.9) (135, 80.9) (180, 78.6) (225, 80.2) (270, 81.6) (315, 80.9)
            }; \addlegendentry{Gemini 1.5 F 8B}

            \addplot+[color=darkOlive, fill=darkOlive] coordinates {
                (0, 40.6) (45, 40.7) (90, 39.5) (135, 40.5) (180, 39.4) (225, 40.2) (270, 41.5) (315, 41.1)
            }; \addlegendentry{LlaVa Mistral 7B}

            \addplot+[color=darkCyan, fill=darkCyan] coordinates {
                (0, 62.5) (45, 62.2) (90, 61.0) (135, 62.6) (180, 60.7) (225, 61.3) (270, 63.5) (315, 62.5)
            }; \addlegendentry{Qwen2VL 7B}

            \addplot+[color=darkOrange, fill=darkOrange] coordinates {
                (0, 42.2) (45, 41.7) (90, 41.1) (135, 42.1) (180, 39.8) (225, 41.4) (270, 42.8) (315, 42.1)
            }; \addlegendentry{PALO 7B}

            \addplot+[color=darkLime, fill=darkLime] coordinates {
                (0, 43.3) (45, 43.0) (90, 42.4) (135, 43.4) (180, 41.1) (225, 42.7) (270, 44.1) (315, 43.4)
            }; \addlegendentry{MAYA 8B}

            \addplot+[color=modelBrown, fill=modelBrown] coordinates {
                (0, 49.5) (45, 49.4) (90, 48.6) (135, 49.6) (180, 47.4) (225, 48.5) (270, 50.4) (315, 49.2)
            }; \addlegendentry{LlaVa Mistral 13B}

            \addplot+[color=modelSlateGray, fill=modelSlateGray] coordinates {
                (0, 43.5) (45, 43.3) (90, 42.5) (135, 43.6) (180, 41.3) (225, 42.3) (270, 44.0) (315, 43.0)
            }; \addlegendentry{PALO 13B}

            \end{polaraxis}
        \end{tikzpicture}
    \end{minipage}
    \hfill
    \begin{minipage}[t]{0.48\textwidth}
        \centering
        \begin{tikzpicture}[scale=0.65]
            \begin{polaraxis}[
                title style={font=\small, yshift=-2ex},
                ymin=0, ymax=100,
                ytick={20,40,60,80,100},
                yticklabel style={font=\tiny, color=black!70},
                yticklabel pos=right,
                xtick={0,45,90,135,180,225,270,315},
                xticklabels={
                    Object\\Recognition,
                    Scene\\Understanding,
                    Relation\\Understanding,
                    Semantic\\Segmentation,
                    Image\\Captioning,
                    Image-Text\\Matching,
                    Unrelatedness,
                    Visual Question\\Answering
                },
                xticklabel style={font=\scriptsize, text width=2.2cm, align=center},
                ymajorgrids=true,
                xmajorgrids=true,
                grid style={dashed, gray!30},
                every axis plot/.append style={
                    thick,
                    mark=*,
                    mark size=1.2pt,
                    fill opacity=0.2,
                    draw opacity=0.75
                },
                legend style={
                    at={(0.5, -0.35)},
                    anchor=north,
                    font=\tiny,
                    draw=none,
                    column sep=0.15cm,
                    row sep=0.1cm
                },
                legend columns=2
            ]

            \addplot+[color=bblue, fill=bblue] coordinates {
                (0, 86.4) (45, 85.4) (90, 85.7) (135, 85.7) (180, 81.5) (225, 84.9) (270, 87.3) (315, 87.9)
            }; \addlegendentry{GPT-4o}

            \addplot+[color=rred, fill=rred] coordinates {
                (0, 79.9) (45, 77.2) (90, 78.4) (135, 77.1) (180, 73.2) (225, 73.4) (270, 76.4) (315, 76.4)
            }; \addlegendentry{GPT-4o-mini}

            \addplot+[color=ggreen, fill=ggreen] coordinates {
                (0, 83.6) (45, 80.7) (90, 82.1) (135, 85.7) (180, 76.9) (225, 84.9) (270, 87.3) (315, 87.9)
            }; \addlegendentry{Gemini 2.0 FL}

            \addplot+[color=ppurple, fill=ppurple] coordinates {
                (0, 78.2) (45, 75.3) (90, 76.7) (135, 85.7) (180, 71.5) (225, 84.9) (270, 87.3) (315, 87.9)
            }; \addlegendentry{Gemini 1.5 F 8B}

            \addplot+[color=darkOlive, fill=darkOlive] coordinates {
                (0, 33.6) (45, 29.8) (90, 31.4) (135, 30.4) (180, 25.7) (225, 26.2) (270, 31.2) (315, 29.3)
            }; \addlegendentry{LlaVa Mistral 7B}

            \addplot+[color=darkCyan, fill=darkCyan] coordinates {
                (0, 65.1) (45, 63.8) (90, 62.7) (135, 61.4) (180, 57.5) (225, 57.2) (270, 62.4) (315, 60.3)
            }; \addlegendentry{Qwen2VL 7B}

            \addplot+[color=darkOrange, fill=darkOrange] coordinates {
                (0, 42.8) (45, 43.1) (90, 40.3) (135, 39.6) (180, 35.7) (225, 35.9) (270, 40.2) (315, 38.9)
            }; \addlegendentry{PALO 7B}

            \addplot+[color=darkLime, fill=darkLime] coordinates {
                (0, 43.9) (45, 40.2) (90, 41.7) (135, 40.4) (180, 36.5) (225, 36.6) (270, 42.4) (315, 39.7)
            }; \addlegendentry{MAYA 8B}

            \addplot+[color=modelBrown, fill=modelBrown] coordinates {
                (0, 37.2) (45, 32.4) (90, 31.4) (135, 31.6) (180, 25.7) (225, 32.0) (270, 32.1) (315, 29.8)
            }; \addlegendentry{LlaVa Mistral 13B}

            \addplot+[color=modelSlateGray, fill=modelSlateGray] coordinates {
                (0, 41.0) (45, 37.7) (90, 38.8) (135, 37.4) (180, 33.6) (225, 38.1) (270, 39.0) (315, 36.9)
            }; \addlegendentry{PALO 13B}

            \end{polaraxis}
        \end{tikzpicture}
    \end{minipage}
    \caption{Comparison of VLM performance on eight VL tasks under the \textit{One-shot, With Rationales} setting. Left: \textbf{English} Input/Output. Right: \textbf{Japanese} Input/Output.}
    \label{fig:radar_vlm_en_jp_final}
\end{figure*}
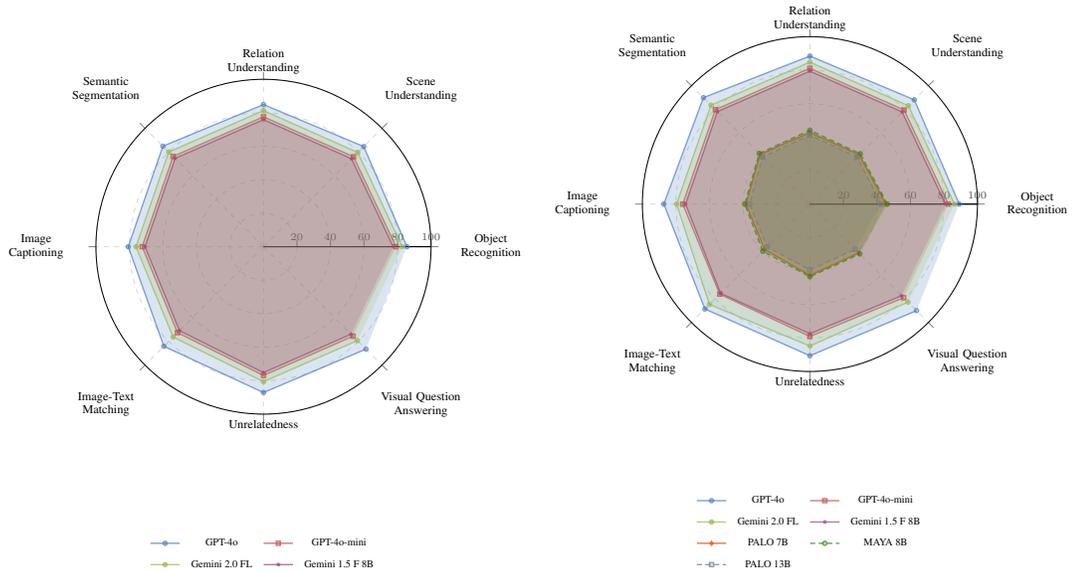
\begin{figure*}[h]
    \centering
    \begin{minipage}[t]{0.48\textwidth}
        \centering
        \begin{tikzpicture}[scale=0.65]
            \begin{polaraxis}[
                title style={font=\small, yshift=-2ex},
                ymin=0, ymax=100,
                ytick={20,40,60,80,100},
                yticklabel style={font=\tiny, color=black!70},
                yticklabel pos=right,
                xtick={0,45,90,135,180,225,270,315},
                xticklabels={
                    Object\\Recognition,
                    Scene\\Understanding,
                    Relation\\Understanding,
                    Semantic\\Segmentation,
                    Image\\Captioning,
                    Image-Text\\Matching,
                    Unrelatedness,
                    Visual Question\\Answering
                },
                xticklabel style={font=\scriptsize, text width=2.2cm, align=center},
                ymajorgrids=true,
                xmajorgrids=true,
                grid style={dashed, gray!30},
                every axis plot/.append style={
                    thick,
                    mark=*,
                    mark size=1.2pt,
                    fill opacity=0.2,
                    draw opacity=0.75
                },
                legend style={
                    at={(0.5, -0.35)},
                    anchor=north,
                    font=\tiny,
                    draw=none,
                    column sep=0.15cm,
                    row sep=0.1cm
                },
                legend columns=2
            ]

            \addplot+[color=bblue, fill=bblue] coordinates {(0,85.6) (45,84.6) (90,84.9) (135,84.9) (180,80.7) (225,84.1) (270,87.1) (315,86.5)}; \addlegendentry{GPT-4o}
            \addplot+[color=rred, fill=rred] coordinates {(0,79.1) (45,76.0) (90,77.6) (135,76.3) (180,72.4) (225,72.6) (270,76.9) (315,75.6)}; \addlegendentry{GPT-4o-mini}
            \addplot+[color=ggreen, fill=ggreen] coordinates {(0,82.8) (45,79.6) (90,81.3) (135,80.0) (180,76.1) (225,76.3) (270,80.6) (315,79.3)}; \addlegendentry{Gemini 2.0 FL}
            \addplot+[color=ppurple, fill=ppurple] coordinates {(0,77.4) (45,74.3) (90,75.9) (135,74.6) (180,70.7) (225,70.9) (270,75.2) (315,73.9)}; \addlegendentry{Gemini 1.5 F 8B}

            \end{polaraxis}
        \end{tikzpicture}
    \end{minipage}
    \hfill
    \begin{minipage}[t]{0.48\textwidth}
        \centering
        \begin{tikzpicture}[scale=0.65]
            \begin{polaraxis}[
                title style={font=\small, yshift=-2ex},
                ymin=0, ymax=100,
                ytick={20,40,60,80,100},
                yticklabel style={font=\tiny, color=black!70},
                yticklabel pos=right,
                xtick={0,45,90,135,180,225,270,315},
                xticklabels={
                    Object\\Recognition,
                    Scene\\Understanding,
                    Relation\\Understanding,
                    Semantic\\Segmentation,
                    Image\\Captioning,
                    Image-Text\\Matching,
                    Unrelatedness,
                    Visual Question\\Answering
                },
                xticklabel style={font=\scriptsize, text width=2.2cm, align=center},
                ymajorgrids=true,
                xmajorgrids=true,
                grid style={dashed, gray!30},
                every axis plot/.append style={
                    thick,
                    mark=*,
                    mark size=1.2pt,
                    fill opacity=0.2,
                    draw opacity=0.75
                },
                legend style={
                    at={(0.5, -0.35)},
                    anchor=north,
                    font=\tiny,
                    draw=none,
                    column sep=0.15cm,
                    row sep=0.1cm
                },
                legend columns=2
            ]

            \addplot+[color=bblue, fill=bblue] coordinates {(0,89.1) (45,88.1) (90,88.4) (135,89.9) (180,87.2) (225,88.6) (270,90.6) (315,90.0)}; \addlegendentry{GPT-4o}
            \addplot+[color=rred, fill=rred] coordinates {(0,82.6) (45,79.5) (90,81.1) (135,79.8) (180,75.9) (225,76.1) (270,79.1) (315,79.1)}; \addlegendentry{GPT-4o-mini}
            \addplot+[color=ggreen, fill=ggreen] coordinates {(0,86.3) (45,83.1) (90,84.8) (135,83.5) (180,79.6) (225,84.8) (270,84.8) (315,82.8)}; \addlegendentry{Gemini 2.0 FL}
            \addplot+[color=ppurple, fill=ppurple] coordinates {(0,80.9) (45,77.8) (90,79.4) (135,78.1) (180,74.2) (225,75.4) (270,77.4) (315,77.4)}; \addlegendentry{Gemini 1.5 F 8B}
            \addplot+[color=darkOrange, fill=darkOrange] coordinates {(0,45.3) (45,41.6) (90,42.7) (135,42.0) (180,38.1) (225,38.3) (270,42.6) (315,41.3)}; \addlegendentry{PALO 7B}
            \addplot+[color=darkLime, fill=darkLime] coordinates {(0,46.2) (45,42.5) (90,44.1) (135,42.8) (180,38.9) (225,39.8) (270,43.4) (315,42.1)}; \addlegendentry{MAYA 8B}
            \addplot+[color=modelSlateGray, fill=modelSlateGray] coordinates {(0,42.6) (45,39.7) (90,41.2) (135,39.8) (180,36.0) (225,36.1) (270,39.3) (315,38.2)}; \addlegendentry{PALO 13B}

            \end{polaraxis}
        \end{tikzpicture}
    \end{minipage}

    \caption{Comparison of VLM performance on eight VL tasks under the \textit{One-shot, With Rationales} setting. Left: \textbf{Swahili} Input/Output. Right: \textbf{Urdu} Input/Output.}
    \label{fig:radar_vlm_sw_ur}
\end{figure*}

\clearpage
To enhance the understanding of the open models of the tasks in \emph{VLURes} and improve performance, we fine-tuned the open models, (that is, LlaVa Mistral 7B, PALO 7B, Qwen2VL 7B, MAYA 8B, LlaVa Mistral 13B, and PALO 13B) using the data in \emph{VLURes}. Again, the \textit{one-shot with rationales} setting resulted in the best performance for all models.
Figures \ref{fig:radar_vlm_en_jp_finetuned}, \ref{fig:radar_vlm_ur_finetuned} show the performance of six fine-tuned open models on En, Jp, and Ur data, respectively. 
Whereas all six models reasonably understand En and Jp data, only five models strongly support Ur data. 
In Ur results, we excluded Qwen2VL 7B  because the Ur responses generated by Qwen2VL 7B were unintelligible or nonsensical in our preliminary studies. 
Lastly, none of the six models generated intelligible Sw responses; thus, we do not report Sw results of fine-tuning LlaVa Mistral 7B, PALO 7B, Qwen2VL 7B, MAYA 8B, LlaVa Mistral 13B, and PALO 13B. 
\begin{figure*}[h] %
    \centering
    \begin{minipage}[t]{0.48\textwidth}
        \centering
        \begin{tikzpicture}[scale=0.65] %
            \begin{polaraxis}[
                title style={font=\small, yshift=-1.5ex}, %
                ymin=0, ymax=100,
                ytick={20,40,60,80,100},
                yticklabel style={font=\tiny, color=black!70},
                yticklabel pos=right,
                xtick={0,45,90,135,180,225,270,315},
                xticklabels={
                    Object\\Recognition,
                    Scene\\Understanding,
                    Relation\\Understanding,
                    Semantic\\Segmentation,
                    Image\\Captioning,
                    Image-Text\\Matching,
                    Unrelatedness,
                    Visual Question\\Answering
                },
                xticklabel style={font=\scriptsize, text width=2.2cm, align=center},
                ymajorgrids=true,
                xmajorgrids=true,
                grid style={dashed, gray!30},
                every axis plot/.append style={
                    thick,
                    mark=*,
                    mark size=1.2pt,
                    fill opacity=0.2,
                    draw opacity=0.75
                },
                legend style={
                    at={(0.5, -0.38)}, %
                    anchor=north,
                    font=\tiny,
                    draw=none,
                    column sep=0.15cm,
                    row sep=0.1cm
                },
                legend columns=2 %
            ]

            \addplot+[color=darkOlive, fill=darkOlive] coordinates {
                (0, 50.9) (45, 51.0) (90, 49.8) (135, 50.8) (180, 49.7) (225, 50.5) (270, 51.8) (315, 51.4)
            }; \addlegendentry{LlaVa Mistral 7B}

            \addplot+[color=darkCyan, fill=darkCyan] coordinates {
                (0, 71.3) (45, 70.9) (90, 69.8) (135, 71.4) (180, 65.2) (225, 64.9) (270, 70.1) (315, 68.0)
            }; \addlegendentry{Qwen2VL 7B}

            \addplot+[color=darkOrange, fill=darkOrange] coordinates {
                (0, 51.0) (45, 50.5) (90, 49.9) (135, 50.9) (180, 48.6) (225, 50.2) (270, 58.1) (315, 50.9)
            }; \addlegendentry{PALO 7B}

            \addplot+[color=darkLime, fill=darkLime] coordinates {
                (0, 52.1) (45, 51.8) (90, 51.2) (135, 52.2) (180, 49.9) (225, 51.5) (270, 57.7) (315, 52.2)
            }; \addlegendentry{MAYA 8B}

            \addplot+[color=modelBrown, fill=modelBrown] coordinates {
                (0, 57.4) (45, 57.3) (90, 56.5) (135, 57.5) (180, 55.3) (225, 56.4) (270, 58.3) (315, 57.1)
            }; \addlegendentry{LlaVa Mistral 13B}

            \addplot+[color=modelSlateGray, fill=modelSlateGray] coordinates {
                (0, 52.3) (45, 52.1) (90, 51.3) (135, 52.4) (180, 50.1) (225, 51.1) (270, 64.8) (315, 51.8)
            }; \addlegendentry{PALO 13B}

            \end{polaraxis}
        \end{tikzpicture}
    \end{minipage}
    \hfill %
    \begin{minipage}[t]{0.48\textwidth}
        \centering
        \begin{tikzpicture}[scale=0.65] %
            \begin{polaraxis}[
                title style={font=\small, yshift=-1.5ex}, %
                ymin=0, ymax=100,
                ytick={20,40,60,80,100},
                yticklabel style={font=\tiny, color=black!70},
                yticklabel pos=right,
                xtick={0,45,90,135,180,225,270,315},
                xticklabels={
                    Object\\Recognition,
                    Scene\\Understanding,
                    Relation\\Understanding,
                    Semantic\\Segmentation,
                    Image\\Captioning,
                    Image-Text\\Matching,
                    Unrelatedness,
                    Visual Question\\Answering
                },
                xticklabel style={font=\scriptsize, text width=2.2cm, align=center},
                ymajorgrids=true,
                xmajorgrids=true,
                grid style={dashed, gray!30},
                every axis plot/.append style={
                    thick,
                    mark=*,
                    mark size=1.2pt,
                    fill opacity=0.2,
                    draw opacity=0.75
                },
                legend style={
                    at={(0.5, -0.38)}, %
                    anchor=north,
                    font=\tiny,
                    draw=none,
                    column sep=0.15cm,
                    row sep=0.1cm
                },
                legend columns=2 %
            ]

            \addplot+[color=darkOlive, fill=darkOlive] coordinates {
                (0, 52.1) (45, 48.8) (90, 50.0) (135, 49.1) (180, 43.8) (225, 44.9) (270, 50.1) (315, 48.0)
            }; \addlegendentry{LlaVa Mistral 7B}

            \addplot+[color=darkCyan, fill=darkCyan] coordinates {
                (0, 77.2) (45, 74.0) (90, 75.2) (135, 74.0) (180, 70.0) (225, 69.7) (270, 75.0) (315, 72.8)
            }; \addlegendentry{Qwen2VL 7B}

            \addplot+[color=darkOrange, fill=darkOrange] coordinates {
                (0, 55.4) (45, 51.7) (90, 52.8) (135, 52.1) (180, 48.2) (225, 48.4) (270, 52.7) (315, 51.4)
            }; \addlegendentry{PALO 7B}

            \addplot+[color=darkLime, fill=darkLime] coordinates {
                (0, 56.3) (45, 52.6) (90, 54.2) (135, 52.9) (180, 49.0) (225, 49.1) (270, 53.5) (315, 52.2)
            }; \addlegendentry{MAYA 8B}

            \addplot+[color=modelBrown, fill=modelBrown] coordinates {
                (0, 58.1) (45, 55.1) (90, 56.5) (135, 55.5) (180, 51.3) (225, 51.2) (270, 56.0) (315, 53.7)
            }; \addlegendentry{LlaVa Mistral 13B}

            \addplot+[color=modelSlateGray, fill=modelSlateGray] coordinates {
                (0, 52.7) (45, 49.8) (90, 51.3) (135, 50.0) (180, 46.1) (225, 45.6) (270, 51.5) (315, 49.4)
            }; \addlegendentry{PALO 13B}

            \end{polaraxis}
        \end{tikzpicture}
    \end{minipage}
    \caption{Comparison of VLM performance on eight VL tasks \textbf{after finetuning}, under the \textit{One-shot, With Rationales} setting. Left: \textbf{English} Input/Output. Right: \textbf{Japanese} Input/Output.}
    \label{fig:radar_vlm_en_jp_finetuned}
\end{figure*}
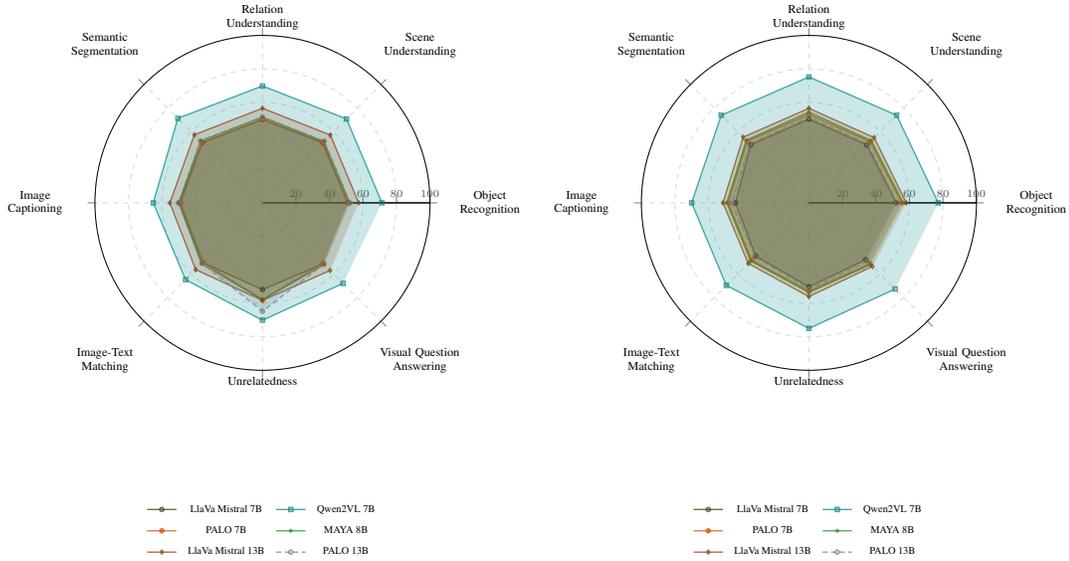
\begin{figure}[h] 
    \centering
    \begin{tikzpicture}[scale=0.65] %
        \begin{polaraxis}[
            title style={font=\small, yshift=-1.5ex},
            ymin=0, ymax=100,
            ytick={20,40,60,80,100},
            yticklabel style={font=\tiny, color=black!70},
            yticklabel pos=right,
            xtick={0,45,90,135,180,225,270,315},
            xticklabels={
                Object\\Recognition,
                Scene\\Understanding,
                Relation\\Understanding,
                Semantic\\Segmentation,
                Image\\Captioning,
                Image-Text\\Matching,
                Unrelatedness,
                Visual Question\\Answering
            },
            xticklabel style={font=\scriptsize, text width=2.2cm, align=center},
            ymajorgrids=true,
            xmajorgrids=true,
            grid style={dashed, gray!30},
            every axis plot/.append style={
                thick,
                mark=*,
                mark size=1.2pt,
                fill opacity=0.2,
                draw opacity=0.75
            },
            legend style={
                at={(0.5, -0.38)}, 
                anchor=north,
                font=\tiny,
                draw=none,
                column sep=0.15cm,
                row sep=0.1cm
            },
            legend columns=2 %
        ]

        \addplot+[color=darkOrange, fill=darkOrange] coordinates {
            (0, 55.4) (45, 51.7) (90, 52.8) (135, 52.1) (180, 48.2) (225, 48.4) (270, 52.7) (315, 51.4)
        }; \addlegendentry{PALO 7B}

        \addplot+[color=darkLime, fill=darkLime] coordinates {
            (0, 56.3) (45, 52.6) (90, 54.2) (135, 52.9) (180, 49.0) (225, 49.1) (270, 53.5) (315, 52.2)
        }; \addlegendentry{MAYA 8B}

        \addplot+[color=modelSlateGray, fill=modelSlateGray] coordinates {
            (0, 52.7) (45, 49.8) (90, 51.3) (135, 50.0) (180, 46.1) (225, 45.6) (270, 51.5) (315, 49.4)
        }; \addlegendentry{PALO 13B}

        \end{polaraxis}
    \end{tikzpicture}
    \caption{Performance of VLMs on eight VL tasks \textbf{after finetuning}, under the \textit{One-shot, With Rationales} setting for \textbf{Urdu} Input/Output.} %
    \label{fig:radar_vlm_ur_finetuned}
\end{figure}
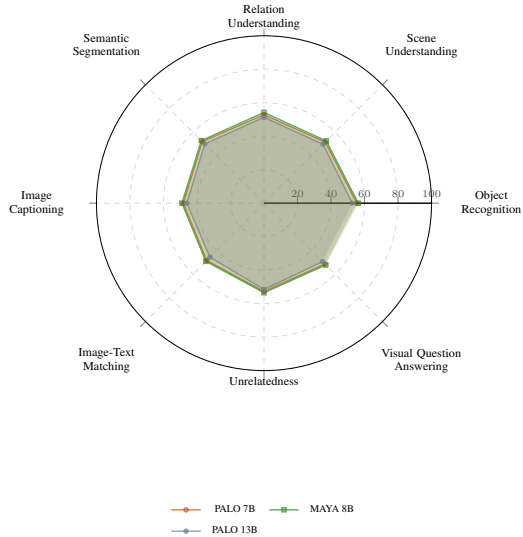

\clearpage
\subsection{More Related Works}
\label{App:MoreRelatedWorks}
In addition to the previous works presented in \S \ref{sec:RelatedWork}, there are task-specific datasets from which we draw our inspiration.  

The datasets cover VL tasks, such as image captioning (IC), reference expression generation (REG), reference expression comprehension (REC), visual question answering (VQA), \textit{inter alia}. We have shown a summary in Table \ref{Table:VLMBenchmarksCompared}.  

\textbf{Image Captioning.} Popular datasets are: MSCOCO~\cite{lin2015microsoft}, Flickr30k~\cite{plummer2016flickr30k}, Conceptual 12M~\cite{changpinyo2021cc12m}, Conceptual Captions~\cite{sharma2018conceptual}, all of which are in English. In response to the growing need for image captions in other languages, the Multi30K~\cite{elliott-etal-2016-multi30k} dataset extended Flickr30k to German, English, French, and Czech. Then, \cite{ilahi2021efficient} extended a portion of Flickr8k~\cite{hodosh2013framing} to Urdu. MSCOCO was extended to Japanese in the STAIR Captions~\cite{Yoshikawa2017} and  YJ Captions~\cite{P16-1168} datasets. AI Challenger~\cite{Wu_2019} is available in Chinese. Crossmodal-3600~\cite{thapliyal-etal-2022-crossmodal} is available in 36 languages, including English, Japanese, and Swahili. However, significant datasets are in high-resource languages, leaving room for further exploration in low-resource languages. The Swahili data in Crossmodal-3600 contains only 100 images and two captions per image. The dataset by \cite{ilahi2021efficient} contains 700 images, limited only to images containing ``people''. \textit{In both datasets, Crossmodal-3600 and \cite{ilahi2021efficient}, which include Swahili and Urdu, respectively, the captions do not give thorough details about objects in the image}. Another problem is the lack of a balanced representation of image categories.

\textbf{Reference Expression Generation.} Most notable REG\footnote{The REG task aims to provide a natural language description for a \textit{specific object} within the image, which is contrary to the IC task, which aims to condition the generation of natural language on all the visual features contained in the image. On the other hand, REC aims to \textit{identify} the referred object in the image when the input is an expression in natural language.} datasets are RefCOCO \cite{kazemzadeh-etal-2014-referitgame}, RefCOCO+ \cite{kazemzadeh-etal-2014-referitgame}, and RefCOCOg \cite{mao2016generation}, all of which suffer from the limitation that the description for the target object in the image is too short. 

\textbf{Visual Question Answering.} In the same vein as REG, datasets developed for VQA contain limited descriptive information \cite{goyal2017making, okvqa, mishraICDAR19, hudson2019gqa, Krishna2016VisualGC} about the objects contained in the image. Additionally, the datasets are predominantly \textit{English}, leaving room for developing datasets in underserved languages, such as Swahili and Urdu.

\clearpage
\subsection{Vision and Language Task Definitions}
\label{subsec:VisionAndLanguageTasks}
\begin{table}[h] %
    \centering
    \caption{Definitions of the eight VL tasks used in \textbf{\emph{VLU-RES}.}} 
    \resizebox{0.85\textwidth}{!}{
    \begin{tabular}{@{}p{0.28\linewidth}p{0.68\linewidth}@{}} %
        \toprule
        \textbf{Task} & \textbf{Task Definition} \\
        \midrule
        \textbf{Object Recognition (OR)} & 
        The object recognition or object detection task involves confirming the presence of an object \cite{lin2015microsoft}. In this study, we investigated the ability of GPT-4V to identify and categorize objects within an image based on their visual features. \textit{For example cat, bottle, car, etc.} \\
        \midrule
        \textbf{Scene Understanding (SU)} & 
        The task involves interpreting the context and the overall scene beyond just individual objects \cite{kafle2017analysis}. \textit{For example, A girl is sitting on a bench in a park.} \\
        \midrule
        \textbf{Relationship Understanding (RU)} & 
        This task requires models to identify, characterize, and reason about relationships between different objects within an image \cite{Krishna2016VisualGC}, which includes relationships like spatial proximity, interaction, ownership, causality, social, and more. \textit{For example, A girl sitting on a desk is feeding the cat, and there is a pet-owner relationship between the girl and the cat.} \\
        \midrule
        \textbf{Semantic Segmentation (SS)} & 
        This task involves dividing an image into parts with a semantic meaning \cite{lin2015microsoft}, such as identifying roads, buildings, and people in a street scene. \\
        \midrule
        \textbf{Image Captioning (IC)} & 
        In this task, the VLM is required to generate a natural language description of an image \cite{lin2015microsoft}, and such a description captures the scene's content, including objects, actions, relationships, emotions, and atmosphere. \textit{For example, This image contains a girl wearing a pink color skirt and feeding a white color cat.} \\
        \midrule
        \textbf{Image-Text Matching (ITM)} & 
        This task requires the VLM to comprehend which parts of the text correspond to the image \cite{xu-etal-2023-multiinstruct}. Given an image-text pair, we prompted the VLM to select the exact part of the text that best describes the image. \\
        \midrule
        \textbf{Unrelatedness (U)} & 
        This is a new task that we introduced. Herein, we prompted the VLM to select the exact part of the text that is \textit{not relevant} to the image when given an image-text pair. \\
        \midrule
        \textbf{Visual Question Answering (VQA)} & 
        Under this task, the VLM needs to understand a natural language question about an image and generate appropriate answers \cite{agrawal2016vqa} by integrating visual understanding, language comprehension, and reasoning abilities. \\
        \bottomrule
    \end{tabular}
    }
    \label{tab:task-definitions} 
\end{table}
The tasks are related to each other as follows.

\textbf{Hierarchical Dependency:} Image-centric tasks like OR and SS provide foundational data (objects and pixel-level details) that feed into higher-level tasks like SU and RU, which interpret context and relationships. These, in turn, support multimodal tasks like IC and VQA, which require both visual and linguistic synthesis.

\textbf{Complementary:} Each task probes a distinct aspect of comprehension--OR focuses on ``what,'' RU on ``how,'' SU on ``where,'' and SS on ``to what extent''--while ITM, U, and VQA test how well this understanding aligns with text.

\textbf{Shared Input, Diverse Outputs:} A single image-text pair (e.g., a car outside a hotel) can be analyzed across all eight tasks. For instance, OR identifies the car, SU places it in a parking context, IC describes it, and VQA answers questions about it, with ITM ensuring text-image consistency throughout.

\textbf{Evaluation Synergy:} Together, they form a comprehensive framework. Success in one task (e.g., accurate OR) enhances performance in others (e.g., precise IC), while failure in a foundational task (e.g., missing an object in OR) cascades to higher-level tasks like VQA.

\subsection{Formal Definition: Vision and Language Tasks}
\label{subsec:DefinitionsVLTasks}
To study the visual and linguistic abilities of VLMs, we introduce eight tasks where the input comprises a combination of \textit{images, texts, and prompts}, and the outputs are \textit{responses} generated by VLMs (as illustrated in Table~\ref{Table:VLMoutputsIntro}).
To provide a formal structure for our benchmark, \emph{VLURes}, we define its task space. Let $\mathcal{I}$ denote the space of images and $\mathcal{X}_{\text{txt}}$ denote the space of article-length text descriptions relevant to an image. The benchmark comprises a set of eight distinct vision-language tasks, denoted by $\mathcal{T}$:
\begin{equation}
    \mathcal{T} = \{ \tau_k \}_{k=1}^8,
    \label{eq:task_set}
\end{equation}
where each task $\tau_k = (X_k, Y_k, \Phi_k)$ is characterized by:
\begin{itemize}
    \item $X_k$: The input data space for task $\tau_k$.
    \item $Y_k$: The output space, representing the expected response format (e.g., object lists, object relations, segmentation masks, image captions).
    \item $\Phi_k$: The evaluation function (or metric) used to score the performance on task $\tau_k$, mapping predicted outputs and ground truth labels to a scalar score.
\end{itemize}

The tasks in \emph{VLURes} are broadly categorized based on their primary input modality requirements during visual reasoning. We formally define these subsets as follows:
\textbf{(1) Image-Only Reasoning Tasks ($\mathcal{T}_{\text{img}}$):} These tasks primarily rely on visual understanding of the image content.
\begin{equation}
    \mathcal{T}_{\text{img}} = \{ \tau_{\text{OR}}, \tau_{\text{SU}}, \tau_{\text{RU}}, \tau_{\text{SS}}, \tau_{\text{IC}} \}.
    \label{eq:task_set_img}
\end{equation}
For any task $\tau_k \in \mathcal{T}_{\text{img}}$, the input space is predominantly $X_k \subseteq \mathcal{I}$.

\textbf{(2) Image-Text Reasoning Tasks ($\mathcal{T}_{\text{img+txt}}$):} These tasks require joint reasoning over both the visual information in the image and the semantic content of the accompanying text.
\begin{equation}
    \mathcal{T}_{\text{img+txt}} = \{ \tau_{\text{ITM}}, \tau_{\text{U}}, \tau_{\text{VQA}} \}.
    \label{eq:task_set_img_txt}
\end{equation}
For any task $\tau_k \in \mathcal{T}_{\text{img+txt}}$, the input space involves pairs of images and texts, $X_k \subseteq \mathcal{I} \times \mathcal{X}_{\text{txt}}$. 

\textbf{Task Interdependencies:} \emph{VLURes} is designed such that tasks exhibit hierarchical and complementary relationships.
This structure can be conceptualized as a Directed Acyclic Graph (DAG), $G = (\mathcal{T}, E)$, where an edge $(\tau_i, \tau_j) \in E$ signifies that the successful execution of task $\tau_i$ potentially provides necessary or beneficial information for task $\tau_j$. For instance, accurate Object Recognition ($\tau_{\text{OR}}$) and Semantic Segmentation ($\tau_{\text{SS}}$) form foundations for Scene Understanding ($\tau_{\text{SU}}$) and Relationship Understanding ($\tau_{\text{RU}}$). In turn, these contribute to complex generative tasks like Image Captioning ($\tau_{\text{IC}}$) and reasoning tasks such as Visual Question Answering ($\tau_{\text{VQA}}$) and Image-Text Matching ($\tau_{\text{ITM}}$).
Quantifying the strength of these dependencies can be approached using information-theoretic measures. For example, the conditional mutual information between the outputs $Y_i$ and $Y_j$ of two tasks, given their respective inputs $X_i$ and $X_j$ (which are often derived from the same underlying image-text pair), indicates how much information one task's output provides about the other, conditional on the inputs:
\begin{equation}
    I(Y_i; Y_j \mid X_i, X_j) = \mathbb{E}_{p(x_i, x_j, y_i, y_j)} \left[ \log \frac{p(y_i, y_j | x_i, x_j)}{p(y_i | x_i) p(y_j | x_j)} \right].
    \label{eq:mutual_info}
\end{equation}
Here, the expectation is taken over the joint distribution of inputs and outputs. High mutual information between tasks like $\tau_{\text{OR}}$ and $\tau_{\text{IC}}$ formally supports the intuition that identifying objects is crucial for generating accurate descriptions. Computing these values empirically across the benchmark could further highlight the structured nature of visual and linguistic understanding required by \emph{VLURes}.

\clearpage
\subsection{Introducing the \emph{Unrelatedness} Task}
Advances in VLM development have enabled researchers to investigate a plethora of tasks embedded in the many recently proposed benchmarks. 
Looking at the multimodal benchmarks, that is, benchmarks which evaluate VLM performance when the input to the foundation model is \textbf{ \emph{text + Y}}, where \textbf{Y = \{ image, video, audio, speech, music, molecule, etc.\}}, we observe that there are no benchmarks which explicitly studied the foundation model's abilities to distinguish the information signals from one modality, which are not relevant to the other modality. 

\begin{figure*}[h]
\centering
\includegraphics[width=0.95\textwidth]{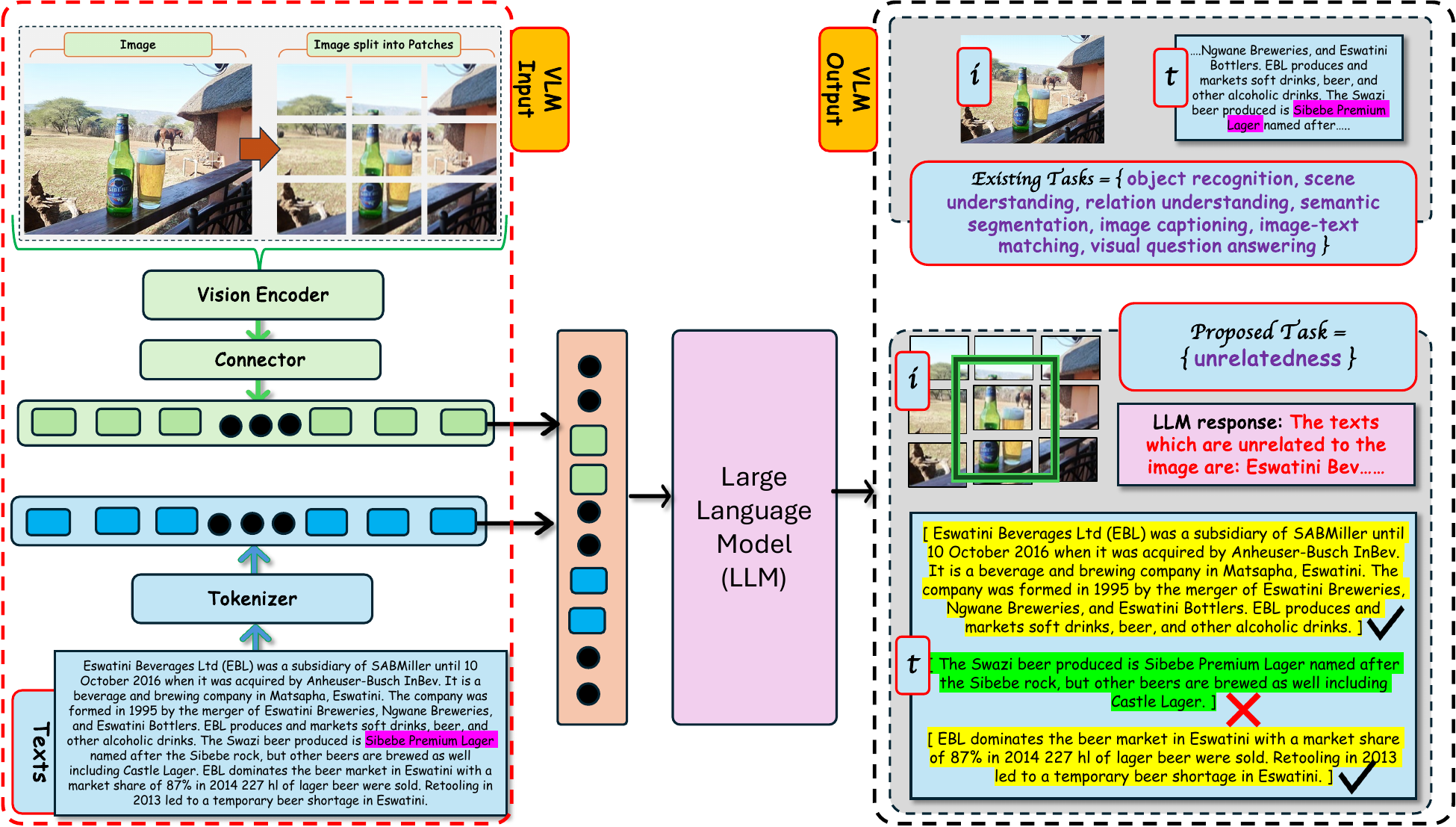} 
\caption{ The proposed \textbf{ \emph{Unrelatedness}} task. 
\textbf{Left:} The VLM inputs consists of two \textit{modalities}, a pair of \textbf{image} and \textbf{texts}. 
The image goes through a series of transformations in the \emph{vision encoder} and \emph{connector}, generating visual tokens ready for alignment in a shared embedding space.
Similarly, the text is tokenized by a tokenizer, and textual tokens are generated. The textual and visual tokens are aligned in a shared embedding space and fed as input to the LLM. 
\textbf{Right.} The LLM uses its \emph{multimodal understanding} to decide what textual information is relevant to different parts of the image. 
We see that the text painted green (marked with a cross sign) is directly related to the region of the image shown inside a green square box. That is, the text matches the image part shown in green. 
But in this task, we are interested in the text \emph{unrelated} to the image. Hence, the text painted yellow (marked with a check sign) is the answer to our \emph{Unrelatedness} task.  
}
\label{fig:UnrelatednessTask}
\end{figure*}

From this point of view, we deliberate on this effort and define a new task called \textbf{ \emph{unrelatedness}}. 
The \emph{unrelatedness} task encourages models to first understand the information available from all modalities in the input. 
Then, via a shared embedding space, the model leverages its multi-modal alignment abilities to align the textual information with the information contained in the other modality, such as image or video. 
After that, the model must discard the textual information related to the image or video in the shared embedding space and focus on the textual information not relevant to the image or video.
Finally, the model generates the answer, highlighting all textual information unrelated to the image. 
Figure \ref{fig:UnrelatednessTask} illustrates the entire process.

\clearpage
\subsection{Detailed Table 1 from Page 2}
\begin{table}[h]
    \centering
    \caption{ \emph{VLU-RES} includes eight \colorbox{red!10}{tasks}, and \textcolor{violet}{prompts} specific to each task needed to generate the \textcolor{teal}{responses} from the VLM.}
    \resizebox{0.91\textwidth}{!}{
    \begin{tabular}{@{}p{2.5cm}p{4cm}p{15cm}@{}}
        \toprule
        \textbf{Task} & \textbf{Prompt} & \textbf{Response from Vision Language Model} \\
        \midrule
        \textbf{1. Object Recognition (OR)} &
        \textcolor{violet}{Analyze this image and list all objects present.} Categorize each object into groups such as furniture, electronic devices, clothing, etc. Be thorough and specific in your identification. &
        Image Analysis: \textcolor{teal}{Objects Present: 1. Infrastructure: - Bridge - Railings - Streetlights - Road markings 2. Buildings: - High-rise buildings - Residential or commercial structures 3. People:  - Individuals walking or cycling 4. Transportation: - Bicycles. Categories: Infrastructure: Bridge, railings, streetlights, road markings. Buildings: High-rise buildings. People: Individuals. Transportation: Bicycles.} \\
        \midrule
        \textbf{2. Scene Understanding (SU)} &
        \textcolor{violet}{Describe the overall scene in this image.} What is the setting, and what activities or events are taking place? Provide a comprehensive overview of the environment and any actions occurring. &
        Image Analysis: \textcolor{teal}{The scene depicts an urban setting with a modern bridge leading towards a cityscape of tall buildings. The bridge has a dedicated pathway for pedestrians and cyclists, as indicated by the markings on the ground. There are a few people walking and cycling along this path, suggesting it is a popular route for commuting or leisure activities...... 
        The bridge is flanked by metal railings and streetlights, and it appears to span over a body of water, visible on the right side. The sky is mostly clear with some clouds, indicating a pleasant day. The overall atmosphere is calm and orderly, typical of a well-maintained urban environment. } \\
        \midrule
        \textbf{3. Relationship Understanding (RU)} &
        \textcolor{violet}{Identify any interactions or relationships between objects or entities in this image.} How are they related or interacting with each other? Explain any spatial, functional, or social connections you observe. &
        Image Analysis: \textcolor{teal}{Interactions and Relationships. 1. Spatial Relationships: - The image shows a bridge or elevated pathway leading towards a cityscape with several tall buildings in the background. The pathway is divided into lanes, likely for pedestrians and cyclists. - The bridge is flanked by railings on both sides, providing safety and separation from the surrounding environment. 2. Functional Connections: - The bridge serves as a connection between different parts of the city, facilitating movement for pedestrians and cyclists. It likely connects residential or commercial areas with the city center. - The presence of streetlights along the bridge suggests it is used during nighttime as well, enhancing safety and visibility. 
        3. Social Interactions: - There are people walking and cycling on the bridge, indicating it is a shared space for different modes of non-motorized transport. - The design of the bridge promotes social interaction by providing a shared space for pedestrians and cyclists, encouraging a community atmosphere. Overall, the image depicts a well-integrated urban infrastructure that supports efficient and safe movement within the city, fostering both functional and social interactions. } \\
        \midrule
        \textbf{4. Semantic Segmentation (SS)} &
        \textcolor{violet}{Divide this image into different semantic regions.} Label each region (e.g., sky, buildings, people, street) and briefly describe its contents. Provide a clear breakdown of the image’s composition. &
        Image Analysis: \textcolor{teal}{Semantic Regions. 1. Sky: - Location: Top portion of the image. - Description: The sky is mostly clear with some scattered clouds, providing a bright backdrop to the scene. 
        2. Buildings: - Location: Middle section of the image, in the background.  - Description: A series of modern high-rise buildings with glass facades. They vary in height and design, contributing to an urban skyline. 
        3. Bridge/Pathway: Location: Foreground, extending from the bottom to the middle of the image - Description: A wide pedestrian and cycling pathway on a bridge. It is flanked by metal railings on both sides, with a clear division for different lanes. 
        4. People: Location: Middle of the pathway - Description: Several individuals are walking or cycling along the pathway, indicating active use of the bridge. 
        5. Street Lamps: - Location: Left side of the bridge - Description: Modern street lamps are positioned along the bridge, providing lighting for the pathway. 
        6. Water: - Location: Right side, below the bridge - Description: A body of water is visible beneath the bridge, suggesting the bridge spans over a river or canal. 
        This breakdown highlights the urban setting and infrastructure, emphasizing the interaction between architecture and public spaces. } \\
        \midrule
        \textbf{5. Image Captioning (IC)} &
        \textcolor{violet}{Provide a detailed, natural language description of what is happening in this image.} Narrate the scene as if you were explaining it to someone who cannot see it, including all relevant details and actions. &
        Image Analysis: \textcolor{teal}{The image depicts a modern urban scene featuring a pedestrian and cycling path on a bridge. The path is divided by a dashed white line, indicating separate lanes for cyclists and pedestrians. The bridge is flanked by metal railings on both sides, providing safety for those using the path. In the background, a skyline of tall, contemporary buildings is visible, suggesting a city setting. These buildings are mostly glass and steel, reflecting a modern architectural style. The sky above is partly cloudy, with patches of blue visible, indicating a clear day. On the path, a few people are visible. One person is cycling towards the city, while another is walking. The bridge appears to span over a body of water, as suggested by the glimpse of water visible through the railings on the right side. Overall, the scene conveys a sense of urban life, with infrastructure designed to accommodate both pedestrians and cyclists, set against the backdrop of a bustling cityscape. } \\
        \midrule
        \textbf{6. Image-Text Matching (ITM)} &
        \textcolor{violet}{Extract and list the specific parts of the text that closely match or directly reference entities, objects, or scenes depicted in the image.} Be precise in identifying these connections and explain the visual evidence that supports each textual reference. &
        \textcolor{teal}{The image depicts a modern urban area with high-rise buildings, which aligns with the description of Wentworth Point in the text. Here are the specific connections between the text and the image: 
        1. Wentworth Point: The text mentions that the incident occurred in Wentworth Point, a suburb with high-rise buildings. The image shows a skyline with several tall buildings, which is characteristic of urban areas like Wentworth Point. 
        2. Apartment Buildings: The text references Musa attempting to enter an apartment building. The image shows multiple high-rise residential buildings, which could represent the type of apartment complex mentioned. 
        3. Urban Setting: The image’s urban setting with modern infrastructure supports the text’s context of a suburban area with secured garages and apartment complexes. 
        These elements in the image visually support textual references to locations and settings of events described. } \\
        \midrule
        \textbf{7. Unrelatedness (U)} &
        \textcolor{violet}{Identify which parts of the text are not relevant to or not represented in the image.} Explain why these elements are unrelated by describing what is missing in the image that would be needed to illustrate these textual elements. &
        \textcolor{teal}{The image shows a view of a bridge or walkway leading towards a cluster of modern buildings, likely in an urban area. 
        Here’s how the text relates to the image: 
        1. Location Context: - The text mentions Wentworth Point, which is likely the area depicted in the image. The buildings in the image could be part of this suburb. 
        2. Unrelated Text Elements: 
        - Specific Events: The text describes a specific incident involving a killing, a text message, and interactions between individuals. None of these events or interactions are depicted in the image. 
        - Individuals and Vehicles: The text mentions specific people, a Toyota Kluger SUV, and a garage. The image does not show any people, vehicles, or a garage. 
        - Interior Details: The text includes details about entering a building and a damaged roller door. The image does not show any building interiors or specific architectural features like a roller door. 
        3. Missing Elements for Illustration: - To illustrate the textual elements, the image would need to show the specific individuals involved, the Toyota Kluger, the garage, or the interior of the building mentioned in the text. 
        Overall, while the image provides a general sense of the location, it does not depict the specific events or details described in the text. } \\
        \midrule
        \textbf{8. Visual Question Answering (VQA)} &
        \textcolor{violet}{What places are mentioned in the text or shown in the image?} For each place identified, indicate whether it appears in the text, the image, or both. If any of these places are famous or well-known locations, explain why they are significant. &
        \textcolor{teal}{The text mentions the following places: 
        1. Wentworth Point: This is where the incident occurred, specifically at a garage along Hill Road and near Half Street. It is a suburb in New South Wales, Australia. 
        2. Granville: Musa searched for Thagipur in several cafés and restaurants in this area. It is another suburb in New South Wales, Australia. 
        3. Toyota Kluger: Mentioned as the type of vehicle Musa was driving. The image shows a modern urban area with high-rise buildings, which could be representative of a suburb like Wentworth Point. However, without specific landmarks or signs, it’s not possible to definitively identify the location in the image as Wentworth Point or any other specific place mentioned in the text. 
        Significance: - Wentworth Point: Known for its residential developments and proximity to Sydney Olympic Park, making it a notable suburb in Sydney. 
        - Granville: Known for its diverse community and historical significance in Sydney. 
        In summary, Wentworth Point and Granville are mentioned in the text, while the image could potentially depict a similar urban environment but does not explicitly show any identifiable location from the text. } \\
        \bottomrule
    \end{tabular}
    }
\end{table}

\clearpage
\subsection{Topic modeling with LDA}
\begin{table*}[h]
\centering
\caption{Topic modeling with LDA outputs for, k=50 topics, when trained on all the data samples of En, Jp, Sw, Ur. In the initial stages, we removed stop words from each dataset. 
Then, for each language, the topic frequencies are determined by calculating the expected count of each word within each topic over all documents in the dataset. 
Finally, we prompt GPT-4o to map the LDA top words to the semantic topics in English, Japanese, Swahili, and Urdu. Each row shows the language, matched semantic topic, and top words.}
\resizebox{\textwidth}{!}{%
\begin{tabular}{llll}
\toprule
\textbf{Language} & \textbf{Topic name} & \textbf{Top Words} \\
\hline

\multirow{10}{*}{En}
& Animals        & animal, animals, meat, fish, beef, shark, sharks \\
& Products       & product, products, currency, coin, coins, note, banknotes \\
& Buildings      & building, buildings, house, architecture, \\
& Locations      & city, state, country, united, republic \\
& Events         & event, events, festival, war, revolution, treaty, series, final, game, games \\
& Food           & food, milk, meat, rice, sugar, banana, dish, flour, cheese, gado \\
& Drinks         & drink, drinks, tea, coffee, beverage, water, \\
& Hobbies        & hobby, hobbies, sports, game, games, fun \\
& Works of Art   & art, works, painting, collecting, history\\
& Organization   & organization, government, party, state, parliament, bank \\
\hline

\multirow{10}{*}{Sw} 
& Animals       & wanyama, mifugo, samahani, maziwa \\
& Products      & bidhaa, kiwanda, mafuta, kampuni, biashara, shilingi\\
& Buildings     & jengo, ujenzi, barabara, nyumba, \\
& Locations     & jijini, mjini, tanzania, afrika, eneo, mkoa, wilaya\\
& Events        & mashindano, mbio, fainali, sikukuu\\
& Food          & maziwa, korosho, kilimo, dawa, mazao\\
& Drinks        & maji, pombe\\
& Hobbies       &  michezo, mchezo, timu, kucheza\\
& Works of Art  & sanaa, muziki, wasanii, msanii, utamaduni\\
& Organization  & serikali, bunge, wizara, taasisi, rais, chama, baraza\\
\hline

\multirow{10}{*}{Ur}
& Animals       & \begin{UrduText}جانور، جانوروں، گوشت، مچھلی، سانپ، پرندے، کتا، بلی\end{UrduText} \textcolor{teal}{\textbf{ [Animals]}} \\
& Products      & \begin{UrduText}فون، موبائل، آلے، صارفین، نوٹ، سکہ، سککوں، گاڑی، کمپیوٹر\end{UrduText} \textcolor{teal}{\textbf{ [Products]}} \\
& Buildings     & \begin{UrduText}عمارت، عمارتیں، محل، مزار، مسجد، ریلوے، کالج، مقبرہ، اسٹیشن\end{UrduText} \textcolor{teal}{\textbf{ [Buildings]}} \\
& Locations     & \begin{UrduText}شہر، مقام، پشاور، کراچی، لاہور، سیالکوٹ، اسلام آباد، ملک\end{UrduText} \textcolor{teal}{\textbf{ [Locations]}} \\
& Events        & \begin{UrduText}تہوار، عید، یوم، دن، جشن، تاریخ، تقریب، واقعہ، محفل\end{UrduText} \textcolor{teal}{\textbf{ [Events]}} \\
& Food          & \begin{UrduText}کھانا، دودھ، جلیبی، گوشت، چاول، روٹی، سبزی، پھل\end{UrduText} \textcolor{teal}{\textbf{ [Food]}} \\
& Drinks        & \begin{UrduText}پانی، مشروبات، چائے، دودھ، جوس، کولا\end{UrduText} \textcolor{teal}{\textbf{ [Drinks]}} \\
& Hobbies       & \begin{UrduText}کھیل، کھلاڑی، میدان، فٹ بال، کرکٹ، ہاکی، تیراکی\end{UrduText} \textcolor{teal}{\textbf{ [Hobbies]}} \\
& Works of Art  & \begin{UrduText}فن، فنون، ثقافت، عجائب، مصوری، مجسمہ، خطاطی، شاعری\end{UrduText} \textcolor{teal}{\textbf{ [Works of Art]}} \\
& Organization  & \begin{UrduText}حکومت، ریاست، سلطنت، وزیر، فوج، بینک، ادارہ، کمپنی\end{UrduText} \textcolor{teal}{\textbf{ [Organization]}} \\
\bottomrule
\end{tabular}%
}
\label{tab:lda-category-mapping}
\end{table*}

\clearpage
\subsection{Experiments with \emph{VLURes} Japanese Data}
\label{Appendix_ExperimentsJapaneseData}
\begin{table*}[h] 
    \centering
    \caption{Performance of VLMs on \textbf{eight VL tasks} under \colorbox{violet!20}{zero-shot} and \colorbox{teal!20}{one-shot} settings, measured by Accuracy (\%).
            \textbf{Input:} Japanese Texts + Images; \textbf{Output:} En, Jp responses.
            \colorbox{orange!20}{Shaded columns represent \{Jp\} in input and \{Jp\} in output VLM results.}
            En $\Delta$Acc. = \{En score from Table \ref{tab:ZerooneshotAccuracyEn} (En-Input) -- En score from this Table (Jp-Input)\}.
            Jp $\Delta$Acc. = \{Jp score from Table \ref{tab:ZerooneshotAccuracyEn} (En-Input) -- Jp score from this Table (Jp-Input).\}
            \textcolor{blue}{Positive} $\Delta$ in blue, \textcolor{red}{negative} $\Delta$ in red.
            }
    \resizebox{\textwidth}{!}{
\begin{tabular}{r|| c c >{\columncolor{apricot}}c >{\columncolor{apricot}}c | c c >{\columncolor{apricot}}c >{\columncolor{apricot}}c | c c >{\columncolor{apricot}}c >{\columncolor{apricot}}c | c c >{\columncolor{apricot}}c >{\columncolor{apricot}}c | c c >{\columncolor{apricot}}c >{\columncolor{apricot}}c || c c >{\columncolor{apricot}}c >{\columncolor{apricot}}c | c c >{\columncolor{apricot}}c >{\columncolor{apricot}}c | c c >{\columncolor{apricot}}c >{\columncolor{apricot}}c }
        \toprule
        \multirow{2}{*}{\textbf{Model}} & \multicolumn{4}{c|}{\textbf{Object Recognition}} & \multicolumn{4}{c|}{\textbf{Scene Understanding}} & \multicolumn{4}{c|}{\textbf{Relation Understanding}} & \multicolumn{4}{c|}{\textbf{Semantic Segmentation}} & \multicolumn{4}{c|}{\textbf{Image Captioning}} & \multicolumn{4}{c||}{\textbf{Image-Text Matching}} & \multicolumn{4}{c|}{\textbf{Unrelatedness}} & \multicolumn{4}{c}{\textbf{Visual Question Answering}} \\
        \cmidrule(lr){2-5} \cmidrule(lr){6-9} \cmidrule(lr){10-13} \cmidrule(lr){14-17} \cmidrule(lr){18-21} \cmidrule(lr){22-25} \cmidrule(lr){26-29} \cmidrule(lr){30-33}
        & En & $\Delta_{\text{En}}$ & Jp & $\Delta_{\text{Jp}}$ & En & $\Delta_{\text{En}}$ & Jp & $\Delta_{\text{Jp}}$ & En & $\Delta_{\text{En}}$ & Jp & $\Delta_{\text{Jp}}$ & En & $\Delta_{\text{En}}$ & Jp & $\Delta_{\text{Jp}}$ & En & $\Delta_{\text{En}}$ & Jp & $\Delta_{\text{Jp}}$ & En & $\Delta_{\text{En}}$ & Jp & $\Delta_{\text{Jp}}$ & En & $\Delta_{\text{En}}$ & Jp & $\Delta_{\text{Jp}}$ & En & $\Delta_{\text{En}}$ & Jp & $\Delta_{\text{Jp}}$ \\
        \midrule
        \rowcolor{gray!20}
        \multicolumn{33}{c}{\textcolor{violet}{\textbf{Zero-shot, \textit{Without Rationales}}}} \\
\bottomrule
    \end{tabular}
    }
    \label{tab:ZerooneshotAccuracyJp} 
\end{table*}
\begin{table*}[h] 
    \centering
    \caption{Performance of VLMs on \textbf{eight VL tasks} under \colorbox{blue!20}{finetuning} settings, measured by Accuracy (\%).
            \textbf{Input:} Japanese Texts + Images; \textbf{Output:} En, Jp responses.
            \colorbox{orange!20}{Shaded columns represent \{Jp\} in input and \{Jp\} in output VLM results.}
            En $\Delta$Acc. = \{En score from Table \ref{tab:FinetuneAccuracyEn} (En-Input) -- En score from this Table (Jp-Input)\}.
            Jp $\Delta$Acc. = \{Jp score from Table \ref{tab:FinetuneAccuracyEn} (En-Input) -- Jp score from this Table (Jp-Input).\}
            \textcolor{blue}{Positive} $\Delta$ in blue, \textcolor{red}{negative} $\Delta$ in red.
            }
    \resizebox{\textwidth}{!}{
    \begin{tabular}{r|| c c >{\columncolor{apricot}}c >{\columncolor{apricot}}c | c c >{\columncolor{apricot}}c >{\columncolor{apricot}}c | c c >{\columncolor{apricot}}c >{\columncolor{apricot}}c | c c >{\columncolor{apricot}}c >{\columncolor{apricot}}c | c c >{\columncolor{apricot}}c >{\columncolor{apricot}}c || c c >{\columncolor{apricot}}c >{\columncolor{apricot}}c | c c >{\columncolor{apricot}}c >{\columncolor{apricot}}c | c c >{\columncolor{apricot}}c >{\columncolor{apricot}}c }
        \toprule
        \multirow{2}{*}{\textbf{Model}} & \multicolumn{4}{c|}{\textbf{Object Recognition}} & \multicolumn{4}{c|}{\textbf{Scene Understanding}} & \multicolumn{4}{c|}{\textbf{Relation Understanding}} & \multicolumn{4}{c|}{\textbf{Semantic Segmentation}} & \multicolumn{4}{c|}{\textbf{Image Captioning}} & \multicolumn{4}{c||}{\textbf{Image-Text Matching}} & \multicolumn{4}{c|}{\textbf{Unrelatedness}} & \multicolumn{4}{c}{\textbf{Visual Question Answering}} \\
        \cmidrule(lr){2-5} \cmidrule(lr){6-9} \cmidrule(lr){10-13} \cmidrule(lr){14-17} \cmidrule(lr){18-21} \cmidrule(lr){22-25} \cmidrule(lr){26-29} \cmidrule(lr){30-33}
        & En & $\Delta_{\text{En}}$ & Jp & $\Delta_{\text{Jp}}$ & En & $\Delta_{\text{En}}$ & Jp & $\Delta_{\text{Jp}}$ & En & $\Delta_{\text{En}}$ & Jp & $\Delta_{\text{Jp}}$ & En & $\Delta_{\text{En}}$ & Jp & $\Delta_{\text{Jp}}$ & En & $\Delta_{\text{En}}$ & Jp & $\Delta_{\text{Jp}}$ & En & $\Delta_{\text{En}}$ & Jp & $\Delta_{\text{Jp}}$ & En & $\Delta_{\text{En}}$ & Jp & $\Delta_{\text{Jp}}$ & En & $\Delta_{\text{En}}$ & Jp & $\Delta_{\text{Jp}}$ \\
        \midrule
        \rowcolor{blue!20} 
        \multicolumn{33}{c}{  \textcolor{violet}{\textbf{Zero-shot, \textit{With Rationales}}}  } \\ 
        \midrule
        LlaVa Mistral 7B       & 37.4 & \textcolor{red}{-12.1} & 44.9 & \textcolor{red}{-6.5} & 37.4 & \textcolor{red}{-11.7} & 42.0 & \textcolor{red}{-6.5} & 35.6 & \textcolor{red}{-11.2} & 37.4 & \textcolor{red}{-6.8} & 34.6 & \textcolor{red}{-12.1} & 35.0 & \textcolor{red}{-6.4} & 35.6 & \textcolor{red}{-11.5} & 34.0 & \textcolor{red}{-6.8} & 35.5 & \textcolor{red}{-12.6} & 39.5 & \textcolor{red}{-7.6} & 40.4 & \textcolor{red}{-11.8} & 44.1 & \textcolor{red}{-6.9} & 38.4 & \textcolor{red}{-12.5} & 43.4 & \textcolor{red}{-6.7} \\
        Qwen2VL 7B             & 68.6 & \textcolor{blue}{0.6} & 76.4 & \textcolor{blue}{5.1} & 69.7 & \textcolor{red}{-2.0} & 73.4 & \textcolor{blue}{5.8} & 67.3 & \textcolor{blue}{0.7} & 68.8 & \textcolor{blue}{5.1} & 66.1 & \textcolor{red}{-0.3} & 66.4 & \textcolor{blue}{5.5} & 67.1 & \textcolor{red}{-0.5} & 65.4 & \textcolor{blue}{5.4} & 66.7 & \textcolor{red}{-0.6} & 70.9 & \textcolor{blue}{4.9} & 71.9 & \textcolor{red}{-0.5} & 75.5 & \textcolor{blue}{5.3} & 69.8 & \textcolor{red}{-0.8} & 75.0 & \textcolor{blue}{5.0} \\
        PALO 7B                & 48.6 & \textcolor{red}{-0.7} & 54.4 & \textcolor{blue}{5.2} & 48.6 & \textcolor{red}{-0.8} & 51.4 & \textcolor{blue}{5.2} & 47.2 & \textcolor{red}{-0.5} & 46.8 & \textcolor{blue}{5.8} & 45.7 & \textcolor{red}{-0.5} & 44.4 & \textcolor{blue}{5.1} & 46.7 & \textcolor{red}{-0.5} & 43.4 & \textcolor{blue}{5.0} & 46.2 & \textcolor{red}{-0.3} & 48.9 & \textcolor{blue}{5.2} & 51.0 & \textcolor{red}{-7.3} & 53.5 & \textcolor{blue}{2.4} & 49.3 & \textcolor{red}{-0.4} & 52.9 & \textcolor{blue}{5.2} \\
        MAYA 8B                & 49.5 & \textcolor{red}{-0.3} & 55.2 & \textcolor{red}{-0.4} & 49.7 & \textcolor{red}{-0.4} & 52.4 & \textcolor{blue}{0.3} & 48.3 & \textcolor{red}{-0.5} & 47.6 & \textcolor{red}{-0.1} & 46.6 & \textcolor{red}{-0.1} & 45.2 & \textcolor{red}{-0.3} & 47.8 & \textcolor{red}{-0.5} & 44.2 & \textcolor{blue}{0.5} & 47.3 & \textcolor{red}{-0.1} & 49.7 & \textcolor{blue}{2.4} & 49.8 & \textcolor{red}{-8.7} & 54.3 & \textcolor{blue}{2.5} & 50.7 & \textcolor{red}{-0.5} & 53.7 & \textcolor{red}{-0.5} \\
        LlaVa Mistral 13B      & 45.8 & \textcolor{red}{-8.4} & 48.9 & \textcolor{red}{-8.5} & 46.1 & \textcolor{red}{-8.3} & 46.0 & \textcolor{red}{-8.0} & 44.7 & \textcolor{red}{-8.1} & 41.4 & \textcolor{red}{-8.6} & 43.8 & \textcolor{red}{-8.9} & 39.0 & \textcolor{red}{-8.1} & 44.4 & \textcolor{red}{-8.7} & 38.0 & \textcolor{red}{-8.2} & 43.9 & \textcolor{red}{-8.4} & 43.5 & \textcolor{red}{-8.2} & 49.3 & \textcolor{red}{-8.9} & 48.1 & \textcolor{red}{-8.8} & 47.5 & \textcolor{red}{-8.9} & 47.4 & \textcolor{red}{-9.1} \\
        PALO 13B               & 49.7 & \textcolor{blue}{0.4} & 52.5 & \textcolor{blue}{0.2} & 50.0 & \textcolor{red}{-0.5} & 49.4 & \textcolor{blue}{0.4} & 48.7 & \textcolor{red}{-0.5} & 44.8 & \textcolor{blue}{0.4} & 47.0 & \textcolor{red}{-0.6} & 42.4 & \textcolor{red}{-0.5} & 47.6 & \textcolor{blue}{0.4} & 41.4 & \textcolor{blue}{0.4} & 47.9 & \textcolor{red}{-0.7} & 46.9 & \textcolor{blue}{0.3} & 52.9 & \textcolor{red}{-13.0} & 51.5 & \textcolor{red}{-0.4} & 50.8 & \textcolor{red}{-0.9} & 51.0 & \textcolor{red}{-0.3} \\
        \midrule
        \rowcolor{blue!20} %
        \multicolumn{33}{c}{  \textcolor{violet}{\textbf{Zero-shot, \textit{With Rationales}}}    }    \\ 
        \midrule
        LlaVa Mistral 7B      & 37.4 & \textcolor{red}{-12.9} & 45.6 & \textcolor{red}{-6.3} & 37.2 & \textcolor{red}{-11.8} & 42.1 & \textcolor{red}{-6.4} & 36.6 & \textcolor{red}{-11.4} & 41.8 & \textcolor{red}{-6.5} & 37.6 & \textcolor{red}{-11.3} & 43.0 & \textcolor{red}{-6.3} & 35.4 & \textcolor{red}{-11.3} & 38.1 & \textcolor{red}{-6.8} & 36.5 & \textcolor{red}{-11.4} & 39.0 & \textcolor{red}{-7.5} & 41.1 & \textcolor{red}{-12.0} & 45.0 & \textcolor{red}{-6.9} & 33.9 & \textcolor{red}{-12.1} & 38.8 & \textcolor{red}{-6.7} \\
        Qwen2VL 7B            & 68.9 & \textcolor{red}{-0.3} & 76.7 & \textcolor{blue}{4.5} & 67.7 & \textcolor{red}{-0.3} & 73.5 & \textcolor{blue}{4.5} & 68.6 & \textcolor{red}{-0.5} & 73.3 & \textcolor{blue}{4.2} & 69.2 & \textcolor{red}{-0.3} & 74.0 & \textcolor{blue}{4.4} & 67.6 & \textcolor{red}{-0.5} & 69.4 & \textcolor{blue}{4.5} & 68.2 & \textcolor{red}{-0.3} & 72.1 & \textcolor{blue}{4.5} & 72.0 & \textcolor{red}{-0.6} & 76.0 & \textcolor{blue}{4.5} & 65.3 & \textcolor{red}{-0.3} & 70.5 & \textcolor{blue}{4.2} \\
        PALO 7B               & 48.1 & \textcolor{red}{-0.2} & 55.1 & \textcolor{blue}{4.8} & 47.9 & \textcolor{red}{-0.2} & 50.8 & \textcolor{blue}{4.8} & 47.7 & \textcolor{red}{-0.2} & 51.5 & \textcolor{blue}{4.8} & 48.1 & \textcolor{red}{-0.2} & 52.3 & \textcolor{blue}{4.8} & 46.8 & \textcolor{red}{-0.6} & 46.9 & \textcolor{blue}{4.3} & 47.8 & \textcolor{red}{-0.5} & 49.6 & \textcolor{blue}{5.1} & 51.4 & \textcolor{red}{-4.0} & 53.7 & \textcolor{blue}{2.6} & 44.4 & \textcolor{red}{-0.5} & 48.2 & \textcolor{blue}{5.2} \\
        MAYA 8B               & 49.4 & \textcolor{red}{-0.2} & 55.4 & \textcolor{red}{-0.2} & 49.2 & \textcolor{red}{-0.2} & 52.3 & \textcolor{red}{-0.2} & 49.0 & \textcolor{red}{-0.2} & 52.1 & \textcolor{red}{-0.2} & 49.4 & \textcolor{red}{-0.2} & 53.1 & \textcolor{red}{-0.2} & 47.7 & \textcolor{red}{-0.3} & 47.8 & \textcolor{red}{-0.4} & 49.1 & \textcolor{red}{-0.5} & 50.4 & \textcolor{red}{-0.5} & 52.9 & \textcolor{red}{-2.5} & 54.5 & \textcolor{red}{-0.1} & 45.6 & \textcolor{red}{-0.4} & 49.0 & \textcolor{red}{-0.4} \\
        LlaVa Mistral 13B     & 55.2 & \textcolor{red}{-0.3} & 58.0 & \textcolor{blue}{0.0} & 53.8 & \textcolor{blue}{0.0} & 53.9 & \textcolor{blue}{0.0} & 54.8 & \textcolor{red}{-0.4} & 54.5 & \textcolor{red}{-0.2} & 55.1 & \textcolor{blue}{0.0} & 55.4 & \textcolor{blue}{0.0} & 53.3 & \textcolor{blue}{0.0} & 49.9 & \textcolor{blue}{0.0} & 54.4 & \textcolor{blue}{0.0} & 53.0 & \textcolor{blue}{0.0} & 57.9 & \textcolor{blue}{0.0} & 57.1 & \textcolor{red}{-0.4} & 50.9 & \textcolor{blue}{0.0} & 51.0 & \textcolor{blue}{0.0} \\
        PALO 13B              & 49.5 & \textcolor{red}{-0.1} & 52.6 & \textcolor{red}{-0.3} & 49.4 & \textcolor{red}{-0.2} & 49.3 & \textcolor{red}{-0.2} & 49.3 & \textcolor{red}{-0.3} & 49.3 & \textcolor{red}{-0.6} & 50.1 & \textcolor{red}{-0.3} & 50.1 & \textcolor{red}{-0.5} & 48.1 & \textcolor{red}{-0.4} & 45.0 & \textcolor{red}{-0.5} & 49.1 & \textcolor{red}{-0.3} & 48.1 & \textcolor{red}{-0.4} & 52.5 & \textcolor{blue}{1.3} & 52.2 & \textcolor{red}{-0.4} & 46.2 & \textcolor{red}{-0.3} & 46.3 & \textcolor{red}{-0.4} \\
        \midrule
        \rowcolor{blue!20} 
        \multicolumn{33}{c}{    \textcolor{teal}{\textbf{One-shot, \textit{Without Rationales}}}    } \\ 
        \midrule
        LlaVa Mistral 7B      & 45.3 & \textcolor{red}{-6.4} & 46.9 & \textcolor{red}{-1.5} & 45.8 & \textcolor{red}{-6.3} & 46.0 & \textcolor{red}{-2.0} & 43.1 & \textcolor{red}{-6.8} & 43.2 & \textcolor{red}{-3.2} & 45.0 & \textcolor{red}{-6.7} & 44.5 & \textcolor{red}{-2.1} & 42.8 & \textcolor{red}{-6.2} & 39.0 & \textcolor{red}{-3.8} & 42.5 & \textcolor{red}{-6.0} & 41.9 & \textcolor{red}{-4.1} & 46.5 & \textcolor{red}{-6.2} & 50.5 & \textcolor{red}{-2.0} & 44.2 & \textcolor{red}{-5.9} & 45.8 & \textcolor{red}{-1.8} \\
        Qwen2VL 7B            & 70.8 & \textcolor{red}{-0.6} & 72.7 & \textcolor{blue}{5.1} & 70.9 & \textcolor{red}{-0.4} & 73.0 & \textcolor{blue}{5.1} & 68.5 & \textcolor{red}{-0.5} & 71.7 & \textcolor{blue}{5.1} & 70.6 & \textcolor{red}{-0.5} & 71.5 & \textcolor{blue}{5.2} & 69.2 & \textcolor{red}{-0.5} & 66.6 & \textcolor{blue}{5.2} & 69.2 & \textcolor{red}{-0.5} & 70.1 & \textcolor{blue}{5.0} & 72.5 & \textcolor{red}{-0.4} & 76.1 & \textcolor{blue}{5.0} & 69.9 & \textcolor{red}{-0.4} & 72.1 & \textcolor{blue}{5.4} \\
        PALO 7B               & 50.1 & \textcolor{red}{-0.6} & 51.3 & \textcolor{blue}{4.7} & 50.6 & \textcolor{red}{-0.6} & 50.8 & \textcolor{blue}{4.4} & 47.8 & \textcolor{red}{-0.3} & 49.6 & \textcolor{blue}{5.0} & 49.9 & \textcolor{red}{-0.5} & 49.3 & \textcolor{blue}{5.3} & 48.3 & \textcolor{red}{-0.4} & 45.0 & \textcolor{blue}{5.0} & 48.7 & \textcolor{red}{-0.6} & 48.4 & \textcolor{blue}{4.2} & 51.3 & \textcolor{red}{-3.1} & 54.1 & \textcolor{blue}{2.5} & 50.2 & \textcolor{red}{-0.6} & 50.5 & \textcolor{blue}{5.1} \\
        MAYA 8B               & 51.6 & \textcolor{red}{-0.4} & 51.6 & \textcolor{red}{-0.4} & 51.8 & \textcolor{red}{-0.5} & 51.6 & \textcolor{red}{-0.1} & 49.3 & \textcolor{red}{-0.5} & 50.4 & \textcolor{red}{-0.4} & 51.4 & \textcolor{red}{-0.7} & 50.4 & \textcolor{red}{-0.3} & 49.7 & \textcolor{red}{-0.5} & 45.4 & \textcolor{red}{-0.4} & 50.0 & \textcolor{red}{-0.6} & 49.3 & \textcolor{red}{-0.1} & 52.9 & \textcolor{red}{-1.7} & 54.9 & \textcolor{red}{-0.7} & 51.6 & \textcolor{red}{-0.7} & 51.3 & \textcolor{red}{-0.4} \\
        LlaVa Mistral 13B     & 57.0 & \textcolor{red}{-0.2} & 54.1 & \textcolor{red}{-0.2} & 56.9 & \textcolor{red}{-0.2} & 54.0 & \textcolor{red}{-0.2} & 54.0 & \textcolor{red}{-0.2} & 53.1 & \textcolor{red}{-0.2} & 56.0 & \textcolor{red}{-0.2} & 52.3 & \textcolor{red}{-0.2} & 55.1 & \textcolor{blue}{0.0} & 48.0 & \textcolor{red}{-0.4} & 55.4 & \textcolor{red}{-0.5} & 51.1 & \textcolor{red}{-0.5} & 58.4 & \textcolor{red}{-0.6} & 57.0 & \textcolor{red}{-0.4} & 55.8 & \textcolor{red}{-0.4} & 53.1 & \textcolor{red}{-0.5} \\
        PALO 13B              & 52.6 & \textcolor{red}{-0.5} & 48.8 & \textcolor{red}{-0.5} & 52.3 & \textcolor{red}{-0.6} & 48.8 & \textcolor{red}{-0.3} & 49.4 & \textcolor{red}{-0.4} & 47.7 & \textcolor{red}{-0.4} & 50.9 & \textcolor{red}{-0.1} & 47.1 & \textcolor{red}{-0.7} & 50.1 & \textcolor{red}{-0.6} & 42.6 & \textcolor{red}{-0.5} & 50.0 & \textcolor{red}{-0.4} & 46.2 & \textcolor{red}{-0.7} & 53.3 & \textcolor{blue}{11.9} & 52.3 & \textcolor{red}{-0.7} & 51.3 & \textcolor{red}{-0.5} & 48.3 & \textcolor{red}{-0.3} \\
        \midrule
        \rowcolor{blue!20} 
        \multicolumn{33}{c}{   \textcolor{teal}{\textbf{One-shot, \textit{With Rationales}}}   } \\ 
        \midrule
        LlaVa Mistral 7B      & 44.7 & \textcolor{red}{-5.8} & 52.1 & \textcolor{red}{-0.7} & 44.7 & \textcolor{red}{-5.7} & 48.8 & \textcolor{red}{-0.5} & 43.7 & \textcolor{red}{-5.9} & 50.0 & \textcolor{red}{-0.4} & 44.5 & \textcolor{red}{-5.7} & 49.1 & \textcolor{red}{-0.7} & 43.5 & \textcolor{red}{-5.8} & 43.8 & \textcolor{red}{-2.0} & 44.8 & \textcolor{red}{-6.3} & 44.9 & \textcolor{red}{-0.7} & 46.3 & \textcolor{red}{-6.5} & 50.1 & \textcolor{red}{-0.4} & 45.1 & \textcolor{red}{-5.7} & 48.0 & \textcolor{red}{-0.7} \\
        Qwen2VL 7B            & \textbf{71.3} & \textcolor{red}{-0.4} & \textbf{77.2} & \textcolor{blue}{5.3} & \textbf{71.2} & \textcolor{red}{-0.5} & \textbf{74.0} & \textcolor{blue}{4.6} & \textbf{70.1} & \textcolor{red}{-0.7} & \textbf{75.2} & \textcolor{blue}{4.8} & \textbf{71.5} & \textcolor{red}{-0.5} & \textbf{74.0} & \textcolor{blue}{5.2} & \textbf{69.5} & \textcolor{red}{-0.4} & \textbf{70.0} & \textcolor{blue}{5.0} & \textbf{70.2} & \textcolor{red}{-0.3} & \textbf{69.7} & \textcolor{blue}{5.0} & \textbf{72.5} & \textcolor{red}{-0.6} & \textbf{75.0} & \textcolor{blue}{3.7} & \textbf{71.6} & \textcolor{red}{-0.7} & \textbf{72.8} & \textcolor{blue}{5.1} \\
        PALO 7B               & 50.8 & \textcolor{red}{-0.2} & 55.4 & \textcolor{blue}{5.6} & 50.5 & \textcolor{red}{-0.4} & 51.7 & \textcolor{blue}{5.1} & 50.0 & \textcolor{red}{-0.5} & 52.8 & \textcolor{blue}{5.1} & 50.8 & \textcolor{red}{-0.3} & 52.1 & \textcolor{blue}{4.6} & 48.6 & \textcolor{red}{-0.4} & 48.2 & \textcolor{blue}{5.3} & 50.5 & \textcolor{red}{-0.7} & 48.4 & \textcolor{blue}{4.9} & 51.3 & \textcolor{red}{-6.6} & 52.7 & \textcolor{blue}{0.9} & 51.2 & \textcolor{red}{-0.7} & 51.4 & \textcolor{blue}{4.8} \\
        MAYA 8B               & 52.5 & \textcolor{red}{-0.8} & 56.3 & \textcolor{red}{-0.8} & 51.7 & \textcolor{red}{-0.3} & 52.6 & \textcolor{red}{-0.3} & 51.7 & \textcolor{red}{-0.9} & 54.2 & \textcolor{red}{-0.9} & 52.7 & \textcolor{red}{-0.9} & 52.9 & \textcolor{red}{-0.8} & 50.3 & \textcolor{red}{-0.8} & 49.0 & \textcolor{red}{-0.9} & 52.0 & \textcolor{red}{-0.9} & 49.1 & \textcolor{red}{-0.9} & 52.7 & \textcolor{red}{-5.0} & 53.5 & \textcolor{red}{-0.7} & 52.7 & \textcolor{red}{-0.9} & 52.2 & \textcolor{red}{-0.3} \\
        LlaVa Mistral 13B     & 57.6 & \textcolor{red}{-0.6} & 58.1 & \textcolor{red}{-0.8} & 57.6 & \textcolor{red}{-0.7} & 55.1 & \textcolor{red}{-0.2} & 57.0 & \textcolor{red}{-0.9} & 56.5 & \textcolor{red}{-0.6} & 57.8 & \textcolor{red}{-0.7} & 55.5 & \textcolor{red}{-0.5} & 55.4 & \textcolor{red}{-0.5} & 51.3 & \textcolor{red}{-0.5} & 56.7 & \textcolor{red}{-0.7} & 51.2 & \textcolor{red}{-0.8} & 58.7 & \textcolor{red}{-0.8} & 56.0 & \textcolor{red}{-0.5} & 57.5 & \textcolor{red}{-0.8} & 53.7 & \textcolor{red}{-0.7} \\
        PALO 13B              & 52.3 & \textcolor{red}{-0.4} & 52.7 & \textcolor{red}{-0.6} & 52.0 & \textcolor{red}{-0.3} & 49.8 & \textcolor{red}{-0.6} & 51.3 & \textcolor{red}{-0.4} & 51.3 & \textcolor{red}{-0.5} & 52.4 & \textcolor{red}{-0.4} & 50.0 & \textcolor{red}{-0.5} & 50.4 & \textcolor{red}{-0.5} & 46.1 & \textcolor{red}{-0.4} & 51.3 & \textcolor{red}{-0.6} & 45.6 & \textcolor{red}{-0.3} & 52.2 & \textcolor{red}{-11.8} & 51.5 & \textcolor{red}{-0.6} & 51.5 & \textcolor{red}{-0.1} & 49.4 & \textcolor{red}{-0.5} \\
        \bottomrule
    \end{tabular}
    }
    \label{tab:FinetuneAccuracyJp} 
\end{table*}

\clearpage
\subsection{Experiments with \emph{VLURes} Swahili Data}
\label{Appendix_ExperimentsSwahiliData}
From our initial experiments and analysis, we found that these open-source models: LlaVa Mistral 7B, PALO 7B, MAYA 7B, Qwen2VL 7B , LlaVa Mistral 13B, and PALO 13B, have little understanding of Sw. 
Therefore, we excluded the models from further experiments.  
\begin{table*}[h]
    \centering
    \caption{Performance of VLMs on \textbf{eight VL tasks} under \colorbox{violet!20}{zero-shot} and \colorbox{teal!20}{one-shot} settings, measured by Accuracy (\%).
            \textbf{Input:} Swahili Texts + Images; \textbf{Output:} En, Sw responses.
            \colorbox{orange!20}{Shaded columns represent \{Sw\} in input and \{Sw\} in output VLM results.}
            En $\Delta$Acc. = \{En score from Table \ref{tab:ZerooneshotAccuracyEn} (En-Input) -- En score from this Table (Sw-Input)\}.
            Sw $\Delta$Acc. = \{Sw score from Table \ref{tab:ZerooneshotAccuracyEn} (En-Input) -- Sw score from this Table (Sw-Input).\}
            \textcolor{blue}{Positive} $\Delta$ in blue, \textcolor{red}{negative} $\Delta$ in red.
            }
    \resizebox{\textwidth}{!}{
    \begin{tabular}{r|| c c >{\columncolor{apricot}}c >{\columncolor{apricot}}c | c c >{\columncolor{apricot}}c >{\columncolor{apricot}}c | c c >{\columncolor{apricot}}c >{\columncolor{apricot}}c | c c >{\columncolor{apricot}}c >{\columncolor{apricot}}c | c c >{\columncolor{apricot}}c >{\columncolor{apricot}}c || c c >{\columncolor{apricot}}c >{\columncolor{apricot}}c | c c >{\columncolor{apricot}}c >{\columncolor{apricot}}c | c c >{\columncolor{apricot}}c >{\columncolor{apricot}}c }
        \toprule
        \multirow{2}{*}{\textbf{Model}} & \multicolumn{4}{c|}{\textbf{Object Recognition}} & \multicolumn{4}{c|}{\textbf{Scene Understanding}} & \multicolumn{4}{c|}{\textbf{Relation Understanding}} & \multicolumn{4}{c|}{\textbf{Semantic Segmentation}} & \multicolumn{4}{c|}{\textbf{Image Captioning}} & \multicolumn{4}{c||}{\textbf{Image-Text Matching}} & \multicolumn{4}{c|}{\textbf{Unrelatedness}} & \multicolumn{4}{c}{\textbf{Visual Question Answering}} \\
        \cmidrule(lr){2-5} \cmidrule(lr){6-9} \cmidrule(lr){10-13} \cmidrule(lr){14-17} \cmidrule(lr){18-21} \cmidrule(lr){22-25} \cmidrule(lr){26-29} \cmidrule(lr){30-33}
        & En & $\Delta_{\text{En}}$ & Sw & $\Delta_{\text{Sw}}$ & En & $\Delta_{\text{En}}$ & Sw & $\Delta_{\text{Sw}}$ & En & $\Delta_{\text{En}}$ & Sw & $\Delta_{\text{Sw}}$ & En & $\Delta_{\text{En}}$ & Sw & $\Delta_{\text{Sw}}$ & En & $\Delta_{\text{En}}$ & Sw & $\Delta_{\text{Sw}}$ & En & $\Delta_{\text{En}}$ & Sw & $\Delta_{\text{Sw}}$ & En & $\Delta_{\text{En}}$ & Sw & $\Delta_{\text{Sw}}$ & En & $\Delta_{\text{En}}$ & Sw & $\Delta_{\text{Sw}}$ \\
        \midrule
        \rowcolor{gray!20}
        \multicolumn{33}{c}{\textcolor{violet}{\textbf{Zero-shot, \textit{Without Rationales}}}} \\
        \midrule
        GPT-4o                 & 84.9 & \textcolor{red}{-4.5} & 84.3 & \textcolor{red}{-2.1} & 84.3 & \textcolor{red}{-4.7} & 83.8 & \textcolor{red}{-4.2} & 83.9 & \textcolor{red}{-4.6} & 79.7 & \textcolor{red}{-7.0} & 80.6 & \textcolor{red}{-5.1} & 76.5 & \textcolor{red}{-6.9} & 81.0 & \textcolor{red}{-5.3} & 74.8 & \textcolor{red}{-8.9} & 85.4 & \textcolor{red}{-5.0} & 84.8 & \textcolor{red}{-4.4} & 86.8 & \textcolor{red}{-4.8} & 86.6 & \textcolor{red}{-4.9} & 84.2 & \textcolor{red}{-4.2} & 85.1 & \textcolor{red}{-2.0} \\
        GPT-4o-mini            & 75.1 & \textcolor{red}{-4.5} & 78.6 & \textcolor{blue}{0.6} & 75.9 & \textcolor{red}{-5.2} & 75.6 & \textcolor{red}{-3.2} & 74.5 & \textcolor{red}{-5.1} & 71.0 & \textcolor{red}{-5.2} & 73.0 & \textcolor{red}{-5.1} & 68.6 & \textcolor{red}{-5.4} & 73.1 & \textcolor{red}{-4.2} & 67.6 & \textcolor{red}{-6.1} & 73.6 & \textcolor{red}{-5.0} & 73.1 & \textcolor{red}{-7.2} & 78.8 & \textcolor{red}{-5.2} & 77.7 & \textcolor{red}{-4.1} & 76.6 & \textcolor{red}{-5.0} & 77.1 & \textcolor{blue}{0.8} \\
        Gemini 2.0 Flash Lite  & 79.5 & \textcolor{red}{-5.2} & 82.3 & \textcolor{blue}{0.5} & 79.8 & \textcolor{red}{-5.4} & 79.3 & \textcolor{red}{-3.1} & 78.2 & \textcolor{red}{-5.1} & 74.7 & \textcolor{red}{-5.0} & 76.6 & \textcolor{red}{-5.0} & 72.3 & \textcolor{red}{-5.0} & 76.7 & \textcolor{red}{-4.2} & 71.3 & \textcolor{red}{-5.3} & 77.6 & \textcolor{red}{-5.3} & 76.8 & \textcolor{red}{-7.1} & 82.3 & \textcolor{red}{-5.0} & 81.4 & \textcolor{red}{-4.2} & 79.8 & \textcolor{red}{-4.5} & 80.8 & \textcolor{blue}{0.5} \\
        Gemini 1.5 Flash 8B    & 74.3 & \textcolor{red}{-5.4} & 76.9 & \textcolor{blue}{0.5} & 74.2 & \textcolor{red}{-5.2} & 73.9 & \textcolor{red}{-3.1} & 72.9 & \textcolor{red}{-5.2} & 69.3 & \textcolor{red}{-5.3} & 71.3 & \textcolor{red}{-5.1} & 66.9 & \textcolor{red}{-5.3} & 72.2 & \textcolor{red}{-5.0} & 65.9 & \textcolor{red}{-6.1} & 71.2 & \textcolor{red}{-4.3} & 71.4 & \textcolor{red}{-7.0} & 77.4 & \textcolor{red}{-5.5} & 76.0 & \textcolor{red}{-4.0} & 75.0 & \textcolor{red}{-5.1} & 75.4 & \textcolor{blue}{0.6} \\
        \midrule
        \rowcolor{gray!20}
        \multicolumn{33}{c}{\textcolor{violet}{\textbf{Zero-shot, \textit{With Rationales}}}} \\
        \midrule
        GPT-4o                & 84.8 & \textcolor{red}{-5.1} & 82.3 & \textcolor{red}{-5.4} & 85.3 & \textcolor{red}{-5.0} & 84.1 & \textcolor{red}{-2.8} & 83.4 & \textcolor{red}{-4.8} & 81.7 & \textcolor{red}{-4.5} & 83.5 & \textcolor{red}{-4.6} & 82.0 & \textcolor{red}{-4.7} & 82.1 & \textcolor{red}{-5.0} & 78.0 & \textcolor{red}{-5.4} & 85.1 & \textcolor{red}{-5.0} & 82.5 & \textcolor{red}{-5.0} & 86.5 & \textcolor{red}{-4.6} & 86.5 & \textcolor{red}{-4.2} & 83.0 & \textcolor{red}{-4.2} & 82.5 & \textcolor{red}{-3.6} \\
        GPT-4o-mini           & 75.5 & \textcolor{red}{-4.9} & 78.8 & \textcolor{blue}{1.5} & 75.3 & \textcolor{red}{-4.9} & 75.7 & \textcolor{red}{-3.3} & 75.3 & \textcolor{red}{-5.1} & 75.6 & \textcolor{red}{-3.1} & 75.0 & \textcolor{red}{-4.4} & 76.5 & \textcolor{red}{-0.7} & 73.2 & \textcolor{red}{-4.4} & 71.2 & \textcolor{red}{-0.6} & 75.0 & \textcolor{red}{-5.0} & 73.8 & \textcolor{red}{-2.3} & 78.7 & \textcolor{red}{-5.0} & 77.9 & \textcolor{red}{-3.3} & 71.2 & \textcolor{red}{-4.6} & 72.4 & \textcolor{red}{-4.5} \\
        Gemini 2.0 Flash Lite & 79.1 & \textcolor{red}{-4.8} & 82.5 & \textcolor{blue}{1.2} & 79.1 & \textcolor{red}{-5.0} & 79.4 & \textcolor{red}{-3.0} & 78.9 & \textcolor{red}{-5.0} & 79.4 & \textcolor{red}{-2.4} & 78.6 & \textcolor{red}{-4.3} & 80.2 & \textcolor{red}{-0.5} & 77.4 & \textcolor{red}{-4.9} & 74.8 & \textcolor{red}{-0.8} & 77.8 & \textcolor{red}{-4.2} & 77.5 & \textcolor{red}{-2.0} & 82.4 & \textcolor{red}{-5.0} & 81.6 & \textcolor{red}{-3.4} & 75.4 & \textcolor{red}{-5.1} & 76.1 & \textcolor{red}{-5.0} \\
        Gemini 1.5 Flash 8B   & 73.9 & \textcolor{red}{-5.0} & 77.1 & \textcolor{blue}{1.1} & 73.7 & \textcolor{red}{-5.0} & 74.0 & \textcolor{red}{-3.0} & 72.7 & \textcolor{red}{-4.2} & 74.2 & \textcolor{red}{-2.3} & 74.1 & \textcolor{red}{-5.2} & 74.8 & \textcolor{red}{-0.8} & 72.2 & \textcolor{red}{-5.1} & 69.5 & \textcolor{red}{-0.6} & 73.5 & \textcolor{red}{-5.2} & 72.1 & \textcolor{red}{-1.2} & 76.4 & \textcolor{red}{-4.4} & 76.2 & \textcolor{red}{-4.0} & 67.6 & \textcolor{red}{-2.7} & 70.7 & \textcolor{red}{-3.6} \\
        \midrule
        \rowcolor{gray!20}
        \multicolumn{33}{c}{\textcolor{teal}{\textbf{One-shot, \textit{Without Rationales}}}} \\
        \midrule
        GPT-4o                & 86.2 & \textcolor{red}{-5.1} & 83.0 & \textcolor{red}{-4.6} & 85.1 & \textcolor{red}{-4.0} & 83.9 & \textcolor{red}{-5.0} & 84.4 & \textcolor{red}{-4.4} & 83.2 & \textcolor{red}{-3.6} & 84.2 & \textcolor{red}{-4.5} & 81.1 & \textcolor{red}{-5.2} & 83.7 & \textcolor{red}{-5.1} & 74.7 & \textcolor{red}{-10.4} & 84.1 & \textcolor{red}{-4.4} & 82.6 & \textcolor{red}{-4.4} & 86.4 & \textcolor{red}{-4.8} & 86.3 & \textcolor{red}{-4.3} & 83.8 & \textcolor{red}{-4.0} & 82.8 & \textcolor{red}{-5.6} \\
        GPT-4o-mini           & 77.1 & \textcolor{red}{-4.5} & 75.0 & \textcolor{red}{-4.4} & 77.0 & \textcolor{red}{-4.3} & 75.0 & \textcolor{red}{-4.4} & 74.4 & \textcolor{red}{-4.2} & 73.9 & \textcolor{red}{-3.5} & 76.4 & \textcolor{red}{-4.3} & 73.7 & \textcolor{red}{-4.0} & 75.4 & \textcolor{red}{-4.8} & 68.8 & \textcolor{red}{-6.0} & 76.2 & \textcolor{red}{-5.4} & 72.7 & \textcolor{red}{-5.4} & 78.7 & \textcolor{red}{-5.1} & 78.3 & \textcolor{red}{-4.7} & 76.6 & \textcolor{red}{-4.3} & 74.7 & \textcolor{red}{-3.3} \\
        Gemini 2.0 Flash Lite & 81.1 & \textcolor{red}{-4.8} & 78.7 & \textcolor{red}{-4.2} & 80.7 & \textcolor{red}{-4.3} & 78.7 & \textcolor{red}{-4.2} & 78.8 & \textcolor{red}{-4.9} & 77.6 & \textcolor{red}{-3.1} & 80.8 & \textcolor{red}{-5.0} & 77.4 & \textcolor{red}{-4.0} & 78.8 & \textcolor{red}{-4.5} & 72.5 & \textcolor{red}{-6.6} & 79.5 & \textcolor{red}{-5.0} & 76.4 & \textcolor{red}{-5.0} & 81.7 & \textcolor{red}{-4.4} & 82.0 & \textcolor{red}{-4.6} & 80.4 & \textcolor{red}{-4.4} & 78.4 & \textcolor{red}{-3.2} \\
        Gemini 1.5 Flash 8B   & 75.9 & \textcolor{red}{-5.0} & 73.3 & \textcolor{red}{-3.9} & 76.3 & \textcolor{red}{-5.3} & 73.3 & \textcolor{red}{-4.4} & 73.5 & \textcolor{red}{-5.0} & 72.3 & \textcolor{red}{-3.2} & 75.4 & \textcolor{red}{-5.0} & 72.0 & \textcolor{red}{-4.1} & 73.3 & \textcolor{red}{-4.4} & 67.1 & \textcolor{red}{-5.5} & 73.9 & \textcolor{red}{-4.8} & 71.0 & \textcolor{red}{-5.5} & 76.9 & \textcolor{red}{-5.0} & 76.6 & \textcolor{red}{-3.8} & 75.8 & \textcolor{red}{-5.2} & 73.0 & \textcolor{red}{-5.9} \\
        \midrule
        \rowcolor{gray!20}
        \multicolumn{33}{c}{\textcolor{teal}{\textbf{One-shot, \textit{With Rationales}}}} \\
        \midrule
        GPT-4o                & \textbf{86.4} & \textcolor{red}{-5.0} & \textbf{85.6} & \textcolor{red}{-5.0} & \textbf{85.5} & \textcolor{red}{-4.2} & \textbf{84.6} & \textcolor{red}{-4.5} & \textbf{85.0} & \textcolor{red}{-4.0} & \textbf{84.9} & \textcolor{red}{-4.0} & \textbf{86.4} & \textcolor{red}{-5.0} & \textbf{84.9} & \textcolor{red}{-5.4} & \textbf{84.2} & \textcolor{red}{-5.1} & \textbf{80.7} & \textcolor{red}{-5.0} & \textbf{85.4} & \textcolor{red}{-4.5} & \textbf{84.1} & \textcolor{red}{-4.5} & \textbf{87.3} & \textcolor{red}{-5.0} & \textbf{87.1} & \textcolor{red}{-4.3} & \textbf{86.4} & \textcolor{red}{-5.0} & \textbf{86.5} & \textcolor{red}{-3.0} \\
        GPT-4o-mini           & 78.1 & \textcolor{red}{-5.0} & 79.1 & \textcolor{red}{-2.0} & 78.1 & \textcolor{red}{-5.2} & 76.0 & \textcolor{red}{-4.0} & 74.8 & \textcolor{red}{-2.6} & 77.6 & \textcolor{red}{-2.0} & 77.1 & \textcolor{red}{-3.9} & 76.3 & \textcolor{red}{-2.8} & 75.6 & \textcolor{red}{-4.7} & 72.4 & \textcolor{red}{-3.8} & 77.4 & \textcolor{red}{-4.9} & 72.6 & \textcolor{red}{-5.8} & 78.9 & \textcolor{red}{-5.0} & 76.9 & \textcolor{red}{-5.4} & 78.2 & \textcolor{red}{-5.0} & 75.6 & \textcolor{red}{-4.0} \\
        Gemini 2.0 Flash Lite & 81.1 & \textcolor{red}{-4.3} & 82.8 & \textcolor{red}{-2.1} & 81.7 & \textcolor{red}{-5.2} & 79.6 & \textcolor{red}{-3.5} & 78.6 & \textcolor{red}{-2.7} & 81.3 & \textcolor{red}{-2.3} & 80.7 & \textcolor{red}{-3.8} & 80.0 & \textcolor{red}{-3.6} & 79.0 & \textcolor{red}{-4.4} & 76.1 & \textcolor{red}{-3.5} & 81.7 & \textcolor{red}{-5.5} & 76.3 & \textcolor{red}{-5.8} & 82.6 & \textcolor{red}{-5.0} & 80.6 & \textcolor{red}{-4.8} & 81.8 & \textcolor{red}{-4.9} & 79.3 & \textcolor{red}{-3.4} \\
        Gemini 1.5 Flash 8B   & 76.3 & \textcolor{red}{-4.9} & 77.4 & \textcolor{red}{-2.0} & 76.1 & \textcolor{red}{-5.0} & 74.3 & \textcolor{red}{3.4} & 75.5 & \textcolor{red}{-5.0} & 75.9 & \textcolor{red}{-2.4} & 77.2 & \textcolor{red}{-5.7} & 74.6 & \textcolor{red}{-4.0} & 73.9 & \textcolor{red}{-4.7} & 70.7 & \textcolor{red}{-3.9} & 75.4 & \textcolor{red}{-4.6} & 70.9 & \textcolor{red}{-5.8} & 77.2 & \textcolor{red}{-5.0} & 75.2 & \textcolor{red}{-5.0} & 76.8 & \textcolor{red}{-5.3} & 73.9 & \textcolor{red}{-3.7} \\
        \midrule
        \midrule
        \textbf{Human Performance} & -- & -- & 95.5 & -- & -- & -- & 95.1 & -- & -- & -- & 94.7 & -- & -- & -- & 94.1 & -- & -- & -- & 95.7 & -- & -- & -- & 93.4 & -- & -- & -- & 89.9 & -- & -- & -- & 80.8 & -- \\
        \bottomrule
    \end{tabular}
    }
    \label{tab:ZerooneshotAccuracySw}
\end{table*}

\clearpage
\subsection{Experiments with \emph{VLURes} Urdu Data}
\label{Appendix_ExperimentsUrduData}
\begin{table*}[h] 
    \centering
    \caption{Performance of VLMs on \textbf{eight VL tasks} under \colorbox{violet!20}{zero-shot} and \colorbox{teal!20}{one-shot} settings, measured by Accuracy (\%).
            \textbf{Input:} Urdu Texts + Images; \textbf{Output:} En, Ur responses.
            \colorbox{orange!20}{Shaded columns represent \{Ur\} in input and \{Ur\} in output VLM results.}
            En $\Delta$Acc. = \{En score from Table \ref{tab:ZerooneshotAccuracyEn} (En-Input) -- En score from this Table (Ur-Input)\}.
            Ur $\Delta$Acc. = \{Ur score from Table \ref{tab:ZerooneshotAccuracyEn} (En-Input) -- Ur score from this Table (Ur-Input).\}
            \textcolor{blue}{Positive} $\Delta$ in blue, \textcolor{red}{negative} $\Delta$ in red.
            We exclude Qwen2VL 7B, and LlaVa Mistral 7B/13B from experiments due to limited Urdu support, from our initial experiments.
            }
    \resizebox{\textwidth}{!}{
    \begin{tabular}{r|| c c >{\columncolor{apricot}}c >{\columncolor{apricot}}c | c c >{\columncolor{apricot}}c >{\columncolor{apricot}}c | c c >{\columncolor{apricot}}c >{\columncolor{apricot}}c | c c >{\columncolor{apricot}}c >{\columncolor{apricot}}c | c c >{\columncolor{apricot}}c >{\columncolor{apricot}}c || c c >{\columncolor{apricot}}c >{\columncolor{apricot}}c | c c >{\columncolor{apricot}}c >{\columncolor{apricot}}c | c c >{\columncolor{apricot}}c >{\columncolor{apricot}}c }
        \toprule
        \multirow{2}{*}{\textbf{Model}} & \multicolumn{4}{c|}{\textbf{Object Recognition}} & \multicolumn{4}{c|}{\textbf{Scene Understanding}} & \multicolumn{4}{c|}{\textbf{Relation Understanding}} & \multicolumn{4}{c|}{\textbf{Semantic Segmentation}} & \multicolumn{4}{c|}{\textbf{Image Captioning}} & \multicolumn{4}{c||}{\textbf{Image-Text Matching}} & \multicolumn{4}{c|}{\textbf{Unrelatedness}} & \multicolumn{4}{c}{\textbf{Visual Question Answering}} \\
        \cmidrule(lr){2-5} \cmidrule(lr){6-9} \cmidrule(lr){10-13} \cmidrule(lr){14-17} \cmidrule(lr){18-21} \cmidrule(lr){22-25} \cmidrule(lr){26-29} \cmidrule(lr){30-33}
        & En & $\Delta_{\text{En}}$ & Ur & $\Delta_{\text{Ur}}$ & En & $\Delta_{\text{En}}$ & Ur & $\Delta_{\text{Ur}}$ & En & $\Delta_{\text{En}}$ & Ur & $\Delta_{\text{Ur}}$ & En & $\Delta_{\text{En}}$ & Ur & $\Delta_{\text{Ur}}$ & En & $\Delta_{\text{En}}$ & Ur & $\Delta_{\text{Ur}}$ & En & $\Delta_{\text{En}}$ & Ur & $\Delta_{\text{Ur}}$ & En & $\Delta_{\text{En}}$ & Ur & $\Delta_{\text{Ur}}$ & En & $\Delta_{\text{En}}$ & Ur & $\Delta_{\text{Ur}}$ \\
        \midrule
        \rowcolor{gray!20}
        \multicolumn{33}{c}{ \textcolor{violet}{\textbf{Zero-shot, \textit{Without Rationales}}}} \\
        \midrule
        GPT-4o                 & 88.9 & \textcolor{red}{-1.5} & 87.8 & 0.0 & 87.6 & \textcolor{red}{-1.0} & 87.3 & \textcolor{red}{-0.6} & 87.6 & \textcolor{red}{-1.1} & 83.2 & \textcolor{blue}{0.9} & 83.3 & \textcolor{red}{-0.8} & 80.0 & \textcolor{red}{-0.5} & 84.5 & \textcolor{red}{-1.8}  & 78.3 & \textcolor{red}{-0.5} & 89.1 & \textcolor{red}{-1.7} & \textbf{88.3} & \textcolor{blue}{2.4} & \textbf{91.0} & \textcolor{red}{-2.0} & 90.1 & \textcolor{red}{-0.5} & 88.6 & \textcolor{red}{-1.6} & 88.6 & \textcolor{blue}{1.6} \\
        GPT-4o-mini            & 79.3 & \textcolor{red}{-1.7} & 82.1 & \textcolor{blue}{2.5} & 79.7 & \textcolor{red}{-2.0} & 79.1 & \textcolor{blue}{0.6} & 78.4 & \textcolor{red}{-2.0} & 74.5 & \textcolor{red}{-1.8} & 76.4 & \textcolor{red}{-1.5} & 72.1 & \textcolor{red}{-2.7} & 77.4 & \textcolor{red}{-1.5} & 71.1 & \textcolor{red}{-1.2} & 76.6 & \textcolor{red}{-1.0} & 76.6 & \textcolor{red}{-3.0} & 82.1 & \textcolor{red}{-1.5} & 81.2 & \textcolor{red}{-0.7} & 80.1 & \textcolor{red}{-1.5} & 80.6 & \textcolor{blue}{3.0} \\
        Gemini 2.0 Flash Lite  & 82.5 & \textcolor{red}{-1.2} & 85.8 & \textcolor{blue}{1.9} & 82.6 & \textcolor{red}{-1.2} & 82.8 & \textcolor{blue}{0.2} & 81.3 & \textcolor{red}{-1.2} & 78.2 & \textcolor{red}{-1.3} & 79.8 & \textcolor{red}{-1.2} & 75.8 & \textcolor{red}{-2.4} & 81.0 & \textcolor{red}{-1.5} & 74.8 & \textcolor{red}{-1.5} & 80.8 & \textcolor{red}{-1.5} & 80.3 & \textcolor{red}{-2.6} & 86.6 & \textcolor{red}{-2.3} & 84.9 & \textcolor{red}{-0.9} & 83.3 & \textcolor{red}{-1.0} & 84.3 & \textcolor{blue}{4.0} \\
        Gemini 1.5 Flash 8B    & 77.5 & \textcolor{red}{-1.6} & 80.4 & \textcolor{blue}{1.5} & 77.4 & \textcolor{red}{-1.4} & 77.4 & \textcolor{blue}{0.6} & 76.0 & \textcolor{red}{-1.3} & 72.8 & \textcolor{red}{-1.7} & 74.1 & \textcolor{red}{-0.9} & 70.4 & \textcolor{red}{-2.8} & 75.8 & \textcolor{red}{-1.6} & 69.4 & \textcolor{red}{-1.7} & 74.6 & \textcolor{red}{-0.7} & 74.9 & \textcolor{red}{-2.6} & 80.9 & \textcolor{red}{-2.0} & 79.5 & \textcolor{red}{-0.9} & 79.0 & \textcolor{red}{-2.1} & 78.9 & \textcolor{blue}{3.5} \\
        PALO 7B                & 38.6 & \textcolor{red}{-2.1} & 44.3 & \textcolor{red}{-5.9} & 38.6 & \textcolor{red}{-2.0} & 41.3 & \textcolor{red}{-8.2} & 37.4 & \textcolor{red}{-2.1} & 36.7 & \textcolor{red}{-8.8} & 35.8 & \textcolor{red}{-2.0} & 34.3 & \textcolor{red}{-9.8} & 36.6 & \textcolor{red}{-1.8} & 33.3 & \textcolor{red}{-8.9} & 36.3 & \textcolor{red}{-1.8} & 38.8 & \textcolor{red}{-11.1} & 41.6 & \textcolor{red}{-2.1} & 43.4 & \textcolor{red}{-9.0} & 39.8 & \textcolor{red}{-2.3} & 42.8 & \textcolor{red}{-3.6} \\
        MAYA 8B                & 40.2 & \textcolor{red}{-2.4} & 45.1 & \textcolor{red}{-5.1} & 39.9 & \textcolor{red}{-2.0} & 42.1 & \textcolor{red}{-5.9} & 38.6 & \textcolor{red}{-2.0} & 37.5 & \textcolor{red}{-7.9} & 37.4 & \textcolor{red}{-2.3} & 35.1 & \textcolor{red}{-8.9} & 38.2 & \textcolor{red}{-2.3} & 34.1 & \textcolor{red}{-8.5} & 38.3 & \textcolor{red}{-2.5} & 39.6 & \textcolor{red}{-10.2} & 43.2 & \textcolor{red}{-2.4} & 44.2 & \textcolor{red}{-6.7} & 40.4 & \textcolor{red}{-1.6} & 43.6 & \textcolor{red}{-3.6} \\
        PALO 13B               & 40.2 & \textcolor{red}{-1.5} & 42.4 & \textcolor{red}{-13.8} & 40.6 & \textcolor{red}{-2.5} & 39.3 & \textcolor{red}{-16.1} & 38.8 & \textcolor{red}{-2.0} & 34.7 & \textcolor{red}{-17.0} & 37.3 & \textcolor{red}{-2.3} & 32.3 & \textcolor{red}{-17.3} & 38.6 & \textcolor{red}{-2.0} & 31.3 & \textcolor{red}{-16.9} & 38.0 & \textcolor{red}{-2.2} & 36.8 & \textcolor{red}{-18.8} & 43.0 & \textcolor{red}{-2.3} & 41.4 & \textcolor{red}{-18.0} & 40.9 & \textcolor{red}{-2.4} & 40.9 & \textcolor{red}{-12.9} \\
        \midrule
        \rowcolor{gray!20}
        \multicolumn{33}{c}{\textcolor{violet}{\textbf{Zero-shot, \textit{With Rationales}}}} \\
        \midrule
        GPT-4o                & 88.0 & \textcolor{red}{-1.3} & 85.8 & \textcolor{red}{-1.2} & 88.7 & \textcolor{red}{-2.0} & 87.6 & \textcolor{blue}{1.6} & 87.2 & \textcolor{red}{-1.6} & 85.2 & \textcolor{red}{-0.0} & 87.2 & \textcolor{red}{-1.3} & 85.5 & \textcolor{blue}{2.0} & 85.6 & \textcolor{red}{-1.5} & 81.5 & \textcolor{red}{-1.3} & 88.8 & \textcolor{red}{-1.7} & 86.0 & \textcolor{red}{-0.3} & 90.9 & \textcolor{red}{-2.0} & 90.0 & \textcolor{red}{-0.7} & 86.9 & \textcolor{red}{-2.1} & 86.0 & \textcolor{red}{-0.2} \\
        GPT-4o-mini           & 79.1 & \textcolor{red}{-1.5} & 82.3 & \textcolor{blue}{0.3} & 78.9 & \textcolor{red}{-1.5} & 79.2 & \textcolor{blue}{0.2} & 79.1 & \textcolor{red}{-1.9} & 79.1 & \textcolor{blue}{3.2} & 78.5 & \textcolor{red}{-0.9} & 80.0 & \textcolor{blue}{4.2} & 77.4 & \textcolor{red}{-1.6} & 74.7 & \textcolor{blue}{2.3} & 78.4 & \textcolor{red}{-1.4} & 77.3 & \textcolor{blue}{1.3} & 82.3 & \textcolor{red}{-1.6} & 81.4 & \textcolor{red}{-1.4} & 75.3 & \textcolor{red}{-1.7} & 75.9 & \textcolor{red}{-2.9} \\
        Gemini 2.0 Flash Lite & 82.9 & \textcolor{red}{-1.6} & 86.0 & \textcolor{blue}{1.4} & 82.8 & \textcolor{red}{-1.7} & 82.9 & \textcolor{blue}{1.2} & 82.4 & \textcolor{red}{-1.5} & 82.9 & \textcolor{blue}{3.3} & 82.8 & \textcolor{red}{-1.5} & 83.7 & \textcolor{blue}{3.8} & 81.1 & \textcolor{red}{-1.6} & 78.3 & \textcolor{blue}{1.5} & 81.7 & \textcolor{red}{-1.1} & 81.0 & \textcolor{blue}{1.5} & 86.0 & \textcolor{red}{-1.6} & 85.1 & \textcolor{red}{-1.7} & 79.0 & \textcolor{red}{-1.7} & 79.6 & \textcolor{red}{-2.7} \\
        Gemini 1.5 Flash 8B   & 77.6 & \textcolor{red}{-1.7} & 80.6 & \textcolor{blue}{1.0} & 77.2 & \textcolor{red}{-1.5} & 77.5 & \textcolor{blue}{1.2} & 77.5 & \textcolor{red}{-2.0} & 77.7 & \textcolor{blue}{4.0} & 78.0 & \textcolor{red}{-2.1} & 78.3 & \textcolor{blue}{5.0} & 76.1 & \textcolor{red}{-2.0} & 73.0 & \textcolor{blue}{2.3} & 77.3 & \textcolor{red}{-2.0} & 75.6 & \textcolor{blue}{1.5} & 80.5 & \textcolor{red}{-1.5} & 79.7 & \textcolor{red}{-1.3} & 73.5 & \textcolor{red}{-1.6} & 74.2 & \textcolor{red}{-2.7} \\
        PALO 7B               & 39.0 & \textcolor{red}{-2.5} & 45.0 & \textcolor{red}{-6.4} & 38.4 & \textcolor{red}{-2.1} & 40.7 & \textcolor{red}{-6.8} & 38.4 & \textcolor{red}{-2.3} & 41.4 & \textcolor{red}{-4.7} & 38.5 & \textcolor{red}{-2.0} & 42.2 & \textcolor{red}{-3.1} & 36.8 & \textcolor{red}{-2.0} & 36.8 & \textcolor{red}{-5.5} & 37.9 & \textcolor{red}{-2.0} & 39.5 & \textcolor{red}{-6.3} & 42.1 & \textcolor{red}{-2.5} & 43.6 & \textcolor{red}{-9.2} & 35.3 & \textcolor{red}{-2.8} & 38.1 & \textcolor{red}{-9.7} \\
        MAYA 8B               & 39.0 & \textcolor{red}{-1.2} & 45.3 & \textcolor{red}{-6.0} & 39.4 & \textcolor{red}{-1.8} & 42.2 & \textcolor{red}{-6.2} & 39.2 & \textcolor{red}{-1.8} & 42.0 & \textcolor{red}{-3.8} & 39.8 & \textcolor{red}{-2.0} & 43.0 & \textcolor{red}{-2.4} & 37.9 & \textcolor{red}{-1.9} & 37.7 & \textcolor{red}{-5.0} & 39.5 & \textcolor{red}{-2.3} & 40.3 & \textcolor{red}{-4.9} & 42.9 & \textcolor{red}{-2.0} & 44.4 & \textcolor{red}{-8.1} & 35.8 & \textcolor{red}{-2.0} & 38.9 & \textcolor{red}{-8.6} \\
        PALO 13B              & 40.3 & \textcolor{red}{-2.3} & 42.5 & \textcolor{red}{-15.9} & 39.8 & \textcolor{red}{-2.0} & 39.2 & \textcolor{red}{-15.8} & 39.6 & \textcolor{red}{-2.0} & 39.2 & \textcolor{red}{-13.8} & 40.9 & \textcolor{red}{-2.5} & 40.0 & \textcolor{red}{-10.8} & 38.6 & \textcolor{red}{-2.3} & 34.9 & \textcolor{red}{-13.7} & 39.4 & \textcolor{red}{-2.0} & 38.0 & \textcolor{red}{-14.9} & 43.4 & \textcolor{red}{-2.1} & 42.1 & \textcolor{red}{-16.9} & 35.3 & \textcolor{red}{-1.0} & 36.2 & \textcolor{red}{-19.0} \\
        \midrule
        \rowcolor{gray!20}
        \multicolumn{33}{c}{\textcolor{teal}{\textbf{One-shot, \textit{Without Rationales}}}} \\
        \midrule
        GPT-4o                & \textbf{89.5} & \textcolor{red}{-1.4} & 86.5 & \textcolor{red}{-0.6} & 89.8 & \textcolor{red}{-1.7} & 87.4 & \textcolor{red}{-0.5} & 88.6 & \textcolor{red}{-1.6} & 86.7 & \textcolor{blue}{2.7} & 87.8 & \textcolor{red}{-1.1} & 84.6 & \textcolor{red}{-0.5} & 86.8 & \textcolor{red}{-1.2} & 78.2 & \textcolor{red}{-2.1} & 88.7 & \textcolor{red}{-2.0} & 86.1 & \textcolor{blue}{0.5} & 89.7 & \textcolor{red}{-1.1} & 89.8 & \textcolor{red}{-0.7} & 88.3 & \textcolor{red}{-1.5} & 86.3 & \textcolor{blue}{0.4} \\
        GPT-4o-mini           & 80.6 & \textcolor{red}{-1.0} & 78.5 & \textcolor{blue}{0.5} & 81.3 & \textcolor{red}{-1.6} & 78.5 & \textcolor{red}{-0.6} & 78.2 & \textcolor{red}{-1.0} & 77.4 & \textcolor{blue}{1.8} & 80.1 & \textcolor{red}{-1.0} & 77.2 & \textcolor{blue}{1.2} & 78.6 & \textcolor{red}{-1.0} & 72.3 & \textcolor{red}{-1.3} & 79.2 & \textcolor{red}{-1.4} & 76.2 & \textcolor{blue}{1.4} & 82.0 & \textcolor{red}{-1.4} & 81.8 & \textcolor{red}{-0.6} & 80.3 & \textcolor{red}{-1.0} & 78.2 & \textcolor{blue}{1.8} \\
        Gemini 2.0 Flash Lite & 85.3 & \textcolor{red}{-2.0} & 82.2 & \textcolor{blue}{0.7} & 84.7 & \textcolor{red}{-1.3} & 82.2 & \textcolor{red}{-0.5} & 81.9 & \textcolor{red}{-1.0} & 81.1 & \textcolor{blue}{2.0} & 84.4 & \textcolor{red}{-1.6} & 80.9 & \textcolor{blue}{2.1} & 82.8 & \textcolor{red}{-1.5} & 76.0 & \textcolor{red}{-1.9} & 82.9 & \textcolor{red}{-1.4} & 79.9 & \textcolor{blue}{1.7} & 85.3 & \textcolor{red}{-1.0} & 85.5 & \textcolor{red}{-0.8} & 84.6 & \textcolor{red}{-1.6} & 81.9 & \textcolor{blue}{1.6} \\
        Gemini 1.5 Flash 8B   & 79.6 & \textcolor{red}{-1.7} & 76.8 & \textcolor{blue}{0.8} & 79.6 & \textcolor{red}{-1.6} & 76.8 & \textcolor{red}{-0.5} & 76.2 & \textcolor{red}{-0.7} & 75.8 & \textcolor{blue}{2.0} & 78.3 & \textcolor{red}{-0.9} & 75.5 & \textcolor{blue}{2.0} & 77.6 & \textcolor{red}{-1.7} & 70.6 & \textcolor{red}{-1.3} & 77.7 & \textcolor{red}{-1.6} & 74.5 & \textcolor{blue}{1.2} & 79.6 & \textcolor{red}{-0.7} & 80.1 & \textcolor{red}{-0.6} & 79.3 & \textcolor{red}{-1.7} & 76.5 & \textcolor{blue}{1.9} \\
        PALO 7B               & 40.5 & \textcolor{red}{-2.0} & 41.2 & \textcolor{red}{-6.7} & 40.6 & \textcolor{red}{-2.0} & 40.7 & \textcolor{red}{-8.0} & 38.1 & \textcolor{red}{-2.0} & 39.5 & \textcolor{red}{-6.2} & 40.3 & \textcolor{red}{-2.3} & 39.2 & \textcolor{red}{-6.0} & 38.3 & \textcolor{red}{-1.8} & 34.9 & \textcolor{red}{-6.4} & 38.7 & \textcolor{red}{-2.0} & 38.3 & \textcolor{red}{-10.9} & 42.0 & \textcolor{red}{-2.5} & 44.0 & \textcolor{red}{-8.4} & 39.7 & \textcolor{red}{-1.5} & 40.4 & \textcolor{red}{-7.1} \\
        MAYA 8B               & 41.1 & \textcolor{red}{-1.3} & 41.5 & \textcolor{red}{-6.0} & 42.4 & \textcolor{red}{-2.5} & 41.5 & \textcolor{red}{-5.7} & 39.8 & \textcolor{red}{-2.4} & 40.3 & \textcolor{red}{-5.2} & 41.7 & \textcolor{red}{-2.4} & 40.3 & \textcolor{red}{-4.7} & 40.0 & \textcolor{red}{-2.2} & 35.3 & \textcolor{red}{-5.7} & 39.8 & \textcolor{red}{-1.8} & 39.2 & \textcolor{red}{-10.2} & 42.1 & \textcolor{red}{-1.3} & 44.8 & \textcolor{red}{-7.2} & 41.3 & \textcolor{red}{-1.8} & 41.2 & \textcolor{red}{-5.2} \\
        PALO 13B              & 41.7 & \textcolor{red}{-1.2} & 38.7 & \textcolor{red}{-16.2} & 42.1 & \textcolor{red}{-1.8} & 38.7 & \textcolor{red}{-15.8} & 38.5 & \textcolor{red}{-0.9} & 37.6 & \textcolor{red}{-14.6} & 41.3 & \textcolor{red}{-1.9} & 37.0 & \textcolor{red}{-14.9} & 40.7 & \textcolor{red}{-2.6} & 32.5 & \textcolor{red}{-17.6} & 40.2 & \textcolor{red}{-2.0} & 36.1 & \textcolor{red}{-16.0} & 42.8 & \textcolor{red}{-1.2} & 42.2 & \textcolor{red}{-16.5} & 41.2 & \textcolor{red}{-1.8} & 38.2 & \textcolor{red}{-16.0} \\
        \midrule
        \rowcolor{gray!20}
        \multicolumn{33}{c}{ \textcolor{teal}{\textbf{One-shot, \textit{With Rationales}}}} \\
        \midrule
        GPT-4o                & 89.4 & \textcolor{red}{-1.0} & \textbf{89.1} & \textcolor{blue}{1.3} & \textbf{90.0} & \textcolor{red}{-1.7} & \textbf{88.1} & \textcolor{blue}{0.6} & \textbf{90.1} & \textcolor{red}{-2.1} & \textbf{88.4} & \textcolor{blue}{1.8} & \textbf{89.9} & \textcolor{red}{-1.5} & \textbf{88.4} & \textcolor{blue}{2.3} & \textbf{87.2} & \textcolor{red}{-1.1} & \textbf{84.2} & \textcolor{blue}{0.8} & \textbf{88.6} & \textcolor{red}{-0.7} & \textbf{87.6} & \textcolor{red}{-0.7} & 90.4 & \textcolor{red}{-1.1} & \textbf{90.6} & \textcolor{red}{-0.5} & \textbf{90.3} & \textcolor{red}{-1.9} & \textbf{90.0} & \textcolor{blue}{2.6} \\
        GPT-4o-mini           & 81.6 & \textcolor{red}{-1.5} & 82.6 & \textcolor{blue}{2.3} & 81.6 & \textcolor{red}{-1.7} & 79.5 & \textcolor{blue}{1.7} & 80.7 & \textcolor{red}{-1.5} & 81.1 & \textcolor{blue}{2.8} & 81.7 & \textcolor{red}{-1.5} & 79.8 & \textcolor{blue}{2.3} & 78.8 & \textcolor{red}{-0.9} & 75.9 & \textcolor{blue}{1.9} & 80.5 & \textcolor{red}{-1.0} & 76.1 & \textcolor{red}{-3.0} & 81.8 & \textcolor{red}{-0.9} & 80.4 & \textcolor{red}{-2.5} & 82.2 & \textcolor{red}{-2.0} & 79.1 & \textcolor{red}{-0.7} \\
        Gemini 2.0 Flash Lite & 85.4 & \textcolor{red}{-1.6} & 86.3 & \textcolor{blue}{2.2} & 84.5 & \textcolor{red}{-1.0} & 83.1 & \textcolor{blue}{1.7} & 83.9 & \textcolor{red}{-1.0} & 84.8 & \textcolor{blue}{3.1} & 84.8 & \textcolor{red}{-0.9} & 83.5 & \textcolor{blue}{2.3} & 82.3 & \textcolor{red}{-0.7} & 79.6 & \textcolor{blue}{2.0} & 84.8 & \textcolor{red}{-1.6} & 79.8 & \textcolor{red}{-3.5} & 86.5 & \textcolor{red}{-1.9} & 84.1 & \textcolor{red}{-2.1} & 84.8 & \textcolor{red}{-0.9} & 82.8 & \textcolor{red}{-0.5} \\
        Gemini 1.5 Flash 8B   & 79.9 & \textcolor{red}{-1.5} & 80.9 & \textcolor{blue}{3.2} & 80.1 & \textcolor{red}{-2.0} & 77.8 & \textcolor{blue}{1.7} & 78.9 & \textcolor{red}{-1.4} & 79.4 & \textcolor{blue}{2.4} & 80.5 & \textcolor{red}{-2.0} & 78.1 & \textcolor{blue}{2.7} & 77.1 & \textcolor{red}{-0.9} & 74.2 & \textcolor{blue}{2.0} & 79.8 & \textcolor{red}{-2.0} & 74.4 & \textcolor{red}{-2.9} & 81.3 & \textcolor{red}{-2.1} & 78.7 & \textcolor{red}{-1.8} & 79.3 & \textcolor{red}{-0.8} & 77.4 & \textcolor{red}{-0.7} \\
        PALO 7B               & 41.3 & \textcolor{red}{-2.1} & 45.3 & \textcolor{red}{-4.8} & 40.7 & \textcolor{red}{-2.0} & 41.6 & \textcolor{red}{-6.7} & 40.2 & \textcolor{red}{-2.1} & 42.7 & \textcolor{red}{-6.0} & 41.4 & \textcolor{red}{-2.3} & 42.0 & \textcolor{red}{-7.0} & 38.6 & \textcolor{red}{-1.8} & 38.1 & \textcolor{red}{-6.5} & 40.4 & \textcolor{red}{-2.0} & 38.3 & \textcolor{red}{-10.4} & 41.8 & \textcolor{red}{-2.0} & 42.6 & \textcolor{red}{-9.4} & 41.2 & \textcolor{red}{-2.1} & 41.3 & \textcolor{red}{-7.5} \\
        MAYA 8B               & 42.3 & \textcolor{red}{-2.0} & 46.2 & \textcolor{red}{-4.1} & 42.0 & \textcolor{red}{-2.0} & 42.5 & \textcolor{red}{-4.8} & 41.2 & \textcolor{red}{-1.8} & 44.1 & \textcolor{red}{-4.3} & 42.4 & \textcolor{red}{-2.0} & 42.8 & \textcolor{red}{-4.5} & 40.2 & \textcolor{red}{-2.1} & 38.9 & \textcolor{red}{-4.9} & 42.2 & \textcolor{red}{-2.5} & 39.0 & \textcolor{red}{-8.9} & 42.1 & \textcolor{red}{-1.0} & 43.4 & \textcolor{red}{-7.8} & 41.6 & \textcolor{red}{-1.2} & 42.1 & \textcolor{red}{-6.2} \\
        PALO 13B              & 41.7 & \textcolor{red}{-1.2} & 42.6 & \textcolor{red}{-14.6} & 42.5 & \textcolor{red}{-2.2} & 39.7 & \textcolor{red}{-14.5} & 41.8 & \textcolor{red}{-2.3} & 41.2 & \textcolor{red}{-13.5} & 42.4 & \textcolor{red}{-1.8} & 39.8 & \textcolor{red}{-14.7} & 40.4 & \textcolor{red}{-2.1} & 36.0 & \textcolor{red}{-14.2} & 41.3 & \textcolor{red}{-2.0} & 35.5 & \textcolor{red}{-16.2} & 43.3 & \textcolor{red}{-2.3} & 41.4 & \textcolor{red}{-17.7} & 42.5 & \textcolor{red}{-2.5} & 39.3 & \textcolor{red}{-15.4} \\
        \midrule
        \midrule
        \textbf{Human Performance} & -- & -- & 98.4 & -- & -- & -- & 99.4  & -- & -- & -- & 98.8 & -- & -- & -- & 99.2 & -- & -- & -- & 98.7 & -- & -- & -- & 84.9 & -- & -- & -- & 87.4 & -- & -- & -- & 90.5 & -- \\
        \bottomrule
    \end{tabular}
    }
    \label{tab:ZerooneshotAccuracyUr} 
\end{table*}
\begin{table*}[h] 
    \centering
    \caption{Performance of VLMs on \textbf{eight VL tasks} under \colorbox{blue!20}{finetuning} settings, measured by Accuracy (\%).
            \textbf{Input:} Urdu Texts + Images; \textbf{Output:} En, Ur responses.
            \colorbox{orange!20}{Shaded columns represent \{Ur\} in input and \{Ur\} in output VLM results.}
            En $\Delta$Acc. = \{En score from Table \ref{tab:FinetuneAccuracyEn} (En-Input) -- En score from this Table (Ur-Input)\}.
            Ur $\Delta$Acc. = \{Ur score from Table \ref{tab:FinetuneAccuracyEn} (En-Input) -- Ur score from this Table (Ur-Input).\}
            \textcolor{blue}{Positive} $\Delta$ in blue, \textcolor{red}{negative} $\Delta$ in red.
            We exclude Qwen2VL 7B, and LlaVa Mistral 7B/13B from experiments due to limited Ur support, from our initial experiments.
            }
    \resizebox{\textwidth}{!}{
    \begin{tabular}{r|| c c >{\columncolor{apricot}}c >{\columncolor{apricot}}c | c c >{\columncolor{apricot}}c >{\columncolor{apricot}}c | c c >{\columncolor{apricot}}c >{\columncolor{apricot}}c | c c >{\columncolor{apricot}}c >{\columncolor{apricot}}c | c c >{\columncolor{apricot}}c >{\columncolor{apricot}}c || c c >{\columncolor{apricot}}c >{\columncolor{apricot}}c | c c >{\columncolor{apricot}}c >{\columncolor{apricot}}c | c c >{\columncolor{apricot}}c >{\columncolor{apricot}}c }
        \toprule
        \multirow{2}{*}{\textbf{Model}} & \multicolumn{4}{c|}{\textbf{Object Recognition}} & \multicolumn{4}{c|}{\textbf{Scene Understanding}} & \multicolumn{4}{c|}{\textbf{Relation Understanding}} & \multicolumn{4}{c|}{\textbf{Semantic Segmentation}} & \multicolumn{4}{c|}{\textbf{Image Captioning}} & \multicolumn{4}{c||}{\textbf{Image-Text Matching}} & \multicolumn{4}{c|}{\textbf{Unrelatedness}} & \multicolumn{4}{c}{\textbf{Visual Question Answering}} \\
        \cmidrule(lr){2-5} \cmidrule(lr){6-9} \cmidrule(lr){10-13} \cmidrule(lr){14-17} \cmidrule(lr){18-21} \cmidrule(lr){22-25} \cmidrule(lr){26-29} \cmidrule(lr){30-33}
        & En & $\Delta_{\text{En}}$ & Ur & $\Delta_{\text{Ur}}$ & En & $\Delta_{\text{En}}$ & Ur & $\Delta_{\text{Ur}}$ & En & $\Delta_{\text{En}}$ & Ur & $\Delta_{\text{Ur}}$ & En & $\Delta_{\text{En}}$ & Ur & $\Delta_{\text{Ur}}$ & En & $\Delta_{\text{En}}$ & Ur & $\Delta_{\text{Ur}}$ & En & $\Delta_{\text{En}}$ & Ur & $\Delta_{\text{Ur}}$ & En & $\Delta_{\text{En}}$ & Ur & $\Delta_{\text{Ur}}$ & En & $\Delta_{\text{En}}$ & Ur & $\Delta_{\text{Ur}}$ \\
        \midrule
        \rowcolor{blue!20}
        \multicolumn{33}{c}{  \textcolor{violet}{\textbf{Zero-shot, \textit{Without Rationales}}} } \\
        \midrule
        PALO 7B                & 42.3 & \textcolor{red}{-3.0} & 48.1 & \textcolor{red}{-11.4} & 42.2 & \textcolor{red}{-3.2} & 45.1 & \textcolor{red}{-12.6} & 40.3 & \textcolor{red}{-3.8} & 40.5 & \textcolor{red}{-13.9} & 39.0 & \textcolor{red}{-3.6} & 38.1 & \textcolor{red}{-14.7} & 39.9 & \textcolor{red}{-3.7} & 37.1 & \textcolor{red}{-13.2} & 39.5 & \textcolor{red}{-3.8} & 42.6 & \textcolor{red}{-15.4} & 45.0 & \textcolor{red}{-3.3} & 47.2 & \textcolor{red}{-12.6} & 42.4 & \textcolor{red}{-3.9} & 46.6 & \textcolor{red}{-9.6} \\
        MAYA 8B                & 48.0 & \textcolor{blue}{4.7} & 51.3 & \textcolor{red}{-7.3} & 47.6 & \textcolor{blue}{4.7} & 48.3 & \textcolor{red}{-9.0} & 44.8 & \textcolor{blue}{4.7} & 43.7 & \textcolor{red}{-10.6} & 46.5 & \textcolor{blue}{4.7} & 41.3 & \textcolor{red}{-11.6} & 42.8 & \textcolor{blue}{5.7} & 40.3 & \textcolor{red}{-11.6} & 43.3 & \textcolor{blue}{4.9} & 45.8 & \textcolor{red}{-12.6} & 48.3 & \textcolor{blue}{4.5} & 50.4 & \textcolor{red}{-10.2} & 46.9 & \textcolor{blue}{4.3} & 49.8 & \textcolor{red}{-5.7} \\
        PALO 13B               & 49.7 & \textcolor{blue}{0.4} & 52.5 & \textcolor{blue}{0.2} & 50.0 & \textcolor{red}{-0.5} & 49.4 & \textcolor{blue}{0.4} & 48.7 & \textcolor{red}{-0.5} & 44.8 & \textcolor{blue}{0.4} & 47.0 & \textcolor{red}{-0.6} & 42.4 & \textcolor{red}{-0.5} & 47.6 & \textcolor{blue}{0.4} & 41.4 & \textcolor{blue}{0.4} & 47.9 & \textcolor{red}{-0.7} & 46.9 & \textcolor{blue}{0.3} & 52.9 & \textcolor{red}{-13.0} & 51.5 & \textcolor{red}{-0.4} & 50.8 & \textcolor{red}{-0.9} & 51.0 & \textcolor{red}{-0.3} \\
        \midrule
        \rowcolor{blue!20}
        \multicolumn{33}{c}{ \textcolor{violet}{\textbf{Zero-shot, \textit{With Rationales}}}   } \\
        \midrule
        PALO 7B               & 48.1 & \textcolor{red}{-0.2} & 55.1 & \textcolor{blue}{4.8} & 47.9 & \textcolor{red}{-0.2} & 50.8 & \textcolor{blue}{4.8} & 47.7 & \textcolor{red}{-0.2} & 51.5 & \textcolor{blue}{4.8} & 48.1 & \textcolor{red}{-0.2} & 52.3 & \textcolor{blue}{4.8} & 46.8 & \textcolor{red}{-0.6} & 46.9 & \textcolor{blue}{4.3} & 47.8 & \textcolor{red}{-0.5} & 49.6 & \textcolor{blue}{5.1} & 51.4 & \textcolor{red}{-4.0} & 53.7 & \textcolor{blue}{2.6} & 44.4 & \textcolor{red}{-0.5} & 48.2 & \textcolor{blue}{5.2} \\
        MAYA 8B               & 49.4 & \textcolor{red}{-0.2} & 55.4 & \textcolor{red}{-0.2} & 49.2 & \textcolor{red}{-0.2} & 52.3 & \textcolor{red}{-0.2} & 49.0 & \textcolor{red}{-0.2} & 52.1 & \textcolor{red}{-0.2} & 49.4 & \textcolor{red}{-0.2} & 53.1 & \textcolor{red}{-0.2} & 47.7 & \textcolor{red}{-0.3} & 47.8 & \textcolor{red}{-0.4} & 49.1 & \textcolor{red}{-0.5} & 50.4 & \textcolor{red}{-0.5} & 52.9 & \textcolor{red}{-2.5} & 54.5 & \textcolor{red}{-0.1} & 45.6 & \textcolor{red}{-0.4} & 49.0 & \textcolor{red}{-0.4} \\
        PALO 13B              & 49.5 & \textcolor{red}{-0.1} & 52.6 & \textcolor{red}{-0.3} & 49.4 & \textcolor{red}{-0.2} & 49.3 & \textcolor{red}{-0.2} & 49.3 & \textcolor{red}{-0.3} & 49.3 & \textcolor{red}{-0.6} & 50.1 & \textcolor{red}{-0.3} & 50.1 & \textcolor{red}{-0.5} & 48.1 & \textcolor{red}{-0.4} & 45.0 & \textcolor{red}{-0.5} & 49.1 & \textcolor{red}{-0.3} & 48.1 & \textcolor{red}{-0.4} & 52.5 & \textcolor{blue}{1.3} & 52.2 & \textcolor{red}{-0.4} & 46.2 & \textcolor{red}{-0.3} & 46.3 & \textcolor{red}{-0.4} \\
        \midrule
        \rowcolor{blue!20}
        \multicolumn{33}{c}{ \textcolor{teal}{\textbf{One-shot, \textit{Without Rationales}}} } \\
        \midrule
        PALO 7B               & 50.1 & \textcolor{red}{-0.6} & 51.3 & \textcolor{blue}{4.7} & 50.6 & \textcolor{red}{-0.6} & 50.8 & \textcolor{blue}{4.4} & 47.8 & \textcolor{red}{-0.3} & 49.6 & \textcolor{blue}{5.0} & 49.9 & \textcolor{red}{-0.5} & 49.3 & \textcolor{blue}{5.3} & 48.3 & \textcolor{red}{-0.4} & 45.0 & \textcolor{blue}{5.0} & 48.7 & \textcolor{red}{-0.6} & 48.4 & \textcolor{blue}{4.2} & 51.3 & \textcolor{red}{-3.1} & 54.1 & \textcolor{blue}{2.5} & 50.2 & \textcolor{red}{-0.6} & 50.5 & \textcolor{blue}{5.1} \\
        MAYA 8B               & 51.6 & \textcolor{red}{-0.4} & 51.6 & \textcolor{red}{-0.4} & 51.8 & \textcolor{red}{-0.5} & 51.6 & \textcolor{red}{-0.1} & 49.3 & \textcolor{red}{-0.5} & 50.4 & \textcolor{red}{-0.4} & 51.4 & \textcolor{red}{-0.7} & 50.4 & \textcolor{red}{-0.3} & 49.7 & \textcolor{red}{-0.5} & 45.4 & \textcolor{red}{-0.4} & 50.0 & \textcolor{red}{-0.6} & 49.3 & \textcolor{red}{-0.1} & 52.9 & \textcolor{red}{-1.7} & 54.9 & \textcolor{red}{-0.7} & 51.6 & \textcolor{red}{-0.7} & 51.3 & \textcolor{red}{-0.4} \\
        PALO 13B              & 52.6 & \textcolor{red}{-0.5} & 48.8 & \textcolor{red}{-0.5} & 52.3 & \textcolor{red}{-0.6} & 48.8 & \textcolor{red}{-0.3} & 49.4 & \textcolor{red}{-0.4} & 47.7 & \textcolor{red}{-0.4} & 50.9 & \textcolor{red}{-0.1} & 47.1 & \textcolor{red}{-0.7} & 50.1 & \textcolor{red}{-0.6} & 42.6 & \textcolor{red}{-0.5} & 50.0 & \textcolor{red}{-0.4} & 46.2 & \textcolor{red}{-0.7} & 53.3 & \textcolor{blue}{11.9} & 52.3 & \textcolor{red}{-0.7} & 51.3 & \textcolor{red}{-0.5} & 48.3 & \textcolor{red}{-0.3} \\
        \midrule
        \rowcolor{blue!20}
        \multicolumn{33}{c}{ \textcolor{teal}{\textbf{One-shot, \textit{With Rationales}}} } \\
        \midrule
        PALO 7B               & 50.8 & \textcolor{red}{-0.2} & 55.4 & \textcolor{blue}{5.6} & 50.5 & \textcolor{red}{-0.4} & 51.7 & \textcolor{blue}{5.1} & 50.0 & \textcolor{red}{-0.5} & 52.8 & \textcolor{blue}{5.1} & 50.8 & \textcolor{red}{-0.3} & 52.1 & \textcolor{blue}{4.6} & 48.6 & \textcolor{red}{-0.4} & 48.2 & \textcolor{blue}{5.3} & 50.5 & \textcolor{red}{-0.7} & 48.4 & \textcolor{blue}{4.9} & 51.3 & \textcolor{red}{-6.6} & 52.7 & \textcolor{blue}{0.9} & 51.2 & \textcolor{red}{-0.7} & 51.4 & \textcolor{blue}{4.8} \\
        MAYA 8B               & 52.5 & \textcolor{red}{-0.8} & 56.3 & \textcolor{red}{-0.8} & 51.7 & \textcolor{red}{-0.3} & 52.6 & \textcolor{red}{-0.3} & 51.7 & \textcolor{red}{-0.9} & 54.2 & \textcolor{red}{-0.9} & 52.7 & \textcolor{red}{-0.9} & 52.9 & \textcolor{red}{-0.8} & 50.3 & \textcolor{red}{-0.8} & 49.0 & \textcolor{red}{-0.9} & 52.0 & \textcolor{red}{-0.9} & 49.1 & \textcolor{red}{-0.9} & 52.7 & \textcolor{red}{-5.0} & 53.5 & \textcolor{red}{-0.7} & 52.7 & \textcolor{red}{-0.9} & 52.2 & \textcolor{red}{-0.3} \\
        PALO 13B              & 52.3 & \textcolor{red}{-0.4} & 52.7 & \textcolor{red}{-0.6} & 52.0 & \textcolor{red}{-0.3} & 49.8 & \textcolor{red}{-0.6} & 51.3 & \textcolor{red}{-0.4} & 51.3 & \textcolor{red}{-0.5} & 52.4 & \textcolor{red}{-0.4} & 50.0 & \textcolor{red}{-0.5} & 50.4 & \textcolor{red}{-0.5} & 46.1 & \textcolor{red}{-0.4} & 51.3 & \textcolor{red}{-0.6} & 45.6 & \textcolor{red}{-0.3} & 52.2 & \textcolor{red}{-11.8} & 51.5 & \textcolor{red}{-0.6} & 51.5 & \textcolor{red}{-0.1} & 49.4 & \textcolor{red}{-0.5} \\
        \bottomrule
    \end{tabular}
    }
    \label{tab:FinetuneAccuracyUr} 
\end{table*}

\clearpage
\subsection{Comparison with MaRVL Dataset}
\label{subsec:MaRVL}

The MaRVL dataset \cite{liu-etal-2021-visually} contains image-text pairs in five languages: Indonesian (Id), Mandarin Chinese (Zh), Swahili (Sw), Tamil (Ta), and Turkish (Tr). 
We collect the Sw subset of images from this dataset and the captions accompanying those images.\footnote{MaRVL dataset is available at the URL \url{https://marvl-challenge.github.io/download}.}

Because each `text` in MaRVL contains several associated images, we deploy CLIP to align the text with the most relevant caption. Hence, we use CLIP-aligned image-text pairs for all the analysis in this section, and there are 78 image-text pairs.
We provide input prompts to the VLM in Sw, and the VLM generates responses in Sw. 
The results are shown in the Table \ref{tab:ZerooneshotAccuracyMaRVL}.

Under zero-shot settings, the best accuracy previously reported was \textbf{55.5\%}, achieved by xUNITER, a variant of the UNITER \cite{Chen2019UNITERUI} model. 
However, we observe a dramatic increase in the accuracy on Sw, demonstrating the impressive abilities of VLMs. The GPT-4o model achieved the highest accuracy, per task. 
\begin{table*}[h]
    \centering
    \caption{Performance of VLMs on \textbf{eight VL tasks} under \colorbox{violet!20}{zero-shot} and \colorbox{teal!20}{one-shot} settings, measured by Accuracy (\%).
            \textbf{Input:} Swahili Texts + Images; \textbf{Output:} Swahili responses.
            We do not include LlaVa Mistral 7B, PALO 7B, MAYA 7B, LlaVa Mistral 13B, PALO 13B, because they showed very little understanding of Sw in our initial experiments.} 
    \resizebox{\textwidth}{!}{ 
    \begin{tabular}{r|| c|c|c|c|c||c|c|c}
        \toprule
        \multirow{2}{*}{\textbf{Model}} & \textbf{Object Recognition} & \textbf{Scene Understanding} & \textbf{Relation Understanding} & \textbf{Semantic Segmentation} & \textbf{Image Captioning} & \textbf{Image-Text Matching} & \textbf{Unrelatedness} & \textbf{Visual Quesiton Answering} \\ 
        & &  & & & & & & \\
        \midrule
        \rowcolor{gray!20}
        \multicolumn{9}{c}{\textcolor{violet}{\textbf{Zero-shot, \textit{Without Rationales}}}} \\
        \midrule
        GPT-4o                 & 87.0 & 86.8 & 82.6 & 77.0 & 82.1 & 89.9 & 97.3 & 92.9 \\
        GPT-4o-mini            & 81.3 & 78.6 & 73.9 & 69.1 & 74.9 & 78.2 & 88.4 & 84.9 \\
        Gemini 2.0 Flash Lite  & 85.0 & 82.3 & 77.6 & 72.8 & 78.6 & 81.9 & 92.1 & 88.6 \\
        Gemini 1.5 Flash 8B    & 79.6 & 76.9 & 72.2 & 67.4 & 73.2 & 76.5 & 86.7 & 83.2 \\
        \midrule
        \rowcolor{gray!20}
        \multicolumn{9}{c}{\textcolor{violet}{\textbf{Zero-shot, \textit{With Rationales}}}} \\
        \midrule
        GPT-4o                & 85.0 & 87.1 & 84.6 & 82.5 & 85.3 & 87.6 & 97.2 & 90.3 \\
        GPT-4o-mini           & 81.5 & 78.7 & 78.5 & 77.0 & 78.5 & 78.9 & 88.6 & 80.2 \\
        Gemini 2.0 Flash Lite & 85.2 & 82.4 & 82.3 & 80.7 & 82.1 & 82.6 & 92.3 & 83.9 \\
        Gemini 1.5 Flash 8B   & 79.8 & 77.0 & 77.1 & 75.3 & 76.8 & 77.2 & 86.9 & 78.5 \\
        \midrule
        \rowcolor{gray!20}
        \multicolumn{9}{c}{\textcolor{teal}{\textbf{One-shot, \textit{Without Rationales}}}} \\
        \midrule
        GPT-4o                & 85.7 & 86.9 & 86.1 & 81.6 & 82.0 & 87.7 & 97.0 & 90.6 \\
        GPT-4o-mini           & 77.7 & 78.0 & 76.8 & 74.2 & 76.1 & 77.8 & 89.0 & 82.5 \\
        Gemini 2.0 Flash Lite & 81.4 & 81.7 & 80.5 & 77.9 & 79.8 & 81.5 & 92.7 & 86.2 \\
        Gemini 1.5 Flash 8B   & 76.0 & 76.3 & 75.2 & 72.5 & 74.4 & 76.1 & 87.3 & 80.8 \\
        \midrule
        \rowcolor{gray!20}
        \multicolumn{9}{c}{\textcolor{teal}{\textbf{One-shot, \textit{With Rationales}}}} \\
        \midrule
        GPT-4o                & \textbf{88.3} & \textbf{87.6} & \textbf{87.8} & \textbf{85.4} & \textbf{88.0} & \textbf{89.2} & \textbf{97.8} & \textbf{94.3} \\
        GPT-4o-mini           & 81.8 & 79.0 & 80.5 & 76.8 & 79.7 & 77.7 & 87.6 & 83.4 \\
        Gemini 2.0 Flash Lite & 85.5 & 82.6 & 84.2 & 80.5 & 83.4 & 81.4 & 91.3 & 87.1 \\
        Gemini 1.5 Flash 8B   & 80.1 & 77.3 & 78.8 & 75.1 & 77.9 & 75.7 & 85.9 & 81.7 \\
        \bottomrule
    \end{tabular}
    } 
    \label{tab:ZerooneshotAccuracyMaRVL} 
\end{table*}

\clearpage
\subsection{Tables of Robustness Scores}
\label{subsec:Robustness}
\begin{table*}[h]
    \centering
    \caption{Average Task Difficulty, i.e., \(|1 - \text{Avg. Accuracy}|\), per VLM across Languages.
             Scores are obtained from the \textit{One-shot, With Rationales} setting.
             The average is calculated over available languages for each VLM-task pair (up to 4: En, Jp, Sw, Ur).
             A $^{(n)}$ superscript indicates the number of languages included in the average if less than 4.
             \textbf{Lower scores indicate lower average difficulty (higher average accuracy)}. Values rounded to three decimal places.}
    \resizebox{\textwidth}{!}{%
    \begin{tabular}{l || c c c c c c c c}
        \toprule
        \multirow{2}{*}{\textbf{Model}} & \multicolumn{8}{c}{\textbf{Average Difficulty per Task}} \\
        \cmidrule(lr){2-9}
        & \textbf{OR} & \textbf{SU} & \textbf{RU} & \textbf{SS} & \textbf{IC} & \textbf{ITM} & \textbf{U} & \textbf{VQA} \\
        \midrule
        GPT-4o                 & 0.120 & 0.128 & 0.127 & 0.126 & 0.163 & 0.133 & 0.107 & 0.114 \\
        GPT-4o-mini            & 0.190 & 0.213 & 0.203 & 0.211 & 0.246 & 0.240 & 0.204 & 0.216 \\
        Gemini 2.0 Flash Lite  & 0.153 & 0.177 & 0.166 & 0.174 & 0.209 & 0.203 & 0.167 & 0.179 \\
        Gemini 1.5 Flash 8B    & 0.207 & 0.230 & 0.220 & 0.228 & 0.263 & 0.257 & 0.221 & 0.233 \\
        LlaVa Mistral 7B       & 0.630$^{(2)}$ & 0.654$^{(2)}$ & 0.648$^{(2)}$ & 0.651$^{(2)}$ & 0.686$^{(2)}$ & 0.680$^{(2)}$ & 0.643$^{(2)}$ & 0.657$^{(2)}$ \\
        Qwen2VL 7B             & 0.362$^{(2)}$ & 0.370$^{(2)}$ & 0.382$^{(2)}$ & 0.380$^{(2)}$ & 0.409$^{(2)}$ & 0.408$^{(2)}$ & 0.371$^{(2)}$ & 0.386$^{(2)}$ \\
        PALO 7B                & 0.566$^{(3)}$ & 0.579$^{(3)}$ & 0.586$^{(3)}$ & 0.588$^{(3)}$ & 0.621$^{(3)}$ & 0.615$^{(3)}$ & 0.581$^{(3)}$ & 0.592$^{(3)}$ \\
        MAYA 8B                & 0.555$^{(3)}$ & 0.581$^{(3)}$ & 0.573$^{(3)}$ & 0.578$^{(3)}$ & 0.612$^{(3)}$ & 0.606$^{(3)}$ & 0.572$^{(3)}$ & 0.583$^{(3)}$ \\
        LlaVa Mistral 13B      & 0.579$^{(2)}$ & 0.606$^{(2)}$ & 0.603$^{(2)}$ & 0.606$^{(2)}$ & 0.642$^{(2)}$ & 0.639$^{(2)}$ & 0.600$^{(2)}$ & 0.620$^{(2)}$ \\
        PALO 13B               & 0.576$^{(3)}$ & 0.598$^{(3)}$ & 0.592$^{(3)}$ & 0.597$^{(3)}$ & 0.630$^{(3)}$ & 0.630$^{(3)}$ & 0.585$^{(3)}$ & 0.603$^{(3)}$ \\
        \bottomrule
    \end{tabular}%
    }
    \label{tab:avg_task_difficulty_per_vlm_simple}
\end{table*}
\begin{table}[h]
    \centering
    \caption{Overall Average Difficulty per Language.
             Calculated from \textit{One-shot, With Rationales} accuracies, averaged over 8 tasks using N=4 common VLMs (GPT-4o, GPT-4o-mini, Gemini 2.0 Flash Lite, Gemini 1.5 Flash 8B).
             Difficulty =  \(|1 - \text{Avg. Accuracy}|\).
             \textbf{Lower scores indicate lower average difficulty.}
             }
    \begin{tabular}{lc}
        \toprule
        \textbf{Language} & \textbf{Average Difficulty} \\
        \midrule
        English           & 0.153 \\
        Japanese          & 0.206 \\
        Swahili*           & 0.214 \\
        Urdu              & 0.179 \\
        \bottomrule
    \end{tabular}
    \label{tab:avg_difficulty_per_language}
\end{table}
\begin{table}[h]
    \centering
    \caption{Overall Average Difficulty per Language (all available models).
             Calculated from \textit{One-shot, With Rationales} accuracies.
             For each language, the average difficulty (1 - Avg. Accuracy of a model on that language)
             is first computed per model, then these model-specific average difficulties are averaged
             across all models evaluated for that language.
             The number of models included in each language's average is noted.
             \textbf{Lower scores indicate lower average difficulty}.
             }
    \begin{tabular}{lcc}
        \toprule
        \textbf{Language} & \textbf{No. of Models (N)} & \textbf{Average Difficulty} \\
        \midrule
        English           & 10                         & 0.382 \\
        Japanese          & 10                         & 0.443 \\
        Swahili*           & 4                          & 0.214 \\
        Urdu              & 9                          & 0.422 \\
        \bottomrule
    \end{tabular}
    \label{tab:avg_difficulty_per_language_all_models}
\end{table}

\clearpage
\begin{figure}[ht]
  \centering
  \pgfplotstableread[col sep=space]{
    model                    ZS_noR   ZS_R    OS_noR  OS_R
    GPT-4o                   -4.1     -3.8    -4.1    -3.7
    GPT-4o-mini              -3.9     -4.3    -4.2    -3.5
    Gemini-2.0-Flash-Lite    -3.8     -3.9    -3.8    -3.2
    Gemini-1.5-Flash-8B      -3.5     -3.7    -4.5    -3.7
  }\jptableA

  \begin{subfigure}[b]{0.92\textwidth}
    \centering
    \resizebox{\linewidth}{!}{%
      \begin{tikzpicture}
        \begin{axis}[
          width=1.1\linewidth, height=3cm,
          ybar=2pt, bar width=7.5pt,
          cycle list name=mybarlist,
          ylabel={ΔAcc (pp)},
          symbolic x coords={GPT-4o,GPT-4o-mini,Gemini-2.0-Flash-Lite,Gemini-1.5-Flash-8B},
          xtick=data,
          xticklabel style={font=\normalsize,rotate=-25,anchor=north west},
          ytick={-6,-4,-2,0},
          ymin=-6, ymax=0,
          scaled y ticks=false,
          legend style={
            at={(0.5,1.12)}, anchor=south, 
            legend columns=4, font=\normalsize,
            cells={anchor=west},
          }
        ]
          \addplot table[x=model,y=ZS_noR] {\jptableA}; \addlegendentry{ZS w/o rat.}
          \addplot table[x=model,y=ZS_R]   {\jptableA}; \addlegendentry{ZS w/ rat.}
          \addplot table[x=model,y=OS_noR] {\jptableA}; \addlegendentry{OS w/o rat.}
          \addplot table[x=model,y=OS_R]   {\jptableA}; \addlegendentry{OS w/ rat.}
        \end{axis}
      \end{tikzpicture}
    }
    \caption{Models 1–4}
  \end{subfigure}%
  \hfill
  \pgfplotstableread[col sep=space]{
    model                   ZS_noR   ZS_R    OS_noR  OS_R
    LlaVa-v1.6-Vicuna-7B    -14.3    -14.5   -14.8   -14.0
    Qwen2VL-7B              -3.5     -3.7    -4.0    -4.9
    PALO-7B                 -3.7     -3.6    -3.7    -3.5
    MAYA-8B                 -4.0     -4.3    -4.1    -4.2
  }\jptableB

  \begin{subfigure}[b]{0.92\textwidth}
    \centering
    \resizebox{\linewidth}{!}{%
      \begin{tikzpicture}
        \begin{axis}[
          width=1.1\linewidth, height=3cm,
          ybar=2pt, bar width=7.5pt,
          cycle list name=mybarlist,
          ylabel={ΔAcc (pp)},
          symbolic x coords={LlaVa-v1.6-Vicuna-7B,Qwen2VL-7B,PALO-7B,MAYA-8B},
          xtick=data,
          xticklabel style={font=\normalsize,rotate=-25,anchor=north west},
          ytick={-16,-12,-8,-4,0},
          ymin=-16, ymax=0,
          scaled y ticks=false,
          legend style={draw=none, font=\scriptsize} %
        ]
          \addplot table[x=model,y=ZS_noR] {\jptableB};
          \addplot table[x=model,y=ZS_R]   {\jptableB};
          \addplot table[x=model,y=OS_noR] {\jptableB};
          \addplot table[x=model,y=OS_R]   {\jptableB};
        \end{axis}
      \end{tikzpicture}
    }
    \caption{Models 5–8}
  \end{subfigure}%
  \hfill
  \pgfplotstableread[col sep=space]{
    model                     ZS_noR   ZS_R    OS_noR  OS_R
    LlaVa-v1.6-Vicuna-13B     -16.3    -16.0   -16.4   -15.7
    PALO-13B                  -3.8     -3.8    -4.1    -3.6
  }\jptableC

  \begin{subfigure}[b]{0.92\textwidth}
    \centering
    \resizebox{\linewidth}{!}{%
      \begin{tikzpicture}
        \begin{axis}[
          width=1.1\linewidth, height=3cm,
          ybar=2pt, bar width=7.5pt,
          cycle list name=mybarlist,
          ylabel={ΔAcc (pp)},
          symbolic x coords={LlaVa-v1.6-Vicuna-13B,PALO-13B},
          xtick=data,
          xticklabel style={font=\normalsize,rotate=-25,anchor=north west},
          ytick={-16,-12,-8,-4,0},
          ymin=-17, ymax=0,
          scaled y ticks=false,
          legend style={draw=none, font=\scriptsize}
        ]
          \addplot table[x=model,y=ZS_noR] {\jptableC};
          \addplot table[x=model,y=ZS_R]   {\jptableC};
          \addplot table[x=model,y=OS_noR] {\jptableC};
          \addplot table[x=model,y=OS_R]   {\jptableC};
        \end{axis}
      \end{tikzpicture}
    }
    \caption{Models 9–10}
  \end{subfigure}

  \caption{  \textbf{ΔAccuracy (En→Jp)} for all ten VLMs under four prompt settings, organized into three rows of four, four, and two models.
        We observe that all models are sensitive to language inputs. Specifically, the \textbf{accuracy decreased under all prompting settings for all VLMs}. LlaVa v1.6 Vicuna models experienced the sharpest decline in accuracy (hence the most sensitive to language changes). 
        }
  \label{fig:XlingualDeltaJp}
\end{figure}
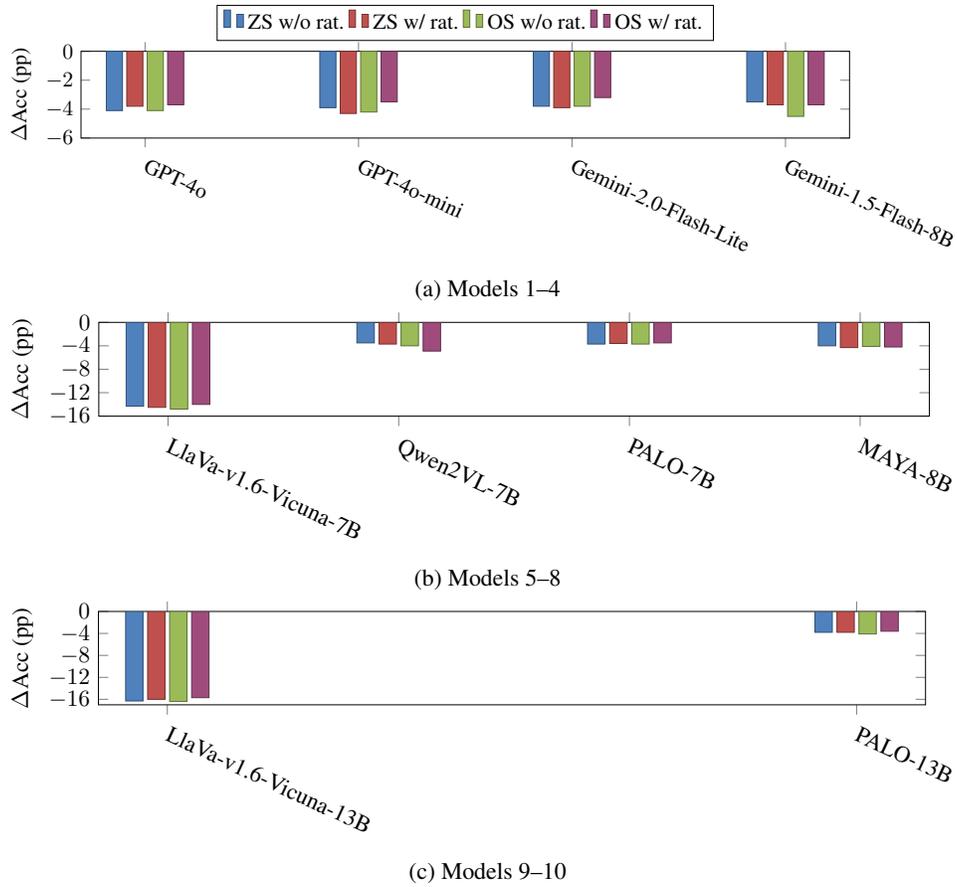

\begin{figure}[htb]
  \centering
  \pgfplotstableread[col sep=space]{
    model                    ZS_noR   ZS_R    OS_noR  OS_R
    GPT-4o                   -4.5     -5.1    -5.1    -5.0
    GPT-4o-mini              -4.5     -4.9    -4.5    -5.0
    Gemini-2.0-Flash-Lite    -5.2     -4.8    -4.8    -4.3
    Gemini-1.5-Flash-8B      -5.4     -4.8    -4.8    -4.2
  }\swptable

  \begin{subfigure}[b]{0.92\textwidth}
    \centering
    \begin{tikzpicture}
      \begin{axis}[
        width=\linewidth, height=3.2cm,
        ybar=2pt, bar width=8pt,
        cycle list name=mybarlist,
        ylabel={ΔAcc (pp)},
        symbolic x coords={GPT-4o,GPT-4o-mini,Gemini-2.0-Flash-Lite,Gemini-1.5-Flash-8B},
        xtick=data,
        xticklabel style={font=\normalsize,rotate=-25,anchor=north west},
        ytick={-8,-6,-4,-2,0},
        ymin=-8, ymax=0,
        scaled y ticks=false,
        legend style={
          at={(0.5,1.08)}, anchor=south, 
          legend columns=4, font=\normalsize,
          cells={anchor=west},
        }
      ]
        \addplot table[x=model,y=ZS_noR] {\swptable}; \addlegendentry{ZS w/o rat.}
        \addplot table[x=model,y=ZS_R]   {\swptable}; \addlegendentry{ZS w/ rat.}
        \addplot table[x=model,y=OS_noR] {\swptable}; \addlegendentry{OS w/o rat.}
        \addplot table[x=model,y=OS_R]   {\swptable}; \addlegendentry{OS w/ rat.}
      \end{axis}
    \end{tikzpicture}
    \caption{ΔAcc (En→Sw) for four VLMs.}
  \end{subfigure}

  \caption{\textbf{Swahili-input ΔAccuracy} under four prompting conditions, for models that support Swahili.}
  \label{fig:XlingualDeltaSw}
\end{figure}
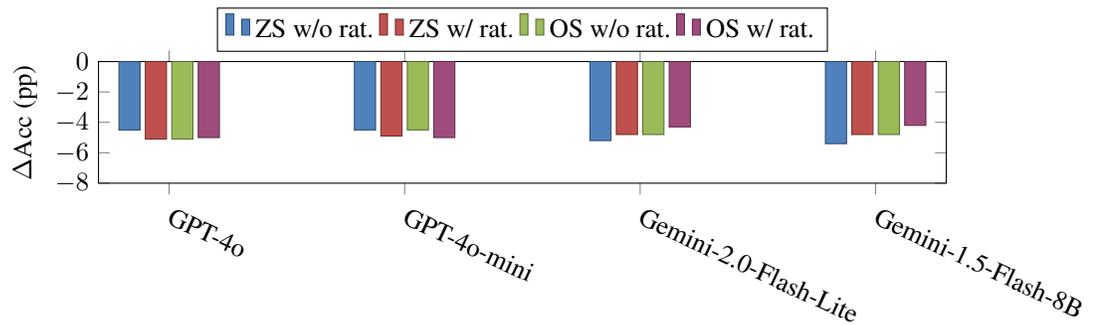
\clearpage
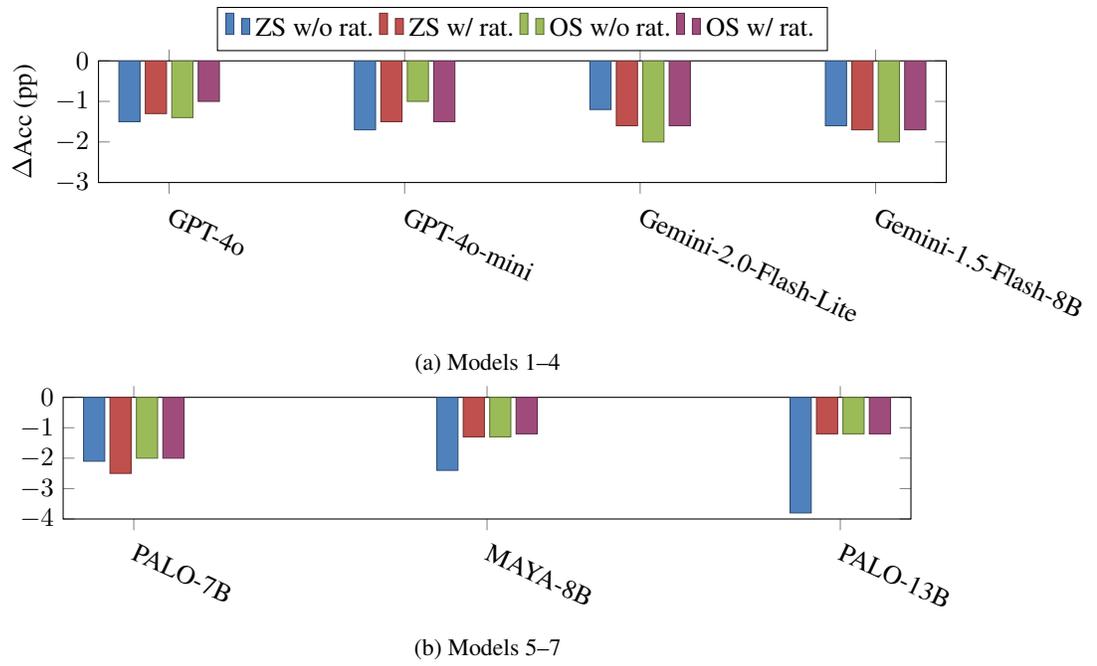
\begin{figure}[ht]
  \centering
  \pgfplotstableread[col sep=space]{
    model                    ZS_noR   ZS_R    OS_noR  OS_R
    GPT-4o                   -1.5     -1.3    -1.4    -1.0
    GPT-4o-mini              -1.7     -1.5    -1.0    -1.5
    Gemini-2.0-Flash-Lite    -1.2     -1.6    -2.0    -1.6
    Gemini-1.5-Flash-8B      -1.6     -1.7    -2.0    -1.7
  }\urptableA

  \begin{subfigure}[b]{0.92\textwidth}
    \centering
    \begin{tikzpicture}
      \begin{axis}[
        width=\linewidth, height=3.2cm,
        ybar=2pt, bar width=8pt,
        cycle list name=mybarlist,
        ylabel={ΔAcc (pp)},
        symbolic x coords={GPT-4o,GPT-4o-mini,Gemini-2.0-Flash-Lite,Gemini-1.5-Flash-8B},
        xtick=data,
        xticklabel style={font=\normalsize,rotate=-25,anchor=north west},
        ytick={-3,-2,-1,0},
        ymin=-3, ymax=0,
        scaled y ticks=false,
        legend style={at={(0.5,1.08)}, anchor=south, legend columns=4, font=\normalsize, cells={anchor=west}}
      ]
        \addplot table[x=model,y=ZS_noR] {\urptableA}; \addlegendentry{ZS w/o rat.}
        \addplot table[x=model,y=ZS_R]   {\urptableA}; \addlegendentry{ZS w/ rat.}
        \addplot table[x=model,y=OS_noR] {\urptableA}; \addlegendentry{OS w/o rat.}
        \addplot table[x=model,y=OS_R]   {\urptableA}; \addlegendentry{OS w/ rat.}
      \end{axis}
    \end{tikzpicture}
    \caption{Models 1–4}
  \end{subfigure}%
  \hfill

  \pgfplotstableread[col sep=space]{
    model                    ZS_noR   ZS_R    OS_noR  OS_R
    PALO-7B                  -2.1     -2.5    -2.0    -2.0
    MAYA-8B                  -2.4     -1.3    -1.3    -1.2
    PALO-13B                 -3.8     -1.2    -1.2    -1.2
  }\urptableB

  \begin{subfigure}[b]{0.92\textwidth}
    \centering
    \begin{tikzpicture}
      \begin{axis}[
        width=\linewidth, height=3.2cm,
        ybar=2pt, bar width=8pt,
        cycle list name=mybarlist,
        symbolic x coords={PALO-7B,MAYA-8B,PALO-13B},
        xtick=data,
        xticklabel style={font=\normalsize,rotate=-25,anchor=north west},
        ytick={-4,-3,-2,-1,0},
        ymin=-4, ymax=0,
        scaled y ticks=false,
        legend style={draw=none, font=\scriptsize}
      ]
        \addplot table[x=model,y=ZS_noR] {\urptableB};
        \addplot table[x=model,y=ZS_R]   {\urptableB};
        \addplot table[x=model,y=OS_noR] {\urptableB};
        \addplot table[x=model,y=OS_R]   {\urptableB};
      \end{axis}
    \end{tikzpicture}
    \caption{Models 5–7}
  \end{subfigure}

  \caption{\textbf{ΔAccuracy (En→Ur)} for all seven VLMs under four prompt settings, organized into rows of 4–3. Across all settings, the drops remain modest.}
  \label{fig:XlingualDeltaUr}
\end{figure}

\clearpage
\subsubsection{Detailed Cross-lingual Results: En→Jp, En→Sw, En→Ur.}
\begin{table*}[ht]
  \centering
  \captionsetup{singlelinecheck=off}%
  \caption{%
              Δ English accuracy (En→Jp, En→Sw, En→Ur) for each VLM on \textbf{Object Recognition} under four prompting settings: zero-shot without rationales (\textbf{ZS w/o rat.}), zero-shot with rationales (\textbf{ZS w/ rat.}), one-shot without rationales, (\textbf{OS w/o rat.}), one-shot with rationales (\textbf{OS w/ rat.}). 
              \\[0.5ex]\footnotesize
            All Δ$_{\mathrm{En},\mathrm{lang}}$ are computed as
            $\Delta_{\mathrm{En},\mathrm{lang}}
              = \mathrm{Acc}_{\mathrm{En\text{-}input}}
              - \mathrm{Acc}_{\mathrm{lang\text{-}input}}$.%
            }
  \resizebox{\textwidth}{!}{%
    \begin{tabular}{l
      ccc ccc ccc ccc
    }
      \toprule
      \multirow{2}{*}{\textbf{Model}}
       & \multicolumn{3}{c}{\textbf{ZS w/o rat.}}
       & \multicolumn{3}{c}{\textbf{ZS w/ rat.}}
       & \multicolumn{3}{c}{\textbf{OS w/o rat.}}
       & \multicolumn{3}{c}{\textbf{OS w/ rat.}} \\
      \cmidrule(lr){2-4}\cmidrule(lr){5-7}\cmidrule(lr){8-10}\cmidrule(lr){11-13}
       & Jp & Sw & Ur
       & Jp & Sw & Ur
       & Jp & Sw & Ur
       & Jp & Sw & Ur \\
      \midrule
      GPT-4o                & \textcolor{red}{-4.1} & \textcolor{red}{-4.5} & \textcolor{red}{-1.5} & \textcolor{red}{-3.8} & \textcolor{red}{-5.1} & \textcolor{red}{-1.3} & \textcolor{red}{-4.1} & \textcolor{red}{-5.1} & \textcolor{red}{-1.4} & \textcolor{red}{-3.7} & \textcolor{red}{-5.0} & \textcolor{red}{-1.0} \\
      GPT-4o-mini           & \textcolor{red}{-3.9} & \textcolor{red}{-4.5} & \textcolor{red}{-1.7} & \textcolor{red}{-4.3} & \textcolor{red}{-4.9} & \textcolor{red}{-1.5} & \textcolor{red}{-4.2} & \textcolor{red}{-4.5} & \textcolor{red}{-1.0} & \textcolor{red}{-3.5} & \textcolor{red}{-5.0} & \textcolor{red}{-1.5} \\
      Gemini 2.0 Flash Lite & \textcolor{red}{-3.8} & \textcolor{red}{-5.2} & \textcolor{red}{-1.2} & \textcolor{red}{-3.9} & \textcolor{red}{-4.8} & \textcolor{red}{-1.6} & \textcolor{red}{-3.8} & \textcolor{red}{-4.8} & \textcolor{red}{-2.0} & \textcolor{red}{-3.2} & \textcolor{red}{-4.3} & \textcolor{red}{-1.6} \\
      Gemini 1.5 Flash 8B   & \textcolor{red}{-3.5} & \textcolor{red}{-5.4} & \textcolor{red}{-1.6} & \textcolor{red}{-3.7} & \textcolor{red}{-4.8} & \textcolor{red}{-1.6} & \textcolor{red}{-4.5} & \textcolor{red}{-4.8} & \textcolor{red}{-2.0} & \textcolor{red}{-3.7} & \textcolor{red}{-4.3} & \textcolor{red}{-1.6} \\
      \midrule
      LlaVa Mistral 7B      & \textcolor{red}{-14.3} & 0.0 & 0.0 & \textcolor{red}{-14.5} & 0.0 & 0.0 & \textcolor{red}{-14.8} & 0.0 & 0.0 & \textcolor{red}{-14.0} & 0.0 & 0.0 \\
      Qwen2VL 7B            & \textcolor{red}{-3.5} & 0.0 & 0.0 & \textcolor{red}{-3.7} & 0.0 & 0.0 & \textcolor{red}{-4.0} & 0.0 & 0.0 & \textcolor{red}{-4.9} & 0.0 & 0.0 \\
      PALO 7B               & \textcolor{red}{-3.7} & 0.0 & \textcolor{red}{-2.1} & \textcolor{red}{-3.6} & 0.0 & \textcolor{red}{-1.8} & \textcolor{red}{-3.7} & 0.0 & \textcolor{red}{-2.5} & \textcolor{red}{-3.5} & 0.0 & \textcolor{red}{-2.0} \\
      MAYA 8B               & \textcolor{red}{-4.0} & 0.0 & \textcolor{red}{-2.4} & \textcolor{red}{-4.3} & 0.0 & \textcolor{red}{-2.0} & \textcolor{red}{-4.1} & 0.0 & \textcolor{red}{-1.8} & \textcolor{red}{-4.2} & 0.0 & \textcolor{red}{-1.9} \\
      LlaVa Mistral 13B     & \textcolor{red}{-16.3} & 0.0 & 0.0 & \textcolor{red}{-16.0} & 0.0 & 0.0 & \textcolor{red}{-16.4} & 0.0 & 0.0 & \textcolor{red}{-15.7} & 0.0 & 0.0 \\
      PALO 13B              & \textcolor{red}{-3.8} & 0.0 & \textcolor{red}{-1.2} & \textcolor{red}{-3.8} & 0.0 & \textcolor{red}{-1.2} & \textcolor{red}{-4.1} & 0.0 & \textcolor{red}{-1.8} & \textcolor{red}{-3.6} & 0.0 & \textcolor{red}{-1.5} \\
      \bottomrule
    \end{tabular}%
  }
  \label{tab:DetailedXLingualTask1}
\end{table*}
\begin{table*}[ht]
  \centering
  \captionsetup{singlelinecheck=off}%
  \caption{%
    Δ English accuracy (En→Jp, En→Sw, En→Ur) for each VLM on \textbf{Scene Understanding} under four prompting settings: 
    zero-shot without rationales (\textbf{ZS w/o rat.}), 
    zero-shot with rationales (\textbf{ZS w/ rat.}), 
    one-shot without rationales (\textbf{OS w/o rat.}), 
    one-shot with rationales (\textbf{OS w/ rat.}). 
    \\[0.5ex]\footnotesize
            All Δ$_{\mathrm{En},\mathrm{lang}}$ are computed as
            $\Delta_{\mathrm{En},\mathrm{lang}}
              = \mathrm{Acc}_{\mathrm{En\text{-}input}}
              - \mathrm{Acc}_{\mathrm{lang\text{-}input}}$.%
  }
  \resizebox{\textwidth}{!}{%
    \begin{tabular}{l
      ccc ccc ccc ccc
    }
      \toprule
      \multirow{2}{*}{\textbf{Model}}
       & \multicolumn{3}{c}{\textbf{ZS w/o rat.}}
       & \multicolumn{3}{c}{\textbf{ZS w/ rat.}}
       & \multicolumn{3}{c}{\textbf{OS w/o rat.}}
       & \multicolumn{3}{c}{\textbf{OS w/ rat.}} \\
      \cmidrule(lr){2-4}\cmidrule(lr){5-7}\cmidrule(lr){8-10}\cmidrule(lr){11-13}
       & Jp & Sw & Ur
       & Jp & Sw & Ur
       & Jp & Sw & Ur
       & Jp & Sw & Ur \\
      \midrule
      GPT-4o                & \textcolor{red}{-3.8} & \textcolor{red}{-4.7} & \textcolor{red}{-1.0} & \textcolor{red}{-4.0} & \textcolor{red}{-5.0} & \textcolor{red}{-2.0} & \textcolor{red}{-3.8} & \textcolor{red}{-4.0} & \textcolor{red}{-1.7} & \textcolor{red}{-3.7} & \textcolor{red}{-4.2} & \textcolor{red}{-1.7} \\
      GPT-4o-mini           & \textcolor{red}{-4.1} & \textcolor{red}{-5.4} & \textcolor{red}{-1.5} & \textcolor{red}{-3.7} & \textcolor{red}{-4.8} & \textcolor{red}{-1.5} & \textcolor{red}{-3.5} & \textcolor{red}{-4.3} & \textcolor{red}{-1.6} & \textcolor{red}{-3.6} & \textcolor{red}{-4.1} & \textcolor{red}{-1.6} \\
      Gemini 2.0 Flash Lite & \textcolor{red}{-3.4} & \textcolor{red}{-5.4} & \textcolor{red}{-1.7} & \textcolor{red}{-3.2} & \textcolor{red}{-4.0} & \textcolor{red}{-1.6} & \textcolor{red}{-3.9} & \textcolor{red}{-4.1} & \textcolor{red}{-2.0} & \textcolor{red}{-3.0} & \textcolor{red}{-3.7} & \textcolor{red}{-1.5} \\
      Gemini 1.5 Flash 8B   & \textcolor{red}{-3.8} & \textcolor{red}{-5.2} & \textcolor{red}{-1.9} & \textcolor{red}{-3.7} & \textcolor{red}{-4.8} & \textcolor{red}{-1.8} & \textcolor{red}{-4.2} & \textcolor{red}{-4.1} & \textcolor{red}{-2.2} & \textcolor{red}{-3.0} & \textcolor{red}{-3.9} & \textcolor{red}{-1.6} \\
      \midrule
      LlaVa Mistral 7B      & \textcolor{red}{-8.3} & 0.0 & 0.0 & \textcolor{red}{-8.4} & 0.0 & 0.0 & \textcolor{red}{-9.0} & 0.0 & 0.0 & \textcolor{red}{-9.1} & 0.0 & 0.0 \\
      Qwen2VL 7B            & \textcolor{red}{-4.6} & 0.0 & 0.0 & \textcolor{red}{-4.0} & 0.0 & 0.0 & \textcolor{red}{-3.5} & 0.0 & 0.0 & \textcolor{red}{-3.8} & 0.0 & 0.0 \\
      PALO 7B               & \textcolor{red}{-3.4} & 0.0 & \textcolor{red}{-2.0} & \textcolor{red}{-3.4} & 0.0 & \textcolor{red}{-2.2} & \textcolor{red}{-3.7} & 0.0 & \textcolor{red}{-2.5} & \textcolor{red}{-3.6} & 0.0 & \textcolor{red}{-2.1} \\
      MAYA 8B               & \textcolor{red}{-3.7} & 0.0 & \textcolor{red}{-2.5} & \textcolor{red}{-3.8} & 0.0 & \textcolor{red}{-2.4} & \textcolor{red}{-3.6} & 0.0 & \textcolor{red}{-2.0} & \textcolor{red}{-3.8} & 0.0 & \textcolor{red}{-1.9} \\
      LlaVa Mistral 13B     & \textcolor{red}{-8.2} & 0.0 & 0.0 & \textcolor{red}{-7.8} & 0.0 & 0.0 & \textcolor{red}{-8.0} & 0.0 & 0.0 & \textcolor{red}{-9.1} & 0.0 & 0.0 \\
      PALO 13B              & \textcolor{red}{-3.8} & 0.0 & \textcolor{red}{-2.2} & \textcolor{red}{-3.6} & 0.0 & \textcolor{red}{-2.3} & \textcolor{red}{-3.8} & 0.0 & \textcolor{red}{-2.4} & \textcolor{red}{-3.6} & 0.0 & \textcolor{red}{-2.0} \\
      \bottomrule
    \end{tabular}
  }
  \label{tab:DetailedXLingualTask2}
\end{table*}
\begin{table*}[ht]
  \centering
  \captionsetup{singlelinecheck=off}%
  \caption{%
    Δ English accuracy (En→Jp, En→Sw, En→Ur) for each VLM on \textbf{Relation Understanding} under four prompting settings: 
    zero-shot without rationales (\textbf{ZS w/o rat.}), 
    zero-shot with rationales (\textbf{ZS w/ rat.}), 
    one-shot without rationales (\textbf{OS w/o rat.}), 
    one-shot with rationales (\textbf{OS w/ rat.}). 
    \\[0.5ex]\footnotesize
            All Δ$_{\mathrm{En},\mathrm{lang}}$ are computed as
            $\Delta_{\mathrm{En},\mathrm{lang}}
              = \mathrm{Acc}_{\mathrm{En\text{-}input}}
              - \mathrm{Acc}_{\mathrm{lang\text{-}input}}$.%
  }
  \resizebox{\textwidth}{!}{%
    \begin{tabular}{l
      ccc ccc ccc ccc
    }
      \toprule
      \multirow{2}{*}{\textbf{Model}}
       & \multicolumn{3}{c}{\textbf{ZS w/o rat.}}
       & \multicolumn{3}{c}{\textbf{ZS w/ rat.}}
       & \multicolumn{3}{c}{\textbf{OS w/o rat.}}
       & \multicolumn{3}{c}{\textbf{OS w/ rat.}} \\
      \cmidrule(lr){2-4}\cmidrule(lr){5-7}\cmidrule(lr){8-10}\cmidrule(lr){11-13}
       & Jp & Sw & Ur
       & Jp & Sw & Ur
       & Jp & Sw & Ur
       & Jp & Sw & Ur \\
      \midrule
      GPT-4o                & \textcolor{red}{-4.5} & \textcolor{red}{-4.6} & \textcolor{red}{-1.1} & \textcolor{red}{-3.4} & \textcolor{red}{-4.8} & \textcolor{red}{-1.6} & \textcolor{red}{-3.5} & \textcolor{red}{-4.4} & \textcolor{red}{-1.6} & \textcolor{red}{-4.4} & \textcolor{red}{-4.0} & \textcolor{red}{-2.1} \\
      GPT-4o-mini           & \textcolor{red}{-3.5} & \textcolor{red}{-4.5} & \textcolor{red}{-1.7} & \textcolor{red}{-3.7} & \textcolor{red}{-4.9} & \textcolor{red}{-1.5} & \textcolor{red}{-3.5} & \textcolor{red}{-4.5} & \textcolor{red}{-1.6} & \textcolor{red}{-3.7} & \textcolor{red}{-5.0} & \textcolor{red}{-1.6} \\
      Gemini 2.0 Flash Lite & \textcolor{red}{-3.7} & \textcolor{red}{-5.2} & \textcolor{red}{-1.2} & \textcolor{red}{-3.7} & \textcolor{red}{-4.8} & \textcolor{red}{-1.6} & \textcolor{red}{-3.5} & \textcolor{red}{-4.8} & \textcolor{red}{-2.0} & \textcolor{red}{-3.7} & \textcolor{red}{-4.3} & \textcolor{red}{-1.6} \\
      Gemini 1.5 Flash 8B   & \textcolor{red}{-4.2} & \textcolor{red}{-5.4} & \textcolor{red}{-1.8} & \textcolor{red}{-3.7} & \textcolor{red}{-4.8} & \textcolor{red}{-1.6} & \textcolor{red}{-3.1} & \textcolor{red}{-4.8} & \textcolor{red}{-2.0} & \textcolor{red}{-3.7} & \textcolor{red}{-4.3} & \textcolor{red}{-1.6} \\
      \midrule
      LlaVa Mistral 7B      & \textcolor{red}{-11.8} & 0.0 & 0.0 & \textcolor{red}{-13.0} & 0.0 & 0.0 & \textcolor{red}{-13.5} & 0.0 & 0.0 & \textcolor{red}{-13.0} & 0.0 & 0.0 \\
      Qwen2VL 7B            & \textcolor{red}{-3.1} & 0.0 & 0.0 & \textcolor{red}{-3.5} & 0.0 & 0.0 & \textcolor{red}{-3.6} & 0.0 & 0.0 & \textcolor{red}{-2.7} & 0.0 & 0.0 \\
      PALO 7B               & \textcolor{red}{-3.0} & 0.0 & \textcolor{red}{-1.8} & \textcolor{red}{-3.2} & 0.0 & \textcolor{red}{-1.8} & \textcolor{red}{-3.0} & 0.0 & \textcolor{red}{-2.2} & \textcolor{red}{-3.6} & 0.0 & \textcolor{red}{-1.9} \\
      MAYA 8B               & \textcolor{red}{-4.2} & 0.0 & \textcolor{red}{-1.8} & \textcolor{red}{-3.9} & 0.0 & \textcolor{red}{-1.9} & \textcolor{red}{-4.0} & 0.0 & \textcolor{red}{-1.8} & \textcolor{red}{-3.6} & 0.0 & \textcolor{red}{-1.9} \\
      LlaVa Mistral 13B     & \textcolor{red}{-17.0} & 0.0 & 0.0 & \textcolor{red}{-15.9} & 0.0 & 0.0 & \textcolor{red}{-15.0} & 0.0 & 0.0 & \textcolor{red}{-17.0} & 0.0 & 0.0 \\
      PALO 13B              & \textcolor{red}{-4.4} & 0.0 & \textcolor{red}{-2.0} & \textcolor{red}{-4.0} & 0.0 & \textcolor{red}{-2.3} & \textcolor{red}{-4.2} & 0.0 & \textcolor{red}{-2.6} & \textcolor{red}{-3.6} & 0.0 & \textcolor{red}{-2.4} \\
      \bottomrule
    \end{tabular}
  }
  \label{tab:DetailedXLingualTask3}
\end{table*}
\begin{table*}[ht]
  \centering
  \captionsetup{singlelinecheck=off}%
  \caption{%
    Δ English accuracy on \textbf{Semantic Segmentation} (En→Jp, En→Sw, En→Ur) for each VLM under four prompting settings: 
    zero-shot w/o rationales (\textbf{ZS w/o rat.}), zero-shot w/ rationales (\textbf{ZS w/ rat.}), 
    one-shot w/o rationales (\textbf{OS w/o rat.}), one-shot w/ rationales (\textbf{OS w/ rat.}).
    \\[0.5ex]\footnotesize
    All Δ$_{\mathrm{En},\mathrm{lang}}$ are computed as
        $\Delta_{\mathrm{En},\mathrm{lang}}
          = \mathrm{Accuracy}_{\mathrm{En\text{-}input}}
          - \mathrm{Accuracy}_{\mathrm{lang\text{-}input}}$.%
  }
  \resizebox{\textwidth}{!}{%
    \begin{tabular}{l
      ccc ccc ccc ccc
    }
      \toprule
      \multirow{2}{*}{\textbf{Model}}
       & \multicolumn{3}{c}{\textbf{ZS w/o rat.}}
       & \multicolumn{3}{c}{\textbf{ZS w/ rat.}}
       & \multicolumn{3}{c}{\textbf{OS w/o rat.}}
       & \multicolumn{3}{c}{\textbf{OS w/ rat.}} \\
      \cmidrule(lr){2-4}\cmidrule(lr){5-7}\cmidrule(lr){8-10}\cmidrule(lr){11-13}
       & Jp & Sw & Ur
       & Jp & Sw & Ur
       & Jp & Sw & Ur
       & Jp & Sw & Ur \\
      \midrule
      GPT-4o                & \textcolor{red}{-5.0} & \textcolor{red}{-5.1} & \textcolor{red}{-0.8} & \textcolor{red}{-4.6} & \textcolor{red}{-5.2} & \textcolor{red}{-0.9} & \textcolor{red}{-5.3} & \textcolor{red}{-4.8} & \textcolor{red}{-1.0} & \textcolor{red}{-4.5} & \textcolor{red}{-4.9} & \textcolor{red}{-0.7} \\
      GPT-4o-mini           & \textcolor{red}{-3.2} & \textcolor{red}{-5.1} & \textcolor{red}{-1.5} & \textcolor{red}{-3.0} & \textcolor{red}{-5.2} & \textcolor{red}{-1.0} & \textcolor{red}{-3.4} & \textcolor{red}{-4.0} & \textcolor{red}{-0.8} & \textcolor{red}{-3.6} & \textcolor{red}{-5.1} & \textcolor{red}{-1.2} \\
      Gemini 2.0 Flash Lite & \textcolor{red}{-2.7} & \textcolor{red}{-5.0} & \textcolor{red}{-1.2} & \textcolor{red}{-3.0} & \textcolor{red}{-5.0} & \textcolor{red}{-1.3} & \textcolor{red}{-2.4} & \textcolor{red}{-4.3} & \textcolor{red}{-1.1} & \textcolor{red}{-3.0} & \textcolor{red}{-5.2} & \textcolor{red}{-1.0} \\
      Gemini 1.5 Flash 8B   & \textcolor{red}{-3.3} & \textcolor{red}{-5.0} & \textcolor{red}{-1.8} & \textcolor{red}{-2.9} & \textcolor{red}{-5.1} & \textcolor{red}{-1.3} & \textcolor{red}{-3.0} & \textcolor{red}{-5.1} & \textcolor{red}{-1.5} & \textcolor{red}{-3.1} & \textcolor{red}{-5.1} & \textcolor{red}{-1.6} \\
      \midrule
      LlaVa Mistral 7B      & \textcolor{red}{-13.9} & 0.0 & 0.0 & \textcolor{red}{-12.5} & 0.0 & 0.0 & \textcolor{red}{-13.8} & 0.0 & 0.0 & \textcolor{red}{-12.2} & 0.0 & 0.0 \\
      Qwen2VL 7B            & \textcolor{red}{-4.1} & 0.0 & 0.0 & \textcolor{red}{-4.2} & 0.0 & 0.0 & \textcolor{red}{-3.8} & 0.0 & 0.0 & \textcolor{red}{-4.0} & 0.0 & 0.0 \\
      PALO 7B               & \textcolor{red}{-4.0} & 0.0 & \textcolor{red}{-2.4} & \textcolor{red}{-4.3} & 0.0 & \textcolor{red}{-2.0} & \textcolor{red}{-4.1} & 0.0 & \textcolor{red}{-1.8} & \textcolor{red}{-4.2} & 0.0 & \textcolor{red}{-1.9} \\
      MAYA 8B               & \textcolor{red}{-3.7} & 0.0 & \textcolor{red}{-2.0} & \textcolor{red}{-3.6} & 0.0 & \textcolor{red}{-1.9} & \textcolor{red}{-3.7} & 0.0 & \textcolor{red}{-1.8} & \textcolor{red}{-3.5} & 0.0 & \textcolor{red}{-1.7} \\
      LlaVa Mistral 13B     & \textcolor{red}{-15.0} & 0.0 & 0.0 & \textcolor{red}{-15.9} & 0.0 & 0.0 & \textcolor{red}{-15.7} & 0.0 & 0.0 & \textcolor{red}{-15.0} & 0.0 & 0.0 \\
      PALO 13B              & \textcolor{red}{-4.0} & 0.0 & \textcolor{red}{-2.8} & \textcolor{red}{-4.1} & 0.0 & \textcolor{red}{-2.0} & \textcolor{red}{-4.3} & 0.0 & \textcolor{red}{-1.8} & \textcolor{red}{-3.9} & 0.0 & \textcolor{red}{-1.9} \\
      \bottomrule
    \end{tabular}%
  }
  \label{tab:DetailedXLingualTask4}
\end{table*}
\begin{table*}[ht]
  \centering
  \captionsetup{singlelinecheck=off}%
  \caption{%
    Δ English accuracy on \textbf{Image Captioning} (En→Jp, En→Sw, En→Ur) for each VLM under four prompting settings:
    zero-shot without rationales (\textbf{ZS w/o rat.}), zero-shot with rationales (\textbf{ZS w/ rat.}),
    one-shot without rationales (\textbf{OS w/o rat.}), one-shot with rationales (\textbf{OS w/ rat.}).
    \\[0.5ex]\footnotesize
    All Δ$_{\mathrm{En},\mathrm{lang}}$ are computed as
    $\Delta_{\mathrm{En},\mathrm{lang}}
      = \mathrm{Acc}_{\mathrm{En\text{-}input}}
      - \mathrm{Acc}_{\mathrm{lang\text{-}input}}$.%
  }
  \resizebox{\textwidth}{!}{%
    \begin{tabular}{l
      ccc ccc ccc ccc
    }
      \toprule
      \multirow{2}{*}{\textbf{Model}}
       & \multicolumn{3}{c}{\textbf{ZS w/o rat.}}
       & \multicolumn{3}{c}{\textbf{ZS w/ rat.}}
       & \multicolumn{3}{c}{\textbf{OS w/o rat.}}
       & \multicolumn{3}{c}{\textbf{OS w/ rat.}} \\
      \cmidrule(lr){2-4}\cmidrule(lr){5-7}\cmidrule(lr){8-10}\cmidrule(lr){11-13}
       & Jp & Sw & Ur
       & Jp & Sw & Ur
       & Jp & Sw & Ur
       & Jp & Sw & Ur \\
      \midrule
      GPT-4o                & \textcolor{red}{-3.8} & \textcolor{red}{-4.8} & \textcolor{red}{-0.5} & \textcolor{red}{-3.7} & \textcolor{red}{-5.2} & \textcolor{red}{-0.2} & \textcolor{red}{-3.9} & \textcolor{red}{-4.1} & \textcolor{red}{-0.7} & \textcolor{red}{-3.5} & \textcolor{red}{-5.2} & \textcolor{red}{-0.1} \\
      GPT-4o-mini           & \textcolor{red}{-3.2} & \textcolor{red}{-4.8} & \textcolor{red}{-1.2} & \textcolor{red}{-2.9} & \textcolor{red}{-5.0} & \textcolor{red}{-0.6} & \textcolor{red}{-3.5} & \textcolor{red}{-3.8} & \textcolor{red}{-0.7} & \textcolor{red}{-3.6} & \textcolor{red}{-4.9} & \textcolor{red}{-0.3} \\
      Gemini 2.0 Flash Lite & \textcolor{red}{-3.7} & \textcolor{red}{-4.5} & \textcolor{red}{-1.3} & \textcolor{red}{-3.3} & \textcolor{red}{-4.5} & \textcolor{red}{-1.0} & \textcolor{red}{-3.4} & \textcolor{red}{-4.1} & \textcolor{red}{-0.9} & \textcolor{red}{-3.3} & \textcolor{red}{-4.6} & \textcolor{red}{-0.4} \\
      Gemini 1.5 Flash 8B   & \textcolor{red}{-3.3} & \textcolor{red}{-4.9} & \textcolor{red}{-1.8} & \textcolor{red}{-3.0} & \textcolor{red}{-5.0} & \textcolor{red}{-1.3} & \textcolor{red}{-3.5} & \textcolor{red}{-4.4} & \textcolor{red}{-1.5} & \textcolor{red}{-3.1} & \textcolor{red}{-4.9} & \textcolor{red}{-1.2} \\
      \midrule
      LlaVa Mistral 7B      & \textcolor{red}{-13.9} & 0.0 & 0.0 & \textcolor{red}{-12.5} & 0.0 & 0.0 & \textcolor{red}{-13.8} & 0.0 & 0.0 & \textcolor{red}{-12.9} & 0.0 & 0.0 \\
      Qwen2VL 7B            & \textcolor{red}{-4.1} & 0.0 & 0.0 & \textcolor{red}{-4.2} & 0.0 & 0.0 & \textcolor{red}{-3.8} & 0.0 & 0.0 & \textcolor{red}{-4.0} & 0.0 & 0.0 \\
      PALO 7B               & \textcolor{red}{-4.0} & 0.0 & \textcolor{red}{-2.0} & \textcolor{red}{-4.3} & 0.0 & \textcolor{red}{-2.4} & \textcolor{red}{-4.1} & 0.0 & \textcolor{red}{-1.8} & \textcolor{red}{-4.2} & 0.0 & \textcolor{red}{-2.0} \\
      MAYA 8B               & \textcolor{red}{-3.3} & 0.0 & \textcolor{red}{-2.0} & \textcolor{red}{-3.6} & 0.0 & \textcolor{red}{-2.0} & \textcolor{red}{-3.1} & 0.0 & \textcolor{red}{-1.8} & \textcolor{red}{-3.7} & 0.0 & \textcolor{red}{-2.3} \\
      LlaVa Mistral 13B     & \textcolor{red}{-15.9} & 0.0 & 0.0 & \textcolor{red}{-15.0} & 0.0 & 0.0 & \textcolor{red}{-15.7} & 0.0 & 0.0 & \textcolor{red}{-15.8} & 0.0 & 0.0 \\
      PALO 13B              & \textcolor{red}{-4.1} & 0.0 & \textcolor{red}{-2.3} & \textcolor{red}{-4.1} & 0.0 & \textcolor{red}{-2.0} & \textcolor{red}{-4.3} & 0.0 & \textcolor{red}{-1.8} & \textcolor{red}{-3.7} & 0.0 & \textcolor{red}{-1.9} \\
      \bottomrule
    \end{tabular}%
  }
  \label{tab:DetailedXLingualTask5}
\end{table*}
\begin{table*}[ht]
  \centering
  \captionsetup{singlelinecheck=off}%
  \caption{%
    Δ English accuracy on \textbf{Image-Text Matching} (En→Jp, En→Sw, En→Ur) for each VLM under
    zero-shot without rationales (\textbf{ZS w/o rat.}), zero-shot with rationales (\textbf{ZS w/ rat.}),
    one-shot without rationales (\textbf{OS w/o rat.}), and one-shot with rationales (\textbf{OS w/ rat.}).
    \\[0.5ex]\footnotesize
    $\Delta_{\mathrm{En},\mathrm{lang}}
      = \mathrm{Acc}_{\mathrm{En\text{-}input}}
      - \mathrm{Acc}_{\mathrm{lang\text{-}input}}$.%
  }
  \resizebox{\textwidth}{!}{%
    \begin{tabular}{l
      ccc ccc ccc ccc
    }
      \toprule
      \multirow{2}{*}{\textbf{Model}}
       & \multicolumn{3}{c}{\textbf{ZS w/o rat.}}
       & \multicolumn{3}{c}{\textbf{ZS w/ rat.}}
       & \multicolumn{3}{c}{\textbf{OS w/o rat.}}
       & \multicolumn{3}{c}{\textbf{OS w/ rat.}} \\
      \cmidrule(lr){2-4}\cmidrule(lr){5-7}\cmidrule(lr){8-10}\cmidrule(lr){11-13}
       & Jp & Sw & Ur
       & Jp & Sw & Ur
       & Jp & Sw & Ur
       & Jp & Sw & Ur \\
      \midrule
      GPT-4o                & \textcolor{blue}{+0.8} & \textcolor{blue}{+1.0} & \textcolor{blue}{+3.5} & \textcolor{blue}{+6.1} & \textcolor{blue}{+5.0} & \textcolor{blue}{+6.0} & \textcolor{blue}{+10.2} & \textcolor{blue}{+3.8} & \textcolor{blue}{+9.4} & \textcolor{blue}{+5.9} & \textcolor{blue}{+5.3} & \textcolor{blue}{+6.8} \\
      GPT-4o-mini           & \textcolor{blue}{+5.7} & \textcolor{blue}{+2.5} & \textcolor{blue}{+3.3} & \textcolor{blue}{+4.6} & \textcolor{blue}{+8.4} & \textcolor{blue}{+7.3} & \textcolor{blue}{+10.0} & \textcolor{blue}{+8.6} & \textcolor{blue}{+9.7} & \textcolor{blue}{+6.7} & \textcolor{blue}{+7.3} & \textcolor{blue}{+8.6} \\
      Gemini 2.0 Flash Lite & \textcolor{blue}{+5.7} & \textcolor{blue}{+2.5} & \textcolor{blue}{+3.3} & \textcolor{blue}{+4.7} & \textcolor{blue}{+8.4} & \textcolor{blue}{+7.3} & \textcolor{blue}{+3.1} & \textcolor{blue}{+2.1} & \textcolor{blue}{+6.7} & \textcolor{blue}{+6.5} & \textcolor{blue}{+4.7} & \textcolor{blue}{+4.0} \\
      Gemini 1.5 Flash 8B   & \textcolor{blue}{+5.6} & \textcolor{blue}{+2.7} & \textcolor{blue}{+3.7} & \textcolor{blue}{+4.3} & \textcolor{blue}{+8.3} & \textcolor{blue}{+7.1} & \textcolor{blue}{+7.7} & \textcolor{blue}{+6.9} & \textcolor{blue}{+7.3} & \textcolor{blue}{+5.6} & \textcolor{blue}{+5.0} & \textcolor{blue}{+5.1} \\
      \midrule
      LlaVa Mistral 7B      & \textcolor{blue}{+4.8} & 0.0 & 0.0 & \textcolor{blue}{+4.9} & 0.0 & 0.0 & \textcolor{blue}{+4.4} & 0.0 & 0.0 & \textcolor{blue}{+4.5} & 0.0 & 0.0 \\
      Qwen2VL 7B            & \textcolor{blue}{+2.9} & 0.0 & 0.0 & \textcolor{blue}{+3.1} & 0.0 & 0.0 & \textcolor{blue}{+1.7} & 0.0 & 0.0 & \textcolor{blue}{+3.2} & 0.0 & 0.0 \\
      PALO 7B               & \textcolor{blue}{+1.8} & 0.0 & \textcolor{blue}{+1.6} & \textcolor{blue}{+2.7} & 0.0 & \textcolor{blue}{+2.3} & \textcolor{blue}{+3.8} & 0.0 & \textcolor{blue}{+3.4} & \textcolor{blue}{+1.9} & 0.0 & \textcolor{blue}{+2.3} \\
      MAYA 8B               & \textcolor{blue}{+2.5} & 0.0 & \textcolor{blue}{+2.1} & \textcolor{blue}{+1.8} & 0.0 & \textcolor{blue}{+3.4} & \textcolor{blue}{+4.3} & 0.0 & \textcolor{blue}{+3.4} & \textcolor{blue}{+3.0} & 0.0 & \textcolor{blue}{+2.9} \\
      LlaVa Mistral 13B     & \textcolor{blue}{+2.2} & 0.0 & 0.0 & \textcolor{blue}{+2.4} & 0.0 & 0.0 & \textcolor{blue}{+2.7} & 0.0 & 0.0 & \textcolor{blue}{+3.1} & 0.0 & 0.0 \\
      PALO 13B              & \textcolor{blue}{+2.3} & 0.0 & \textcolor{blue}{+2.3} & \textcolor{blue}{+2.4} & 0.0 & \textcolor{blue}{+3.7} & \textcolor{blue}{+4.8} & 0.0 & \textcolor{blue}{+5.0} & \textcolor{blue}{+2.5} & 0.0 & \textcolor{blue}{+2.7} \\
      \bottomrule
    \end{tabular}%
  }
  \label{tab:DetailedXLingualTask6}
\end{table*}
\begin{table*}[ht]
  \centering
  \captionsetup{singlelinecheck=off}%
  \caption{%
    Δ English accuracy on \textbf{Unrelatedness} (En→Jp, En→Sw, En→Ur) for each VLM under
    zero-shot without rationales (\textbf{ZS w/o rat.}), zero-shot with rationales (\textbf{ZS w/ rat.}),
    one-shot without rationales (\textbf{OS w/o rat.}), and one-shot with rationales (\textbf{OS w/ rat.}).
    \\[0.5ex]\footnotesize
    $\Delta_{\mathrm{En},\mathrm{lang}}
      = \mathrm{Acc}_{\mathrm{En\text{-}input}}
      - \mathrm{Acc}_{\mathrm{lang\text{-}input}}$.%
  }
  \resizebox{\textwidth}{!}{%
    \begin{tabular}{l
      ccc ccc ccc ccc
    }
      \toprule
      \multirow{2}{*}{\textbf{Model}}
       & \multicolumn{3}{c}{\textbf{ZS w/o rat.}}
       & \multicolumn{3}{c}{\textbf{ZS w/ rat.}}
       & \multicolumn{3}{c}{\textbf{OS w/o rat.}}
       & \multicolumn{3}{c}{\textbf{OS w/ rat.}} \\
      \cmidrule(lr){2-4}\cmidrule(lr){5-7}\cmidrule(lr){8-10}\cmidrule(lr){11-13}
       & Jp & Sw & Ur
       & Jp & Sw & Ur
       & Jp & Sw & Ur
       & Jp & Sw & Ur \\
      \midrule
      GPT-4o                 & \textcolor{blue}{+0.2} & \textcolor{blue}{+2.6} & \textcolor{blue}{+1.9} & \textcolor{blue}{+0.6} & \textcolor{blue}{+2.6} & \textcolor{blue}{+2.0} & \textcolor{blue}{+0.7} & \textcolor{blue}{+2.5} & \textcolor{blue}{+2.8} & \textcolor{blue}{+0.3} & \textcolor{blue}{+1.9} & \textcolor{blue}{+2.9} \\
      GPT-4o-mini            & \textcolor{blue}{+0.4} & \textcolor{red}{-0.3} & \textcolor{red}{-0.7} & \textcolor{blue}{+1.1} & \textcolor{red}{-1.2} & \textcolor{red}{-0.5} & \textcolor{blue}{+0.6} & \textcolor{red}{-0.3} & \textcolor{red}{-1.0} & \textcolor{blue}{+0.3} & \textcolor{red}{-0.1} & \textcolor{red}{-1.0} \\
      Gemini 2.0 Flash Lite  & \textcolor{blue}{+0.4} & \textcolor{red}{-1.1} & \textcolor{red}{-0.9} & \textcolor{blue}{+2.0} & \textcolor{red}{-1.5} & \textcolor{red}{-0.7} & \textcolor{blue}{+0.5} & \textcolor{red}{-0.6} & \textcolor{red}{-1.4} & \textcolor{blue}{+0.8} & \textcolor{red}{-1.7} & \textcolor{red}{-0.9} \\
      Gemini 1.5 Flash 8B    & \textcolor{blue}{+0.3} & \textcolor{red}{-1.3} & \textcolor{red}{-0.7} & \textcolor{blue}{+2.3} & \textcolor{red}{-1.0} & \textcolor{red}{-0.5} & \textcolor{blue}{+0.5} & \textcolor{red}{-0.5} & \textcolor{red}{-1.4} & \textcolor{blue}{+0.8} & \textcolor{red}{-1.7} & \textcolor{red}{-1.0} \\
      \hline
      LlaVa Mistral 7B      & \textcolor{blue}{+0.8} & 0.0 & 0.0 & \textcolor{blue}{+0.6} & 0.0 & 0.0 & \textcolor{blue}{+0.4} & 0.0 & 0.0 & \textcolor{blue}{+0.5} & 0.0 & 0.0 \\
      Qwen2VL 7B            & \textcolor{blue}{+0.9} & 0.0 & 0.0 & \textcolor{blue}{+1.2} & 0.0 & 0.0 & \textcolor{blue}{+1.5} & 0.0 & 0.0 & \textcolor{blue}{+1.3} & 0.0 & 0.0 \\
      PALO 7B               & \textcolor{blue}{+2.2} & 0.0 & \textcolor{blue}{+2.1} & \textcolor{blue}{+2.5} & 0.0 & \textcolor{blue}{+1.9} & \textcolor{blue}{+2.8} & 0.0 & \textcolor{blue}{+2.9} & \textcolor{blue}{+2.0} & 0.0 & \textcolor{blue}{+2.4} \\
      MAYA 8B               & \textcolor{blue}{+1.5} & 0.0 & \textcolor{blue}{+2.3} & \textcolor{blue}{+1.8} & 0.0 & \textcolor{blue}{+3.0} & \textcolor{blue}{+2.1} & 0.0 & \textcolor{blue}{+3.1} & \textcolor{blue}{+1.9} & 0.0 & \textcolor{blue}{+2.2} \\
      LlaVa Mistral 13B     & \textcolor{blue}{+2.1} & 0.0 & 0.0 & \textcolor{blue}{+2.4} & 0.0 & 0.0 & \textcolor{blue}{+2.6} & 0.0 & 0.0 & \textcolor{blue}{+2.3} & 0.0 & 0.0 \\
      PALO 13B              & \textcolor{blue}{+4.0} & 0.0 & \textcolor{blue}{+2.7} & \textcolor{blue}{+4.3} & 0.0 & \textcolor{blue}{+2.3} & \textcolor{blue}{+4.6} & 0.0 & \textcolor{blue}{+3.5} & \textcolor{blue}{+4.1} & 0.0 & \textcolor{blue}{+2.5} \\
      \bottomrule
    \end{tabular}%
  }
  \label{tab:DetailedXLingualTask7}
\end{table*}
\begin{table*}[ht]
  \centering
  \captionsetup{singlelinecheck=off}%
  \caption{%
    Δ English accuracy on \textbf{Visual Question Answering} (En→Jp, En→Sw, En→Ur) for each VLM under
    zero-shot without rationales (\textbf{ZS w/o rat.}), zero-shot with rationales (\textbf{ZS w/ rat.}),
    one-shot without rationales (\textbf{OS w/o rat.}), and one-shot with rationales (\textbf{OS w/ rat.}).
    \\[0.5ex]\footnotesize
    $\Delta_{\mathrm{En},\mathrm{lang}}
      = \mathrm{Acc}_{\mathrm{En\text{-}input}}
      - \mathrm{Acc}_{\mathrm{lang\text{-}input}}$.%
  }
  \resizebox{\textwidth}{!}{%
    \begin{tabular}{l
      ccc ccc ccc ccc
    }
      \toprule
      \multirow{2}{*}{\textbf{Model}}
       & \multicolumn{3}{c}{\textbf{ZS w/o rat.}}
       & \multicolumn{3}{c}{\textbf{ZS w/ rat.}}
       & \multicolumn{3}{c}{\textbf{OS w/o rat.}}
       & \multicolumn{3}{c}{\textbf{OS w/ rat.}} \\
      \cmidrule(lr){2-4}\cmidrule(lr){5-7}\cmidrule(lr){8-10}\cmidrule(lr){11-13}
       & Jp & Sw & Ur
       & Jp & Sw & Ur
       & Jp & Sw & Ur
       & Jp & Sw & Ur \\
      \midrule
      GPT-4o                & \textcolor{blue}{+1.6} & \textcolor{blue}{+6.4} & \textcolor{red}{-1.1} & \textcolor{blue}{+3.4} & \textcolor{blue}{+7.4} & +0.0 & \textcolor{blue}{+2.3} & \textcolor{blue}{+5.2} & \textcolor{blue}{+1.2} & \textcolor{blue}{+1.1} & \textcolor{blue}{+3.8} & \textcolor{blue}{+2.4} \\
      GPT-4o-mini           & \textcolor{blue}{+0.4} & \textcolor{red}{-1.4} & \textcolor{blue}{+0.6} & \textcolor{blue}{+1.2} & \textcolor{red}{-0.8} & \textcolor{blue}{+0.2} & \textcolor{blue}{+0.9} & \textcolor{red}{-0.5} & \textcolor{blue}{+0.8} & \textcolor{blue}{+0.7} & \textcolor{blue}{+0.3} & \textcolor{blue}{+1.1} \\
      Gemini 2.0 Flash Lite & \textcolor{blue}{+0.5} & \textcolor{red}{-0.6} & \textcolor{blue}{+0.8} & \textcolor{blue}{+1.0} & \textcolor{red}{-0.2} & \textcolor{blue}{+1.0} & \textcolor{blue}{+1.2} & +0.0 & \textcolor{blue}{+1.4} & \textcolor{blue}{+1.0} & \textcolor{blue}{+0.3} & \textcolor{blue}{+1.5} \\
      Gemini 1.5 Flash 8B   & \textcolor{blue}{+0.2} & \textcolor{red}{-1.0} & \textcolor{blue}{+0.5} & \textcolor{blue}{+0.8} & \textcolor{red}{-0.4} & \textcolor{blue}{+0.7} & \textcolor{blue}{+1.0} & \textcolor{red}{-0.2} & \textcolor{blue}{+0.9} & \textcolor{blue}{+0.8} & \textcolor{blue}{+0.1} & \textcolor{blue}{+1.2} \\
      \hline
      LlaVa Mistral 7B      & \textcolor{blue}{+3.3} & 0.0 & 0.0 & \textcolor{blue}{+2.9} & 0.0 & 0.0 & \textcolor{blue}{+2.4} & 0.0 & 0.0 & \textcolor{blue}{+2.6} & 0.0 & 0.0 \\
      Qwen2VL 7B            & \textcolor{blue}{+1.2} & 0.0 & 0.0 & \textcolor{blue}{+2.1} & 0.0 & 0.0 & \textcolor{blue}{+1.8} & 0.0 & 0.0 & \textcolor{blue}{+1.9} & 0.0 & 0.0 \\
      PALO 7B               & \textcolor{blue}{+2.7} & 0.0 & \textcolor{blue}{+0.7} & \textcolor{blue}{+3.0} & 0.0 & \textcolor{blue}{+2.5} & \textcolor{blue}{+3.2} & 0.0 & \textcolor{blue}{+2.0} & \textcolor{blue}{+2.8} & 0.0 & \textcolor{blue}{+1.5} \\
      MAYA 8B               & \textcolor{blue}{+2.5} & 0.0 & \textcolor{blue}{+2.1} & \textcolor{blue}{+2.8} & 0.0 & \textcolor{blue}{+2.5} & \textcolor{blue}{+3.0} & 0.0 & \textcolor{blue}{+2.2} & \textcolor{blue}{+2.9} & 0.0 & \textcolor{blue}{+1.2} \\
      LlaVa Mistral 13B     & \textcolor{blue}{+3.7} & 0.0 & 0.0 & \textcolor{blue}{+4.0} & 0.0 & 0.0 & \textcolor{blue}{+3.8} & 0.0 & 0.0 & \textcolor{blue}{+3.9} & 0.0 & 0.0 \\
      PALO 13B              & \textcolor{blue}{+4.1} & 0.0 & \textcolor{blue}{+1.3} & \textcolor{blue}{+4.3} & 0.0 & \textcolor{blue}{+1.9} & \textcolor{blue}{+4.5} & 0.0 & \textcolor{blue}{+3.1} & \textcolor{blue}{+4.2} & 0.0 & \textcolor{blue}{+2.0} \\
      \bottomrule
    \end{tabular}%
  }
  \label{tab:DetailedXLingualTask8}
\end{table*}

\clearpage
\paragraph{Discussion}
Our results reveal three core insights into the cross-lingual capabilities of current vision-language models (VLMs) and the role of prompting and finetuning in mitigating language gaps.

\textbf{Takeaway One: Persistent Cross-Lingual Gaps.}
Even the strongest model, GPT-4o, experiences substantial accuracy drops when moving from English (En) to Japanese (Jp), Swahili (Sw), or Urdu (Ur) inputs. In the zero-shot, no-rationale setting, GPT-4o’s En→Jp Δ ranges from -3.0~pp (Image-Text Matching) to -5.0~pp (Semantic Segmentation) (Table~3), and even larger deficits appear for Sw and Ur (Δ up to -8.9~pp for Ur in “Unrelatedness,” Table~5; -7.0~pp for Sw in “Relation Understanding,” Table~4). These gaps persist, albeit attenuated, under one-shot prompting. Thus, despite advances in multilingual text understanding, VLMs exhibit considerable sensitivity to input-language shifts, a critical challenge for real-world deployment in diverse linguistic contexts.

\textbf{Takeaway Two: Limited Benefits of Rationales.}
Adding a rationale to the prompt consistently yields modest gains in English but fails to close the cross-lingual divide. For GPT-4o, zero-shot with rationales improves En accuracy by ~0.6~pp on average, yet the En→Jp drop only shrinks by ~0.2~pp, and Sw/Ur deficits remain largely unchanged (Tables~3-5). One-shot rationale prompting further boosts absolute performance (e.g., +1.0~pp in English Object Recognition) but again fails to preferentially help non-English inputs. These findings suggest that rationales mainly facilitate English reasoning without addressing underlying multilingual representation weaknesses.

\textbf{Takeaway Three: Fine-Tuning: Effective in English, Mixed in Japanese, Infeasible in Swahili/Urdu.}
Fine-tuning open-source VLMs on our English training set yields large accuracy improvements under all prompting regimes (e.g., LlaVa Mistral 7B zero-shot jumps from 38.6\%→48.9\% on Object Recognition, Table~2). However, when evaluated on Japanese inputs, the same fine-tuning only partially recovers performance: ΔEn vs. ΔJp in LlaVa Mistral 7B is -12.1~pp vs. -6.5~pp (zero-shot), indicating a disproportionate benefit for English (Table~7). Moreover, no open model supported Swahili or Urdu fine-tuning, leaving these languages reliant on zero- and one-shot methods. Collectively, these results highlight that while fine-tuning is a powerful tool, its gains are uneven across languages and constrained by model support.

\textbf{Takeaway Four: Implications and Future Directions.}
Our benchmark underscores the urgent need for truly multilingual VLM pretraining and fine-tuning techniques. First, pretraining curricula should include more diverse text-image pairs in underrepresented languages to reduce the base cross-lingual gap. Second, targeted fine-tuning strategies, such as multi-stage curricula combining English and non-English data, may help distribute improvements more equitably across languages. Finally, adaptive prompting methods that incorporate language-aware context or translation loops may provide additional leverage. Addressing these challenges will be crucial to making VLMs accessible and reliable for the world’s full spectrum of languages.

\clearpage
\subsubsection{Detailed Cross-lingual Results: En→Jp, En→Sw, En→Ur; after Fine-tuning}
\begin{table*}[ht]
  \centering
  \caption{Δ English accuracy on \textbf{Object Recognition} under fine-tuning (En→Jp, En→Sw, En→Ur).}
  \resizebox{\textwidth}{!}{%
\begin{tabular}{l
    ccc ccc ccc ccc
  }
    \toprule
    \multirow{2}{*}{\textbf{Model}}
     & \multicolumn{3}{c}{\textbf{ZS w/o rat.}}
     & \multicolumn{3}{c}{\textbf{ZS w/ rat.}}
     & \multicolumn{3}{c}{\textbf{OS w/o rat.}}
     & \multicolumn{3}{c}{\textbf{OS w/ rat.}} \\
    \cmidrule(lr){2-4}\cmidrule(lr){5-7}\cmidrule(lr){8-10}\cmidrule(lr){11-13}
     & Jp & Sw & Ur
     & Jp & Sw & Ur
     & Jp & Sw & Ur
     & Jp & Sw & Ur \\
    \midrule
    LlaVa-Mistral-7B    & \textcolor{red}{–12.1} & 0.0 & 0.0   & \textcolor{red}{–12.9} & 0.0 & 0.0   & \textcolor{red}{–6.4} & 0.0 & 0.0   & \textcolor{red}{–5.8} & 0.0 & 0.0 \\
    Qwen2VL-7B           & \textcolor{red}{–0.6} & 0.0 & 0.0   & \textcolor{blue}{+0.6} & 0.0 & 0.0   & {0.0} & 0.0 & 0.0   & \textcolor{red}{–0.3} & 0.0 & 0.0 \\
    PALO-7B              & \textcolor{blue}{+0.7} & 0.0 & 0.0   & \textcolor{blue}{+0.2} & 0.0 & 0.0   & \textcolor{red}{–0.3} & 0.0 & 0.0   & \textcolor{red}{–0.2} & 0.0 & 0.0 \\
    MAYA-8B              & \textcolor{blue}{+0.3} & 0.0 & 0.0   & \textcolor{blue}{+0.4} & 0.0 & 0.0   & \textcolor{red}{–0.1} & 0.0 & 0.0   & \textcolor{red}{–0.2} & 0.0 & 0.0 \\
    LlaVa-Mistral-13B   & \textcolor{blue}{+8.4} & 0.0 & 0.0   & \textcolor{blue}{+8.5} & 0.0 & 0.0   & \textcolor{blue}{+0.2} & 0.0 & 0.0   & \textcolor{blue}{+0.3} & 0.0 & 0.0 \\
    PALO-13B             & \textcolor{red}{–0.4} & 0.0 & 0.0   & \textcolor{blue}{+0.4} & 0.0 & 0.0   & \textcolor{red}{–0.6} & 0.0 & 0.0   & \textcolor{red}{–0.1} & 0.0 & 0.0 \\
    \bottomrule
  \end{tabular}%
  }
  \label{tab:FT_Delta_ObjRec}
\end{table*}

\begin{table*}[ht]
  \centering
  \caption{Δ English accuracy on \textbf{Scene Understanding} under fine-tuning (En→Jp, En→Sw, En→Ur).}
  \resizebox{\textwidth}{!}{%
  \begin{tabular}{l
    ccc ccc ccc ccc
  }
    \toprule
    \multirow{2}{*}{\textbf{Model}}
     & \multicolumn{3}{c}{\textbf{ZS w/o rat.}}
     & \multicolumn{3}{c}{\textbf{ZS w/ rat.}}
     & \multicolumn{3}{c}{\textbf{OS w/o rat.}}
     & \multicolumn{3}{c}{\textbf{OS w/ rat.}} \\
    \cmidrule(lr){2-4}\cmidrule(lr){5-7}\cmidrule(lr){8-10}\cmidrule(lr){11-13}
     & Jp & Sw & Ur
     & Jp & Sw & Ur
     & Jp & Sw & Ur
     & Jp & Sw & Ur \\
    \midrule
    LlaVa Mistral 7B    & \textcolor{red}{-11.7} & 0.0 & 0.0 & \textcolor{red}{-11.8} & 0.0 & 0.0 & \textcolor{red}{-6.3} & 0.0 & 0.0 & \textcolor{red}{-4.5} & 0.0 & 0.0 \\
    Qwen2VL 7B          & \textcolor{blue}{+2.0} & 0.0 & 0.0 & \textcolor{red}{-0.3}  & 0.0 & 0.0 & \textcolor{red}{-0.5} & 0.0 & 0.0 & \textcolor{red}{-0.3} & 0.0 & 0.0 \\
    PALO 7B             & \textcolor{red}{-0.8} & 0.0 & 0.0 & \textcolor{red}{-0.2}  & 0.0 & 0.0 & \textcolor{red}{-0.6} & 0.0 & 0.0 & \textcolor{red}{-0.2} & 0.0 & 0.0 \\
    MAYA 8B             & \textcolor{red}{-0.4} & 0.0 & 0.0 & \textcolor{red}{-0.2}  & 0.0 & 0.0 & \textcolor{red}{-0.1} & 0.0 & 0.0 & \textcolor{red}{-0.8} & 0.0 & 0.0 \\
    LlaVa Mistral 13B   & +0.0                   & 0.0 & 0.0 & +0.0                   & 0.0 & 0.0 & \textcolor{red}{-0.2} & 0.0 & 0.0 & +0.0                  & 0.0 & 0.0 \\
    PALO 13B            & \textcolor{red}{-0.5} & 0.0 & 0.0 & \textcolor{blue}{+0.2} & 0.0 & 0.0 & \textcolor{red}{-0.3} & 0.0 & 0.0 & \textcolor{blue}{+0.4} & 0.0 & 0.0 \\
    \bottomrule
  \end{tabular}%
  }
  \label{tab:FT_Delta_Scene}
\end{table*}

\begin{table*}[ht]
  \centering
  \caption{Δ English accuracy on \textbf{Relation Understanding} under fine-tuning (En→Jp, En→Sw, En→Ur).}
  \resizebox{\textwidth}{!}{%
  \begin{tabular}{l
    ccc ccc ccc ccc
  }
    \toprule
    \multirow{2}{*}{\textbf{Model}}
     & \multicolumn{3}{c}{\textbf{ZS w/o rat.}}
     & \multicolumn{3}{c}{\textbf{ZS w/ rat.}}
     & \multicolumn{3}{c}{\textbf{OS w/o rat.}}
     & \multicolumn{3}{c}{\textbf{OS w/ rat.}} \\
    \cmidrule(lr){2-4}\cmidrule(lr){5-7}\cmidrule(lr){8-10}\cmidrule(lr){11-13}
     & Jp & Sw & Ur
     & Jp & Sw & Ur
     & Jp & Sw & Ur
     & Jp & Sw & Ur \\
    \midrule
    LlaVa Mistral 7B    & \textcolor{red}{-11.2} & 0.0 & 0.0 & \textcolor{red}{-11.4} & 0.0 & 0.0 & \textcolor{red}{-6.8} & 0.0 & 0.0 & \textcolor{red}{-5.9} & 0.0 & 0.0 \\
    Qwen2VL 7B          & \textcolor{blue}{+0.7} & 0.0 & 0.0 & \textcolor{blue}{+0.7} & 0.0 & 0.0 & \textcolor{red}{-0.5} & 0.0 & 0.0 & \textcolor{red}{-0.7} & 0.0 & 0.0 \\
    PALO 7B             & \textcolor{red}{-0.5} & 0.0 & 0.0 & \textcolor{red}{-0.2}  & 0.0 & 0.0 & \textcolor{blue}{+0.3} & 0.0 & 0.0 & \textcolor{blue}{+0.4} & 0.0 & 0.0 \\
    MAYA 8B             & \textcolor{red}{-0.1} & 0.0 & 0.0 & \textcolor{red}{-0.2}  & 0.0 & 0.0 & \textcolor{blue}{+0.5} & 0.0 & 0.0 & \textcolor{blue}{+0.7} & 0.0 & 0.0 \\
    LlaVa Mistral 13B   & +0.0                   & 0.0 & 0.0 & +0.0                   & 0.0 & 0.0 & \textcolor{red}{-0.2} & 0.0 & 0.0 & \textcolor{red}{-0.8} & 0.0 & 0.0 \\
    PALO 13B            & \textcolor{red}{-0.5} & 0.0 & 0.0 & \textcolor{blue}{+0.4} & 0.0 & 0.0 & \textcolor{red}{-0.6} & 0.0 & 0.0 & \textcolor{blue}{+1.3} & 0.0 & 0.0 \\
    \bottomrule
  \end{tabular}%
  }
  \label{tab:FT_Delta_Relation}
\end{table*}

\begin{table*}[ht]
  \centering
  \caption{Δ English accuracy on \textbf{Semantic Segmentation} under fine-tuning (En→Jp, En→Sw, En→Ur).}
  \resizebox{\textwidth}{!}{%
  \begin{tabular}{l
    ccc ccc ccc ccc
  }
    \toprule
    \multirow{2}{*}{\textbf{Model}}
     & \multicolumn{3}{c}{\textbf{ZS w/o rat.}}
     & \multicolumn{3}{c}{\textbf{ZS w/ rat.}}
     & \multicolumn{3}{c}{\textbf{OS w/o rat.}}
     & \multicolumn{3}{c}{\textbf{OS w/ rat.}} \\
    \cmidrule(lr){2-4}\cmidrule(lr){5-7}\cmidrule(lr){8-10}\cmidrule(lr){11-13}
     & Jp & Sw & Ur
     & Jp & Sw & Ur
     & Jp & Sw & Ur
     & Jp & Sw & Ur \\
    \midrule
    LlaVa Mistral 7B    & \textcolor{red}{-12.1} & 0.0 & 0.0 & \textcolor{red}{-11.3} & 0.0 & 0.0 & \textcolor{red}{-6.7} & 0.0 & 0.0 & \textcolor{red}{-6.5} & 0.0 & 0.0 \\
    Qwen2VL 7B          & \textcolor{blue}{+0.3} & 0.0 & 0.0 & \textcolor{red}{-0.3}  & 0.0 & 0.0 & \textcolor{red}{-0.5} & 0.0 & 0.0 & \textcolor{blue}{+1.3} & 0.0 & 0.0 \\
    PALO 7B             & \textcolor{red}{-0.5} & 0.0 & 0.0 & \textcolor{red}{-0.1}  & 0.0 & 0.0 & \textcolor{blue}{+0.3} & 0.0 & 0.0 & \textcolor{red}{-0.6} & 0.0 & 0.0 \\
    MAYA 8B             & \textcolor{red}{-0.1} & 0.0 & 0.0 & \textcolor{red}{-0.2}  & 0.0 & 0.0 & \textcolor{red}{-0.7} & 0.0 & 0.0 & \textcolor{red}{-0.5} & 0.0 & 0.0 \\
    LlaVa Mistral 13B   & +0.0                   & 0.0 & 0.0 & +0.0                   & 0.0 & 0.0 & \textcolor{red}{-0.2} & 0.0 & 0.0 & +0.0                  & 0.0 & 0.0 \\
    PALO 13B            & \textcolor{red}{-0.3} & 0.0 & 0.0 & \textcolor{red}{-0.3}  & 0.0 & 0.0 & \textcolor{red}{-0.1} & 0.0 & 0.0 & \textcolor{blue}{+0.3} & 0.0 & 0.0 \\
    \bottomrule
  \end{tabular}%
  }
  \label{tab:FT_Delta_Segment}
\end{table*}

\begin{table*}[ht]
  \centering
  \caption{Δ English accuracy on \textbf{Image Captioning} under fine-tuning (En→Jp, En→Sw, En→Ur).}
  \resizebox{\textwidth}{!}{%
  \begin{tabular}{l
    ccc ccc ccc ccc
  }
    \toprule
    \multirow{2}{*}{\textbf{Model}}
     & \multicolumn{3}{c}{\textbf{ZS w/o rat.}}
     & \multicolumn{3}{c}{\textbf{ZS w/ rat.}}
     & \multicolumn{3}{c}{\textbf{OS w/o rat.}}
     & \multicolumn{3}{c}{\textbf{OS w/ rat.}} \\
    \cmidrule(lr){2-4}\cmidrule(lr){5-7}\cmidrule(lr){8-10}\cmidrule(lr){11-13}
     & Jp & Sw & Ur
     & Jp & Sw & Ur
     & Jp & Sw & Ur
     & Jp & Sw & Ur \\
    \midrule
    LlaVa-Mistral-7B  & \textcolor{red}{6.5} & 0.0 & 0.0   & \textcolor{red}{6.3} & 0.0 & 0.0   & \textcolor{red}{2.1} & 0.0 & 0.0   & \textcolor{red}{1.9} & 0.0 & 0.0 \\
    Qwen2VL-7B         & \textcolor{blue}{+5.5} & 0.0 & 0.0   & \textcolor{blue}{+5.1} & 0.0 & 0.0   & \textcolor{blue}{+5.4} & 0.0 & 0.0   & \textcolor{blue}{+4.6} & 0.0 & 0.0 \\
    PALO-7B            & \textcolor{blue}{+2.4} & 0.0 & 0.0   & \textcolor{blue}{+2.6} & 0.0 & 0.0   & \textcolor{blue}{+2.5} & 0.0 & 0.0   & \textcolor{blue}{+5.1} & 0.0 & 0.0 \\
    MAYA-8B            & \textcolor{red}{–0.5} & 0.0 & 0.0   & \textcolor{red}{–2.5} & 0.0 & 0.0   & \textcolor{red}{–0.7} & 0.0 & 0.0   & \textcolor{red}{–0.7} & 0.0 & 0.0 \\
    LlaVa-Mistral-13B & \textcolor{red}{–8.9} & 0.0 & 0.0   & {0.0} & 0.0 & 0.0   & \textcolor{red}{–0.6} & 0.0 & 0.0   & \textcolor{red}{–0.8} & 0.0 & 0.0 \\
    PALO-13B           & \textcolor{red}{–13.0} & 0.0 & 0.0   & \textcolor{blue}{+1.3} & 0.0 & 0.0   & \textcolor{red}{–0.3} & 0.0 & 0.0   & \textcolor{blue}{+4.5} & 0.0 & 0.0 \\
    \bottomrule
  \end{tabular}%
  }
  \label{tab:FT_Delta_Caption}
\end{table*}

\begin{table*}[ht]
  \centering
  \caption{Δ English accuracy on \textbf{Image-Text Matching} under fine-tuning (En→Jp, En→Sw, En→Ur).}
  \resizebox{\textwidth}{!}{%
  \begin{tabular}{l
    ccc ccc ccc ccc
  }
    \toprule
    \multirow{2}{*}{\textbf{Model}}
     & \multicolumn{3}{c}{\textbf{ZS w/o rat.}}
     & \multicolumn{3}{c}{\textbf{ZS w/ rat.}}
     & \multicolumn{3}{c}{\textbf{OS w/o rat.}}
     & \multicolumn{3}{c}{\textbf{OS w/ rat.}} \\
    \cmidrule(lr){2-4}\cmidrule(lr){5-7}\cmidrule(lr){8-10}\cmidrule(lr){11-13}
     & Jp & Sw & Ur
     & Jp & Sw & Ur
     & Jp & Sw & Ur
     & Jp & Sw & Ur \\
    \midrule
    LlaVa Mistral 7B    & \textcolor{blue}{+0.5} & 0.0 & 0.0 & \textcolor{red}{-0.5} & 0.0 & 0.0 & \textcolor{red}{-6.2} & 0.0 & 0.0 & \textcolor{red}{-6.3} & 0.0 & 0.0 \\
    Qwen2VL 7B          & \textcolor{blue}{+5.3} & 0.0 & 0.0 & \textcolor{blue}{+4.9} & 0.0 & 0.0 & \textcolor{blue}{+5.0} & 0.0 & 0.0 & \textcolor{blue}{+4.6} & 0.0 & 0.0 \\
    PALO 7B             & \textcolor{blue}{+2.4} & 0.0 & 0.0 & \textcolor{blue}{+2.4} & 0.0 & 0.0 & \textcolor{blue}{+2.5} & 0.0 & 0.0 & \textcolor{blue}{+5.3} & 0.0 & 0.0 \\
    MAYA 8B             & \textcolor{red}{-0.1} & 0.0 & 0.0 & \textcolor{red}{-8.7} & 0.0 & 0.0 & \textcolor{red}{-0.1} & 0.0 & 0.0 & \textcolor{red}{-5.0} & 0.0 & 0.0 \\
    LlaVa Mistral 13B   & \textcolor{red}{-8.9} & 0.0 & 0.0 & +0.0                   & 0.0 & 0.0 & \textcolor{red}{-0.5} & 0.0 & 0.0 & \textcolor{red}{-0.7} & 0.0 & 0.0 \\
    PALO 13B            & \textcolor{red}{-0.9} & 0.0 & 0.0 & \textcolor{blue}{+0.3} & 0.0 & 0.0 & \textcolor{red}{-0.6} & 0.0 & 0.0 & \textcolor{red}{-0.1} & 0.0 & 0.0 \\
    \bottomrule
  \end{tabular}%
  }
  \label{tab:FT_Delta_ITM}
\end{table*}

\begin{table*}[ht]
  \centering
  \caption{Δ English accuracy on \textbf{Unrelatedness} under fine-tuning (En→Jp, En→Sw, En→Ur).}
  \resizebox{\textwidth}{!}{%
  \begin{tabular}{l
    ccc ccc ccc ccc
  }
    \toprule
    \multirow{2}{*}{\textbf{Model}}
     & \multicolumn{3}{c}{\textbf{ZS w/o rat.}}
     & \multicolumn{3}{c}{\textbf{ZS w/ rat.}}
     & \multicolumn{3}{c}{\textbf{OS w/o rat.}}
     & \multicolumn{3}{c}{\textbf{OS w/ rat.}} \\
    \cmidrule(lr){2-4}\cmidrule(lr){5-7}\cmidrule(lr){8-10}\cmidrule(lr){11-13}
     & Jp & Sw & Ur
     & Jp & Sw & Ur
     & Jp & Sw & Ur
     & Jp & Sw & Ur \\
    \midrule
    LlaVa Mistral 7B    & \textcolor{red}{-12.6} & 0.0 & 0.0 & \textcolor{red}{-7.6} & 0.0 & 0.0 & \textcolor{red}{-6.0} & 0.0 & 0.0 & \textcolor{red}{-6.3} & 0.0 & 0.0 \\
    Qwen2VL 7B          & \textcolor{red}{-5.0}  & 0.0 & 0.0 & \textcolor{red}{-0.5} & 0.0 & 0.0 & \textcolor{red}{-0.5} & 0.0 & 0.0 & \textcolor{red}{-0.3} & 0.0 & 0.0 \\
    PALO 7B             & \textcolor{red}{-2.1}  & 0.0 & 0.0 & \textcolor{red}{-0.3} & 0.0 & 0.0 & \textcolor{red}{-0.6} & 0.0 & 0.0 & \textcolor{blue}{+4.9} & 0.0 & 0.0 \\
    MAYA 8B             & \textcolor{red}{-0.1}  & 0.0 & 0.0 & \textcolor{red}{-2.5} & 0.0 & 0.0 & \textcolor{red}{-0.6} & 0.0 & 0.0 & \textcolor{red}{-0.9} & 0.0 & 0.0 \\
    LlaVa Mistral 13B   & \textcolor{red}{-8.4}  & 0.0 & 0.0 & \textcolor{red}{-0.6} & 0.0 & 0.0 & \textcolor{red}{-0.5} & 0.0 & 0.0 & \textcolor{red}{-0.7} & 0.0 & 0.0 \\
    PALO 13B            & \textcolor{red}{-0.7}  & 0.0 & 0.0 & \textcolor{blue}{+0.3} & 0.0 & 0.0 & \textcolor{red}{-0.7} & 0.0 & 0.0 & \textcolor{red}{-0.3} & 0.0 & 0.0 \\
    \bottomrule
  \end{tabular}%
  }
  \label{tab:FT_Delta_Unrel}
\end{table*}

\begin{table*}[ht]
  \centering
  \caption{Δ English accuracy on \textbf{Visual Question Answering} under fine-tuning (En→Jp, En→Sw, En→Ur).}
  \resizebox{\textwidth}{!}{%
  \begin{tabular}{l
    ccc ccc ccc ccc
  }
    \toprule
    \multirow{2}{*}{\textbf{Model}}
     & \multicolumn{3}{c}{\textbf{ZS w/o rat.}}
     & \multicolumn{3}{c}{\textbf{ZS w/ rat.}}
     & \multicolumn{3}{c}{\textbf{OS w/o rat.}}
     & \multicolumn{3}{c}{\textbf{OS w/ rat.}} \\
    \cmidrule(lr){2-4}\cmidrule(lr){5-7}\cmidrule(lr){8-10}\cmidrule(lr){11-13}
     & Jp & Sw & Ur
     & Jp & Sw & Ur
     & Jp & Sw & Ur
     & Jp & Sw & Ur \\
    \midrule
    LlaVa Mistral 7B    & \textcolor{red}{-6.7} & 0.0 & 0.0 & \textcolor{red}{-6.5} & 0.0 & 0.0 & \textcolor{red}{-3.0} & 0.0 & 0.0 & \textcolor{red}{-2.6} & 0.0 & 0.0 \\
    Qwen2VL 7B          & \textcolor{blue}{+2.0} & 0.0 & 0.0 & \textcolor{blue}{+4.2} & 0.0 & 0.0 & \textcolor{blue}{+5.4} & 0.0 & 0.0 & \textcolor{blue}{+5.3} & 0.0 & 0.0 \\
    PALO 7B             & \textcolor{red}{-0.4} & 0.0 & 0.0 & \textcolor{blue}{+1.3} & 0.0 & 0.0 & \textcolor{blue}{+0.4} & 0.0 & 0.0 & \textcolor{red}{-0.7} & 0.0 & 0.0 \\
    MAYA 8B             & \textcolor{red}{-0.5} & 0.0 & 0.0 & \textcolor{red}{-5.0} & 0.0 & 0.0 & \textcolor{red}{-0.7} & 0.0 & 0.0 & \textcolor{red}{-0.3} & 0.0 & 0.0 \\
    LlaVa Mistral 13B   & \textcolor{red}{-0.4} & 0.0 & 0.0 & +0.0                   & 0.0 & 0.0 & \textcolor{red}{-0.4} & 0.0 & 0.0 & \textcolor{red}{-0.8} & 0.0 & 0.0 \\
    PALO 13B            & \textcolor{red}{-0.6} & 0.0 & 0.0 & \textcolor{blue}{+1.3} & 0.0 & 0.0 & \textcolor{red}{-0.6} & 0.0 & 0.0 & \textcolor{red}{-0.5} & 0.0 & 0.0 \\
    \bottomrule
  \end{tabular}%
  }
  \label{tab:FT_Delta_VQA}
\end{table*}

\clearpage
\subsubsection{Cross lingual accuracy with Japanese input and English output (i.e., Jp→En) }
\begin{table*}[ht]
    \centering
    \caption{Cross-lingual drop in English accuracy when switching from English input to Japanese input (Jp→En).  $\Delta_{\mathrm{En}}=\mathrm{Acc}_{\mathrm{En\text{-}input}}-\mathrm{Acc}_{\mathrm{Jp\text{-}input}}$.  \textcolor{blue}{Positive} Δs in blue, \textcolor{red}{negative} Δs in red.}
    \resizebox{0.65\textwidth}{!}{%
    \begin{tabular}{l *{8}{c}}
    \toprule
    \textbf{Model} 
      & \textbf{OR} & \textbf{SU} & \textbf{RU} & \textbf{SS} & \textbf{IC} & \textbf{ITM} & \textbf{U} & \textbf{VQA} \\
    \midrule
    \rowcolor{gray!20}
    \multicolumn{9}{c}{\textbf{Zero-shot, Without Rationales}}\\
    \midrule
    GPT-4o            & \textcolor{red}{-4.1} & \textcolor{red}{-3.8} & \textcolor{red}{-4.5} & \textcolor{red}{-5.0} & \textcolor{red}{-3.9} & \textcolor{red}{-3.5} & \textcolor{red}{-4.0} & \textcolor{red}{-4.0} \\
    GPT-4o-mini       & \textcolor{red}{-3.9} & \textcolor{red}{-3.9} & \textcolor{red}{-3.5} & \textcolor{red}{-3.2} & \textcolor{red}{-4.0} & \textcolor{red}{-3.3} & \textcolor{red}{-3.5} & \textcolor{red}{-3.5} \\
    Gemini 2.0        & \textcolor{red}{-3.8} & \textcolor{red}{-3.8} & \textcolor{red}{-3.7} & \textcolor{red}{-2.7} & \textcolor{red}{-3.7} & \textcolor{red}{-3.0} & \textcolor{red}{-3.0} & \textcolor{red}{-3.0} \\
    Gemini 1.5        & \textcolor{red}{-3.5} & \textcolor{red}{-3.8} & \textcolor{red}{-4.2} & \textcolor{red}{-3.3} & \textcolor{red}{-3.8} & \textcolor{red}{-4.1} & \textcolor{red}{-4.0} & \textcolor{red}{-4.0} \\
    LlaVa Mistral 7B  & \textcolor{red}{-14.3}& \textcolor{red}{-13.5}& \textcolor{red}{-11.8}& \textcolor{red}{-13.9}& \textcolor{red}{-8.6} & \textcolor{red}{-13.6}& \textcolor{red}{-8.5} & \textcolor{red}{-8.7} \\
    Qwen2VL 7B        & \textcolor{red}{-3.5} & \textcolor{red}{-4.6} & \textcolor{red}{-3.1} & \textcolor{red}{-4.1} & \textcolor{blue}{+1.9}& \textcolor{blue}{+2.4}& \textcolor{blue}{+3.0}& \textcolor{blue}{+2.1}\\
\bottomrule
    \end{tabular}
    }
    \label{tab:DeltaEn_JpInput}
\end{table*}
\clearpage
\subsubsection{Cross lingual accuracy with Swahili input and English output (i.e., Sw→En) }
\begin{table*}[ht]
    \centering
    \caption{Cross-lingual drop in English accuracy when switching from English-input to Swahili-input (Sw→En).  
             $\Delta_{\mathrm{En}}=\mathrm{Acc}_{\mathrm{En\text{-}input}}-\mathrm{Acc}_{\mathrm{Sw\text{-}input}}$.  
             \textcolor{blue}{Positive} Δs in blue, \textcolor{red}{negative} Δs in red.}
    \resizebox{0.65\textwidth}{!}{%
    \begin{tabular}{l *{8}{c}}
    \toprule
    \textbf{Model} 
      & \textbf{OR} & \textbf{SU} & \textbf{RU} & \textbf{SS} & \textbf{IC} & \textbf{ITM} & \textbf{U} & \textbf{VQA} \\
    \midrule
    \rowcolor{gray!20}
    \multicolumn{9}{c}{\textbf{Zero-shot, Without Rationales}}\\
    \midrule
    GPT-4o             & \textcolor{red}{-4.5} & \textcolor{red}{-4.7} & \textcolor{red}{-4.6} & \textcolor{red}{-5.1} & \textcolor{red}{-6.9} & \textcolor{red}{-5.3} & \textcolor{red}{-4.8} & \textcolor{red}{-4.2} \\
    GPT-4o-mini        & \textcolor{red}{-4.5} & \textcolor{red}{-4.1} & \textcolor{red}{-4.3} & \textcolor{red}{-4.1} & \textcolor{red}{-5.4} & \textcolor{red}{-4.2} & \textcolor{red}{-5.2} & \textcolor{blue}{+0.8}\\
    Gemini 2.0 Flash   & \textcolor{red}{-5.2} & \textcolor{red}{-5.4} & \textcolor{red}{-5.1} & \textcolor{red}{-5.0} & \textcolor{red}{-5.0} & \textcolor{red}{-4.2} & \textcolor{red}{-5.0} & \textcolor{blue}{+0.5}\\
    Gemini 1.5 Flash   & \textcolor{red}{-5.4} & \textcolor{red}{-5.2} & \textcolor{red}{-5.2} & \textcolor{red}{-5.1} & \textcolor{red}{-5.3} & \textcolor{red}{-5.0} & \textcolor{red}{-5.5} & \textcolor{blue}{+0.6}\\
    LlaVa Mistral 7B   & 0.0 & 0.0 & 0.0 & 0.0 & 0.0 & 0.0 & 0.0 & 0.0 \\
    Qwen2VL 7B         & 0.0 & 0.0 & 0.0 & 0.0 & 0.0 & 0.0 & 0.0 & 0.0 \\
\bottomrule
    \end{tabular}
    }
    \label{tab:DeltaEn_SwInput}
\end{table*}
\clearpage
\subsubsection{Cross lingual accuracy with Urdu input and English output (i.e., Ur→En) }
\begin{table*}[ht]
    \centering
    \caption{Cross-lingual drop in English accuracy when switching from English-input to Urdu-input (Ur→En).  
             $\Delta_{\mathrm{En}}=\mathrm{Acc}_{\mathrm{En\text{-}input}}-\mathrm{Acc}_{\mathrm{Ur\text{-}input}}$.  
             \textcolor{blue}{Positive} Δs in blue, \textcolor{red}{negative} Δs in red.}
    \resizebox{0.65\textwidth}{!}{%
    \begin{tabular}{l *{8}{c}}
    \toprule
    \textbf{Model} 
      & \textbf{OR} & \textbf{SU} & \textbf{RU} & \textbf{SS} & \textbf{IC} & \textbf{ITM} & \textbf{U} & \textbf{VQA} \\
    \midrule
    \rowcolor{gray!20}
    \multicolumn{9}{c}{\textbf{Zero-shot, Without Rationales}}\\
    \midrule
    GPT-4o            & \textcolor{red}{-1.5} & \textcolor{red}{-1.0} & \textcolor{red}{-1.1} & \textcolor{red}{-0.8} & \textcolor{red}{-0.5} & \textcolor{red}{-1.8} & \textcolor{blue}{+2.4} & \textcolor{blue}{+1.6} \\
    GPT-4o-mini       & \textcolor{red}{-1.7} & \textcolor{red}{-2.0} & \textcolor{red}{-2.0} & \textcolor{red}{-1.5} & \textcolor{red}{-2.7} & \textcolor{red}{-1.5} & \textcolor{red}{-1.0} & \textcolor{blue}{+3.0} \\
    Gemini 2.0 Flash  & \textcolor{red}{-1.2} & \textcolor{red}{-1.2} & \textcolor{red}{-1.2} & \textcolor{red}{-1.2} & \textcolor{red}{-2.4} & \textcolor{red}{-1.5} & \textcolor{red}{-1.5} & \textcolor{blue}{+4.0} \\
    Gemini 1.5 Flash  & \textcolor{red}{-1.6} & \textcolor{red}{-1.4} & \textcolor{red}{-1.3} & \textcolor{red}{-0.9} & \textcolor{red}{-2.8} & \textcolor{red}{-1.6} & \textcolor{red}{-1.7} & \textcolor{blue}{+3.5} \\
    LlaVa Mistral 7B  & 0.0 & 0.0 & 0.0 & 0.0 & 0.0 & 0.0 & 0.0 & 0.0 \\
    Qwen2VL 7B        & 0.0 & 0.0 & 0.0 & 0.0 & 0.0 & 0.0 & 0.0 & 0.0 \\
\bottomrule
    \end{tabular}
    }
    \label{tab:DeltaEn_UrInput}
\end{table*}

\clearpage
\subsubsection{Finetuning: Jp→En}
\begin{table*}[ht]
  \centering
  \caption{Δ in English accuracy after fine-tuning, Jp→En.  
           $\Delta_{\mathrm{En}}=\mathrm{Acc}_{\mathrm{En\text{-}input}}^{\mathrm{ft}}-\mathrm{Acc}_{\mathrm{Jp\text{-}input}}^{\mathrm{ft}}$.  
           \textcolor{blue}{Positive} Δs in blue, \textcolor{red}{negative} Δs in red.}
  \resizebox{0.65\textwidth}{!}{%
  \begin{tabular}{l *{8}{c}}
    \toprule
    \textbf{Model} 
      & \textbf{OR} & \textbf{SU} & \textbf{RU} & \textbf{SS} & \textbf{IC} & \textbf{ITM} & \textbf{U} & \textbf{VQA} \\
    \midrule
    \rowcolor{gray!20}
    \multicolumn{9}{c}{\textbf{Zero-shot, Without Rationales}}\\
    \midrule
    LlaVa Mistral 7B  & \textcolor{red}{-11.5} & \textcolor{red}{-8.5} & \textcolor{red}{-11.2} & \textcolor{red}{-8.3} & \textcolor{red}{-8.0}  & \textcolor{red}{-11.8} & \textcolor{red}{-8.9} & \textcolor{red}{-9.1} \\
    Qwen2VL 7B        & \textcolor{blue}{+0.6} & \textcolor{blue}{+5.1} & \textcolor{blue}{+0.7} & \textcolor{red}{-0.3} & \textcolor{blue}{+5.3} & \textcolor{red}{-0.5}  & \textcolor{blue}{+5.3} & \textcolor{blue}{+5.0} \\
    PALO 7B           & \textcolor{red}{-0.7} & \textcolor{blue}{+5.2} & \textcolor{red}{-0.5} & \textcolor{red}{-0.5} & \textcolor{blue}{+5.1} & \textcolor{red}{-0.3}  & \textcolor{blue}{+5.2} & \textcolor{blue}{+5.2} \\
    MAYA 8B           & \textcolor{red}{-0.3} & \textcolor{red}{-0.4} & \textcolor{red}{-0.1} & \textcolor{red}{-0.1} & \textcolor{red}{-0.3} & \textcolor{blue}{+2.4}  & \textcolor{blue}{+2.4} & \textcolor{red}{-0.5} \\
    LlaVa Mistral 13B & \textcolor{red}{-8.4} & \textcolor{red}{-8.5} & \textcolor{red}{-8.1} & \textcolor{red}{-8.9} & \textcolor{red}{-8.1} & \textcolor{red}{-8.7}  & \textcolor{red}{-8.4} & \textcolor{red}{-9.1} \\
    PALO 13B          & \textcolor{blue}{+0.4} & \textcolor{blue}{+0.2} & \textcolor{red}{-0.5} & \textcolor{red}{-0.6} & \textcolor{blue}{+0.4} & \textcolor{red}{-0.7}  & \textcolor{blue}{+0.3} & \textcolor{red}{-0.3} \\
\bottomrule
  \end{tabular}%
  }
  \label{tab:DeltaEn_JpInput_ft}
\end{table*}
\clearpage
\subsubsection{Finetuning: Ur→En}
\begin{table*}[ht]
  \centering
  \caption{Change in English accuracy after fine-tuning on Urdu input (Ur → En).  
           $\Delta_{\mathrm{En}}=\mathrm{Acc}_{\mathrm{En\text{-}input}}^{\mathrm{ft}}-\mathrm{Acc}_{\mathrm{Ur\text{-}input}}^{\mathrm{ft}}$.  
           \textcolor{blue}{Positive} Δs in blue, \textcolor{red}{negative} Δs in red.}
  \resizebox{0.65\textwidth}{!}{%
  \begin{tabular}{l *{8}{c}}
    \toprule
    \textbf{Model} 
      & \textbf{OR} & \textbf{SU} & \textbf{RU} & \textbf{SS} & \textbf{IC} & \textbf{ITM} & \textbf{U} & \textbf{VQA} \\
    \midrule
    \rowcolor{gray!20}
    \multicolumn{9}{c}{\textbf{Zero-shot, Without Rationales}}\\
    \midrule
    LlaVa Mistral 7B  & 0.0 & 0.0 & 0.0 & 0.0 & 0.0 & 0.0 & 0.0 & 0.0 \\
    Qwen2VL 7B        & 0.0 & 0.0 & 0.0 & 0.0 & 0.0 & 0.0 & 0.0 & 0.0 \\
    PALO 7B           & \textcolor{red}{-3.0} & \textcolor{red}{-3.2} & \textcolor{red}{-3.8} & \textcolor{red}{-3.6} & \textcolor{red}{-3.7}  & \textcolor{red}{-3.8}  & \textcolor{red}{-3.3}  & \textcolor{red}{-3.9}  \\
    MAYA 8B           & \textcolor{blue}{+4.7} & \textcolor{blue}{+4.7} & \textcolor{blue}{+4.7} & \textcolor{blue}{+4.7} & \textcolor{blue}{+5.7} & \textcolor{blue}{+4.9} & \textcolor{blue}{+4.5} & \textcolor{blue}{+4.3} \\
    LlaVa Mistral 13B & 0.0 & 0.0 & 0.0 & 0.0 & 0.0 & 0.0 & 0.0 & 0.0 \\
    PALO 13B          & \textcolor{blue}{+0.4} & \textcolor{blue}{+0.2} & \textcolor{red}{-0.5} & \textcolor{red}{-0.6} & \textcolor{blue}{+0.4} & \textcolor{red}{-0.7}  & \textcolor{blue}{+0.3} & \textcolor{red}{-0.3} \\
\bottomrule
  \end{tabular}%
  }
  \label{tab:DeltaEn_UrInput_ft}
\end{table*}

\clearpage
\subsection{Language support of Open Vision Language Models}
\label{subsec:OpenVLMlanguageSupport}
\begin{table*}[h]
    \centering
    \small
    \caption{Language Support and Base LLMs for Open-Source Vision-Language Models.}
    \label{tab:VLMlanguageSupport}
    \resizebox{\textwidth}{!}{
    \begin{tabular}{@{}l | p{4.5cm} p{8cm}@{}}
        \toprule
        \textbf{Model} & \textbf{Base LLM} & \textbf{Language Support (Vision-Language Tasks)} \\
        \midrule
        \multirow{2}{*}{LLaVA 7B/13B Mistral} & \multirow{2}{*}{\parbox{4cm}{Mistral (strong multilingual text capabilities)}} & Primarily English for vision-language alignment. VL performance in other languages varies without specific multilingual VL fine-tuning. \\
        \midrule
        \multirow{2}{*}{PALO 7B/13B} & \multirow{2}{*}{\parbox{4cm}{Typically Llama-based (multilingual text understanding)}} & Primarily English for vision-language alignment. Performance on VL tasks in other languages depends on specific fine-tuning efforts. \\
        \midrule
        \multirow{2}{*}{MAYA 8B} & \multirow{2}{*}{\parbox{4cm}{Aya-based foundation (supports a broad set of languages for text)}} & Strong support for English and multiple Indic languages (e.g., Hindi, Tamil, Telugu, etc.). VL alignment across all base LLM languages is extensive. \\
        \midrule
        \multirow{2}{*}{Qwen2-VL} & \multirow{2}{*}{\parbox{4cm}{Qwen series (broad multilingual capabilities)}} & Strong support for English and Chinese. Good to extensive support for many other languages reported (e.g. Fr, Es, De, Ar, Jp, Kr, Sw, Ur). \\
        \bottomrule
    \end{tabular}%
    }
\end{table*}

\clearpage
\subsection{Open-source VLM Pretraining Data}
The table below contains the data, per task, used to pre-train the open-source models.

\begin{table*}[h]
  \centering
  \caption{Overview of pretraining datasets used by different models.}
  \label{tab:ModelPretrainingData}
  \resizebox{0.8\textwidth}{!}{%
    \begin{tabular}{@{} l | l c c c c @{}}
      \toprule
      \multirow{2}{*}{\textbf{Data Type}} & \multirow{2}{*}{\textbf{Data Name}} &
        \multicolumn{4}{c}{\textbf{Model}} \\
      \cmidrule(lr){3-6}
      & & \shortstack{LLaVA 7B/13B\\Vicuna} & \shortstack{PALO\\7B/13B} &
            \shortstack{MAYA\\8B} & \shortstack{Qwen2-\\VL} \\
      \midrule

      \multirow{5}{*}{Text}
        & ShareGPT-4V \cite{chen2024sharegpt4v}        & \ding{51} & \ding{51} &  & \ding{51} \\
        & LAION-GPT-V \cite{gpt4v}                     & \ding{51} & \ding{51} &  & \ding{51} \\
        & SlimOrca \cite{SlimOrca}                     &           &           &  & \ding{51} \\
        & Aya multilingual dataset \cite{singh-etal-2024-aya} &    &           & \ding{51} &  \\
        & In-house Data                                 &           &           &  & \ding{51} \\
      \midrule

      \multirow{8}{*}{Dialogue Caption}
        & COCO \cite{Lin2014MicrosoftCC}               &           &           &  & \ding{51} \\
\bottomrule
    \end{tabular}%
  }
\end{table*}
\clearpage
\subsection{Details of Model Parameters and Finetuning Settings}
The hyperparameters for VLMs used in the experiments in \S \ref{sec:Experiments} are shown in Table \ref{tab:ModelParameters}.
\begin{table*}[h]
    \centering
    \caption{Model Generation Settings.}
    \resizebox{0.95\textwidth}{!}{
    \begin{tabular}{@{}l | l@{}}
        \toprule
        \textbf{Model} & \textbf{Generation Settings} \\
        \midrule
        GPT-4o                & model = \texttt{gpt-4o-2024-08-06}, temperature = 0, max\_tokens = \texttt{1024}, batch\_size = \texttt{100} \\
        GPT-4o-mini           & model = \texttt{gpt-4o-mini-2024-07-18}, temperature = 0, max\_tokens = \texttt{1024}, batch\_size = \texttt{100} \\
        Gemini 2.0 Flash-Lite & model = \texttt{gemini-2.0-flash-lite}, temperature = 0, max\_tokens = \texttt{1024} \\
        Gemini 1.5 Flash-8B   & model = \texttt{gemini-1.5-flash-8b}, temperature = 0, max\_tokens = \texttt{1024}  \\
        LLaVA-NeXT-Vicuna-7B  & model = \texttt{llava-hf/llava-v1.6-vicuna-7b-hf}, max\_tokens=1024 \\
        LLaVA-NeXT-Vicuna-13B & model = \texttt{llava-hf/llava-v1.6-vicuna-13b-hf}, max\_tokens=1024 \\ 
        MAYA 8B               & model = \texttt{maya}, max\_tokens=1024 \\ 
        PALO 7B               & model = \texttt{palo}, max\_tokens=1024 \\
        PALO 13B               & model = \texttt{palo}, max\_tokens=1024 \\ 
        Qwen2-VL-7B-Instruct  & model = \texttt{Qwen/Qwen2-VL-7B-Instruct}, max\_tokens=1024 \\
        \bottomrule
    \end{tabular}
    }
    \label{tab:ModelParameters}
\end{table*}
\subsubsection{Inference Settings} 
All the experiments are conducted on two NVIDIA A100 80GB GPUs. 
The complete list of VLMs used includes: GPT-4o, GPT-4o-mini, Gemini 2.0 Flash-Lite, Gemini 1.5 Flash-8B, LLaVA-NeXT-Vicuna-7B \footnote{https://huggingface.co/llava-hf/llava-v1.6-vicuna-7b-hf}, LLaVA-NeXT-Vicuna-13B \footnote{https://huggingface.co/llava-hf/llava-v1.6-vicuna-13b-hf}, PALO 7B/13B \footnote{https://github.com/mbzuai-oryx/PALO}, MAYA 8B \footnote{https://github.com/nahidalam/maya}, and Qwen2VL 7B \footnote{https://huggingface.co/Qwen/Qwen2-VL-7B-Instruct}.
We deploy all the open models via the HuggingFace API, and employ 8-bit quantization for model generation. 
We deploy GPT \textit{batch API}\footnote{OpenIA batch API settings: https://platform.openai.com/docs/guides/batch?lang=python.} settings for both \texttt{gpt-4o-2024-08-06} and \texttt{gpt-4o-mini-2024-07-18} models. Moreover, we deploy \textit{concurrent processing} for Gemini models: \texttt{gemini-2.0-flash-lite, gemini-1.5-flash-8b}. 

\subsubsection{Fine-tuning Settings}
During fine-tuning, we set the number of training epochs to 10, per device train batch size to 2, we deploy gradient accumulation and gradient checkpointing, we use the Adam optimizer, and the learning rate is set to $2^{e-4}$. We deploy a constant learning rate scheduler.
Our LoRA configuration includes LoRA matrices of rank 8, and the dropout probability is set to 0.05.
We conduct three fine-tuning and evaluation runs for each experiment and average the results over three runs. 
Details about major environment settings used during our experiments are shown in Table \ref{tab:VLM_hyperparameters_Envt}.
\begin{table}[h]
    \centering
    \caption{VLM experiment settings.}
    \resizebox{0.35\textwidth}{!}{
    \begin{tabular}{r |l}
    \toprule
    \textbf{Dependency/Library} & \textbf{Version} \\
    \hline
     accelerate   &  0.34.2  \\
     bitsandbytes &  0.44.1 \\
     datasets & 3.0. \\
     deepspeed  &  0.15.1 \\
     CUDA & 12.1 \\
     flash-attn & 2.5.8 \\ 
     huggingface\_hub & 0.24.6 \\
     ninja & 1.12.1 \\
     pillow  & 10.4.0 \\
     requests & 2.32.3 \\
     safetensors & 0.4.4 \\
     torch & 2.4.0 \\
     torchaudio &  2.4.0 \\
     tqdm & 4.66.5 \\
     transformers & 4.45.1 \\
     trl & 0.10.1 \\
     wandb &  0.17.7 \\
     \bottomrule
    \end{tabular}
    }
    \label{tab:VLM_hyperparameters_Envt}
\end{table}

\subsubsection{Finetuning Data Splits}
The detailed image-text information used for training the VLMs is shown in Table \ref{tab:FinetuneDataSplits}. We split the image-text data using the 80:10:10 ratio. 
\begin{table}[h]
    \centering
    \caption{Data Splits for VLURes Finetuning (80\% Train / 10\% Dev / 10\% Test). Counts are identical for image-text pairs.}
    \begin{tabular}{l|| r|rrr} %
        \toprule
        \multirow{2}{*}{\textbf{Language}} & \textbf{Total} & \textbf{Train} & \textbf{Dev} & \textbf{Test} \\
        & \textbf{Pairs} & \textbf{(80\%)} & \textbf{(10\%)} & \textbf{(10\%)} \\
        \midrule
        English (En)    & 1000 & 800 & 100 & 100 \\
        Japanese (Jp)   & 1000 & 800 & 100 & 100 \\
        Swahili (Sw)    & 1130 & 904 & 113 & 113 \\
        Urdu (Ur)       & 996  & 796 &  99 & 101 \\ 
        \bottomrule
    \end{tabular}
    \label{tab:FinetuneDataSplits}
\end{table}

\clearpage
\subsection{Project Costs}
\subsubsection{API Costs: Inference}
To ensure transparency and facilitate reproducibility, we provide a detailed analysis of the computational costs associated with our experiments. While our study includes several open-source models run on local infrastructure, this section focuses on the expenses incurred from using proprietary models accessed via their respective APIs: \texttt{gpt-4o}, \texttt{gpt-4o-mini}, \texttt{gemini-2.0-flash-lite}, and \texttt{gemini-1.5-flash-8b}. Our experimental design encompassed a comprehensive evaluation across our full benchmark, which includes four languages (English, Japanese, Swahili, and Urdu), four distinct prompting settings, and eight vision-language tasks, culminating in 32,000 API calls per model. Table~\ref{tab:cost_analysis} provides a granular breakdown of these expenses, detailing the pricing tiers, the total number of input and output tokens processed, and the resulting cost for each model. The cumulative estimated cost for this extensive inference stage totals approximately \textbf{\$221.04}. We note that this figure exclusively covers the model inference phase of our study. A separate analysis detailing the costs associated with our LLM-as-a-judge evaluation framework is presented in the subsequent section. This detailed breakdown serves as a crucial data point for the community, highlighting the resource investment required for large-scale, multilingual VLM benchmarking.
\begin{table}[h]
  \centering
  \caption{ Estimated inference cost for proprietary models on the entire \emph{VLURes} benchmark. Costs are based on Batch API pricing (Aug 2024) and are calculated across four languages, four prompting settings, and eight tasks for our 1,000-item benchmark.}
  \label{tab:cost_analysis}
  \begin{tabular}{@{}lrrrrr@{}}
    \toprule
    \textbf{Model} & \multicolumn{2}{c}{\textbf{Price per 1M Tokens}} & \textbf{Input Tokens} & \textbf{Output Tokens} & \textbf{Est. Cost} \\
    \cmidrule(lr){2-3}
                   & Input (\$) & Output (\$) & (Millions)$^a$ & (Millions)$^b$ & (\$) \\
    \midrule
    GPT-4o           & 1.2500 & 5.0000 & 33.68 & 30.02 & \$192.20 \\
    GPT-4o-mini      & 0.0750 & 0.3000 & 33.68 & 30.02 & \$11.54  \\
    Gemini 2.0 Flash & 0.0750 & 0.3000 & 33.68 & 30.02 & \$11.54  \\
    Gemini 1.5 Flash & 0.0375 & 0.1500 & 33.68 & 30.02 & \$5.76   \\
    \midrule
    \textbf{Total} & \multicolumn{3}{r}{\textbf{Cumulative Cost:}} & & \textbf{\$221.04} \\
    \bottomrule
  \end{tabular}
  \begin{minipage}{0.95\linewidth}
    \vspace{0.5ex}
    \footnotesize
    \textit{Note:} Open-source models were run on local infrastructure and are excluded from this API cost analysis.
    \\
    $^a$ Total input tokens are estimated as 32,000 calls (1k items × 8 tasks × 4 languages) × (avg. 171 prompt tokens + avg. 93 text tokens) = 33,680,000 tokens per model.
    \\
    $^b$ Total output tokens are estimated as 16,000 calls × 154 tokens (w/o rationales) + 16,000 calls × 315 tokens (w/ rationales) = 30,016,000 tokens per model.
  \end{minipage}
\end{table}
\subsubsection{API Costs: Evaluation}
Following the inference stage, we employed an LLM-as-a-judge framework to systematically evaluate the quality of the generated outputs. This section details the costs associated with this evaluation phase, which was conducted using the \texttt{Gemini 1.5 Pro} model. The evaluation was comprehensive, covering the outputs from all ten models (both proprietary and open-source) across the full experimental matrix: eight tasks, four languages, and four prompting settings for each of our 1,000 benchmark items. This extensive scope resulted in 1,280,000 individual outputs requiring evaluation, with each judged by a separate API call. The input to the \texttt{Gemini 1.5 Pro} judge for each call comprised a detailed evaluation prompt ($\approx$490 tokens) and the specific VLM output being assessed. The size of the VLM output varied depending on whether rationales were included ($\approx$154 tokens without, $\approx$315 tokens with). Summing across all evaluation calls, the total input token count is estimated at 927.36 million. Conversely, the judge's output was designed for efficiency, consisting of a minimal JSON object containing only the score ($\approx$5 tokens), leading to a total of 6.4 million output tokens. As detailed in Table~\ref{tab:cost_analysis_eval}, the total estimated cost for this LLM-as-a-judge evaluation phase is \textbf{\$1,191.20}. While substantial, this investment was crucial for achieving robust, automated, and scalable evaluation of generative model outputs across the vast and diverse landscape of our multilingual benchmark.
\begin{table}[h]
  \centering
  \caption{Estimated cost for the LLM-as-a-judge evaluation phase using Gemini 1.5 Pro. The cost covers the evaluation of 1,280,000 individual model outputs from all 10 VLMs across the entire benchmark.}
  \label{tab:cost_analysis_eval}
  \begin{tabular}{@{}ll@{}}
    \toprule
    \textbf{Metric} & \textbf{Value} \\
    \midrule
    LLM-as-a-Judge Model & \texttt{Gemini 1.5 Pro} \\
    Total Evaluation Calls & 1,280,000 \\
    \midrule
    \multicolumn{2}{@{}l}{\textbf{Input Token Calculation}} \\
    Avg. Input Tokens / Call (w/o rationales) & $\approx$ 644 tokens \\
    Avg. Input Tokens / Call (w/ rationales) & $\approx$ 805 tokens \\
    Total Input Tokens (Millions)$^a$ & 927.36 \\
    \midrule
    \multicolumn{2}{@{}l}{\textbf{Output Token Calculation}} \\
    Avg. Output Tokens / Call & $\approx$ 5 tokens \\
    Total Output Tokens (Millions)$^b$ & 6.40 \\
    \midrule
    \multicolumn{2}{@{}l}{\textbf{Cost Calculation}} \\
    Input Cost (\$) & \$1,159.20 \\
    Output Cost (\$) & \$32.00 \\
    \midrule
    \textbf{Total Estimated Evaluation Cost} & \textbf{\$1,191.20} \\
    \bottomrule
  \end{tabular}
  \begin{minipage}{0.95\linewidth}
    \vspace{0.5ex}
    \footnotesize
    \textit{Note:} Pricing is based on Gemini 1.5 Pro rates ($1.25/1M input, $5.00/1M output for prompts $\le$ 128k tokens).
    \\
    $^a$ Total input tokens = (640,000 calls $\times$ 644 tokens) + (640,000 calls $\times$ 805 tokens) = 927,360,000.
    \\
    $^b$ Total output tokens = 1,280,000 calls $\times$ 5 tokens = 6,400,000.
  \end{minipage}
\end{table}
\subsubsection{Total API Costs}
The total computational cost for this study is composed of two primary phases: model inference and LLM-as-a-judge evaluation. The inference phase, detailed in Part 1, amounted to approximately \$221.04. The subsequent evaluation phase, detailed in Part 2, incurred a significantly larger cost of approximately \$1,191.20, primarily due to the vast number of individual outputs requiring judgment. Cumulatively, as summarized in Table~\ref{tab:total_cost_summary}, the total estimated expenditure for the project is \textbf{\$1,412.24}. This figure underscores the substantial financial investment required to conduct comprehensive, large-scale, and multilingual benchmarking of modern foundation models using automated evaluation frameworks.
\begin{table}[h]
  \centering
  \caption{Summary of total estimated project costs, combining both the model inference and the LLM-as-a-judge evaluation phases.}
  \label{tab:total_cost_summary}
  \begin{tabular}{@{}lr@{}}
    \toprule
    \textbf{Component} & \textbf{Estimated Cost (\$)} \\
    \midrule
    Part 1: Model Inference & \$221.04 \\
    Part 2: LLM-as-a-Judge Evaluation & \$1,191.20 \\
    \midrule
    \textbf{Total} & \textbf{\$1,412.24} \\
    \bottomrule
  \end{tabular}
\end{table}
\subsubsection{Human Evaluation Costs}
We hired two native speakers per language (for En, Jp, Sw, and Ur) to rate the quality of VLM-generated outputs, for eight vision and language tasks. 
The evaluators followed clear guidelines to rate the performances of ten VLMs used in this study on a scale of 1 to 100.
All the evaluators were compensated fairly based on prevailing market prices. 

\clearpage
\subsection{Domain Names used for Data Collection}
Table \ref{tab:dataset_sources} shows the web resources we used to gather image-text pairs, for each language, in \emph{VLURes}.
\begin{table*}[h]
    \centering
    \tiny
    \caption{Overview of Dataset Sources for Benchmark by Language.}
    \resizebox{0.5\textwidth}{!}{%
    \begin{tabular}{@{}l| l@{}}
        \toprule
        \textbf{Language} & \textbf{Domain Name} \\
        \midrule
        \multirow{4}{*}{English}
        & www.wikipedia.org \\
        & www.wikinews.org \\
        & www.stackexchange.com \\
        & www.bbc.com \\
        \midrule
        \multirow{14}{*}{Japanese}
        & ja.wikipedia.org \\
        & www.nhk.or.jp \\
        & www.cookpad.com \\
        & www.tabelog.com \\
        & www.jleague.jp \\
        & www.npb.jp \\
        & www.tnm.jp \\
        & www.tokyogendai.com/ja/ \\
        & www.nikkei.com \\
        & www.tokyo-zoo.net \\
        & www.aij.or.jp \\
        & www.yomiuri.co.jp \\
        & www.meti.go.jp \\
        & www.kakaku.com \\
        \midrule
        \multirow{8}{*}{Swahili}
        & sw.wikipedia.org \\
        & www.bbc.com/swahili \\
        & www.michuzi.co.tz \\
        & www.bongo5.com \\
        & www.mwananchi.co.tz \\
        & www.mwanahalisionline.com \\
        & taifaleo.nation.co.ke \\
        & swahilitimes.co.tz \\
        \midrule
        \multirow{6}{*}{Urdu}
        & ur.wikipedia.org \\
        & www.dawnnews.tv \\
        & www.express.pk \\
        & www.bbc.com/urdu \\
        & www.urdupoint.com \\
        & www.kfoods.com \\
        \bottomrule
    \end{tabular}%
    }
    \label{tab:dataset_sources}
\end{table*}

\clearpage
\subsection{Examples of Outputs generated by VLMs}
\subsubsection{VLM Responses for Object Recognition (Task 1) }
\begin{table*}[ht]
\centering
\caption{
An example from our benchmark showing responses generated by \textbf{GPT-4o} for \emph{Object Recognition}.
In this setting, no rationales are generated by the model.
Given an image-text pair as input, we show the outputs in four languages: En, Jp, Sw, and Ur.
\label{tab:Task1OneshotPromptingNoRationaleGPT4o}
}
\tiny
\begin{tabular}{p{0.28\textwidth} p{0.45\textwidth} p{0.22\textwidth}}
\toprule
\multicolumn{3}{p{0.98\textwidth}}{%
\textbf{\textcolor{olive}{Inputs to the VLM}}
} \\
\midrule
\textbf{Image} & \textbf{Text} & \textbf{Prompt Question} \\
\midrule
\adjustbox{valign=t}{\includegraphics[width=\linewidth]{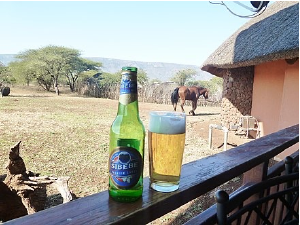}}
&
\texttt{Eswatini Beverages Ltd (EBL) was a subsidiary of SABMiller until 10 October 2016 when it was acquired by Anheuser-Busch InBev. It is a beverage and brewing company in Matsapha, Eswatini. The company was formed in 1995 by the merger of Eswatini Breweries, Ngwane Breweries, and Eswatini Bottlers. EBL produces and markets soft drinks, beer, and other alcoholic drinks. The Swazi beer produced is Sibebe Premium Lager named after the Sibebe rock, but other beers are brewed as well including Castle Lager. EBL dominates the beer market in Eswatini with a market share of 87\% in 2014 227 hl of lager beer were sold. Retooling in 2013 led to a temporary beer shortage in Eswatini.}
&
\adjustbox{valign=t}{%
\parbox{\linewidth}{%
\texttt{\textcolor{red}{Question. Analyze this image and list all objects present. Categorize each object into groups such as furniture, electronic devices, clothing, etc. Be thorough and specific.}}%
}%
}
\\
\midrule
\multicolumn{3}{p{0.98\textwidth}}{%
\textbf{\textcolor{olive}{Responses generated by the VLM}}
} \\
\midrule
\multicolumn{3}{p{0.98\textwidth}}{%
\textbf{\textcolor{blue}{[English Response:]}} \texttt{ Image Analysis: Objects Present: 1. Furniture: - Wooden railing - Chair (partially visible) 2. Beverages: - Beer bottle - Glass of beer 3. Architecture: - Thatched roof building 4. Animals: - Horse 5. Natural Elements: - Trees - Grass - Hills in the background. Each object has been categorized into relevant groups for clarity.}
} \\
\midrule
\multicolumn{3}{p{0.98\textwidth}}{%
\textbf{\textcolor{blue}{[Japanese Response:]}} \begin{CJK*}{UTF8}{min}画像分析：オブジェクトのリストと分類：1. 飲料 − ビールの瓶／ビールのグラス 2. 家具 − 手すり（木製のバルコニーの一部） 3. 動物 − 馬 4. 建物 − 茅葺き屋根の建物 5. 自然 − 木々／草地。これらのオブジェクトが画像に存在しています。\end{CJK*}
} \\
\midrule
\multicolumn{3}{p{0.98\textwidth}}{%
\textbf{\textcolor{blue}{[Swahili Response:]}} \texttt{Katika picha hii, kuna vitu vifuatavyo: Vinywaji: - Chupa ya bia. - Glasi yenye bia. Samani: - Meza ya mbao. Viumbe hai: - Farasi. Majengo: - Nyumba yenye paa la nyasi. Mandhari: - Miti na nyasi kwenye eneo la wazi. Vifaa vya nje: - Uzio wa miti. Picha inaonyesha mazingira ya nje yenye mandhari ya asili. }
} \\
\midrule
\multicolumn{3}{p{0.98\textwidth}}{%
\textbf{\textcolor{blue}{[Urdu Response:]}} \begin{UrduText}اس تصویر میں موجود اشیاء کی فہرست درج ذیل ہے: \textbf{مشروبات}: بوتل (Sibebe) اور گلاس (جس میں مشروب موجود ہے)۔ \textbf{قدرتی مناظر}: گھوڑا، درخت، کھیت یا میدان۔ \textbf{تعمیرات}: چھت والا مکان یا جھونپڑی۔ \textbf{فرنیچر}: ریلنگ جس پر بوتل اور گلاس رکھے ہیں۔ یہ اشیاء مختلف زمروں میں تقسیم کی گئی ہیں، جیسے مشروبات، قدرتی مناظر، تعمیرات اور فرنیچر۔\end{UrduText}
} \\
\bottomrule
\end{tabular}
\end{table*}

\clearpage
\begin{table*}[t]
\centering
\caption{
An example from our benchmark showing responses generated by \textbf{GPT-4o} for \emph{Object Recognition}.
In this setting, \underline{the model generates rationales}, in addition to analyzing the image.
Given an image-text pair as input, we show the outputs in four languages: En, Jp, Sw, and Ur.
\label{tab:Task1OneshotPromptingWithRationaleGPT4o}
}
\tiny
\begin{tabular}{p{0.28\textwidth} p{0.45\textwidth} p{0.22\textwidth}}
\toprule
\multicolumn{3}{p{0.98\textwidth}}{%
\textbf{\textcolor{olive}{Inputs to the VLM}}
} \\
\midrule
\textbf{Image} & \textbf{Text} & \textbf{Prompt Question} \\
\midrule
\adjustbox{valign=t}{\includegraphics[width=\linewidth]{images/TaskExamples/img859.pdf}}
&
\texttt{Eswatini Beverages Ltd (EBL) was a subsidiary of SABMiller until 10 October 2016 when it was acquired by Anheuser-Busch InBev. It is a beverage and brewing company in Matsapha, Eswatini. The company was formed in 1995 by the merger of Eswatini Breweries, Ngwane Breweries, and Eswatini Bottlers. EBL produces and markets soft drinks, beer, and other alcoholic drinks. The Swazi beer produced is Sibebe Premium Lager named after the Sibebe rock, but other beers are brewed as well including Castle Lager. EBL dominates the beer market in Eswatini with a market share of 87\% in 2014 227 hl of lager beer were sold. Retooling in 2013 led to a temporary beer shortage in Eswatini.}
&
\adjustbox{valign=t}{%
\parbox{\linewidth}{%
\texttt{\textcolor{red}{Question. Analyze this image and list all objects present. Categorize each object into groups such as furniture, electronic devices, clothing, etc. Be thorough and specific.} \textcolor{cyan}{Please provide both your analysis and rationale.}}%
}%
}
\\
\midrule
\multicolumn{3}{p{0.98\textwidth}}{%
\textbf{\textcolor{olive}{Responses generated by the VLM}}
} \\
\midrule
\multicolumn{3}{p{0.98\textwidth}}{%
\textbf{\textcolor{blue}{[English Response:]}} \texttt{Image Analysis: Objects Present: 1. Furniture: - Wooden railing 2. Beverages: - Beer bottle - Glass of beer 3. Animals: - Horse 4. Buildings/Structures: - Thatched-roof building 5. Natural Elements: - Trees - Grass - Hills in the background Rationale: - Furniture: The wooden railing is part of the structure where the bottle and glass are placed, indicating an outdoor seating or viewing area. - Beverages: The presence of a beer bottle and a glass of beer suggests a casual, relaxed setting, possibly for leisure or dining. - Animals: The horse in the background adds to the rural or countryside setting, indicating an area where animals are present. - Buildings/Structures: The thatched-roof building is typical of certain rural or traditional architectural styles, contributing to the setting's ambiance. - Natural Elements: The trees, grass, and hills provide context for the location, suggesting a natural, open environment, possibly a farm or countryside retreat.}
} \\
\midrule
\multicolumn{3}{p{0.98\textwidth}}{%
\textbf{\textcolor{blue}{[Japanese Response:]}} \begin{CJK*}{UTF8}{min}画像分析：表示されている物体：1. 家具 — 木製の手すり／屋外用の椅子 2. 飲み物 — ビール瓶／泡のあるビールグラス 3. 動物 — 馬 4. 建物 — 藁ぶき屋根の建物 5. 自然の要素 — 木々／草／背景の丘。\quad\textbf{根拠}: 木製の手すり上に瓶とグラスが置かれており、前景には屋外用の座席が見える。背景には馬と自然景観（木々・草・丘）が確認でき、建物は茅葺き屋根の伝統的な様式であることから、屋外のくつろぎ環境が示唆される。\end{CJK*}
} \\
\midrule
\multicolumn{3}{p{0.98\textwidth}}{%
\textbf{\textcolor{blue}{[Swahili Response:]}} \texttt{Image Analysis: Vitu Vilivyomo: 1. Vinywaji: - Chupa ya bia - Glasi yenye bia 2. Samani: - Meza ya mbao au reli ya mbao 3. Wanyama: - Farasi 4. Majengo: - Nyumba yenye paa la nyasi 5. Mandhari: - Miti - Nyasi - Milima kwa mbali Rationale: - Vinywaji: Chupa ya bia na glasi yenye bia inaonyesha kuwa kuna mtu anayefurahia kinywaji katika eneo hili. Hii inaweza kuashiria mazingira ya kupumzika au burudani. - Samani: Meza au reli ya mbao inatumika kuweka vinywaji, ikionyesha kuwa eneo hili linaweza kuwa sehemu ya kupumzika au kufurahia mandhari. - Wanyama: Farasi anaonekana kwenye mandhari ya nyuma, akionyesha uwepo wa wanyama katika eneo hili, ambalo linaweza kuwa la mashambani au la kitalii. - Majengo: Nyumba yenye paa la nyasi inaonyesha mtindo wa ujenzi wa jadi au wa kiasili, unaoweza kupatikana katika maeneo ya vijijini au ya kitalii. - Mandhari: Miti, nyasi, na milima kwa mbali zinaonyesha mazingira ya asili, yanayoweza kuwa ya kuvutia kwa watalii au kwa wale wanaopenda mandhari ya nje. Picha hii inaonyesha mazingira ya asili na ya kupumzika, ikiashiria uwezekano wa kuwa katika eneo la kitalii au la mashambani.}
} \\
\midrule
\multicolumn{3}{p{0.98\textwidth}}{%
\textbf{\textcolor{blue}{[Urdu Response:]}} \begin{UrduText}تصویر کا تجزیہ: موجود اشیاء میں فرنیچر (لکڑی کی ریلنگ)، مشروبات (بیئر کی بوتل اور بیئر کا گلاس)، جانور (گھوڑا)، عمارات یا ڈھانچے (بھوسے کی چھت والی عمارت)، اور قدرتی عناصر (درخت، گھاس، پس منظر میں پہاڑیاں) شامل ہیں۔ بنیادی وجوہات: فرنیچر کے طور پر ریلنگ پر بوتل اور گلاس رکھے ہیں، جو کھلے ماحول میں بیٹھنے یا منظر دیکھنے کی جگہ کی نشان دہی کرتا ہے۔ مشروبات کی موجودگی پرسکون اور تفریحی ماحول کی طرف اشارہ کرتی ہے۔ پس منظر میں گھوڑا دیہی یا قدرتی ماحول کو اُجاگر کرتا ہے۔ گھاس پھونس کی چھت روایتی یا دیہی طرزِ تعمیر کی علامت ہے۔ درخت، گھاس اور پہاڑیاں کھلا اور قدرتی منظرنامہ واضح کرتی ہیں۔\end{UrduText}
} \\
\bottomrule
\end{tabular}
\end{table*}

\clearpage
\subsubsection{VLM Responses for Scene Understanding (Task 2)}
\begin{table*}[ht]
\centering
\caption{
An example from our benchmark showing responses generated by \textbf{GPT-4o} for \emph{Scene Understanding}.
In this setting, no rationales are generated by the model.
Given an image-text pair as input, we show the outputs in four languages: En, Jp, Sw, and Ur.
\label{tab:Task2OneshotPromptingNoRationaleGPT4o}
}
\tiny
\begin{tabular}{p{0.28\textwidth} p{0.45\textwidth} p{0.22\textwidth}}
\toprule
\multicolumn{3}{p{0.98\textwidth}}{%
\textbf{\textcolor{olive}{Inputs to the VLM}}
} \\
\midrule
\textbf{Image} & \textbf{Text} & \textbf{Prompt Question} \\
\midrule
\adjustbox{valign=t}{\includegraphics[width=\linewidth]{images/TaskExamples/img859.pdf}}
&
\texttt{Eswatini Beverages Ltd (EBL) was a subsidiary of SABMiller until 10 October 2016 when it was acquired by Anheuser-Busch InBev. It is a beverage and brewing company in Matsapha, Eswatini. The company was formed in 1995 by the merger of Eswatini Breweries, Ngwane Breweries, and Eswatini Bottlers. EBL produces and markets soft drinks, beer, and other alcoholic drinks. The Swazi beer produced is Sibebe Premium Lager named after the Sibebe rock, but other beers are brewed as well including Castle Lager. EBL dominates the beer market in Eswatini with a market share of 87\% in 2014 227 hl of lager beer were sold. Retooling in 2013 led to a temporary beer shortage in Eswatini.}
&
\adjustbox{valign=t}{%
\parbox{\linewidth}{%
\texttt{\textcolor{red}{Question: Describe the overall scene in this image. What is the setting and what activities or events are taking place?}}%
}%
}
\\
\midrule
\multicolumn{3}{p{0.98\textwidth}}{%
\textbf{\textcolor{olive}{Responses generated by the VLM}}
} \\
\midrule
\multicolumn{3}{p{0.98\textwidth}}{%
\textbf{\textcolor{blue}{[English Response:]}} \texttt{Image Analysis: The scene is set in a rural or countryside environment, likely in a warm climate given the presence of thatched roofs and open landscape. The setting includes a wooden railing, possibly part of a porch or outdoor seating area, overlooking a grassy field. In the foreground, there is a bottle, and a glass of beer placed on the railing, suggesting a relaxed, leisurely atmosphere. The beer is labeled "Sibebe," indicating a specific brand. In the background, there is a horse grazing near a thatched-roof building, which adds to the rural ambiance. The landscape features sparse trees and a distant view of hills or mountains, enhancing the sense of openness and tranquility. Overall, the image conveys a peaceful, laid-back setting, possibly at a lodge or rural retreat, where one can enjoy a drink while appreciating the natural surroundings.}
} \\
\midrule
\multicolumn{3}{p{0.98\textwidth}}{%
\textbf{\textcolor{blue}{[Japanese Response:]}} \begin{CJK*}{UTF8}{min}画像分析：この画像は、自然が広がる屋外の環境を背景にしています。手前の木製の手すりの上にはビールの瓶とグラスが置かれており、くつろいだ雰囲気がうかがえます。背景には草原が広がり、点在する木々の向こうに、藁葺き屋根の建物と草をはむ馬が見えます。遠景には緩やかな丘（または山並み）が続き、開放感と静けさを強めています。全体として、自然に囲まれたロッジや田園の滞在先で景色を楽しみながら一息ついている場面が描かれています。\end{CJK*}
} \\
\midrule
\multicolumn{3}{p{0.98\textwidth}}{%
\textbf{\textcolor{blue}{[Swahili Response:]}} \texttt{Mandhari na Mazingira: Picha inaonyesha mazingira ya nje yenye mandhari ya asili. Kuna nyumba yenye paa la nyasi upande wa kulia, ambayo inaonekana kuwa ya jadi au ya vijijini. Nyuma ya nyumba hiyo, kuna uzio wa miti na miti ya kijani kibichi inayoongeza uzuri wa mandhari. Mbali kidogo, kuna farasi anayechunga kwenye eneo la wazi. Shughuli na Matukio: Kwenye sehemu ya mbele ya picha, kuna chupa ya bia na glasi iliyojaa bia iliyowekwa juu ya uzio wa mbao. Inaonekana kama mtu alikuwa akifurahia kinywaji huku akitazama mandhari ya nje. Hakuna shughuli nyingi zinazoendelea, na mazingira yanaonekana kuwa tulivu na ya kupumzika. Maelezo ya Ziada: Mandhari hii inaonyesha hali ya utulivu na amani, ikitoa hisia ya kupumzika na kufurahia asili. Farasi anayechunga na nyumba ya jadi vinachangia katika kuonyesha maisha ya vijijini au ya kiasili.}
} \\
\midrule
\multicolumn{3}{p{0.98\textwidth}}{%
\textbf{\textcolor{blue}{[Urdu Response:]}} \begin{UrduText}یہ منظر دیہی اور قدرتی ماحول کی عکاسی کرتا ہے۔ سامنے لکڑی کی ریلنگ پر ایک بوتل اور گلاس رکھا ہے جس میں مشروب موجود ہے، جو پُرسکون اور فراغت بھرا ماحول ظاہر کرتا ہے۔ پس منظر میں کھلا سبزہ زار ہے جہاں ایک گھوڑا چر رہا ہے، جبکہ دائیں جانب گھاس پھوس کی چھت والی جھونپڑی نما عمارت دکھائی دیتی ہے۔ دور تک درخت اور ہلکی پہاڑیاں نظر آتی ہیں جو منظر کے حسن اور کشادگی کو بڑھاتی ہیں۔ مجموعی طور پر تصویر ایک ایسے مقام کو پیش کرتی ہے جہاں فطرت سے لطف اندوز ہوتے ہوئے آرام کیا جا سکتا ہے۔\end{UrduText}
} \\
\bottomrule
\end{tabular}
\end{table*}

\clearpage
\begin{table*}[t]
\centering
\caption{
An example from our benchmark showing responses generated by \textbf{GPT-4o} for \emph{Scene Understanding}.
In this setting, \underline{the model generates rationales}, in addition to analyzing the image.
Given an image-text pair as input, we show the outputs in four languages: En, Jp, Sw, and Ur.
\label{tab:Task2OneshotPromptingWithRationaleGPT4o}
}
\tiny
\begin{tabular}{p{0.28\textwidth} p{0.45\textwidth} p{0.22\textwidth}}
\toprule
\multicolumn{3}{p{0.98\textwidth}}{%
\textbf{\textcolor{olive}{Inputs to the VLM}}
} \\
\midrule
\textbf{Image} & \textbf{Text} & \textbf{Prompt Question} \\
\midrule
\adjustbox{valign=t}{\includegraphics[width=\linewidth]{images/TaskExamples/img859.pdf}}
&
\texttt{Eswatini Beverages Ltd (EBL) was a subsidiary of SABMiller until 10 October 2016 when it was acquired by Anheuser-Busch InBev. It is a beverage and brewing company in Matsapha, Eswatini. The company was formed in 1995 by the merger of Eswatini Breweries, Ngwane Breweries, and Eswatini Bottlers. EBL produces and markets soft drinks, beer, and other alcoholic drinks. The Swazi beer produced is Sibebe Premium Lager named after the Sibebe rock, but other beers are brewed as well including Castle Lager. EBL dominates the beer market in Eswatini with a market share of 87\% in 2014 227 hl of lager beer were sold. Retooling in 2013 led to a temporary beer shortage in Eswatini.}
&
\adjustbox{valign=t}{%
\parbox{\linewidth}{%
\texttt{\textcolor{red}{Question: Describe the overall scene in this image. What is the setting and what activities or events are taking place?} \textcolor{cyan}{Please provide both your analysis and rationale.}}%
}%
}
\\
\midrule
\multicolumn{3}{p{0.98\textwidth}}{%
\textbf{\textcolor{olive}{Responses generated by the VLM}}
} \\
\midrule
\multicolumn{3}{p{0.98\textwidth}}{%
\textbf{\textcolor{blue}{[English Response:]}} \texttt{Image Analysis: The scene is set in a rural outdoor environment, likely in a savanna or grassland area. The landscape features open fields with sparse trees and a distant view of hills or mountains, suggesting a natural and expansive setting. Rationale: 1. Setting: - The image is taken from a vantage point on a porch or deck, as indicated by the wooden railing in the foreground. - A thatched-roof building is partially visible, suggesting a traditional or rustic architectural style, possibly a lodge or a rural home. 2. Objects and Elements: - A bottle and a glass of beer are prominently placed on the railing, indicating a moment of relaxation or leisure. - The label on the bottle reads Sibebe, which is a brand of beer, hinting at a specific regional or cultural context. 3. Activities and Events: - In the background, a horse is grazing, adding to the pastoral and serene atmosphere. - There are no visible human activities, emphasizing a peaceful and quiet setting. 4. Overall Atmosphere: - The scene conveys a sense of tranquility and connection with nature, with the presence of the horse and the open landscape enhancing the feeling of calmness. - The combination of the beer and the natural setting suggests a leisurely moment, possibly enjoyed by someone taking in the view. This comprehensive overview captures the essence of a peaceful rural setting, highlighting elements of relaxation and natural beauty.}
} \\
\midrule
\multicolumn{3}{p{0.98\textwidth}}{%
\textbf{\textcolor{blue}{[Japanese Response:]}} \begin{CJK*}{UTF8}{min}画像分析：自然豊かな屋外環境が舞台です。前景の木製の手すりの上には瓶のビールとグラスが置かれており、落ち着いた余暇の雰囲気を生み出しています。背景には草地が広がり、点在する木々の向こうに藁葺き屋根の建物と草を食む馬が見えます。根拠（Rationale）：手すり（前景）は眺望用のデッキやポーチと判断できます。瓶とグラスはくつろぎや滞在を示唆します。藁葺き屋根と放牧中の馬は伝統的・田園的な環境を示します。遠景の丘陵は開放感と静けさを強調します。\end{CJK*}
} \\
\midrule
\multicolumn{3}{p{0.98\textwidth}}{%
\textbf{\textcolor{blue}{[Swahili Response:]}} \texttt{Uchambuzi wa Picha: 1. Mazingira: - Picha inaonyesha mandhari ya nje, inaonekana kuwa katika eneo la vijijini au la kitalii. - Kuna nyumba yenye paa la nyasi, ambayo ni ya mtindo wa jadi au wa kiasili, ikionyesha uwezekano wa kuwa katika eneo la Kiafrika. - Mandhari ya nyuma inaonyesha miti na nyasi, ikionyesha mazingira ya savanna au eneo lenye hali ya hewa ya joto. 2. Shughuli na Matukio: - Kuna chupa ya bia na glasi iliyojaa bia kwenye uzio wa mbao, ikionyesha kuwa mtu anaweza kuwa anafurahia kinywaji huku akitazama mandhari. - Farasi anaonekana akila nyasi kwa mbali, ikionyesha uwepo wa wanyama na uwezekano wa shughuli za ufugaji au utalii wa wanyama. Rationale: - Chupa ya bia na glasi zinaonyesha kuwa eneo hili linaweza kuwa sehemu ya mapumziko au hoteli ambapo wageni wanaweza kufurahia vinywaji huku wakitazama mandhari. - Nyumba yenye paa la nyasi na farasi zinaweza kuashiria utamaduni wa eneo hilo na shughuli za kitalii zinazohusiana na utamaduni na mazingira ya asili. Kwa ujumla, picha inaonyesha mandhari ya utulivu na ya kitalii, ambapo mtu anaweza kufurahia vinywaji huku akitazama mandhari ya asili na wanyama.}
} \\
\midrule
\multicolumn{3}{p{0.98\textwidth}}{%
\textbf{\textcolor{blue}{[Urdu Response:]}} \begin{UrduText}تصویری تجزیہ: منظر ایک دیہی، قدرتی کھلے علاقے میں دکھائی دیتا ہے۔ سامنے لکڑی کی ریلنگ پر بوتل اور گلاس رکھے ہیں، جو فراغت اور سکون کی فضا کی طرف اشارہ کرتے ہیں۔ دائیں جانب گھاس پھوس کی چھت والی عمارت جزوی طور پر نظر آتی ہے۔ پس منظر میں کھلا سبزہ زار، چھٹ پٹ درخت اور دور پہاڑیاں دکھائی دیتی ہیں۔ وجوہِ بنیاد (Rationale): سیٹنگ: ریلنگ کی موجودگی سے اندازہ ہوتا ہے کہ تصویر کسی پورچ یا ڈیک سے لی گئی ہے۔ اشیاء: بوتل پر Sibebe کا لیبل اور گلاس میں مشروب فرصت کے لمحات کی علامت ہیں۔ سرگرمیاں: پس منظر میں گھاس چرتا گھوڑا دیہی اور پرسکون فضا کو ابھارتا ہے؛ انسانی سرگرمی واضح نہیں۔ مجموعی تاثر: فطرت سے قربت اور ٹھہراؤ کا احساس، گویا کوئی فرد منظر سے لطف اندوز ہو رہا ہو۔\end{UrduText}
} \\
\bottomrule
\end{tabular}
\end{table*}

\clearpage
\subsubsection{VLM Responses for Relation Understanding (Task 3)}
\begin{table*}[ht]
\centering
\caption{
An example from our benchmark showing responses generated by \textbf{GPT-4o-mini} for \emph{Relation Understanding}.
In this setting, no rationales are generated by the model.
Given an image-text pair as input, we show the outputs in four languages: En, Jp, Sw, and Ur.
\label{tab:Task3OneshotPromptingNoRationaleGPT4omini}
}
\tiny
\begin{tabular}{p{0.28\textwidth} p{0.45\textwidth} p{0.22\textwidth}}
\toprule
\multicolumn{3}{p{0.98\textwidth}}{%
\textbf{\textcolor{olive}{Inputs to the VLM}}
} \\
\midrule
\textbf{Image} & \textbf{Text} & \textbf{Prompt Question} \\
\midrule
\adjustbox{valign=t}{\includegraphics[width=\linewidth]{images/TaskExamples/img859.pdf}}
&
\texttt{Eswatini Beverages Ltd (EBL) was a subsidiary of SABMiller until 10 October 2016 when it was acquired by Anheuser-Busch InBev. It is a beverage and brewing company in Matsapha, Eswatini. The company was formed in 1995 by the merger of Eswatini Breweries, Ngwane Breweries, and Eswatini Bottlers. EBL produces and markets soft drinks, beer, and other alcoholic drinks. The Swazi beer produced is Sibebe Premium Lager named after the Sibebe rock, but other beers are brewed as well including Castle Lager. EBL dominates the beer market in Eswatini with a market share of 87\% in 2014 227 hl of lager beer were sold. Retooling in 2013 led to a temporary beer shortage in Eswatini.}
&
\adjustbox{valign=t}{%
\parbox{\linewidth}{%
\texttt{\textcolor{red}{Question: Identify any interactions or relationships between objects or entities in this image. Explain any spatial, functional, or social connections.}}%
}%
}
\\
\midrule
\multicolumn{3}{p{0.98\textwidth}}{%
\textbf{\textcolor{olive}{Responses generated by the VLM}}
} \\
\midrule
\multicolumn{3}{p{0.98\textwidth}}{%
\textbf{\textcolor{blue}{[English Response:]}} \texttt{Analysis of Interactions and Relationships 1. Spatial Relationships: - The beer bottle and glass are positioned on a wooden railing, indicating a setting where someone might be enjoying a drink while observing the surroundings. - The horse is located in the background, suggesting a rural or natural environment. The distance between the horse and the drink indicates that they are not directly interacting but coexist in the same scene. 2. Functional Relationships: - The beer (both the bottle and the glass) serves a recreational purpose, likely intended for enjoyment by a person who is not visible in the image. This suggests a social or leisure activity. - The horse, being a living animal, may represent a connection to nature or rural life, enhancing the ambiance of relaxation that the drink symbolizes. 3. Social Connections: - The presence of the beer suggests a social context, possibly indicating that the setting is meant for relaxation or socializing, such as a gathering or a moment of solitude in nature. - The horse may symbolize companionship or a connection to traditional rural lifestyles, which can evoke feelings of nostalgia or appreciation for nature. Overall, the image captures a serene moment where leisure (represented by the beer) and nature (represented by the horse) coexist, creating a tranquil atmosphere.}
} \\
\midrule
\multicolumn{3}{p{0.98\textwidth}}{%
\textbf{\textcolor{blue}{[Japanese Response:]}} \begin{CJK*}{UTF8}{min}
解析：この画像には、手前の木製の手すりの上に置かれたビールの瓶とグラス、そして背景にいる馬が写っています。
1. 空間的関係：瓶とグラスは互いに近接し、\texttt{Sibebe} のラベルが見えることから飲用の状況が整っていると分かります。馬は離れた草地に位置し、同一の屋外環境に共存しつつも直接の相互作用はありません。
2. 機能的関係：瓶とグラスは「くつろぎ／飲用」という機能を担い、眺望を楽しむ行為を支えます。馬は風景の一部として場面に牧歌的な雰囲気を付与します。
3. 社会的つながり：飲み物の存在は、ひとりの休憩や少人数の語らいなど、リラックスした社会的文脈を示唆します。馬は地域の生活様式や自然との結び付きを象徴します。
総じて、飲用（余暇）と自然が調和して共存する静かな瞬間が切り取られています。
\end{CJK*}
} \\
\midrule
\multicolumn{3}{p{0.98\textwidth}}{%
\textbf{\textcolor{blue}{[Swahili Response:]}} \texttt{Uchambuzi wa Picha. Vitu Vilivyopo: 1. Chupa ya Bia: Chupa ya bia ya "Sibebe" na glasi yenye bia. 2. Farasi: Farasi mmoja anayeonekana nyuma ya chupa ya bia. 3. Mandhari: Nyumba ya jadi na mazingira ya asili, ikiwa na miti na milima ya mbali. Mahusiano na Uhusiano: 1. Uhusiano wa Kijamii: - Chupa ya bia na glasi zinaweza kuashiria mazingira ya kijamii ambapo watu wanakusanyika kufurahia vinywaji. Hii inaweza kuwa ni sehemu ya utamaduni wa kupumzika na kuungana na marafiki au familia. 2. Uhusiano wa Kimwili: - Farasi anayeonekana nyuma ya chupa ya bia anaweza kuwakilisha uhusiano wa karibu kati ya binadamu na wanyama. Katika mazingira ya vijijini, farasi mara nyingi hutumiwa kwa kazi na burudani, na hivyo kuonyesha umuhimu wa wanyama katika maisha ya kila siku. 3. Uhusiano wa Kiutendaji: - Bia inaweza kutumika kama kinywaji cha kupumzika baada ya kazi, na hivyo kuonyesha uhusiano wa kazi na burudani. Farasi, kwa upande mwingine, anaweza kuwa na jukumu katika shughuli za kilimo au usafiri, hivyo kuonyesha jinsi maisha ya watu yanavyohusiana na wanyama. Hitimisho: Picha hii inaonyesha uhusiano wa kijamii na kiutendaji kati ya binadamu, wanyama, na mazingira yao. Inaweza kuwakilisha maisha ya kawaida ya watu katika eneo la vijijini, ambapo burudani na kazi vinakutana.}
} \\
\midrule
\multicolumn{3}{p{0.98\textwidth}}{%
\textbf{\textcolor{blue}{[Urdu Response:]}} \begin{UrduText}اس تصویر میں موجود اشیاء اور عناصر کے درمیان تعلقات یوں واضح ہوتے ہیں:
1. مکانی تعلقات: بیئر کی بوتل اور گلاس لکڑی کی ریلنگ پر ساتھ رکھے ہیں، جو ایک آرام دہ سماجی ماحول کی طرف اشارہ کرتا ہے۔ گھوڑا پس منظر میں فاصلے پر ہے—منظر کا حصہ ہے مگر براہِ راست تعامل میں نہیں۔
2. فعالی تعلقات: بوتل اور گلاس “تفریح/آرام” کے لیے ہیں اور منظر کا لطف لینے کے عمل کو سہارا دیتے ہیں۔ گھوڑا بطور قدرتی عنصر منظر میں دیہی اور پُرسکون تاثر بڑھاتا ہے۔
3. سماجی ربط: مشروب کی موجودگی تنہائی میں وقفہ یا مختصر میل ملاقات جیسے غیررسمی سیاق کو ظاہر کرتی ہے، جبکہ گھوڑا مقامی طرزِ زندگی اور فطرت سے وابستگی کی علامت بنتا ہے۔
مجموعی طور پر یہ تصویر تفریح اور فطرت کے ہم آہنگ امتزاج کو پُرسکون فضا میں پیش کرتی ہے۔\end{UrduText}
} \\
\bottomrule
\end{tabular}
\end{table*}

\clearpage
\begin{table*}[t]
\centering
\caption{
An example from our benchmark showing responses generated by \textbf{GPT-4o-mini} for \emph{Relation Understanding}.
In this setting, \underline{the model generates rationales}, in addition to analyzing the image.
Given an image-text pair as input, we show the outputs in four languages: En, Jp, Sw, and Ur.
\label{tab:Task3OneshotPromptingWithRationaleGPT4omini}
}
\tiny
\begin{tabular}{p{0.28\textwidth} p{0.45\textwidth} p{0.22\textwidth}}
\toprule
\multicolumn{3}{p{0.98\textwidth}}{%
\textbf{\textcolor{olive}{Inputs to the VLM}}
} \\
\midrule
\textbf{Image} & \textbf{Text} & \textbf{Prompt Question} \\
\midrule
\adjustbox{valign=t}{\includegraphics[width=\linewidth]{images/TaskExamples/img859.pdf}}
&
\texttt{Eswatini Beverages Ltd (EBL) was a subsidiary of SABMiller until 10 October 2016 when it was acquired by Anheuser-Busch InBev. It is a beverage and brewing company in Matsapha, Eswatini. The company was formed in 1995 by the merger of Eswatini Breweries, Ngwane Breweries, and Eswatini Bottlers. EBL produces and markets soft drinks, beer, and other alcoholic drinks. The Swazi beer produced is Sibebe Premium Lager named after the Sibebe rock, but other beers are brewed as well including Castle Lager. EBL dominates the beer market in Eswatini with a market share of 87\% in 2014 227 hl of lager beer were sold. Retooling in 2013 led to a temporary beer shortage in Eswatini.}
&
\adjustbox{valign=t}{%
\parbox{\linewidth}{%
\texttt{\textcolor{red}{Question: Identify any interactions or relationships between objects or entities in this image. Explain any spatial, functional, or social connections.} \textcolor{cyan}{Please provide both your analysis and rationale.} }%
}%
}
\\
\midrule
\multicolumn{3}{p{0.98\textwidth}}{%
\textbf{\textcolor{olive}{Responses generated by the VLM}}
} \\
\midrule
\multicolumn{3}{p{0.98\textwidth}}{%
\textbf{\textcolor{blue}{[English Response:]}} \texttt{ Analysis: The image features a bottle of beer and a glass filled with beer placed on a wooden railing, with a scenic background that includes a horse and a natural landscape. The interactions and relationships between the objects and entities in the image can be described as follows: 1. Spatial Arrangement: - The beer bottle and glass are positioned prominently on the railing, suggesting they are intended for immediate consumption. The placement indicates a casual setting, likely for relaxation or enjoyment. - In the background, a horse is seen grazing, which adds a rural or natural context to the scene. The horse is positioned at a distance from the railing, indicating a separation between the human activity (drinking) and the animal. 2. Functional Connections: - The beer bottle and glass serve a functional purpose of providing refreshment. The presence of both suggests that someone is likely enjoying a drink while taking in the view. - The horse in the background may not have a direct functional connection to the beer but contributes to the overall ambiance of the setting, enhancing the experience of relaxation and leisure. 3. Social Interactions: - The image implies a social or leisurely context, as the presence of the beer suggests that it may be part of a gathering or a moment of relaxation. The setting could be indicative of a social event or a personal retreat. - The horse, while not interacting directly with the human elements, adds to the serene atmosphere, which may evoke feelings of tranquility and connection to nature. Overall, the image captures a moment of leisure where the human element (the beer) is juxtaposed with the natural element (the horse and landscape), creating a harmonious scene that suggests relaxation and enjoyment in a rural setting. }
} \\
\midrule
\multicolumn{3}{p{0.98\textwidth}}{%
\textbf{\textcolor{blue}{[Japanese Response:]}} \begin{CJK*}{UTF8}{min}
解析：手前の木製手すりにはビールの瓶とグラスが置かれ、背景には草地と馬、藁葺きの建物が見えます。以下の観点から相互作用・関係を示します。
1. 空間配置（根拠）：瓶とグラスは視覚的に一対として近接配置され、今まさに飲める状態（グラスに注がれている）です。馬は離れた位置におり、同一の屋外環境に共存するが、直接の関与はしません。
2. 機能的関係（根拠）：瓶／グラスはリフレッシュメントという機能を果たし、眺望を楽しむ行為と結び付きます。馬は風景の主体として、場面に牧歌性と静けさを付与します。
3. 社会的文脈（根拠）：飲み物の存在は、独りの休憩や小さな集いといった穏やかな社交の場面を示唆します。馬は地域の暮らしや自然との結び付きを象徴します。
総括：飲用という人間活動と、馬・景観という自然要素が対置されながらも調和し、余暇と自然の共存を表現しています。
\end{CJK*}
} \\
\midrule
\multicolumn{3}{p{0.98\textwidth}}{%
\textbf{\textcolor{blue}{[Swahili Response:]}} \texttt{Uchambuzi: Picha inaonyesha chupa ya bia na glasi ya bia iliyowekwa juu ya meza, huku nyuma ikiwa na mandhari ya asili ambayo ina farasi na nyumba ya jadi. Mwingiliano na uhusiano kati ya vitu na viumbe katika picha hii unaweza kueleweka kama ifuatavyo: 1. Mpangilio wa Nafasi: - Chupa ya bia na glasi ziko karibu na ukingo wa meza, zikionyesha kuwa mtu anayeweza kuwa anatumia bidhaa hizo. Hii inaashiria mazingira ya kupumzika au ya kijamii. - Farasi aliye nyuma anasimama katika eneo la wazi, akionyesha kuwa kuna uhusiano wa asili na maisha ya vijijini. 2. Mifano ya Kazi: - Chupa ya bia na glasi zinaweza kuashiria shughuli ya kijamii, kama vile kukutana na marafiki au familia, ambapo watu wanakunywa na kufurahia wakati pamoja. - Mandhari ya asili inaonyesha kuwa eneo hili linaweza kuwa ni sehemu ya mapumziko au hoteli ya kitalii, ambapo wageni wanaweza kufurahia mazingira ya asili. 3. Mwingiliano wa Kijamii: - Uwepo wa bia unadhihirisha utamaduni wa kunywa pombe katika mazingira ya kijamii, ambapo watu wanakutana na kusherehekea au kupumzika. - Farasi aliye nyuma anaweza kuwakilisha maisha ya vijijini au shughuli za kilimo, akionyesha uhusiano kati ya watu na mazingira yao. Kwa ujumla, picha hii inakazia mwingiliano wa kijamii na utamaduni wa kupumzika katika mazingira ya asili, ambapo watu wanaweza kufurahia bia huku wakitazama mandhari ya kuvutia.}
} \\
\midrule
\multicolumn{3}{p{0.98\textwidth}}{%
\textbf{\textcolor{blue}{[Urdu Response:]}} \begin{UrduText}تحلیل: منظر میں سامنے لکڑی کی ریلنگ پر بیئر کی بوتل اور گلاس رکھے ہیں؛ پس منظر میں گھوڑا اور قدرتی مناظر (درخت، گھاس، دُور کے ٹیلے) دکھائی دیتے ہیں۔
1. مکانی ترتیب (دلائل): بوتل اور گلاس باہم قریب اور استعمال کے لیے تیار حالت میں ہیں؛ گھوڑا فاصلے پر ہے، اس لیے براہِ راست تعامل نہیں مگر ایک ہی فضا میں ہم موجود ہیں۔
2. فعالی ربط (دلائل): بوتل/گلاس تفریح اور سکون کے مقصد سے وابستہ ہیں؛ گھوڑا بطور قدرتی عنصر منظر کو دیہی، پُرسکون اور فطرت آشنا بناتا ہے۔
3. سماجی سیاق (دلائل): مشروب کی موجودگی غیررسمی میل ملاقات یا تنہائی میں آرام کی صورت حال کی طرف اشارہ کرتی ہے؛ گھوڑا مقامی طرزِ زندگی اور فطرت سے وابستگی کی علامت ہے۔
خلاصہ: انسانی سرگرمی (مشروب کا لطف) اور فطری عناصر (گھوڑا و مناظر) ایک متوازن اور ہم آہنگ کیفیت پیدا کرتے ہیں۔\end{UrduText}
} \\
\bottomrule
\end{tabular}
\end{table*}

\clearpage
\subsubsection{VLM Responses for Semantic Segmentation  (Task 4)}
\begin{table*}[ht]
\centering
\caption{
An example from our benchmark showing responses generated by \textbf{GPT-4o-mini} for \emph{Semantic Segmentation}.
In this setting, no rationales are generated by the model.
Given an image-text pair as input, we show the outputs in four languages: En, Jp, Sw, and Ur.
\label{tab:Task4OneshotPromptingNoRationaleGPT4omini}
}
\tiny
\begin{tabular}{p{0.28\textwidth} p{0.45\textwidth} p{0.22\textwidth}}
\toprule
\multicolumn{3}{p{0.98\textwidth}}{%
\textbf{\textcolor{olive}{Inputs to the VLM}}
} \\
\midrule
\textbf{Image} & \textbf{Text} & \textbf{Prompt Question} \\
\midrule
\adjustbox{valign=t}{\includegraphics[width=\linewidth]{images/TaskExamples/img859.pdf}}
&
\texttt{Eswatini Beverages Ltd (EBL) was a subsidiary of SABMiller until 10 October 2016 when it was acquired by Anheuser-Busch InBev. It is a beverage and brewing company in Matsapha, Eswatini. The company was formed in 1995 by the merger of Eswatini Breweries, Ngwane Breweries, and Eswatini Bottlers. EBL produces and markets soft drinks, beer, and other alcoholic drinks. The Swazi beer produced is Sibebe Premium Lager named after the Sibebe rock, but other beers are brewed as well including Castle Lager. EBL dominates the beer market in Eswatini with a market share of 87\% in 2014 227 hl of lager beer were sold. Retooling in 2013 led to a temporary beer shortage in Eswatini.}
&
\adjustbox{valign=t}{%
\parbox{\linewidth}{%
\texttt{\textcolor{red}{Question: Divide this image into different semantic regions. Label each region (e.g., sky, buildings, people, street) and briefly describe its contents.}}%
}%
}
\\
\midrule
\multicolumn{3}{p{0.98\textwidth}}{%
\textbf{\textcolor{olive}{Responses generated by the VLM}}
} \\
\midrule
\multicolumn{3}{p{0.98\textwidth}}{%
\textbf{\textcolor{blue}{[English Response:]}} \texttt{Image Analysis 1. Foreground: - Label: Beverage - Description: A green bottle of beer labeled "Sibebe" is prominently placed on a wooden railing, alongside a glass filled with beer. The glass has a frothy head, indicating it has just been poured. 2. Middle Ground: - Label: Horse - Description: A horse is visible grazing in the field. It is positioned to the right of the image, adding a natural element to the scene. 3. Background: - Label: Landscape - Description: The background features a grassy field with sparse trees and hills in the distance. The landscape appears to be rural and serene. 4. Right Side: - Label: Structure - Description: A thatched-roof building is partially visible on the right side of the image. The building is made of stone, contributing to the rustic atmosphere. 5. Overall Composition: - The image has a balanced composition with the beverage in the foreground drawing attention, while the horse and landscape provide context and depth. The warm colors of the beer contrast with the natural greens and browns of the background, creating a harmonious scene.}
} \\
\midrule
\multicolumn{3}{p{0.98\textwidth}}{%
\textbf{\textcolor{blue}{[Japanese Response:]}} \begin{CJK*}{UTF8}{min}画像のセマンティック領域の区分： 1. 前景（飲み物）— 内容：ブランド名,\textit{Sibebe} のラベルが見えるビール瓶と、泡立ったビールの入ったグラス。木製の手すりの上に置かれ、シーンの主たる焦点となっている。 2. 中景（馬）— 内容：草地で草を食む馬が見える。自然要素が加わり、屋外の穏やかな雰囲気を強めている。 3. 背景（風景）— 内容：まばらな木々と遠景の丘を伴う広い草原。農村的で静かな環境を示す。 4. 右側（建物）— 内容：茅葺き屋根の小屋／ロッジが部分的に見える。素朴な建築要素が場面の文脈を補強している。 5. 補助要素（手すり）— 内容：前景の手すりがアウトドアの設定を明確にし、視覚的なフレーミングを与える。 以上のように、前景・中景・背景・建物・補助要素に分割でき、それぞれが場面理解に寄与している。\end{CJK*}
} \\
\midrule
\multicolumn{3}{p{0.98\textwidth}}{%
\textbf{\textcolor{blue}{[Swahili Response:]}} \texttt{Uchambuzi wa Picha 1. Eneo la Mbele - Maelezo: Katika sehemu hii, kuna chupa ya bia ya "Sibebe" na glasi ya bia iliyojaa. Vitu hivi viko juu ya meza ya mbao, ikionyesha mazingira ya kupumzika au burudani. 2. Eneo la Nyuma - Maelezo: Katika sehemu hii, kuna farasi mmoja anayeonekana akichungulia kwenye uwanja wa majani. Hii inaonyesha mandhari ya asili na maisha ya wanyama. 3. Anga - Maelezo: Anga linaonekana kuwa wazi na lina milima ya mbali, ikionyesha mazingira ya wazi na ya kupumzika. Hali hii inachangia katika mandhari ya jumla ya picha. 4. Majengo - Maelezo: Katika upande wa kulia, kuna nyumba yenye paa la makuti, ikionyesha muundo wa jadi wa makazi. Nyumba hii inaonekana kuwa sehemu ya mazingira ya vijijini. 5. Uwanja - Maelezo: Uwanja una majani ya kijani kibichi, ukionyesha mazingira ya asili na ya shamba. Hapa kuna pia mti mmoja au wawili, wakiongeza uzuri wa mandhari. Muundo wa Picha Picha hii ina muundo wa tabaka, ambapo vitu vya mbele (bia) viko karibu na mtazamaji, wakati vitu vya nyuma (farasi na milima) viko mbali, na kuunda hisia ya kina katika picha.}
} \\
\midrule
\multicolumn{3}{p{0.98\textwidth}}{%
\textbf{\textcolor{blue}{[Urdu Response:]}} \begin{UrduText}تصویر کو درج ذیل معنوی علاقوں میں تقسیم کیا جا سکتا ہے: 1) پیش منظر (مشروبات): لکڑی کی ریلنگ پر \textit{Sibebe} لیبل والی بیئر کی بوتل اور جھاگ کے ساتھ بھرا ہوا گلاس نمایاں ہیں۔ یہی حصہ منظر کی توجہ کا مرکز ہے۔ 2) درمیانی منظر (گھوڑا): گھاس زار میں ایک گھوڑا دکھائی دیتا ہے جو منظر میں قدرتی اور پرسکون تاثر پیدا کرتا ہے۔ 3) پس منظر (قدرتی مناظر): دور تک پھیلا ہوا سبزہ، چھٹا چھٹا درخت اور پہاڑی ابھار ایک دیہی اور پر سکون ماحول کو ظاہر کرتے ہیں۔ 4) دائیں جانب (عمارت): بھوسے کی چھت والی ایک جھونپڑی/لؤج کا حصہ نظر آتا ہے جو منظر کی سادہ دیہی دلکشی کو ابھارتا ہے۔ 5) معاون عنصر (ہینڈ ریل): ریلنگ آؤٹ ڈور سیٹنگ کو واضح کرتی ہے اور فریمینگ کا کام دیتی ہے۔ ان تمام خطوں کی بنا پر تصویر کا مجموعی تاثر ایک پُرسکون، قدرتی اور خوشگوار ماحول ہے۔\end{UrduText}
} \\
\bottomrule
\end{tabular}
\end{table*}

\clearpage
\begin{table*}[t]
\centering
\caption{
An example from our benchmark showing responses generated by \textbf{GPT-4o-mini} for \emph{Semantic Segmentation}.
In this setting, \underline{the model generates rationales}, in addition to analyzing the image.
Given an image-text pair as input, we show the outputs in four languages: En, Jp, Sw, and Ur.
}
\label{tab:Task4OneshotPromptingWithRationaleGPT4omini}
\tiny
\begin{tabular}{p{0.28\textwidth} p{0.45\textwidth} p{0.22\textwidth}}
\toprule
\multicolumn{3}{p{0.98\textwidth}}{%
\textbf{\textcolor{olive}{Inputs to the VLM}}
} \\
\midrule
\textbf{Image} & \textbf{Text} & \textbf{Prompt Question} \\
\midrule
\adjustbox{valign=t}{\includegraphics[width=\linewidth]{images/TaskExamples/img859.pdf}} &
\texttt{Eswatini Beverages Ltd (EBL) was a subsidiary of SABMiller until 10 October 2016 when it was acquired by Anheuser-Busch InBev. It is a beverage and brewing company in Matsapha, Eswatini. The company was formed in 1995 by the merger of Eswatini Breweries, Ngwane Breweries, and Eswatini Bottlers. EBL produces and markets soft drinks, beer, and other alcoholic drinks. The Swazi beer produced is Sibebe Premium Lager named after the Sibebe rock, but other beers are brewed as well including Castle Lager. EBL dominates the beer market in Eswatini with a market share of 87\% in 2014. Retooling in 2013 led to a temporary beer shortage in Eswatini.} &
\adjustbox{valign=t}{\parbox{\linewidth}{\raggedright \texttt{\textcolor{red}{Question: Divide this image into different semantic regions. Label each region (e.g., sky, buildings, people, street) and briefly describe its contents.}
\textcolor{cyan}{Please provide both your analysis and rationale.} }}} \\
\midrule
\multicolumn{3}{p{0.98\textwidth}}{%
\textbf{\textcolor{olive}{Responses generated by the VLM}}
} \\
\midrule
\multicolumn{3}{p{0.98\textwidth}}{\textbf{\textcolor{blue}{[English Response:]}} \texttt{Image Analysis: Semantic Regions 1. Foreground (Beverages) - Contents: A bottle of Sibebe beer and a glass filled with beer. The bottle is prominently displayed, with its label facing the viewer, indicating the brand. The glass has a frothy top, suggesting it has just been poured. 2. Middle Ground (Natural Landscape) - Contents: An open field with sparse vegetation, including a few trees. This area provides a natural setting, enhancing the outdoor ambiance. The ground appears dry, typical of a rural or semi-arid environment. 3. Background (Horse and Structure) - Contents: A horse is visible grazing in the distance, adding a sense of tranquility to the scene. To the right, there is a thatched-roof structure, likely a hut or lodge, which contributes to the rustic feel of the image. 4. Left Side (Natural Elements) - Contents: A partially visible tree and some rocks, which add to the natural landscape. This area helps frame the scene and provides context for the outdoor setting. 5. Right Side (Building Details) - Contents: The wall of the thatched-roof structure is visible, showcasing a warm color that contrasts with the natural greens and browns of the landscape. This adds depth to the image and emphasizes the rural setting. Overall, the image captures a serene outdoor scene featuring a refreshing beverage in the foreground, set against a backdrop of nature and rustic architecture. The combination of elements creates a relaxed atmosphere, ideal for enjoying a drink in a natural setting.}} \\
\midrule
\multicolumn{3}{p{0.98\textwidth}}{\textbf{\textcolor{blue}{[Japanese Response:]}} \begin{CJK*}{UTF8}{min}画像分析：セマンティック領域 1. 前景（飲み物）— 内容：木製の手すり上に、ブランド名 \textit{Sibebe} のビール瓶と、泡の立つビールグラスが目立って置かれている。直近に注がれたことを示唆する描写。— 根拠：最前面に配置され、視線を強く引き付けるため。 2. 中景（自然景観）— 内容：まばらな植生と数本の木がある開けた草地。半乾燥〜農村的な環境を示す。— 根拠：地表の質感と植生密度から環境の性質が読み取れる。 3. 背景（馬と建物）— 内容：遠景に草を食む馬、右手に茅葺き屋根の小屋／ロッジ。— 根拠：動物の存在が寛ぎの情景を補強し、建築要素が土地の文脈を与える。 4. 左側（自然要素）— 内容：画面左端に樹木や岩が部分的に写り、場面のフレーミングとなっている。— 根拠：縁辺部の自然物が奥行きと屋外性を強調。 5. 右側（建物ディテール）— 内容：茅葺き建物の壁面が温かみのある色調で示され、周囲の緑・茶の自然色と対比をなす。— 根拠：色調の対比が素朴な農村的雰囲気を高める。 総括：前景の飲み物、中景の自然、背景の馬と建物が調和し、穏やかな屋外シーンを形成している。\end{CJK*}} \\
\midrule
\multicolumn{3}{p{0.98\textwidth}}{\textbf{\textcolor{blue}{[Swahili Response:]}} \texttt{Uchambuzi wa Picha: Maeneo ya Kimaana 1. Mbele (Chupa ya Bia na Kioo) - Yaliyomo: Chupa ya bia ya Sibebe na kioo kilichojaa bia. Hii inawakilisha sehemu ya burudani au kupumzika, ikionyesha kuwa mtu anafurahia wakati wa kupumzika. 2. Katikati (Mandhari ya Asili) - Yaliyomo: Nyasi za kijani kibichi na miti, zikionyesha mazingira ya asili. Hapa kuna farasi mmoja akichungulia nyuma, akiongeza uzuri wa mandhari na kuonyesha maisha ya wanyama katika eneo hilo. 3. Nyuma (Jengo la Kijadi) - Yaliyomo: Jengo lenye paa la nyasi, linaonyesha usanifu wa jadi. Hii inatoa muktadha wa kitamaduni na mazingira ya kijiji au eneo la kupumzika. 4. Kulia (Mifumo ya Samahani) - Yaliyomo: Sehemu ya mbenyo wa meza, ikionyesha kuwa mtu anatumia meza hiyo kwa ajili ya kunywa au kupumzika. Hii inasisitiza hali ya kupumzika na urahisi wa mazingira. 5. Kushoto (Mandhari ya Mbali) - Yaliyomo: Milima ya mbali inayoonekana nyuma, ikionyesha uzuri wa mandhari na kuimarisha hisia za kuwa katika eneo la mbali na la asili. Muhtasari: Picha hii inaonyesha mazingira ya kupumzika na ya asili, ambapo mtu anafurahia bia huku akitazama mandhari nzuri ya asili na jengo la jadi. Hali hii inatoa hisia ya utulivu na furaha katika mazingira ya nje.}} \\
\midrule
\multicolumn{3}{p{0.98\textwidth}}{\textbf{\textcolor{blue}{[Urdu Response:]}} \begin{UrduText}تصویری تجزیہ: معنوی تقسیم اور دلائل 1) پیش منظر (مشروبات) — مواد: \textit{Sibebe} بیئر کی بوتل اور جھاگ والا بھرا ہوا گلاس لکڑی کی ریلنگ پر نمایاں ہیں۔ دلیل: یہ اشیاء سب سے آگے ہیں اور فوراً توجہ کھینچتی ہیں؛ برانڈ لیبل منظر کے سیاق میں اضافہ کرتا ہے۔ 2) درمیانی منظر (قدرتی مناظر) — مواد: کھلا میدان، کم گھاس اور چند درخت؛ نیم خشک/دیہی ماحول کا اشارہ۔ دلیل: زمینی بافت اور سبزے کی کمی فطری سیاق بتاتی ہے۔ 3) پس منظر (گھوڑا اور عمارت) — مواد: دور چرنے والا گھوڑا اور دائیں جانب بھوسے کی چھت والی جھونپڑی/لؤج۔ دلیل: جانور پرسکون فضا کو ابھارتا ہے جبکہ عمارت کا روایتی طرزِ تعمیر مقامی سیاق فراہم کرتا ہے۔ 4) بائیں کنارہ (قدرتی عناصر) — مواد: درخت/چٹانی ٹکڑے جزوی طور پر دکھائی دیتے ہیں جو فریم بناتے اور باہر کے ماحول کو نمایاں کرتے ہیں۔ دلیل: اطراف کے قدرتی عناصر گہرائی اور کھلے پن کا احساس بڑھاتے ہیں۔ 5) دایاں کنارہ (عمارت کی جزئیات) — مواد: دیوار اور چھپر کی ساخت گرم رنگوں میں دکھائی دیتی ہے جو اردگرد کی سبز و خاکی رنگت سے متوازن تضاد پیدا کرتی ہے۔ دلیل: یہ جزئیات دیہی دلکشی اور سکون کے تاثر کو مضبوط کرتی ہیں۔ خلاصہ: پیش منظر کی تفریحی شے (بیئر) فطرت اور روایتی عمارت کے پس منظر کے ساتھ ہم آہنگ ہو کر ایک پُرسکون، مدعو کرنے والا بیرونی منظر تشکیل دیتی ہے۔\end{UrduText}
} \\
\bottomrule
\end{tabular}
\end{table*}

\clearpage
\subsubsection{VLM Responses for Image Captioning  (Task 5) }
\begin{table*}[ht]
    \centering
    \caption{
    An example from our benchmark showing responses generated by \textbf{GPT-4o} for \emph{Image Captioning}. 
    In this setting, no rationales are generated by the model. 
    Given an image-text pair as input, we show the outputs in four languages: En, Jp, Sw, and Ur.
    \label{tab:Task5OneshotPromptingNoRationaleGPT4o}
    }
    \tiny
    \begin{tabular}{p{0.28\textwidth} p{0.45\textwidth} p{0.22\textwidth}}
    \toprule
    \multicolumn{3}{p{0.98\textwidth}}{%
    \textbf{\textcolor{olive}{Inputs to the VLM}}
    } \\
    \midrule
    \textbf{Image} & \textbf{Text} & \textbf{Prompt Question} \\
    \midrule
    \adjustbox{valign=t}{\includegraphics[width=\linewidth]{images/TaskExamples/img859.pdf}}
    &
    \texttt{Eswatini Beverages Ltd (EBL) was a subsidiary of SABMiller until 10 October 2016 when it was acquired by Anheuser-Busch InBev. It is a beverage and brewing company in Matsapha, Eswatini. The company was formed in 1995 by the merger of Eswatini Breweries, Ngwane Breweries, and Eswatini Bottlers. EBL produces and markets soft drinks, beer, and other alcoholic drinks. The Swazi beer produced is Sibebe Premium Lager named after the Sibebe rock, but other beers are brewed as well including Castle Lager. EBL dominates the beer market in Eswatini with a market share of 87\% in 2014 227 hl of lager beer were sold. Retooling in 2013 led to a temporary beer shortage in Eswatini.}
    & 
    \adjustbox{valign=t}{%
        \parbox{\linewidth}{%
            \texttt{\textcolor{red}{Question: Provide a detailed natural language description of what is happening in this image, as if describing it to someone who cannot see it.}}%
        }%
    }
    \\
    \midrule
     \multicolumn{3}{p{0.98\textwidth}}{%
    \textbf{\textcolor{olive}{Responses generated by the VLM}}
    } \\
    \midrule
    \multicolumn{3}{p{0.98\textwidth}}{%
    \textbf{\textcolor{blue}{[English Response:]}} \texttt{The image shows a bottle and a glass of beer labeled "Sibebe Premium Lager," which directly corresponds to the text mentioning "The Swazi beer produced is Sibebe Premium Lager named after the Sibebe rock." This connection is evident as the label on the bottle in the image clearly reads "Sibebe Premium Lager," confirming the textual reference. Additionally, the setting in the image, which includes a rural landscape with a thatched-roof building, aligns with the context of Eswatini, where Eswatini Beverages Ltd operates. This supports the text's mention of the company being located in Matsapha, Eswatini, although the specific location in the image is not explicitly identified in the text.}
    } \\
    \midrule
    \multicolumn{3}{p{0.98\textwidth}}{%
    \textbf{\textcolor{blue}{[Japanese Response:]}} \begin{CJK*}{UTF8}{min}画像には「Sibebe Premium Lager」と書かれたビールのボトルと、同じビールが注がれたグラスが写っています。これはテキスト中の「The Swazi beer produced is Sibebe Premium Lager …」という記述と一致します。背景には茅葺き屋根の建物や草地、遠くの丘が見え、素朴な田園風景が広がっています。Eswatini Beverages Ltd が所在するエスワティニの文脈とも整合しており、画像はテキストで言及された銘柄とその雰囲気を視覚的に示しています。\end{CJK*}
    } \\
    \midrule
    \multicolumn{3}{p{0.98\textwidth}}{%
    \textbf{\textcolor{blue}{[Swahili Response:]}} \texttt{ Katika picha, kuna chupa ya bia yenye lebo ya "Sibebe Premium Lager" na glasi iliyojaa bia hiyo. Maandishi yanataja kwamba "Sibebe Premium Lager" ni mojawapo ya bia zinazozalishwa na Eswatini Beverages Ltd (EBL). Hii inathibitisha uhusiano wa moja kwa moja kati ya picha na maandishi. Ushahidi wa kuona: 1. Chupa ya Bia: Picha inaonyesha chupa ya "Sibebe Premium Lager," ambayo inatajwa katika maandishi kama mojawapo ya bidhaa zinazozalishwa na EBL. 2. Mandhari ya Asili: Ingawa maandishi hayataji moja kwa moja mandhari, picha inaonyesha mazingira ya nje ambayo yanaweza kuhusishwa na Eswatini, kutokana na maelezo kuwa kampuni iko Matsapha, Eswatini. Kwa hivyo, picha hii inaonyesha moja kwa moja bidhaa inayozalishwa na kampuni iliyotajwa katika maandishi.}
    } \\
     \midrule
    \multicolumn{3}{p{0.98\textwidth}}{%
    \textbf{\textcolor{blue}{[Urdu Response:]}} \begin{UrduText}تصویر میں ایک دیہی منظر دکھایا گیا ہے۔ سامنے لکڑی کی ریلنگ پر “Sibebe” لیبل والی بوتل اور سنہری مشروب سے بھرا گلاس رکھا ہے۔ پس منظر میں کھلا سبزہ زار، کچھ درخت اور دور پہاڑی سلسلہ دکھائی دیتا ہے، جبکہ دائیں جانب گھاس پھوس کی چھت والا ایک چھوٹا سا پتھریلا مکان نظر آتا ہے۔ مجموعی طور پر منظر پُرسکون اور فطری ہے، جیسے کوئی شخص بیرونی ماحول میں مشروب سے لطف اندوز ہو رہا ہو۔\end{UrduText}
    } \\
    \bottomrule
    \end{tabular}
\end{table*}

\clearpage
\begin{table*}[t]
    \centering
    \caption{
    An example from our benchmark showing responses generated by \textbf{GPT-4o} for \emph{Image Captioning}. 
    In this setting, \underline{the model generates rationales}, in addition to analyzing the image.
    Given an image-text pair as input, we show the outputs in four languages: En, Jp, Sw, and Ur.
    }
    \label{tab:Task5OneshotPromptingWithRationaleGPT4o}
    \tiny
    \begin{tabular}{p{0.28\textwidth} p{0.45\textwidth} p{0.22\textwidth}}
    \toprule
    \multicolumn{3}{p{0.98\textwidth}}{%
    \textbf{\textcolor{olive}{Inputs to the VLM}}
    } \\
    \midrule
    \textbf{Image} & \textbf{Text} & \textbf{Prompt Question} \\
    \midrule
    \adjustbox{valign=t}{\includegraphics[width=\linewidth]{images/TaskExamples/img859.pdf}} &
    \texttt{Eswatini Beverages Ltd (EBL) was a subsidiary of SABMiller until 10 October 2016 when it was acquired by Anheuser-Busch InBev. It is a beverage and brewing company in Matsapha, Eswatini. The company was formed in 1995 by the merger of Eswatini Breweries, Ngwane Breweries, and Eswatini Bottlers. EBL produces and markets soft drinks, beer, and other alcoholic drinks. The Swazi beer produced is Sibebe Premium Lager named after the Sibebe rock, but other beers are brewed as well including Castle Lager. EBL dominates the beer market in Eswatini with a market share of 87\% in 2014. Retooling in 2013 led to a temporary beer shortage in Eswatini.} &
    \adjustbox{valign=t}{\parbox{\linewidth}{\raggedright \texttt{\textcolor{red}{Question: Provide a detailed natural language description of what is happening in this image, as if describing it to someone who cannot see it.} \textcolor{cyan}{Please provide both your analysis and rationale.} }}} \\
    \midrule
     \multicolumn{3}{p{0.98\textwidth}}{%
    \textbf{\textcolor{olive}{Responses generated by the VLM}}
    } \\
    \midrule
    \multicolumn{3}{p{0.98\textwidth}}{\textbf{\textcolor{blue}{[English Response:]}} \texttt{ The image shows a bottle and a glass of beer on a railing with a rural landscape in the background. Here's how the image relates to the text: 1. Sibebe Premium Lager: - Text Reference: The text mentions "The Swazi beer produced is Sibebe Premium Lager named after the Sibebe rock." - Visual Evidence: The image clearly shows a bottle labeled "Sibebe Premium Lager," directly matching the text's reference to this specific beer. 2. Beverage Context: - Text Reference: The text discusses EBL's production and marketing of beer, including Sibebe Premium Lager. - Visual Evidence: The presence of the beer bottle and glass in the image supports the text's focus on beer production and consumption. 3. Rural Setting: - Text Reference: While the text does not explicitly mention a rural setting, the company is based in Eswatini, which is known for its natural landscapes. - Visual Evidence: The rural background with trees and a horse suggests a setting that could be typical of Eswatini, aligning with the company's location. These elements in the image directly support the textual references to Sibebe Premium Lager and the context of beer production by EBL in Eswatini.}} \\
    \midrule
    \multicolumn{3}{p{0.98\textwidth}}{\textbf{\textcolor{blue}{[Japanese Response:]}} \begin{CJK*}{UTF8}{min}画像には、木製の手すりの上に「Sibebe Premium Lager」のボトルと、同じビールが入ったグラスが置かれています。背景には草地とまばらな木々、茅葺き屋根の建物、遠景の丘が見え、のどかな田園の雰囲気です。— 分析と根拠 — 1) \textit{銘柄の一致}: テキストにある "Sibebe Premium Lager" と、ボトルのラベル表記が一致します。2) \textit{製品文脈}: テキストは EBL のビール生産を述べており、画像の被写体はまさにその製品です。3) \textit{環境の整合}: 牧歌的な屋外風景はエスワティニの自然環境とも親和的で、テキストの地理的文脈と矛盾しません。これらにより、画像はテキスト記述を具体的に裏づけています。\end{CJK*}} \\
    \midrule
    \multicolumn{3}{p{0.98\textwidth}}{\textbf{\textcolor{blue}{[Swahili Response:]}} \texttt{ Picha inaonyesha chupa ya bia na glasi iliyojaa bia, ikiwa imewekwa juu ya uzio wa mbao. Katika mandhari ya nyuma kuna mazingira ya asili pamoja na farasi. Hapa kuna uchambuzi wa jinsi picha inavyohusiana na maandishi: 1. Bidhaa - Sibebe Premium Lager: - Ushahidi wa Maandishi: Maandishi yanataja kuwa bia ya Swazi inayozalishwa ni Sibebe Premium Lager, ambayo imepewa jina kutokana na mwamba wa Sibebe. - Ushahidi wa Picha: Picha inaonyesha chupa yenye lebo ya Sibebe Premium Lager, ikithibitisha moja kwa moja bidhaa hii kama ilivyoelezwa kwenye maandishi. 2. Kampuni - Eswatini Beverages Ltd (EBL): - Ushahidi wa Maandishi: EBL inatajwa kama kampuni inayozalisha na kuuza bia, ikiwa ni pamoja na Sibebe Premium Lager. - Ushahidi wa Picha: Uwepo wa chupa ya Sibebe Premium Lager katika picha unaonyesha moja ya bidhaa zinazozalishwa na EBL, na hivyo kuthibitisha maelezo ya maandishi kuhusu uzalishaji wa bia. 3. Mandhari ya Asili: - Ushahidi wa Maandishi: Ingawa maandishi hayataji moja kwa moja mandhari ya asili, Eswatini inajulikana kwa mandhari yake ya kuvutia. - Ushahidi wa Picha: Picha inaonyesha mandhari ya asili pamoja na farasi, hali inayoweza kuhusishwa na mazingira ya Eswatini. Kwa hivyo, picha hii inaonyesha moja kwa moja bidhaa inayozalishwa na Eswatini Beverages Ltd, na inalingana kikamilifu na maelezo yaliyotolewa katika maandishi kuhusu uzalishaji wa Sibebe Premium Lager. }} \\
  \midrule
    \multicolumn{3}{p{0.98\textwidth}}{\textbf{\textcolor{blue}{[Urdu Response:]}} \begin{UrduText}تصویر میں لکڑی کی ریلنگ پر "Sibebe" لیبل والی بیئر کی بوتل اور بھرے ہوئے گلاس کا قریبی منظر ہے۔ پس منظر میں سبزہ زار، چند درخت، دور پہاڑیاں اور گھاس پھوس کی چھت والا ایک چھوٹا سا پتھریلا مکان دکھائی دیتا ہے۔ — تجزیہ و جواز — (1) برانڈ کی مطابقت: بوتل پر درج "Sibebe Premium Lager" متن میں مذکور برانڈ سے مطابقت رکھتا ہے۔ (2) مصنوعاتی سیاق: متن EBL کی بیئر سازی کا ذکر کرتا ہے، اور تصویر اسی مصنوعات کو دکھاتی ہے۔ (3) ماحولیاتی ہم آہنگی: دیہی منظرنامہ اس خطے (اسواتینی) کے قدرتی پس منظر سے میل کھاتا ہے۔ یوں تصویر متن میں دی گئی معلومات کو بصری طور پر تقویت دیتی ہے۔\end{UrduText}} \\
    \bottomrule
    \end{tabular}
\end{table*}

\clearpage
\subsubsection{VLM Responses for Image-Text Matching  (Task 6) }
\begin{table*}[ht]
    \centering
    \caption{
    An example from our benchmark showing responses generated by \textbf{GPT-4o} for \emph{Image-Text Matching}. 
    In this setting, no rationales are generated by the model. 
    Given an image-text pair as input, we show the outputs in four languages: En, Jp, Sw, and Ur.
    \label{tab:Task6OneshotPromptingNoRationaleGPT4o}
    }
    \tiny
    \begin{tabular}{p{0.28\textwidth} p{0.45\textwidth} p{0.22\textwidth}}
    \toprule
    \multicolumn{3}{p{0.98\textwidth}}{%
    \textbf{\textcolor{olive}{Inputs to the VLM}}
    } \\
    \midrule
    \textbf{Image} & \textbf{Text} & \textbf{Prompt Question} \\
    \midrule
    \adjustbox{valign=t}{\includegraphics[width=\linewidth]{images/TaskExamples/img859.pdf}}
    &
    \texttt{Eswatini Beverages Ltd (EBL) was a subsidiary of SABMiller until 10 October 2016 when it was acquired by Anheuser-Busch InBev. It is a beverage and brewing company in Matsapha, Eswatini. The company was formed in 1995 by the merger of Eswatini Breweries, Ngwane Breweries, and Eswatini Bottlers. EBL produces and markets soft drinks, beer, and other alcoholic drinks. The Swazi beer produced is Sibebe Premium Lager named after the Sibebe rock, but other beers are brewed as well including Castle Lager. EBL dominates the beer market in Eswatini with a market share of 87\% in 2014 227 hl of lager beer were sold. Retooling in 2013 led to a temporary beer shortage in Eswatini.}
    & 
    \adjustbox{valign=t}{%
        \parbox{\linewidth}{%
            \texttt{\textcolor{red}{Question: Provide a detailed natural language description of what is happening in this image, as if describing it to someone who cannot see it.}}%
        }%
    }
    \\
    \midrule
     \multicolumn{3}{p{0.98\textwidth}}{%
    \textbf{\textcolor{olive}{Responses generated by the VLM}}
    } \\
    \midrule
    \multicolumn{3}{p{0.98\textwidth}}{%
    \textbf{\textcolor{blue}{[English Response:]}} \texttt{Image Analysis: The scene captures a tranquil outdoor setting, likely in a rural or safari-like environment. In the foreground, a wooden railing holds a bottle of Sibebe beer and a glass filled with the same beverage. The glass contains a golden liquid topped with a frothy head, suggesting it was freshly poured. Beyond the railing stretches a grassy area with sparse vegetation and a few scattered trees. In the distance, low hills or mountains sit under a clear sky, implying a warm, sunny day. To the right stands a structure with a thatched roof and stone walls—likely a traditional hut or lodge—adding to the rustic atmosphere. Near this building, a horse stands on the grass, relaxed and grazing or resting. Overall, the image conveys calm and leisure; the natural setting, the presence of the horse, and the refreshing drink together suggest a peaceful moment of relaxation in a scenic location.}
    } \\
    \midrule
    \multicolumn{3}{p{0.98\textwidth}}{%
    \textbf{\textcolor{blue}{[Japanese Response:]}} \begin{CJK*}{UTF8}{min}画像分析：自然の風景を背景にした屋外シーンです。手前の木製手すりには「Sibebe」と書かれたビールのボトルと、泡立つビールが入ったグラスが置かれています。背景には草地が広がり、まばらな木々と遠景の低い丘が見えます。右側には茅葺き屋根の建物があり、そのそばに馬が立っています。全体として、のどかでリラックスした雰囲気が漂う風景です。\end{CJK*}
    } \\
    \midrule
    \multicolumn{3}{p{0.98\textwidth}}{%
    \textbf{\textcolor{blue}{[Swahili Response:]}} \texttt{Maelezo ya Picha: Picha inaonyesha mandhari ya nje yenye utulivu. Mbele, juu ya uzio wa mbao, kuna chupa ya bia yenye lebo "Sibebe" na glasi iliyojaa bia ya rangi ya dhahabu yenye povu—ikiashiria kuwa imemiminwa hivi karibuni. Nyuma yake kuna nyasi, miti michache, na milima ya mbali chini ya anga lenye jua. Upande wa kulia kuna jengo lenye paa la nyasi (kijadi), na karibu nalo farasi amesimama akionekana kutulia. Kwa ujumla, mchanganyiko wa mandhari ya asili, uwepo wa farasi, na kinywaji baridi unaonesha hali ya mapumziko katika eneo la vijijini.}
    } \\
     \midrule
    \multicolumn{3}{p{0.98\textwidth}}{%
    \textbf{\textcolor{blue}{[Urdu Response:]}} \begin{UrduText}تصویر میں ایک پُرسکون بیرونی منظر دکھایا گیا ہے۔ سامنے لکڑی کی ریلنگ پر "Sibebe" لیبل والی بیئر کی بوتل اور سنہری رنگ کے مشروب سے بھرا گلاس رکھا ہے، جس پر جھاگ موجود ہے—گویا ابھی ابھی انڈیلا گیا ہو۔ پس منظر میں کھلا سبزہ زار، چند درخت اور دور ہلکی پہاڑیاں نظر آتی ہیں۔ دائیں جانب گھاس پھوس کی چھت والا پتھریلا ڈھانچہ ہے اور قریب ہی ایک گھوڑا سکون سے کھڑا ہے۔ مجموعی طور پر منظر راحت، قدرتی حسن اور ایک خوشگوار لمحے کی جھلک پیش کرتا ہے۔\end{UrduText}
    } \\
    \bottomrule
    \end{tabular}
\end{table*}

\clearpage
\begin{table*}[t]
    \centering
    \caption{
    An example from our benchmark showing responses generated by \textbf{GPT-4o} for \emph{Image-Text Matching}. 
    In this setting, \underline{the model generates rationales}, in addition to analyzing the image.
    Given an image-text pair as input, we show the outputs in four languages: En, Jp, Sw, and Ur.
    }
    \label{tab:Task6OneshotPromptingWithRationaleGPT4o}
    \tiny
    \begin{tabular}{p{0.28\textwidth} p{0.45\textwidth} p{0.22\textwidth}}
    \toprule
    \multicolumn{3}{p{0.98\textwidth}}{%
    \textbf{\textcolor{olive}{Inputs to the VLM}}
    } \\
    \midrule
    \textbf{Image} & \textbf{Text} & \textbf{Prompt Question} \\
    \midrule
    \adjustbox{valign=t}{\includegraphics[width=\linewidth]{images/TaskExamples/img859.pdf}} &
    \texttt{Eswatini Beverages Ltd (EBL) was a subsidiary of SABMiller until 10 October 2016 when it was acquired by Anheuser-Busch InBev. It is a beverage and brewing company in Matsapha, Eswatini. The company was formed in 1995 by the merger of Eswatini Breweries, Ngwane Breweries, and Eswatini Bottlers. EBL produces and markets soft drinks, beer, and other alcoholic drinks. The Swazi beer produced is Sibebe Premium Lager named after the Sibebe rock, but other beers are brewed as well including Castle Lager. EBL dominates the beer market in Eswatini with a market share of 87\% in 2014. Retooling in 2013 led to a temporary beer shortage in Eswatini.} &
    \adjustbox{valign=t}{\parbox{\linewidth}{\raggedright \texttt{\textcolor{red}{Question: Provide a detailed natural language description of what is happening in this image, as if describing it to someone who cannot see it.} 
    \textcolor{cyan}{Please provide both your analysis and rationale.} }}} \\
    \midrule
     \multicolumn{3}{p{0.98\textwidth}}{%
    \textbf{\textcolor{olive}{Responses generated by the VLM}}
    } \\
    \midrule
    \multicolumn{3}{p{0.98\textwidth}}{\textbf{\textcolor{blue}{[English Response:]}} \texttt{Detailed Narrative Description: 1) Main Focus: A bottle of Sibebe Premium Lager and a glass of beer sit on a wooden railing in the foreground. 2) Context: The scene is outdoors in a rustic, rural setting; a thatched-roof structure and open fields with scattered trees are visible. Distant low hills under a clear sky further suggest a warm day. 3) Actions: No human actions are visible; a horse stands calmly near the structure. 4) Appearance: The bottle (green with a blue-and-silver label) and the golden beer with a foamy head indicate a freshly poured drink. The railing and building look well maintained. 5) Time/Atmosphere: The composition suggests a leisurely moment—perhaps afternoon—enjoyed in nature. \emph{Rationale:} The beverage is central in the frame (thus focal), the rural cues (thatch, open field, horse) ground the setting, and the static arrangement emphasizes calm and relaxation.}} \\
    \midrule
    \multicolumn{3}{p{0.98\textwidth}}{\textbf{\textcolor{blue}{[Japanese Response:]}} \begin{CJK*}{UTF8}{min}詳細記述：1) \textbf{主題}—手前の木製手すりの上に「Sibebe」ラベルのボトルと泡立つビールグラス。2) \textbf{文脈}—背景には茅葺き屋根の建物、草地、点在する木々、遠景の丘が見え、牧歌的な屋外環境を示す。3) \textbf{動き}—人の動作はなく、建物のそばで馬が穏やかに佇む。4) \textbf{外観}—緑色のボトル、黄金色のビール、しっかりした泡から注ぎたてと分かる。5) \textbf{時間・雰囲気}—明るい自然光と静的構図が、ゆったりとした余暇の瞬間を想起させる。— \emph{根拠}：主題が前景で強調されていること、農村的モチーフ（茅葺き・草地・馬）が環境を裏づけること、静止した配置が安らぎを強めていることに基づく解釈。\end{CJK*}} \\
    \midrule
    \multicolumn{3}{p{0.98\textwidth}}{\textbf{\textcolor{blue}{[Swahili Response:]}} \texttt{Maelezo ya Kina: 1) \textit{Lengo Kuu}: Chupa ya “Sibebe Premium Lager” na glasi yenye bia juu ya uzio wa mbao. 2) \textit{Muktadha}: Mandhari ya kijijini—nyasi, miti michache, milima ya mbali, na jengo lenye paa la nyasi. 3) \textit{Vitendo}: Hakuna mwendo wa moja kwa moja; farasi amesimama tulivu karibu na jengo. 4) \textit{Muonekano}: Bia ya rangi ya dhahabu yenye povu (ishara ya kumiminwa karibuni); mazingira yamehifadhiwa vizuri. 5) \textit{Wakati/Hisia}: Nuru ya mchana na utulivu wa eneo vinaashiria muda wa mapumziko. \emph{Hoja}: Uwepo wa kinywaji kama kiini cha muundo, dalili za mazingira ya kijijini, na mpangilio tuli—vyote vinaunda simulizi la mapumziko na utulivu.}} \\
    \midrule
    \multicolumn{3}{p{0.98\textwidth}}{\textbf{\textcolor{blue}{[Urdu Response:]}} \begin{UrduText}تفصیلی بیان: (1) \textit{مرکزی توجہ}—سامنے لکڑی کی ریلنگ پر “Sibebe Premium Lager” کی بوتل اور جھاگ والے بیئر سے بھرا گلاس رکھا ہے۔ (2) \textit{سیاق}—پس منظر میں گھاس پھوس کی چھت والا ڈھانچہ، کھلا سبزہ زار، چند درخت اور دور پہاڑیاں؛ فضا دیہی اور پُرسکون ہے۔ (3) \textit{حرکت}—انسانی سرگرمی نظر نہیں آتی؛ عمارت کے قریب گھوڑا اطمینان سے کھڑا ہے۔ (4) \textit{ظاہری تفصیل}—سبز بوتل، سنہری بیئر اور مضبوط جھاگ تازہ انڈیلا جانے کی علامت ہیں؛ ریلنگ اور عمارت کی حالت اچھی ہے۔ (5) \textit{وقت و ماحول}—روشن دن اور ساکت ترتیب ایک فراغت بھرا لمحہ ظاہر کرتی ہے۔ \emph{جواز}: پیش منظر میں مشروب کی مرکزیت، دیہی قرائن (چھپر، سبزہ، گھوڑا) اور ساکت ترکیب—یہ سب مل کر سکون و راحت کی کیفیت کو ابھارتے ہیں۔\end{UrduText}} \\
    \bottomrule
    \end{tabular}
\end{table*}

\clearpage
\subsubsection{VLM Responses for Unrelatedness  (Task 7) }
\begin{table*}[ht]
    \centering
    \caption{
    An example from our benchmark showing responses generated by \textbf{GPT-4o} for \emph{Unrelatedness}. 
    In this setting, no rationales are generated by the model. 
    Given an image-text pair as input, we show the outputs in four languages: En, Jp, Sw, and Ur.
    \label{tab:Task7OneshotPromptingNoRationaleGPT4o}
    }
    \tiny
    \begin{tabular}{p{0.28\textwidth} p{0.45\textwidth} p{0.22\textwidth}}
    \toprule
    \multicolumn{3}{p{0.98\textwidth}}{%
    \textbf{\textcolor{olive}{Inputs to the VLM}}
    } \\
    \midrule
    \textbf{Image} & \textbf{Text} & \textbf{Prompt Question} \\
    \midrule
    \adjustbox{valign=t}{\includegraphics[width=\linewidth]{images/TaskExamples/img859.pdf}}
    &
    \texttt{Eswatini Beverages Ltd (EBL) was a subsidiary of SABMiller until 10 October 2016 when it was acquired by Anheuser-Busch InBev. It is a beverage and brewing company in Matsapha, Eswatini. The company was formed in 1995 by the merger of Eswatini Breweries, Ngwane Breweries, and Eswatini Bottlers. EBL produces and markets soft drinks, beer, and other alcoholic drinks. The Swazi beer produced is Sibebe Premium Lager named after the Sibebe rock, but other beers are brewed as well including Castle Lager. EBL dominates the beer market in Eswatini with a market share of 87\% in 2014 227 hl of lager beer were sold. Retooling in 2013 led to a temporary beer shortage in Eswatini.}
    & 
    \adjustbox{valign=t}{%
        \parbox{\linewidth}{%
            \texttt{\textcolor{red}{Question: Identify which parts of the text are not represented in the image and explain why they are unrelated.}}%
        }%
    }
    \\
    \midrule
     \multicolumn{3}{p{0.98\textwidth}}{%
    \textbf{\textcolor{olive}{Responses generated by the VLM}}
    } \\
    \midrule
    \multicolumn{3}{p{0.98\textwidth}}{%
    \textbf{\textcolor{blue}{[English Response:]}} \texttt{The image presents a bottle and glass of Sibebe Premium Lager in a rural setting. Elements \emph{not} represented: (1) Company history/ownership (SABMiller $\to$ AB InBev): no corporate logos, facilities, or historical cues. (2) Formation by mergers (1995): no visual hint of the breweries or merger. (3) Other products (soft drinks, Castle Lager, etc.): only Sibebe appears. (4) Market figures (87\% share; 2014 sales): no charts/data. (5) 2013 retooling/shortage: no factory/supply imagery. (6) Specific location (Matsapha): no geographic markers. Hence, the photo showcases a single product, not the broader corporate, historical, or quantitative context described in the text.}
    } \\
    \midrule
    \multicolumn{3}{p{0.98\textwidth}}{%
    \textbf{\textcolor{blue}{[Japanese Response:]}} \begin{CJK*}{UTF8}{min}画像にはSibebe Premium Lagerのボトルとグラスが写っていますが、以下のテキスト要素は画像で表現されていません。1) 企業の歴史・所有権の変遷（SABMillerからAB InBevへの移行）：企業ロゴや沿革を示す手掛かりはなし。2) 1995年の合併による設立：関連施設や合併に関する視覚情報はなし。3) 他製品（ソフトドリンク、Castle Lager等）：Sibebe以外は不在。4) 市場シェア・販売量などの数値（87\%、2014年販売量）：データ可視化はなし。5) 2013年のリツーリングと一時的なビール不足：生産設備や供給状況を示す要素はなし。6) 立地（マツァパ）：地理的手掛かりはなし。— 画像は単一商品の提示に特化しており、テキストが述べる企業背景や統計情報までは視覚的にカバーしていないためです。\end{CJK*}
    } \\
    \midrule
    \multicolumn{3}{p{0.98\textwidth}}{%
    \textbf{\textcolor{blue}{[Swahili Response:]}} \texttt{Picha inaonyesha tu bia ya Sibebe Premium Lager. Vipengele ambavyo haviwakilishwi: (1) Historia na umiliki wa EBL (kubadilishwa kutoka SABMiller hadi AB InBev); (2) Muungano wa makampuni mwaka 1995; (3) Bidhaa nyingine (vinywaji laini, Castle Lager); (4) Takwimu za soko na mauzo (87\%, 2014); (5) Retooling ya 2013 na upungufu wa bia; (6) Maelezo ya kijiografia (Matsapha). Picha imelenga bidhaa moja pekee, si muktadha mpana wa kibiashara au kihistoria.}
    } \\
     \midrule
    \multicolumn{3}{p{0.98\textwidth}}{%
    \textbf{\textcolor{blue}{[Urdu Response:]}} \begin{UrduText}تصویر میں صرف “Sibebe Premium Lager” کی بوتل اور گلاس نظر آتے ہیں، جبکہ متن کے یہ حصے تصویر میں موجود نہیں: (1) کمپنی کی تاریخ اور ملکیت کی تبدیلی (SABMiller سے AB InBev)، (2) 1995 کی انضمامی تشکیل، (3) دیگر مصنوعات جیسے سافٹ ڈرنکس اور Castle Lager، (4) مارکیٹ شیئر/فروخت کے اعدادوشمار (87\%، 2014)، (5) 2013 کی ری ٹولنگ اور عارضی قلت، (6) مخصوص جغرافیہ (ماتسابا، ایسواتینی)۔ وجہ یہ ہے کہ تصویر محض ایک مصنوعات کی بصری نمائش ہے، جب کہ متن وسیع کارپوریٹ، تاریخی اور مقداری سیاق بیان کرتا ہے۔\end{UrduText}
    } \\
    \bottomrule
    \end{tabular}
\end{table*}

\clearpage
\begin{table*}[t]
    \centering
    \caption{
    An example from our benchmark showing responses generated by \textbf{GPT-4o} for \emph{Unrelatedness}. 
    In this setting, \underline{the model generates rationales}, in addition to analyzing the image.
    Given an image-text pair as input, we show the outputs in four languages: En, Jp, Sw, and Ur.
    }
    \label{tab:Task7OneshotPromptingWithRationaleGPT4o}
    \tiny
    \begin{tabular}{p{0.28\textwidth} p{0.45\textwidth} p{0.22\textwidth}}
    \toprule
    \multicolumn{3}{p{0.98\textwidth}}{%
    \textbf{\textcolor{olive}{Inputs to the VLM}}
    } \\
    \midrule
    \textbf{Image} & \textbf{Text} & \textbf{Prompt Question} \\
    \midrule
    \adjustbox{valign=t}{\includegraphics[width=\linewidth]{images/TaskExamples/img859.pdf}} &
    \texttt{Eswatini Beverages Ltd (EBL) was a subsidiary of SABMiller until 10 October 2016 when it was acquired by Anheuser-Busch InBev. It is a beverage and brewing company in Matsapha, Eswatini. The company was formed in 1995 by the merger of Eswatini Breweries, Ngwane Breweries, and Eswatini Bottlers. EBL produces and markets soft drinks, beer, and other alcoholic drinks. The Swazi beer produced is Sibebe Premium Lager named after the Sibebe rock, but other beers are brewed as well including Castle Lager. EBL dominates the beer market in Eswatini with a market share of 87\% in 2014. Retooling in 2013 led to a temporary beer shortage in Eswatini.} &
    \adjustbox{valign=t}{\parbox{\linewidth}{\raggedright \texttt{\textcolor{red}{Question: Identify which parts of the text are not represented in the image and explain why they are unrelated.} 
    \textcolor{cyan}{Please provide both your analysis and rationale.}  }}} \\
    \midrule
     \multicolumn{3}{p{0.98\textwidth}}{%
    \textbf{\textcolor{olive}{Responses generated by the VLM}}
    } \\
    \midrule
    \multicolumn{3}{p{0.98\textwidth}}{\textbf{\textcolor{blue}{[English Response:]}} \texttt{The photo highlights only the product (Sibebe Premium Lager). Unrepresented text parts: corporate history/ownership (SABMiller $\to$ AB InBev), formation by mergers, other product lines (soft drinks, Castle Lager), market-share and 2014 sales figures, 2013 retooling/shortage, and the specific location (Matsapha). \emph{Rationale:} None of these concepts have obvious visual correlates in a single product shot taken outdoors; the image is illustrative, not documentary.}} \\
    \midrule
     \multicolumn{3}{p{0.98\textwidth}}{%
      \textbf{\textcolor{blue}{[Japanese Response:]}}
      \begin{CJK*}{UTF8}{min}
    画像は屋外での Sibebe Premium Lager の静物を示すにとどまります。画像に反映されないテキスト要素： (1) EBL の歴史・所有権（SABMiller から AB InBev）(2) 1995 年の合併による設立 (3) 他製品（ソフトドリンク、Castle Lager）(4) 市場シェアや 2014 年の販売量などの数値 (5) 2013 年のリツーリングと一時的な品薄 (6) 立地（マツァパ）。\emph{根拠}: これらはいずれも企業情報・統計・生産工程・地理に関する抽象的内容であり、単一商品の屋外写真には視覚的手掛かりがないためです。
      \end{CJK*}
    } \\
    \midrule
    \multicolumn{3}{p{0.98\textwidth}}{\textbf{\textcolor{blue}{[Swahili Response:]}} \texttt{Picha inalenga bidhaa moja tu. Sehemu zisizoakisiwa kwenye picha: historia na umiliki wa EBL, kuundwa kupitia muungano (1995), bidhaa nyingine (soft drinks, Castle Lager), takwimu za soko/mauzo, retooling ya 2013 na upungufu wa bia, na mahali mahsusi (Matsapha). \emph{Hoja}: Haya ni maelezo ya kimuundo na kitakwimu ambayo hayaonekani katika picha ya bidhaa pekee.}} \\
    \midrule
    \multicolumn{3}{p{0.98\textwidth}}{\textbf{\textcolor{blue}{[Urdu Response:]}} \begin{UrduText}تصویر میں صرف “Sibebe Premium Lager” بطور مصنوعات دکھائی گئی ہے۔ غیر متعلق/غیر نمایاں حصے: کمپنی کی تاریخ و ملکیت کی تبدیلی، 1995 کا انضمام، دیگر مصنوعات (سافٹ ڈرنکس، Castle Lager)، مارکیٹ شیئر و 2014 کی فروخت، 2013 کی ری ٹولنگ و عارضی قلت، اور مخصوص مقام (ماتسابا)۔ \emph{جواز:} یہ سب معلوماتی/کارپوریٹ یا مقداری نوعیت کی ہیں جن کی بصری علامت ایک سادہ پروڈکٹ شاٹ میں دستیاب نہیں ہوتی، اس لیے تصویر ان پہلوؤں کی نمائندگی نہیں کرتی۔\end{UrduText} 
    } \\
    \bottomrule
    \end{tabular}
\end{table*}

\clearpage
\subsubsection{VLM Responses for Visual Question Answering  (Task 8) }
\begin{table*}[ht]
    \centering
    \caption{
    An example from our benchmark showing responses generated by \textbf{GPT-4o} for \emph{Visual Question Answering}. 
    In this setting, no rationales are generated by the model. 
    Given an image-text pair as input, we show the outputs in four languages: En, Jp, Sw, and Ur.
    \label{tab:Task8OneshotPromptingNoRationaleGPT4o}
    }
    \tiny
    \begin{tabular}{p{0.28\textwidth} p{0.45\textwidth} p{0.22\textwidth}}
    \toprule
    \multicolumn{3}{p{0.98\textwidth}}{%
    \textbf{\textcolor{olive}{Inputs to the VLM}}
    } \\
    \midrule
    \textbf{Image} & \textbf{Text} & \textbf{Prompt Question} \\
    \midrule
    \adjustbox{valign=t}{\includegraphics[width=\linewidth]{images/TaskExamples/img859.pdf}}
    &
    \texttt{Eswatini Beverages Ltd (EBL) was a subsidiary of SABMiller until 10 October 2016 when it was acquired by Anheuser-Busch InBev. It is a beverage and brewing company in Matsapha, Eswatini. The company was formed in 1995 by the merger of Eswatini Breweries, Ngwane Breweries, and Eswatini Bottlers. EBL produces and markets soft drinks, beer, and other alcoholic drinks. The Swazi beer produced is Sibebe Premium Lager named after the Sibebe rock, but other beers are brewed as well including Castle Lager. EBL dominates the beer market in Eswatini with a market share of 87\% in 2014 227 hl of lager beer were sold. Retooling in 2013 led to a temporary beer shortage in Eswatini.}
    & 
    \adjustbox{valign=t}{%
        \parbox{\linewidth}{%
            \texttt{\textcolor{red}{Question: Identify which parts of the text are not represented in the image and explain why they are unrelated.}}%
        }%
    }
    \\
    \midrule
     \multicolumn{3}{p{0.98\textwidth}}{%
    \textbf{\textcolor{olive}{Responses generated by the VLM}}
    } \\
    \midrule
    \multicolumn{3}{p{0.98\textwidth}}{%
    \textbf{\textcolor{blue}{[English Response:]}} \texttt{The text mentions the following locations: 1) Eswatini — Country where Eswatini Beverages Ltd (EBL) operates and its primary market. 2) Matsapha — Specific location of EBL, a major industrial hub. 3) Sibebe Rock — Namesake of Sibebe Premium Lager and a prominent granite formation. In the image: a bottle and glass of Sibebe Premium Lager are clearly visible, linking to the text’s reference to the product and its name. However, the image does not depict identifiable landmarks for Eswatini or Matsapha. Overall, the text provides broader location context, while the image focuses on the product itself and its brand connection to Sibebe.}
    } \\
    \midrule
    \multicolumn{3}{p{0.98\textwidth}}{%
    \textbf{\textcolor{blue}{[Japanese Response:]}} \begin{CJK*}{UTF8}{min}以下はテキストと画像に基づく場所情報の整理です。1) エスワティニ：EBLが拠点とする国。ビール市場で大きなシェアを持つ。2) マツァパ（Matsapha）：EBLの所在地で、国内の主要な工業拠点。3) シベベ（Sibebe Rock）：\emph{Sibebe Premium Lager} の名称由来となった巨大な花崗岩の一枚岩。— 画像には「Sibebe Premium Lager」の瓶とグラスが写っており、製品名との結びつきは明確。一方で、エスワティニやマツァパを特定できる地理的ランドマークは写っていません。テキストは地理・背景情報を補足し、画像は製品自体を強調しています。\end{CJK*}
    } \\
    \midrule
    \multicolumn{3}{p{0.98\textwidth}}{%
    \textbf{\textcolor{blue}{[Swahili Response:]}} \texttt{Maeneo yaliyotajwa: (1) Eswatini — nchi ambako EBL inaendesha shughuli; (2) Matsapha — eneo maalum la kiwanda la EBL; (3) Sibebe Rock — chanzo cha jina la \emph{Sibebe Premium Lager}. Pichani: chupa na glasi ya Sibebe Premium Lager zinaonekana wazi, zikithibitisha uhusiano na bidhaa. Hata hivyo, picha haioneshi alama bainishi za Eswatini au Matsapha; hivyo maandishi yanatoa muktadha wa kijiografia, na picha inalenga bidhaa.}
    } \\
     \midrule
    \multicolumn{3}{p{0.98\textwidth}}{%
    \textbf{\textcolor{blue}{[Urdu Response:]}} \begin{UrduText}متن میں درج ذیل مقامات کا ذکر ہے: (1) ایسواتینی — وہ ملک جہاں EBL کام کرتی ہے اور جس کا بازار مرکزی اہمیت رکھتا ہے۔ (2) ماتسابا — EBL کا مخصوص مقام اور ایک صنعتی مرکز۔ (3) سِبیبی راک — \emph{Sibebe Premium Lager} کے نام کی نسبت اس مشہور سنگی چٹان سے ہے۔ تصویر میں: \emph{Sibebe Premium Lager} کی بوتل اور گلاس واضح ہیں، جو نامی نسبت کو مضبوط کرتے ہیں؛ تاہم تصویر میں ایسواتینی یا ماتسابا کے پہچانے جانے والے نشانات نہیں۔ یوں متن جغرافیائی پس منظر فراہم کرتا ہے جبکہ تصویر محض مصنوعات کو نمایاں کرتی ہے۔\end{UrduText}
    } \\
    \bottomrule
    \end{tabular}
\end{table*}

\clearpage
\begin{table*}[t]
    \centering
    \caption{
    An example from our benchmark showing responses generated by \textbf{GPT-4o} for \emph{Visual Question Answering}. 
    In this setting, \underline{the model generates rationales}, in addition to analyzing the image.
    Given an image-text pair as input, we show the outputs in four languages: En, Jp, Sw, and Ur.
    }
    \label{tab:Task8OneshotPromptingWithRationaleGPT4o}
    \tiny
    \begin{tabular}{p{0.28\textwidth} p{0.45\textwidth} p{0.22\textwidth}}
    \toprule
    \multicolumn{3}{p{0.98\textwidth}}{%
    \textbf{\textcolor{olive}{Inputs to the VLM}}
    } \\
    \midrule
    \textbf{Image} & \textbf{Text} & \textbf{Prompt Question} \\
    \midrule
    \adjustbox{valign=t}{\includegraphics[width=\linewidth]{images/TaskExamples/img859.pdf}} &
    \texttt{Eswatini Beverages Ltd (EBL) was a subsidiary of SABMiller until 10 October 2016 when it was acquired by Anheuser-Busch InBev. It is a beverage and brewing company in Matsapha, Eswatini. The company was formed in 1995 by the merger of Eswatini Breweries, Ngwane Breweries, and Eswatini Bottlers. EBL produces and markets soft drinks, beer, and other alcoholic drinks. The Swazi beer produced is Sibebe Premium Lager named after the Sibebe rock, but other beers are brewed as well including Castle Lager. EBL dominates the beer market in Eswatini with a market share of 87\% in 2014. Retooling in 2013 led to a temporary beer shortage in Eswatini.} &
    \adjustbox{valign=t}{\parbox{\linewidth}{\raggedright \texttt{\textcolor{red}{Question: Identify which parts of the text are not represented in the image and explain why they are unrelated.} 
    \textcolor{cyan}{Please provide both your analysis and rationale.}  }}} \\
    \midrule
     \multicolumn{3}{p{0.98\textwidth}}{%
    \textbf{\textcolor{olive}{Responses generated by the VLM}}
    } \\
    \midrule
    \multicolumn{3}{p{0.98\textwidth}}{\textbf{\textcolor{blue}{[English Response:]}} \texttt{Places referenced or implied: (1) Matsapha — EBL’s location; an industrial town in Eswatini. (2) Eswatini — country of operation and main market; a small Southern African nation with notable natural landscapes. (3) Sibebe Rock — namesake of \emph{Sibebe Premium Lager}, a famous monolith. In the image, a bottle and glass of Sibebe are shown, directly tying to the product. While the setting suggests rural Eswatini, no specific landmarks (e.g., Matsapha or Sibebe Rock) are visible. The natural backdrop complements the text by reinforcing the product’s local origin and cultural resonance.}} \\
    \midrule
     \multicolumn{3}{p{0.98\textwidth}}{\textbf{\textcolor{blue}{[Japanese Response:]}} \begin{CJK*}{UTF8}{min}分析結果：1) エスワティニ — EBLが操業する国で、国内市場の中心。2) マツァパ — EBLの所在都市で主要な工業拠点。3) シベベ・ロック — \emph{Sibebe Premium Lager} の名称由来となる地質学的名所。画像にはSibebeの瓶とグラスが明確に写っており製品との対応は取れますが、マツァパやシベベ・ロックそのものを特定できるランドマークは見当たりません。自然の景観は、製品の土地性や文化的背景を視覚的に補強しています。\end{CJK*}}\\
    \midrule
    \multicolumn{3}{p{0.98\textwidth}}{\textbf{\textcolor{blue}{[Swahili Response:]}} \texttt{Maeneo yaliyotajwa/kuashiriwa: Matsapha (makao ya EBL), Eswatini (nchi ya uendeshaji), na Sibebe Rock (chanzo cha jina la bia). Picha: inaonyesha chupa na glasi ya \emph{Sibebe Premium Lager}, hivyo kuunganisha moja kwa moja na bidhaa. Ingawa mandhari ya asili inaashiria Eswatini, hakuna alama mahsusi za Matsapha au Sibebe Rock. Hivyo, maandishi yanapanua muktadha; picha inasisitiza utambulisho wa bidhaa.}} \\
    \midrule
    \multicolumn{3}{p{0.98\textwidth}}{\textbf{\textcolor{blue}{[Urdu Response:]}} \begin{UrduText}تجزیہ: (1) ماتسابا — EBL کا مقام اور صنعتی مرکز۔ (2) ایسواتینی — کمپنی کی سرگرمیوں کا ملک اور مرکزی مارکیٹ۔ (3) سِبیبی راک — \emph{Sibebe Premium Lager} کے نام کی نسبت اسی چٹان سے ہے۔ تصویر میں سِبیبی کی بوتل اور گلاس واضح ہیں، اس طرح متن میں مذکور مصنوعات سے براہِ راست ربط قائم ہوتا ہے۔ اگرچہ منظر نامہ دیہی ایسواتینی کی طرف اشارہ کرتا ہے، مگر ماتسابا یا سِبیبی راک جیسی مخصوص نشانیاں دکھائی نہیں دیتیں۔ یوں متن جغرافیائی و ثقافتی سیاق مہیا کرتا ہے اور تصویر خود مصنوعات کو نمایاں کرتی ہے۔\end{UrduText}}\\
    \bottomrule
    \end{tabular}
\end{table*}

\clearpage
\subsection{Error Analysis for the VLM Outputs} 
Overall, VLMs generated logical and well-reasoned outputs under all eight VL tasks, utilizing the textual and visual information provided at the input. 
Moreover, the proprietary VLMs did so well, and the average accuracy is more than 80\%. 
Lastly, the proprietary VLMs exhibited strong linguistic understanding of the four languages (English, Japanese, Swahili, Urdu) involved in this study. 
Conversely, we also noticed several errors in the VLM responses, and we discuss those errors below.
\subsubsection{Errors made by Proprietary VLMs} 
Whereas these VLMs accurately interpreted both \textit{texts} and \textit{images}, there were occasional errors, such as 
(i) Incorrect recognition of objects inside the image, under the object recognition task, 
(ii) Under the VQA task, VLMs occasionally hallucinated about the names of the places shown in the image, and
(iii) One of the VLMs, GPT-4o-mini, was inconsistent with generating outputs strictly in the target language. 
Specifically, GPT-4o-mini generated outputs mainly in English even when the target language was Urdu. On such occasions, we repeated the experiment, monitoring for language compliance from the VLM.

\subsubsection{Errors made by Open-source VLMs} 
Unlike the proprietary counterparts, this category of VLMs exhibited many weaknesses, including:
(i) Responses only repeated the instruction instead of providing answers to the question related to a VL task,
(ii) Response is unintelligible and irrelevant to the VL task, and
(iii) Hallucinated response, not relevant to either the textual or image information,
(iv) Complete failure to understand the texts in a language other than English. For example, we observed \texttt{Qwen2VL} failed to generate intelligible responses when Urdu texts are provided as input to the model. 

\clearpage
\subsection{Human Evaluation and the Prompts used in this study}
\label{HumanEvalCriteria}
\subsubsection{Human Evaluation Alignment}
\label{subsec:AlignmentHumanEval}
We assessed the usefulness of \textit{Gemini 1.5 Pro} as the main evaluator in our work. 
First, we selected 100 responses of GPT-4o-mini, from the \textit{relation understanding} task, for appraisal by human evaluators. 
Then, we recruited two human annotators to review the VLM responses and assign their scores for each response. 
The responses are graded on a scale from one to one hundred, the same as with the \textit{Gemini 1.5 Pro judge}. 
We compare the agreement between their evaluation and the \textit{Gemini 1.5 Pro judge}. 
By calculating the intraclass correlation coefficient, the agreement between human annotators and the LLM judge is 82.3\%. 
The high level of agreement suggests \textit{Gemini 1.5 Pro} performs comparably to human annotators in assessing the perception tasks.

We leave the full evaluation across all tasks for future work due to the high costs of recruiting human evaluators and payment of API access to VLMs.  

\clearpage
\subsection{Error Analysis for the VLM Outputs} 
Overall, VLMs generated logical and well-reasoned outputs under all eight VL tasks, utilizing the textual and visual information provided at the input. 
Moreover, the proprietary VLMs did so well, and the average accuracy is more than 80\%. 
Lastly, the proprietary VLMs exhibited strong linguistic understanding of the four languages (English, Japanese, Swahili, Urdu) involved in this study. 
Conversely, we also noticed several errors in the VLM responses, and we discuss those errors below.
\subsubsection{Errors made by Proprietary VLMs} 
Whereas these VLMs accurately interpreted both \textit{texts} and \textit{images}, there were occasional errors, such as 
(i) Incorrect recognition of objects inside the image, under the object recognition task, 
(ii) Under the VQA task, VLMs occasionally hallucinated about the names of the places shown in the image, and
(iii) One of the VLMs, GPT-4o-mini, was inconsistent with generating outputs strictly in the target language. 
Specifically, GPT-4o-mini generated outputs mainly in English even when the target language was Urdu. On such occasions, we repeated the experiment, monitoring for language compliance from the VLM.

\subsubsection{Errors made by Open-source VLMs} 
Unlike the proprietary counterparts, this category of VLMs exhibited many weaknesses, including:
(i) Responses only repeated the instruction instead of providing answers to the question related to a VL task,
(ii) Response is unintelligible and irrelevant to the VL task, and
(iii) Hallucinated response, not relevant to either the textual or image information,
(iv) Complete failure to understand the texts in a language other than English. For example, we observed \texttt{Qwen2VL} failed to generate intelligible responses when Urdu texts are provided as input to the model. 

\clearpage
\subsection{Human Evaluation and the Prompts used in this study}
\label{HumanEvalCriteria}
\subsubsection{Human Evaluation Alignment}
\label{subsec:AlignmentHumanEval}
We assessed the usefulness of \textit{Gemini 1.5 Pro} as the main evaluator in our work. 
First, we selected 100 responses of GPT-4o-mini, from the \textit{relation understanding} task, for appraisal by human evaluators. 
Then, we recruited two human annotators to review the VLM responses and assign their scores for each response. 
The responses are graded on a scale from one to one hundred, the same as with the \textit{Gemini 1.5 Pro judge}. 
We compare the agreement between their evaluation and the \textit{Gemini 1.5 Pro judge}. 
By calculating the intraclass correlation coefficient, the agreement between human annotators and the LLM judge is 82.3\%. 
The high level of agreement suggests \textit{Gemini 1.5 Pro} performs comparably to human annotators in assessing the perception tasks.

We leave the full evaluation across all tasks for future work due to the high costs of recruiting human evaluators and payment of API access to VLMs.

\subsection{Prompts used for Evaluation in this study}
In what follows, we show several prompts we used throughout the study. 
\begin{tcolorbox}[
  enhanced,
  breakable,
  bicolor, %
  colback=promptboxContentBg,    %
  colupper=promptboxTitleBg, %
  colframe=promptboxFrame,       %
  coltitle=white,                %
  fonttitle=\bfseries,           %
  title={The Prompt for Gemini 1.5 Pro to Rate the Performance of VL Tasks}, %
  boxrule=1pt,                   %
  width=\linewidth,              %
  arc=2mm,                       %
  halign title=center, 
]
\small
Below is a vision and language task with responses from both the vision language model and the ground-truth provided by human annotation. Based on the 3 criteria below, rate the model's performance on a scale of 1-100. Only provide the scores without explanations.

\medskip %

\textbf{Accuracy:} The response provided by the LLM is accurate and has no factual errors. Conclusions not made arbitrarily.

\medskip %

\textbf{Helpfulness:} The model's response provides clear information the task.

\medskip %

\textbf{Linguistic Quality:} The response  is logical. The model correctly understands the task, and the expressions smooth and natural. %

\medskip %

Please ensure that you do not let the length of the text influence your judgment, do not have a preference for any AI assistant names that might appear in the dialogue, do not let irrelevant linguistic habits in the conversation influence your judgment, and strive to remain objective. Your scoring should be strict enough and do not give a perfect score easily.
\end{tcolorbox}

\begin{tcolorbox}[
  enhanced,
  breakable,
  bicolor, %
  colback=promptboxContentBg,    %
  colupper=promptboxTitleBg, %
  colframe=promptboxFrame,       %
  coltitle=white,                %
  fonttitle=\bfseries,           %
  title={Human Evaluation Instructions}, %
  boxrule=1pt,                   %
  width=\linewidth,              %
  arc=2mm,                       %
  halign title=center, 
]
\small
Based on the 3 criteria below, rate the model's performance on a scale of 1-100. Only provide the scores without explanations.

\medskip %

\textbf{Accuracy:} The response provided by the LLM is accurate and has no factual errors. Conclusions not made arbitrarily.

\medskip %

\textbf{Helpfulness:} The model's response provides clear information the task.

\medskip %

\textbf{Linguistic Quality:} The response  is logical. The model correctly understands the task, and the expressions smooth and natural. %

\medskip %

Please ensure that you do not let the length of the text influence your judgment, do not have a preference for any AI assistant names that might appear in the dialogue, do not let irrelevant linguistic habits in the conversation influence your judgment, and strive to remain objective. Your scoring should be strict enough and do not give a perfect score easily.
\end{tcolorbox}
\clearpage
\begin{table*}[t]
\footnotesize
\centering
\caption{
Structure of the Evaluation Prompt for the LLM-as-a-Judge, that is \texttt{Gemini 1.5 Pro}, used in scoring the outputs generated by VLMs in this study. 
Each component is designed to provide comprehensive context and ensure consistent, high-quality evaluation.
The full prompt is shown in Table \ref{tab:llm_judge_prompt}.
}
\label{tab:evaluation_prompt_structure}
\begin{tabular}{p{0.95\textwidth}}
\toprule
\textbf{Components of the LLM-as-a-Judge Prompt} \\
\midrule
\vspace{0.5ex}

\textbf{1. Role and Persona} \\
The prompt begins by assigning a specific role to the LLM-judge: \textit{``You are a meticulous evaluator of Vision-Language AI responses.''} 
This sets the context and encourages a detailed, analytical mode of operation.

\vspace{1ex}
\textbf{2. Core Task and Context} \\
The judge is explicitly told what to evaluate, including the specific \textbf{task description}, the \textbf{language of the response}, and the \textbf{experimental setting} under which the response was generated (e.g., \textit{zero-shot with rationales}).

\vspace{1ex}
\textbf{3. Ground Truth Information} \\
The prompt includes the ground truth materials (\textbf{image text pairs}) that the original VLM used to provide full context for the evaluation. 
This consists of:
\begin{itemize}
    \item The original \textbf{text} content associated with the image (\textit{e.g., ``{text\_content}''}).
    \item A reference to the \textbf{image} file itself (\textit{e.g., ``Image file: img859.jpeg''}).
\end{itemize}

\vspace{1ex}
\textbf{4. Candidate Response} \\
The specific \textbf{VLM-generated output} that needs to be scored is clearly demarcated and presented to the judge, labeled with its unique \textbf{Text ID}.

\vspace{1ex}
\textbf{5. Scoring Rubric and Scale} \\
The judge is instructed to rate the response on a scale from 0 (lowest quality) to 100 (highest quality) based on three explicit criteria:
\begin{itemize}
    \item \textbf{Accuracy:} The correctness and factual integrity of the response.
    \item \textbf{Helpfulness:} The clarity, practicality, and relevance of the assistance provided for the given task.
    \item \textbf{Linguistic Quality:} The logical coherence, naturalness, and correct understanding of the task.
\end{itemize}

\vspace{1ex}
\textbf{6. Objectivity and Best Practices} \\
A set of important guidelines is provided to minimize bias and ensure objective scoring. 
The judge is explicitly instructed to disregard response length, not to prefer specific AI assistant names, and to remain strict and objective in its scoring.

\vspace{1ex}
\textbf{7. Strict Output Formatting} \\
Finally, the prompt concludes with strict instructions for the output format to ensure programmatic parsing. 
The judge must return \textbf{only} a valid JSON object containing the numerical score, with no additional explanation:
\texttt{\{``score'': <number>\}} \\
\bottomrule
\end{tabular}
\end{table*}

\clearpage
\begin{table*}[t]
\footnotesize
\centering
\caption{The complete \textbf{Evaluation Prompt} for the \textbf{Gemini 1.5 Pro LLM-as-a-Judge}. 
This structured prompt was used for all 1.28 million evaluation calls. 
Placeholders (highlighted in \textcolor{bblue}{blue}) were programmatically filled with the relevant information for each specific VLM output being judged.}
\label{tab:llm_judge_prompt}
\begin{tabular}{p{0.95\textwidth}}
\toprule
\midrule
\textbf{\textcolor{cyan}{Evaluation Prompt with the Gemini 1.5 Pro LLM Judge}} \\
\midrule

\begin{tcolorbox}[
  colback=gray!10,
  colframe=gray!50,
  title=\textbf{System Persona and Core Task},
  fonttitle=\bfseries,
  boxrule=0.5pt,
  arc=2mm
]
\small
You are a meticulous and objective AI expert specializing in the evaluation of Vision-Language Model (VLM) responses. Your task is to evaluate the quality of a VLM's response based on the provided context and scoring criteria.
\end{tcolorbox}

\vspace{0.3cm}

\begin{tcolorbox}[
  colback=gray!10,
  colframe=gray!50,
  title=\textbf{Evaluation Context and Candidate Response},
  fonttitle=\bfseries,
  boxrule=0.5pt,
  arc=2mm
]
\small
Please evaluate the following VLM response for the task: ``\textcolor{bblue}{\{task\_description\}}'' in \textcolor{bblue}{\{language\}}.

The response was generated \textcolor{bblue}{\{settings\_description\}} and is based on the following input image-text pair.

GROUND TRUTH CONTEXT:
\begin{itemize}
\item Image File: \textcolor{bblue}{\{image\_info\}}
\item Associated Text: ``\textcolor{bblue}{\{text\_info\}}''
\end{itemize}

CANDIDATE VLM RESPONSE TO EVALUATE (Text ID \textcolor{bblue}{\{text\_id\}}):
``\textcolor{bblue}{\{response\}}''
\end{tcolorbox}

\vspace{0.3cm}

\begin{tcolorbox}[
  colback=gray!10,
  colframe=gray!50,
  title=\textbf{Scoring Criteria and Output Format},
  fonttitle=\bfseries,
  boxrule=0.5pt,
  arc=2mm
]
\small
Please rate the quality of the candidate response on a scale from 0 (lowest quality) to 100 (highest quality) based on the following three criteria:

\begin{enumerate}
\item \textbf{Accuracy:} The response is factually correct and logically sound. Conclusions are well-supported and not arbitrary.
\item \textbf{Helpfulness:} The response is clear, instructive, and directly addresses the specific vision-language task.
\item \textbf{Linguistic Quality:} The response correctly interprets the task, and the language is fluent, natural, and coherent.
\end{enumerate}

\textbf{Important Evaluation Guidelines:}
\begin{itemize}
\item Remain strictly objective and do not let the length of the response influence your judgment.
\item Do not show preference for any AI assistant names that might appear in the response.
\item Strive to be strict in your scoring; do not award a perfect score easily.
\end{itemize}

\textbf{REQUIRED OUTPUT FORMAT:}
Return ONLY a single, valid JSON object with the score. Do not include any other text, explanations, or markdown formatting. Your entire response must be a single, valid JSON object.

\textbf{Example:}
\texttt{\{"score": 85\}}
\end{tcolorbox}
\\
\midrule
\bottomrule
\end{tabular}
\end{table*}

\clearpage
\subsection{The Prompts used for Response Generation in this study}
\subsubsection{Prompt Setting 1: Zero-shot, No Rationales}
\begin{tcolorbox}[
  enhanced,
  breakable,
  bicolor,
  colback=lightOlive,
  colupper=darkOlive,
  colframe=darkOlive,
  coltitle=white,
  fonttitle=\bfseries,
  title={English: Zero-shot, No Rationales},
  boxrule=0.8pt,
  width=\linewidth,
  arc=2mm,
  halign title=center,
  sharp corners=south
]
\tiny
LANGUAGE\_CONFIGS = \{\\
\hspace*{1em}"English": \{\\
\hspace*{2em}"code": "En",\\
\hspace*{2em}"system\_prompt": "You are an AI assistant that analyzes images and text.",\\[0.3em]

\hspace*{2em}"prompt\_template\_image\_only": (\\
\hspace*{3em}"You are an intelligent assistant tasked with analyzing an image.",\\
\hspace*{3em}"Task description: \{task\_description\}",\\
\hspace*{3em}"Provide only the analysis for this task, clearly labeled."\\
\hspace*{2em}),\\[0.3em]

\hspace*{2em}"prompt\_template\_image\_text": (\\
\hspace*{3em}"You are an intelligent assistant tasked with analyzing the relationship between an image and text.",\\
\hspace*{3em}"Examine both the image and the provided text carefully.",\\
\hspace*{3em}"Text associated with the image: \{text\_content\}",\\
\hspace*{3em}"Task description: \{task\_description\}",\\
\hspace*{3em}"Provide your analysis based on both sources, citing specific visual and textual evidence."\\
\hspace*{2em}),\\[0.3em]

\hspace*{2em}"tasks": \{\\
\hspace*{3em}1: "Analyze the image and list all objects present. Categorize each object (e.g., furniture, electronic devices, clothing) and be specific.",\\
\hspace*{3em}2: "Describe the overall scene. What is the setting, and what activities or events are taking place?",\\
\hspace*{3em}3: "Identify interactions or relationships between objects or entities. Explain spatial, functional, or social connections.",\\
\hspace*{3em}4: "Divide the image into semantic regions. Label each region (e.g., sky, buildings, people, street) and briefly describe its contents.",\\
\hspace*{3em}5: "Provide a detailed natural-language description of what is happening in the image, as if for someone who cannot see it.",\\
\hspace*{3em}6: "Extract parts of the text that directly reference entities, objects, or scenes in the image, and explain the visual evidence.",\\
\hspace*{3em}7: "Identify which parts of the text are not represented in the image and explain why they are unrelated.",\\
\hspace*{3em}8: "List places mentioned in the text or shown in the image. For each, indicate whether it appears in the text, the image, or both, and note if it is well known."\\
\hspace*{2em}\}\\
\hspace*{1em}\}\\
\}
\end{tcolorbox}

\begin{tcolorbox}[
  enhanced,
  breakable,
  bicolor,
  colback=lightCyan,
  colupper=darkCyan,
  colframe=darkCyan,
  coltitle=white,
  fonttitle=\bfseries,
  title={Japanese: Zero-shot, No Rationales},
  boxrule=1pt,
  width=\linewidth,
  arc=2mm,
  halign title=center
]
\tiny
\begin{CJK*}{UTF8}{min}
LANGUAGE\_CONFIGS = \{\\
\hspace*{1em}"Japanese": \{\\
\hspace*{2em}"code": "Jp",\\
\hspace*{2em}"system\_prompt": "あなたは画像とテキストを分析するAIアシスタントです。",\\[0.3em]

\hspace*{2em}"prompt\_template\_image\_only": (\\
\hspace*{3em}"あなたは画像を分析するインテリジェントなアシスタントです。以下のタスクを与えられた画像に対して実行してください。",\\
\hspace*{3em}"タスクの説明：\{task\_description\}",\\
\hspace*{3em}"このタスクにのみ焦点を当て、明確なラベル付きで分析結果を提示してください。"\\
\hspace*{2em}),\\[0.3em]

\hspace*{2em}"prompt\_template\_image\_text": (\\
\hspace*{3em}"あなたは画像とテキストの関係を分析するインテリジェントなアシスタントです。",\\
\hspace*{3em}"画像と提供されたテキストの両方を注意深く確認してください。",\\
\hspace*{3em}"画像に関連するテキスト：\{text\_content\}",\\
\hspace*{3em}"タスクの説明：\{task\_description\}",\\
\hspace*{3em}"画像とテキスト双方の証拠を具体的に参照しながら分析結果を提示してください。"\\
\hspace*{2em}),\\[0.3em]

\hspace*{2em}"tasks": \{\\
\hspace*{3em}1: "この画像を分析し、存在するすべてのオブジェクトを列挙してください。家具・電子機器・衣類などに分類し、具体的に記述してください。",\\
\hspace*{3em}2: "画像全体のシーンを説明してください。設定は何で、どのような活動や出来事が生じていますか。",\\
\hspace*{3em}3: "画像内のオブジェクト／エンティティ間の相互作用・関係を特定し、空間的・機能的・社会的なつながりを説明してください。",\\
\hspace*{3em}4: "画像を意味（セマンティック）領域に分割し、各領域（例：空・建物・人物・通り）にラベルを付け、内容を簡潔に説明してください。",\\
\hspace*{3em}5: "視覚に頼れない人へ説明するつもりで、この画像で起きていることを自然言語で詳細に記述してください。",\\
\hspace*{3em}6: "画像に描かれた実体・物体・情景を直接参照するテキストの箇所を抽出し、対応する視覚的証拠を説明してください。",\\
\hspace*{3em}7: "テキストにあるが画像に表れていない箇所を特定し、それらが関連しない理由を説明してください。",\\
\hspace*{3em}8: "テキストに記載／画像に表示されている場所を列挙し、それぞれがテキスト・画像・両方のどれに現れるかを示し、有名な地名であれば注記してください。"\\
\hspace*{2em}\}\\
\hspace*{1em}\}\\
\}
\end{CJK*}
\end{tcolorbox}

\clearpage
\begin{tcolorbox}[
  enhanced,
  breakable,
  bicolor,
  colback=Honeydew,
  colupper=darkLime,
  colframe=darkLime,
  coltitle=white,
  fonttitle=\bfseries,
  title={Swahili: Zero-shot, No Rationales},
  boxrule=1pt,
  width=\linewidth,
  arc=2mm,
  halign title=center
]
\tiny
LANGUAGE\_CONFIGS = \{\\
\hspace*{1em}"Swahili": \{\\
\hspace*{2em}"code": "Sw",\\[0.2em]

\hspace*{2em}"system\_prompt": "Wewe ni msaidizi wa AI anayechambua picha na maandishi.",\\[0.3em]

\hspace*{2em}"prompt\_template\_image\_only": (\\
\hspace*{3em}"Wewe ni msaidizi mwerevu uliyepewa jukumu la kuchambua picha.",\\
\hspace*{3em}"Maelezo ya kazi: \{task\_description\}",\\
\hspace*{3em}"Toa uchambuzi kwa kazi hii pekee, ukiuweka wazi kwa lebo iliyo dhahiri."\\
\hspace*{2em}),\\[0.3em]

\hspace*{2em}"prompt\_template\_image\_text": (\\
\hspace*{3em}"Wewe ni msaidizi mwerevu unayeuchambua uhusiano kati ya picha na maandishi.",\\
\hspace*{3em}"Tafadhali kagua kwa uangalifu picha na maandishi yaliyotolewa.",\\
\hspace*{3em}"Maandishi yanayohusishwa na picha: \{text\_content\}",\\
\hspace*{3em}"Maelezo ya kazi: \{task\_description\}",\\
\hspace*{3em}"Toa uchambuzi unaorejea ushahidi mahususi kutoka kwa picha na kutoka kwenye maandishi."\\
\hspace*{2em}),\\[0.3em]

\hspace*{2em}"tasks": \{\\
\hspace*{3em}1: "Chambua picha na orodhesha vitu vyote vilivyopo. Weka kila kipengee katika makundi kama samani, vifaa vya kielektroniki, mavazi, n.k.; kuwa makini na mahususi.",\\
\hspace*{3em}2: "Eleza mandhari ya jumla ya picha. Ni mpangilio upi, na shughuli au matukio gani yanafanyika?",\\
\hspace*{3em}3: "Tambua mwingiliano au uhusiano kati ya vitu/entiti. Eleza miunganiko ya kieneo (nafasi), kiutendaji, au kijamii.",\\
\hspace*{3em}4: "Gawanya picha katika maeneo ya kimaana (semantiki). Wekea kila eneo lebo (mfano: anga, majengo, watu, barabara) na eleza maudhui yake kwa ufupi.",\\
\hspace*{3em}5: "Toa maelezo ya kina kwa lugha asilia ya kinachoendelea kwenye picha, kana kwamba unamwelezea mtu asiyeona.",\\
\hspace*{3em}6: "Tenga na orodhesha sehemu za maandishi zinazorejelea moja kwa moja entiti, vitu, au mandhari yanayoonekana kwenye picha; eleza ushahidi wa kuona.",\\
\hspace*{3em}7: "Tambua sehemu za maandishi ambazo hazionekani kwenye picha na eleza kwa nini hazihusiani.",\\
\hspace*{3em}8: "Orodhesha maeneo yaliyotajwa kwenye maandishi au kuonyeshwa kwenye picha. Kwa kila moja, onyesha kama yapo kwenye maandishi, kwenye picha, au vyote; taja pia iwapo ni maeneo yanayojulikana."\\
\hspace*{2em}\}\\
\hspace*{1em}\}\\
\}
\end{tcolorbox}

\begin{tcolorbox}[
  enhanced,
  breakable,
  colback=lightOrange,
  colframe=darkOrange,
  coltitle=white,
  fonttitle=\bfseries,
  title={Urdu: Zero-shot, No Rationales},
  boxrule=1pt,
  width=\linewidth,
  arc=2mm,
  halign title=center,
]
\tiny
LANGUAGE\_CONFIGS = \{\\
\hspace*{1em}"Urdu": \{\\
\hspace*{2em}"code": "Ur",\\[0.2em]

\hspace*{2em}"system\_prompt": \begin{UrduText}آپ ایک AI معاون ہیں جو تصاویر اور متن کا تجزیہ کرتے ہیں۔\end{UrduText},\\[0.3em]

\hspace*{2em}"prompt\_template\_image\_only": (\\
\hspace*{3em}\begin{UrduText}آپ ایک ذہین معاون ہیں جنہیں دی گئی تصویر کا تجزیہ کرنے کا کام سونپا گیا ہے۔\end{UrduText},\\
\hspace*{3em}\begin{UrduText}کام کی وضاحت:\end{UrduText} \{task\_description\},\\
\hspace*{3em}\begin{UrduText}براہِ کرم صرف اسی کام پر توجہ دیں اور واضح لیبل کے ساتھ اپنا تجزیہ پیش کریں۔\end{UrduText}\\
\hspace*{2em}),\\[0.3em]

\hspace*{2em}"prompt\_template\_image\_text": (\\
\hspace*{3em}\begin{UrduText}آپ ایک ذہین معاون ہیں جو تصویر اور متن کے باہمی تعلق کا تجزیہ کرتے ہیں۔\end{UrduText},\\
\hspace*{3em}\begin{UrduText}براہِ کرم تصویر اور فراہم کردہ متن—دونوں—غور سے دیکھیں۔\end{UrduText},\\
\hspace*{3em}\begin{UrduText}تصویر سے متعلق متن:\end{UrduText} \{text\_content\},\\
\hspace*{3em}\begin{UrduText}کام کی وضاحت:\end{UrduText} \{task\_description\},\\
\hspace*{3em}\begin{UrduText}تصویر اور متن—دونوں—سے حاصل شواہد کا حوالہ دیتے ہوئے اپنا تجزیہ فراہم کریں۔\end{UrduText}\\
\hspace*{2em}),\\[0.3em]

\hspace*{2em}"tasks": \{\\
\hspace*{3em}1: \begin{UrduText}تصویر کا تجزیہ کریں اور موجود تمام اشیاء کی فہرست بنائیں۔ ہر شے کو گروپس (جیسے فرنیچر، برقی آلات، لباس وغیرہ) میں تقسیم کریں اور مختصر مگر واضح وضاحت دیں۔\end{UrduText},\\
\hspace*{3em}2: \begin{UrduText}تصویر کے مجموعی منظر کی وضاحت کریں: یہ سیٹنگ کیا ہے اور کون سی سرگرمیاں یا واقعات ہو رہے ہیں؟\end{UrduText},\\
\hspace*{3em}3: \begin{UrduText}تصویر میں موجود اشیاء یا اداروں کے باہمی تعامل/تعلق کی نشاندہی کریں۔ مکانی، فعالی، یا سماجی ربط کی وضاحت کریں۔\end{UrduText},\\
\hspace*{3em}4: \begin{UrduText}تصویر کو مختلف معنوی (Semantic) حصوں میں تقسیم کریں۔ ہر حصے (مثلاً آسمان، عمارتیں، لوگ، سڑک) کو لیبل کریں اور اس کے مواد کی مختصر وضاحت دیں۔\end{UrduText},\\
\hspace*{3em}5: \begin{UrduText}ایسے شخص کے لیے جو تصویر نہیں دیکھ سکتا، فطری زبان میں تفصیلی بیان فراہم کریں کہ اس تصویر میں کیا ہو رہا ہے۔\end{UrduText},\\
\hspace*{3em}6: \begin{UrduText}متن کے وہ حصے الگ کر کے فہرست کریں جو براہِ راست تصویر میں موجود اشیاء، اداروں یا مناظر کا حوالہ دیتے ہیں، اور بصری شواہد کی وضاحت کریں۔\end{UrduText},\\
\hspace*{3em}7: \begin{UrduText}متن کے وہ حصے شناخت کریں جو تصویر میں نظر نہیں آتے اور بتائیں کہ وہ کیوں غیر متعلق ہیں۔\end{UrduText},\\
\hspace*{3em}8: \begin{UrduText}متن میں مذکور یا تصویر میں دکھائے گئے مقامات کی فہرست بنائیں۔ ہر مقام کے لیے بتائیں کہ وہ متن میں ہے، تصویر میں ہے، یا دونوں میں؛ اور اگر معروف/مشہور ہو تو مختصراً وضاحت کریں۔\end{UrduText}\\
\hspace*{2em}\}\\
\hspace*{1em}\}\\
\}
\end{tcolorbox}

\clearpage
\subsubsection{Prompt Setting 2: Zero-shot, With Rationales}
\begin{tcolorbox}[
  enhanced,
  breakable,
  bicolor,
  colback=lightOlive,
  colupper=darkOlive,
  colframe=darkOlive,
  coltitle=white,
  fonttitle=\bfseries,
  title={English: Zero-shot, With Rationales},
  boxrule=1pt,
  width=\linewidth,
  arc=2mm,
  halign title=center
]
\tiny
LANGUAGE\_CONFIGS = \{\\
\hspace*{1em}``English'': \{\\
\hspace*{2em}``code'': ``En'',\\

\hspace*{2em}``system\_prompt'': ``You are an AI assistant that analyzes images and text.'',\\

\hspace*{2em}``prompt\_template\_image\_only'': (\\
\hspace*{3em}\{task\_description\},\\
\hspace*{3em}``Please provide your analysis along with your rationale. Explain step by step how you arrived at your answer.''\\
\hspace*{2em}),\\

\hspace*{2em}``prompt\_template\_image\_text'': (\\
\hspace*{3em}``Text associated with the image: \{text\_content\}'',\\
\hspace*{3em}\{task\_description\},\\
\hspace*{3em}``Please provide your analysis and rationale. Explain step by step how you derived your response by referencing evidence from both the image and text.''\\
\hspace*{2em}),\\

\hspace*{2em}``tasks'': \{\\
\hspace*{3em}1: (\\
\hspace*{4em}``Let's analyze this image step by step to identify and categorize all objects present.'',\\
\hspace*{4em}``Steps to follow: 1. Scan the image systematically (left to right, top to bottom). 2. List each identified object. 3. Group objects into categories (furniture, electronics, clothing, etc.). 4. Verify even small or partially visible objects are included.'',\\
\hspace*{4em}``After these steps, provide your detailed answer and then your rationale explaining how you identified and categorized the objects. Think step by step.''\\
\hspace*{3em}),\\

\hspace*{3em}2: (\\
\hspace*{4em}``Let's analyze the scene step by step for a comprehensive description.'',\\
\hspace*{4em}``Steps to follow: 1. Identify the primary setting or location. 2. Note the time of day and overall atmosphere. 3. Identify main activities or events occurring. 4. Observe any additional background activities.'',\\
\hspace*{4em}``After these steps, provide your detailed description and then your rationale describing what visual cues led to your interpretation. Think step by step.''\\
\hspace*{3em}),\\

\hspace*{3em}3: (\\
\hspace*{4em}``Let's analyze the interactions and relationships in this image step by step.'',\\
\hspace*{4em}``Steps to follow: 1. Identify every entity (object or person) in the image. 2. Examine spatial relationships among the entities. 3. Look for direct interactions. 4. Consider any functional or implied social connections.'',\\
\hspace*{4em}``After these steps, provide your detailed analysis and then your rationale on how you established these relationships. Think step by step.''\\
\hspace*{3em}),\\

\hspace*{3em}4: (\\
\hspace*{4em}``Let's divide this image into semantic regions step by step.'',\\
\hspace*{4em}``Steps to follow: 1. Identify major spatial divisions in the image. 2. Determine the function or nature of each region. 3. Describe the contents within each region. 4. Observe transitions between regions.'',\\
\hspace*{4em}``After these steps, provide your detailed breakdown and then your rationale explaining how and why you segmented the image this way. Think step by step.''\\
\hspace*{3em}),\\

\hspace*{3em}5: (\\
\hspace*{4em}``Let's create a detailed narrative description of this image step by step.'',\\
\hspace*{4em}``Steps to follow: 1. Establish the main subject or focus. 2. Describe the context and surroundings. 3. Note any actions or movement. 4. Include key details about appearance and condition.'',\\
\hspace*{4em}``After these steps, provide your detailed narrative and then your rationale describing what guided your narrative construction. Think step by step.''\\
\hspace*{3em}),\\

\hspace*{3em}6: (\\
\hspace*{4em}``Let's analyze the relationship between the image and text step by step.'',\\
\hspace*{4em}``Steps to follow: 1. Carefully observe all visual elements in the image. 2. Read the text thoroughly to understand its description. 3. Create a list of key entities or concepts mentioned in the text. 4. For each entity, identify its visual counterpart in the image.'',\\
\hspace*{4em}``After these steps, provide your detailed analysis and then your rationale explaining how the text and visual elements correspond. Think step by step.''\\
\hspace*{3em}),\\

\hspace*{3em}7: (\\
\hspace*{4em}``Let's identify which parts of the text are not related to the image step by step.'',\\
\hspace*{4em}``Steps to follow: 1. Document all key visual elements in the image. 2. Read through the text to list all statements or claims. 3. For each textual element, check if there is visual evidence. 4. Identify text elements that do not correspond to anything in the image.'',\\
\hspace*{4em}``After these steps, provide your detailed analysis and then your rationale explaining why those parts lack visual support. Think step by step.''\\
\hspace*{3em}),\\

\hspace*{3em}8: (\\
\hspace*{4em}``Let's analyze the places mentioned in the image and text step by step.'',\\
\hspace*{4em}``Steps to follow: 1. Identify any locations visible in the image. 2. Look for explicit mentions of places in the text. 3. For each place, determine whether it is present in the image, the text, or both.'',\\
\hspace*{4em}``After these steps, provide your detailed analysis and then your rationale explaining how you verified each location and its significance. Think step by step.''\\
\hspace*{3em})\\
\hspace*{2em}\}\\
\hspace*{1em}\}\\
\}
\end{tcolorbox}

\clearpage
\begin{tcolorbox}[
  enhanced,
  breakable,
  bicolor,
  colback=lightCyan,
  colupper=darkCyan,
  colframe=darkCyan,
  coltitle=white,
  fonttitle=\bfseries,
  title={Japanese: Zero-shot, With Rationales},
  boxrule=1pt,
  width=\linewidth,
  arc=2mm,
  halign title=center
]
\tiny
\begin{CJK*}{UTF8}{min}
LANGUAGE\_CONFIGS = \{\\
\hspace*{1em}"Japanese": \{\\
\hspace*{2em}"code": "Jp",\\[0.2em]

\hspace*{2em}"system\_prompt": "あなたは画像とテキストを分析するAIアシスタントです。",\\[0.3em]

\hspace*{2em}"prompt\_template\_image\_only": (\\
\hspace*{3em}\{task\_description\},\\
\hspace*{3em}"分析結果とその根拠を示してください。どのように結論に至ったかを段階的（ステップバイステップ）に説明してください。"\\
\hspace*{2em}),\\[0.3em]

\hspace*{2em}"prompt\_template\_image\_text": (\\
\hspace*{3em}"画像に関連するテキスト：\{text\_content\}",\\
\hspace*{3em}\{task\_description\},\\
\hspace*{3em}"画像とテキストの双方から得られる証拠を参照しつつ、分析結果と根拠を示してください。どのように結論に至ったかを段階的（ステップバイステップ）に説明してください。"\\
\hspace*{2em}),\\[0.3em]

\hspace*{2em}"tasks": \{\\
\hspace*{3em}1: (\\
\hspace*{4em}"画像を段階的に分析し、存在するすべてのオブジェクトを特定・分類してください。",\\
\hspace*{4em}"手順：1) 左→右・上→下の順で体系的に走査する。 2) 特定した各オブジェクトを列挙する。 3) 家具／電子機器／衣類などのカテゴリに分類する。 4) 小さな物体や部分的にしか見えない要素も漏らさない。",\\
\hspace*{4em}"最終的な一覧と分類根拠を提示してください（段階的説明つき）。"\\
\hspace*{3em}),\\

\hspace*{3em}2: (\\
\hspace*{4em}"シーンを段階的に分析し、包括的な説明を行ってください。",\\
\hspace*{4em}"手順：1) 主たる設定／場所を特定。 2) 時刻感や雰囲気を推定。 3) 主な活動／イベントを特定。 4) 背景の補助的な動きや状況を観察。",\\
\hspace*{4em}"詳細な記述と、参照した視覚的手掛かりを根拠として示してください（段階的説明つき）。"\\
\hspace*{3em}),\\

\hspace*{3em}3: (\\
\hspace*{4em}"画像内の相互作用・関係性を段階的に分析してください。",\\
\hspace*{4em}"手順：1) すべてのエンティティ（物体・人物）を列挙。 2) エンティティ間の空間関係を把握。 3) 直接的な相互作用を探索。 4) 機能的／暗示的な社会的つながりを検討。",\\
\hspace*{4em}"確立した関係とその根拠を提示してください（段階的説明つき）。"\\
\hspace*{3em}),\\

\hspace*{3em}4: (\\
\hspace*{4em}"画像をセマンティック領域に段階的に分割してください。",\\
\hspace*{4em}"手順：1) 主要な空間区分を特定。 2) 各領域の機能・性質を判断。 3) 各領域の内容を記述。 4) 領域間の遷移や境界を観察。",\\
\hspace*{4em}"分割結果とセグメント化の根拠を提示してください（段階的説明つき）。"\\
\hspace*{3em}),\\

\hspace*{3em}5: (\\
\hspace*{4em}"画像の詳細なナラティブ説明を段階的に作成してください。",\\
\hspace*{4em}"手順：1) 主題・焦点を明確化。 2) 文脈と周囲環境を説明。 3) 行動や動きを記述。 4) 外観・状態に関する重要な詳細を含める。",\\
\hspace*{4em}"完成したナラティブと、その構築に至る根拠を提示してください（段階的説明つき）。"\\
\hspace*{3em}),\\

\hspace*{3em}6: (\\
\hspace*{4em}"画像とテキストの対応関係を段階的に分析してください。",\\
\hspace*{4em}"手順：1) 画像の視覚要素を精査。 2) テキストを丁寧に読解。 3) テキストで言及される主要エンティティ／概念を列挙。 4) 各エンティティの画像内対応箇所を特定。",\\
\hspace*{4em}"対応関係の詳細と、その根拠を提示してください（段階的説明つき）。"\\
\hspace*{3em}),\\

\hspace*{3em}7: (\\
\hspace*{4em}"テキストのうち画像と対応しない部分を段階的に特定してください。",\\
\hspace*{4em}"手順：1) 画像の主要視覚要素を記録。 2) テキストの記述・主張を列挙。 3) 各要素に視覚的裏付けがあるか検証。 4) 非対応の要素を抽出。",\\
\hspace*{4em}"非対応となる理由と根拠を提示してください（段階的説明つき）。"\\
\hspace*{3em}),\\

\hspace*{3em}8: (\\
\hspace*{4em}"画像とテキストで言及される場所を段階的に分析してください。",\\
\hspace*{4em}"手順：1) 画像で識別可能な場所を特定。 2) テキストで明示される場所を抽出。 3) 各場所が画像／テキスト／両方のいずれに現れるかを判定。",\\
\hspace*{4em}"各場所の確認方法と重要性を根拠とともに提示してください（段階的説明つき）。"\\
\hspace*{3em})\\
\hspace*{2em}\}\\
\hspace*{1em}\}\\
\}
\end{CJK*}
\end{tcolorbox}

\clearpage
\begin{tcolorbox}[
  enhanced,
  breakable,
  bicolor,
  colback=Honeydew,
  colupper=darkLime,
  colframe=darkLime,
  coltitle=white,
  fonttitle=\bfseries,
  title={Swahili: Zero-shot, With Rationales},
  boxrule=1pt,
  width=\linewidth,
  arc=2mm,
  halign title=center
]
\tiny
LANGUAGE\_CONFIGS = \{\\
\hspace*{1em}``Swahili'': \{\\
\hspace*{2em}``code'': ``Sw'',\\

\hspace*{2em}``system\_prompt'': ``Wewe ni msaidizi wa AI anayechambua picha na maandishi.'',\\

\hspace*{2em}``prompt\_template\_image\_only'': (\\
\hspace*{3em}\{task\_description\},\\
\hspace*{3em}``Tafadhali toa uchambuzi wako pamoja na msingi wake. Eleza hatua kwa hatua jinsi ulivyofikia jibu lako.''\\
\hspace*{2em}),\\

\hspace*{2em}``prompt\_template\_image\_text'': (\\
\hspace*{3em}``Maandishi yanayohusiana na picha: \{text\_content\}'',\\
\hspace*{3em}\{task\_description\},\\
\hspace*{3em}``Tafadhali toa uchambuzi wako pamoja na msingi wake. Eleza hatua kwa hatua jinsi ulivyopata jibu lako kwa kurejelea ushahidi kutoka picha na maandishi.''\\
\hspace*{2em}),\\

\hspace*{2em}``tasks'': \{\\
\hspace*{3em}1: (\\
\hspace*{4em}``Tuchambue picha hii hatua kwa hatua ili kubaini na kupanga vitu vyote vilivyopo.'',\\
\hspace*{4em}``Hatua za kufuata: 1. Changanua picha kwa mpangilio (kushoto kwenda kulia, juu kwenda chini). 2. Orodhesha kila kitu kilichobainika. 3. Pangilia vitu kwenye makundi (samani, vifaa vya kielektroniki, nguo, n.k.). 4. Hakikisha vitu vidogo au vinavyoonekana sehemu tu vimejumuishwa.'',\\
\hspace*{4em}``Baada ya hatua hizi, toa jibu lako la kina na msingi unaoelezea jinsi ulivyobaini na kupanga vitu. Fikiria hatua kwa hatua.''\\
\hspace*{3em}),\\

\hspace*{3em}2: (\\
\hspace*{4em}``Tuchambue mandhari hatua kwa hatua kwa maelezo kamili.'',\\
\hspace*{4em}``Hatua za kufuata: 1. Tambua mazingira kuu au eneo. 2. Elezea wakati wa siku na hali ya jumla. 3. Tambua shughuli au matukio kuu yanayotokea. 4. Angalia shughuli zingine za nyuma.'',\\
\hspace*{4em}``Baada ya hatua hizi, toa maelezo yako ya kina na msingi unaoelezea ni ishara gani za kuona zilizoongoza tafsiri yako. Fikiria hatua kwa hatua.''\\
\hspace*{3em}),\\

\hspace*{3em}3: (\\
\hspace*{4em}``Tuchambue mwingiliano na uhusiano katika picha hii hatua kwa hatua.'',\\
\hspace*{4em}``Hatua za kufuata: 1. Tambua kila kiumbe (kitu au mtu) katika picha. 2. Chunguza mahusiano ya kijiografia kati ya viumbe. 3. Tafuta mwingiliano wa moja kwa moja. 4. Zingatia uhusiano wowote wa kiutendaji au kijamii wa dhumuni.'',\\
\hspace*{4em}``Baada ya hatua hizi, toa uchambuzi wako wa kina na msingi unaoelezea jinsi ulivyounda uhusiano huo. Fikiria hatua kwa hatua.''\\
\hspace*{3em}),\\

\hspace*{3em}4: (\\
\hspace*{4em}``Tugawanye picha hii katika maeneo ya semantiki hatua kwa hatua.'',\\
\hspace*{4em}``Hatua za kufuata: 1. Tambua mgawanyiko mkuu wa nafasi katika picha. 2. Amua kazi au asili ya kila eneo. 3. Elezea yaliyomo katika kila eneo. 4. Angalia mabadiliko kati ya maeneo.'',\\
\hspace*{4em}``Baada ya hatua hizi, toa muhtasari wa kina na msingi unaoelezea jinsi na kwanini uligawa picha kwa njia hii. Fikiria hatua kwa hatua.''\\
\hspace*{3em}),\\

\hspace*{3em}5: (\\
\hspace*{4em}``Tunde maelezo ya kina ya simulizi ya picha hii hatua kwa hatua.'',\\
\hspace*{4em}``Hatua za kufuata: 1. Weka mada kuu au lengo. 2. Elezea muktadha na mazingira. 3. Elezea matendo au mwendo. 4. Jumuisha maelezo muhimu kuhusu muonekano na hali.'',\\
\hspace*{4em}``Baada ya hatua hizi, toa simulizi yako ya kina na msingi unaoelezea nini kilikuongoza katika uundaji wa simulizi. Fikiria hatua kwa hatua.''\\
\hspace*{3em}),\\

\hspace*{3em}6: (\\
\hspace*{4em}``Tuchambue uhusiano kati ya picha na maandishi hatua kwa hatua.'',\\
\hspace*{4em}``Hatua za kufuata: 1. Angalia kwa makini vipengele vyote vya kuona katika picha. 2. Soma maandishi kwa kina kuelewa maelezo. 3. Tengeneza orodha ya viumbe au dhana kuu zilizotajwa kwenye maandishi. 4. Kwa kila kiumbe, tambua sambamba zake katika picha.'',\\
\hspace*{4em}``Baada ya hatua hizi, toa uchambuzi wako wa kina na msingi unaoelezea jinsi maandishi na vipengele vya kuona vinavyolingana. Fikiria hatua kwa hatua.''\\
\hspace*{3em}),\\

\hspace*{3em}7: (\\
\hspace*{4em}``Tambua sehemu za maandishi ambazo hazihusiani na picha hatua kwa hatua.'',\\
\hspace*{4em}``Hatua za kufuata: 1. Rekodi vipengele vyote muhimu vya kuona katika picha. 2. Soma maandishi ili kuorodhesha taarifa zote au madai. 3. Kwa kila kipengele cha maandishi, angalia kama kuna ushahidi wa kuona. 4. Tambua vipengele vya maandishi visivyolingana na picha.'',\\
\hspace*{4em}``Baada ya hatua hizi, toa uchambuzi wako wa kina na msingi unaoelezea kwanini sehemu hizo hazina msaada wa kuona. Fikiria hatua kwa hatua.''\\
\hspace*{3em}),\\

\hspace*{3em}8: (\\
\hspace*{4em}``Tuchambue maeneo yaliyotajwa katika picha na maandishi hatua kwa hatua.'',\\
\hspace*{4em}``Hatua za kufuata: 1. Tambua maeneo yoyote yanayoonekana kwenye picha. 2. Tafuta taarifa za maeneo zilizoelezwa kwa wazi kwenye maandishi. 3. Kwa kila eneo, amua kama lipo kwenye picha, maandishi, au vyote viwili.'',\\
\hspace*{4em}``Baada ya hatua hizi, toa uchambuzi wako wa kina na msingi unaoelezea jinsi ulivyothibitisha kila eneo na umuhimu wake. Fikiria hatua kwa hatua.''\\
\hspace*{3em})\\
\hspace*{2em}\}\\
\hspace*{1em}\}\\
\}
\end{tcolorbox}

\clearpage
\begin{tcolorbox}[
  enhanced,
  breakable,
  colback=lightOrange,
  colframe=darkOrange,
  coltitle=white,
  fonttitle=\bfseries,
  title={Urdu: Zero-shot, With Rationales},
  boxrule=1pt,
  width=\linewidth,
  arc=2mm,
  halign title=center,
]
\tiny
LANGUAGE\_CONFIGS = \{\\
\hspace*{1em}"Urdu": \{\\
\hspace*{2em}"code": "Ur",\\[0.2em]

\hspace*{2em}"system\_prompt": \begin{UrduText}آپ ایک AI معاون ہیں جو تصاویر اور متن کا تجزیہ کرتا/کرتی ہے۔\end{UrduText},\\[0.3em]

\hspace*{2em}"prompt\_template\_image\_only": (\\
\hspace*{3em}\{task\_description\},\\
\hspace*{3em}\begin{UrduText}براہِ کرم اپنا تجزیہ اور اس کی بنیاد پیش کریں، اور مرحلہ وار وضاحت کریں کہ آپ نتیجے تک کیسے پہنچے۔\end{UrduText}\\
\hspace*{2em}),\\[0.3em]

\hspace*{2em}"prompt\_template\_image\_text": (\\
\hspace*{3em}\begin{UrduText}تصویر سے متعلق متن:\end{UrduText} \{text\_content\},\\
\hspace*{3em}\{task\_description\},\\
\hspace*{3em}\begin{UrduText}براہِ کرم اپنا تجزیہ اور اس کی بنیاد پیش کریں۔ تصویر اور متن دونوں سے شواہد کا حوالہ دیتے ہوئے مرحلہ وار بیان کریں کہ آپ نے جواب کیسے اخذ کیا۔\end{UrduText}\\
\hspace*{2em}),\\[0.3em]

\hspace*{2em}"tasks": \{\\
\hspace*{3em}1: (\\
\hspace*{4em}\begin{UrduText}اس تصویر کا مرحلہ وار تجزیہ کریں اور موجود تمام اشیاء کی نشاندہی اور درجہ بندی کریں۔\end{UrduText},\\
\hspace*{4em}\begin{UrduText}اقدامات: 1) تصویر کو منظم انداز میں اسکین کریں (بائیں سے دائیں، اوپر سے نیچے)۔ 2) ہر شناخت شدہ شے کی فہرست بنائیں۔ 3) اشیاء کو زمروں میں تقسیم کریں (جیسے فرنیچر، الیکٹرانکس، لباس وغیرہ)۔ 4) چھوٹی یا جزوی طور پر نظر آنے والی اشیاء بھی شامل کریں۔\end{UrduText},\\
\hspace*{4em}\begin{UrduText}ان اقدامات کے بعد تفصیلی جواب اور درجہ بندی کی بنیاد واضح کریں (مرحلہ وار توضیح کے ساتھ)。\end{UrduText}\\
\hspace*{3em}),\\

\hspace*{3em}2: (\\
\hspace*{4em}\begin{UrduText}منظر (سین) کا مرحلہ وار تجزیہ کریں اور جامع وضاحت فراہم کریں۔\end{UrduText},\\
\hspace*{4em}\begin{UrduText}اقدامات: 1) مرکزی پس منظر/مقام کی نشاندہی کریں۔ 2) وقت اور مجموعی فضا/ماحول کا اندازہ لگائیں۔ 3) جاری سرگرمیوں یا واقعات کو شناخت کریں۔ 4) پس منظر میں اضافی حرکات یا اشاروں کا مشاہدہ کریں۔\end{UrduText},\\
\hspace*{4em}\begin{UrduText}بعد ازاں، تفصیلی بیان اور وہ بصری اشارے بتائیں جن پر تشریح مبنی ہے (مرحلہ وار)。\end{UrduText}\\
\hspace*{3em}),\\

\hspace*{3em}3: (\\
\hspace*{4em}\begin{UrduText}تصویر میں موجود اشیاء/اکائیوں کے باہمی تعاملات اور تعلقات کا مرحلہ وار تجزیہ کریں۔\end{UrduText},\\
\hspace*{4em}\begin{UrduText}اقدامات: 1) تمام اکائیوں (اشیاء/افراد) کی فہرست بنائیں۔ 2) اکائیوں کے درمیان مکانی تعلقات جانچیں۔ 3) براہِ راست تعاملات تلاش کریں۔ 4) فعالی یا مترشح سماجی روابط پر غور کریں۔\end{UrduText},\\
\hspace*{4em}\begin{UrduText}آخر میں تعلقات کی تفصیل اور ان کی بنیاد واضح کریں (مرحلہ وار)。\end{UrduText}\\
\hspace*{3em}),\\

\hspace*{3em}4: (\\
\hspace*{4em}\begin{UrduText}تصویر کو مرحلہ وار مختلف معنوی (سیمینٹک) علاقوں میں تقسیم کریں۔\end{UrduText},\\
\hspace*{4em}\begin{UrduText}اقدامات: 1) بنیادی مکانی حصوں کی نشاندہی کریں۔ 2) ہر حصے کی نوعیت/کام متعین کریں۔ 3) ہر حصے کے مواد کی مختصر وضاحت دیں۔ 4) حصوں کے مابین انتقال/سرحدوں کا مشاہدہ کریں۔\end{UrduText},\\
\hspace*{4em}\begin{UrduText}بعد ازاں تقسیم کی تفصیل اور اس کے جواز/بنیاد پیش کریں (مرحلہ وار)。\end{UrduText}\\
\hspace*{3em}),\\

\hspace*{3em}5: (\\
\hspace*{4em}\begin{UrduText}تصویر کی تفصیلی بیانیہ (نیرئیٹو) وضاحت مرحلہ وار تیار کریں۔\end{UrduText},\\
\hspace*{4em}\begin{UrduText}اقدامات: 1) مرکزی موضوع/مرکزِ توجہ طے کریں۔ 2) سیاق و سباق اور اردگرد کے حالات بیان کریں۔ 3) حرکات/افعال کی نشاندہی کریں۔ 4) ظاہری حالت اور متعلقہ جزئیات شامل کریں۔\end{UrduText},\\
\hspace*{4em}\begin{UrduText}پھر مکمل بیانیہ اور اس کے استدلالی جواز پیش کریں (مرحلہ وار)。\end{UrduText}\\
\hspace*{3em}),\\

\hspace*{3em}6: (\\
\hspace*{4em}\begin{UrduText}متن اور تصویر کے باہمی تقابل/مطابقت کا مرحلہ وار تجزیہ کریں۔\end{UrduText},\\
\hspace*{4em}\begin{UrduText}اقدامات: 1) تصویری عناصر کو بغور دیکھیں۔ 2) متن مکمل پڑھ کر مفہوم سمجھیں۔ 3) متن میں مذکور کلیدی اکائیوں/تصورات کی فہرست بنائیں۔ 4) ہر اکائی کے لیے تصویر میں موزوں ثبوت/مقام شناخت کریں۔\end{UrduText},\\
\hspace*{4em}\begin{UrduText}مطابقت کی تفصیل اور اس کی بنیاد واضح کریں (مرحلہ وار)。\end{UrduText}\\
\hspace*{3em}),\\

\hspace*{3em}7: (\\
\hspace*{4em}\begin{UrduText}متن کے وہ حصے مرحلہ وار شناخت کریں جو تصویر سے غیر متعلق ہیں۔\end{UrduText},\\
\hspace*{4em}\begin{UrduText}اقدامات: 1) تصویر کے تمام نمایاں بصری عناصر درج کریں۔ 2) متن کے بیانات/دعوات کی فہرست بنائیں۔ 3) ہر بیان کے لیے بصری ثبوت تلاش کریں۔ 4) جن بیانات کے ثبوت نہ ملیں انہیں غیر متعلق/غیر مطابق قرار دیں۔\end{UrduText},\\
\hspace*{4em}\begin{UrduText}بعد ازاں وجوہ اور شواہد کی کمی کی بنیاد بیان کریں (مرحلہ وار)。\end{UrduText}\\
\hspace*{3em}),\\

\hspace*{3em}8: (\\
\hspace*{4em}\begin{UrduText}تصویر اور متن میں مذکور مقامات کا مرحلہ وار جائزہ لیں۔\end{UrduText},\\
\hspace*{4em}\begin{UrduText}اقدامات: 1) تصویر میں قابلِ شناخت مقامات نوٹ کریں۔ 2) متن میں واضح طور پر مذکور مقامات اخذ کریں۔ 3) ہر مقام کے بارے میں بتائیں کہ وہ تصویر میں ہے، متن میں ہے یا دونوں میں۔\end{UrduText},\\
\hspace*{4em}\begin{UrduText}ہر مقام کی تصدیق کے طریقے اور اس کی اہمیت کا جواز پیش کریں (مرحلہ وار)。\end{UrduText}\\
\hspace*{3em})\\
\hspace*{2em}\}\\
\hspace*{1em}\}\\
\}
\end{tcolorbox}

\clearpage
\subsubsection{Prompt Setting 3: One-shot, No rationales}
\begin{tcolorbox}[
  enhanced,
  breakable,
  bicolor,
  colback=lightOlive,
  colupper=darkOlive,
  colframe=darkOlive,
  coltitle=white,
  fonttitle=\bfseries,
  title={English: One-shot, No Rationales},
  boxrule=1pt,
  width=\linewidth,
  arc=2mm,
  halign title=center,
  boxsep=1mm
]
\tiny
EXAMPLES = \{\\
\hspace*{1em}1: \{\\
\hspace*{2em}``question'': ``Analyze this image and list all objects present. Categorize each object into groups such as furniture, electronic devices, clothing, etc.'',\\
\hspace*{2em}``example\_response'': (\\
\hspace*{3em}Example Analysis:\\
\hspace*{3em}1. Furniture: Airport benches\\
\hspace*{3em}2. Electronics: Cameras, Smartphones\\
\hspace*{3em}3. Clothing: Jackets, Hats\\
\hspace*{3em}4. Misc: Luggage, Backpacks )\\
\hspace*{1em}\},\\
\hspace*{1em}2: \{\\
\hspace*{2em}``question'': ``Describe the overall scene in this image. What is the setting, and what activities or events are taking place?'',\\
\hspace*{2em}``example\_response'': (\\
\hspace*{3em}Example Analysis:\\
\hspace*{3em}``An airport terminal with a crowd holding protest signs, some seated and others standing.'' )\\
\hspace*{1em}\},\\
\hspace*{1em}3: \{\\
\hspace*{2em}``question'': ``Identify any interactions or relationships between objects or entities in this image. Explain their connections.'',\\
\hspace*{2em}``example\_response'': (\\
\hspace*{3em}Example Analysis:\\
\hspace*{3em}``Protesters clustered around signs indicate coordinated demonstration.'' )\\
\hspace*{1em}\},\\
\hspace*{1em}4: \{\\
\hspace*{2em}``question'': ``Divide this image into semantic regions. Label each region and describe its contents.'',\\
\hspace*{2em}``example\_response'': (\\
\hspace*{3em}Example Analysis:\\
\hspace*{3em}- Foreground: Protesters and signs\\
\hspace*{3em}- Middle: Dense crowd\\
\hspace*{3em}- Background: Airport interior )\\
\hspace*{1em}\},\\
\hspace*{1em}5: \{\\
\hspace*{2em}``question'': ``Provide a detailed natural language description of what is happening in this image.'',\\
\hspace*{2em}``example\_response'': (\\
\hspace*{3em}Example Analysis:\\
\hspace*{3em}``A protest is taking place in an airport terminal; participants hold signs and gather in groups.'' )\\
\hspace*{1em}\},\\
\hspace*{1em}6: \{\\
\hspace*{2em}``question'': ``Extract and list the specific parts of the text that reference entities, objects, or scenes depicted in the image.'',\\
\hspace*{2em}``example\_response'': (\\
\hspace*{3em}Example Analysis:\\
\hspace*{3em}``The text mentions 'protest signs' which are clearly visible in the crowd.'' )\\
\hspace*{1em}\},\\
\hspace*{1em}7: \{\\
\hspace*{2em}``question'': ``Identify which parts of the text are not represented in the image.'',\\
\hspace*{2em}``example\_response'': (\\
\hspace*{3em}Example Analysis:\\
\hspace*{3em}``Mentions of judicial rulings are not depicted in the image.'' )\\
\hspace*{1em}\},\\
\hspace*{1em}8: \{\\
\hspace*{2em}``question'': ``List the places mentioned in the text or shown in the image.'',\\
\hspace*{2em}``example\_response'': (\\
\hspace*{3em}Example Analysis:\\
\hspace*{3em}``JFK Airport is implied by the terminal setting.'' )\\
\hspace*{1em}\}\\
\}\\

LANGUAGE\_CONFIGS = \{\\
\hspace*{1em}``English'': \{\\
\hspace*{2em}``code'': ``En'',\\

\hspace*{2em}``system\_prompt'': ``You are an AI assistant that analyzes images and text.'',\\

\hspace*{2em}``prompt\_template\_image\_only'': (\\
\hspace*{3em}``Example:'',\\
\hspace*{3em}``Question: \{example\_question\}'',\\
\hspace*{3em}``Response: \{example\_response\}'',\\
\hspace*{3em}``\\'',\\
\hspace*{3em}``Now, analyze the following task:'',\\
\hspace*{3em}``\{task\_description\}'',\\
\hspace*{3em}``\\'',\\
\hspace*{3em}``Provide your analysis.''\\
\hspace*{2em}),\\

\hspace*{2em}``prompt\_template\_image\_text'': (\\
\hspace*{3em}``Example:'',\\
\hspace*{3em}``Question: \{example\_question\}'',\\
\hspace*{3em}``Response: \{example\_response\}'',\\
\hspace*{3em}``Text associated with the image:'',\\
\hspace*{3em}``\{text\_content\}'',\\
\hspace*{3em}``Task:'',\\
\hspace*{3em}``\{task\_description\}'',\\
\hspace*{3em}``Provide your analysis based on both the image and the text.''\\
\hspace*{2em}),\\

\hspace*{2em}``tasks'': \{\\
\hspace*{3em}1: ``Analyze this image and list all objects present. Categorize each object into groups such as furniture, electronic devices, and clothing.'',\\
\hspace*{3em}2: ``Describe the overall scene in this image, including the setting and any visible activities.'',\\
\hspace*{3em}3: ``Identify any interactions or relationships between entities in this image and explain their connections.'',\\
\hspace*{3em}4: ``Divide this image into semantic regions. Label each region and describe its contents.'',\\
\hspace*{3em}5: ``Provide a detailed natural language description of what is happening in this image.'',\\
\hspace*{3em}6: ``Extract from the text the parts that reference entities, objects, or scenes depicted in the image and explain the visual evidence.'',\\
\hspace*{3em}7: ``Identify which parts of the text are not represented in the image and explain why.'',\\
\hspace*{3em}8: ``List the places mentioned in the text or shown in the image; for each, indicate if it appears in the text, image, or both and explain its significance.''\\
\hspace*{2em}\},\\

\hspace*{2em}``examples'': EXAMPLES\\
\hspace*{1em}\}\\
\}
\end{tcolorbox}

\clearpage
\begin{tcolorbox}[
  enhanced,
  breakable,
  bicolor,
  colback=lightCyan,
  colupper=darkCyan,
  colframe=darkCyan,
  coltitle=white,
  fonttitle=\bfseries,
  title={Japanese: One-shot, No Rationales},
  boxrule=1pt,
  width=\linewidth,
  arc=2mm,
  halign title=center
]
\tiny
\begin{CJK}{UTF8}{min}
EXAMPLES = \{\\
\hspace*{1em}1: \{\\
\hspace*{2em}``question'': ``この画像を分析し、存在するすべてのオブジェクトをリストアップしてください。家具、電子機器、衣類などのグループに分類してください。'',\\
\hspace*{2em}``example\_response'': (\\
\hspace*{3em}Example Analysis:\\
\hspace*{3em}1. Furniture: 空港のベンチ\\
\hspace*{3em}2. Electronics: カメラ、スマートフォン\\
\hspace*{3em}3. Clothing: ジャケット、帽子\\
\hspace*{3em}4. Misc: 手荷物、バックパック )\\
\hspace*{1em}\},\\
\hspace*{1em}2: \{\\
\hspace*{2em}``question'': ``この画像全体のシーンを説明してください。設定は何で、どのような活動やイベントが行われていますか？'',\\
\hspace*{2em}``example\_response'': (\\
\hspace*{3em}Example Analysis:\\
\hspace*{3em}``空港のターミナルで抗議用のプラカードを持った群衆がいて、一部は座り、一部は立っています。'' )\\
\hspace*{1em}\},\\
\hspace*{1em}3: \{\\
\hspace*{2em}``question'': ``この画像内のオブジェクトやエンティティ間の相互作用や関係を特定し、そのつながりを説明してください。'',\\
\hspace*{2em}``example\_response'': (\\
\hspace*{3em}Example Analysis:\\
\hspace*{3em}``プラカードを囲む抗議者のクラスターは、統制されたデモを示しています。'' )\\
\hspace*{1em}\},\\
\hspace*{1em}4: \{\\
\hspace*{2em}``question'': ``この画像をセマンティック領域に分割してください。各領域にラベルを付け、その内容を説明してください。'',\\
\hspace*{2em}``example\_response'': (\\
\hspace*{3em}Example Analysis:\\
\hspace*{3em}- 前景：抗議者とプラカード\\
\hspace*{3em}- 中景：密集した群衆\\
\hspace*{3em}- 背景：空港の内部 )\\
\hspace*{1em}\},\\
\hspace*{1em}5: \{\\
\hspace*{2em}``question'': ``この画像で何が起こっているのか、詳細に自然言語で説明してください。'',\\
\hspace*{2em}``example\_response'': (\\
\hspace*{3em}Example Analysis:\\
\hspace*{3em}``空港のターミナルで抗議が行われており、参加者はプラカードを持ってグループで集まっています。'' )\\
\hspace*{1em}\},\\
\hspace*{1em}6: \{\\
\hspace*{2em}``question'': ``画像に描かれているエンティティ、オブジェクト、またはシーンを参照するテキストの特定の部分を抽出してリストアップしてください。'',\\
\hspace*{2em}``example\_response'': (\\
\hspace*{3em}Example Analysis:\\
\hspace*{3em}``テキストには「抗議用プラカード」が言及されており、群衆に明確に見えています。'' )\\
\hspace*{1em}\},\\
\hspace*{1em}7: \{\\
\hspace*{2em}``question'': ``テキストのどの部分が画像に表現されていないかを特定してください。'',\\
\hspace*{2em}``example\_response'': (\\
\hspace*{3em}Example Analysis:\\
\hspace*{3em}``司法判断の言及は画像には描かれていません。'' )\\
\hspace*{1em}\},\\
\hspace*{1em}8: \{\\
\hspace*{2em}``question'': ``テキストに記載されている、または画像に示されている場所をリストアップしてください。'',\\
\hspace*{2em}``example\_response'': (\\
\hspace*{3em}Example Analysis:\\
\hspace*{3em}``JFK空港はターミナルの設定から示唆されています。'' )\\
\hspace*{1em}\}\\
\}\\

LANGUAGE\_CONFIGS = \{\\
\hspace*{1em}``Japanese'': \{\\
\hspace*{2em}``code'': ``Jp'',\\

\hspace*{2em}``system\_prompt'': ``あなたは画像とテキストを分析するAIアシスタントです。'',\\

\hspace*{2em}``prompt\_template\_image\_only'': (\\
\hspace*{3em}``Example:'',\\
\hspace*{3em}``Question: \{example\_question\}'',\\
\hspace*{3em}``Response: \{example\_response\}'',\\
\hspace*{3em}``Now, analyze the following task:'',\\
\hspace*{3em}``\{task\_description\}'',\\
\hspace*{3em}``Provide your analysis.''\\
\hspace*{2em}),\\

\hspace*{2em}``prompt\_template\_image\_text'': (\\
\hspace*{3em}``Example:'',\\
\hspace*{3em}``Question: \{example\_question\}'',\\
\hspace*{3em}``Response: \{example\_response\}'',\\
\hspace*{3em}``Text associated with the image:'',\\
\hspace*{3em}``\{text\_content\}'',\\
\hspace*{3em}``Task:'',\\
\hspace*{3em}``\{task\_description\}'',\\
\hspace*{3em}``Provide your analysis based on both the image and the text.''\\
\hspace*{2em}),\\

\hspace*{2em}``tasks'': \{\\
\hspace*{3em}1: ``この画像を分析し、存在するすべてのオブジェクトをリストアップしてください。家具、電子機器、衣類などのグループに分類してください。'',\\
\hspace*{3em}2: ``この画像全体のシーンを説明してください。設定は何で、どのような活動やイベントが行われていますか？'',\\
\hspace*{3em}3: ``この画像内のエンティティ間の相互作用や関係を特定し、そのつながりを説明してください。'',\\
\hspace*{3em}4: ``この画像をセマンティック領域に分割してください。各領域にラベルを付け、その内容を説明してください。'',\\
\hspace*{3em}5: ``この画像で何が起こっているのか、詳細に自然言語で説明してください。'',\\
\hspace*{3em}6: ``テキストから、画像に描かれているエンティティ、オブジェクト、またはシーンを参照する部分を抽出し、視覚的証拠を説明してください。'',\\
\hspace*{3em}7: ``テキストのどの部分が画像に表現されていないかを特定し、その理由を説明してください。'',\\
\hspace*{3em}8: ``テキストに記載されている、または画像に示されている場所をリストアップし、それぞれがテキスト、画像、または両方に表示されているかを示し、その重要性を説明してください。''\\
\hspace*{2em}\},\\

\hspace*{2em}``examples'': EXAMPLES\\
\hspace*{1em}\}\\
\}
\end{CJK}
\end{tcolorbox}

\clearpage
\begin{tcolorbox}[
  enhanced,
  breakable,
  bicolor,
  colback=Honeydew,
  colupper=darkLime,
  colframe=darkLime,
  coltitle=white,
  fonttitle=\bfseries,
  title={Swahili: One-shot, No Rationales},
  boxrule=1pt,
  width=\linewidth,
  arc=2mm,
  halign title=center
]
\tiny
EXAMPLES = \{\\
\hspace*{1em}1: \{\\
\hspace*{2em}``question'': ``Chambua picha hii na orodhesha vitu vyote vilivyopo. Vikundi vipange kama samani, vifaa vya kielektroniki, nguo, n.k.'',\\
\hspace*{2em}``example\_response'': (\\
\hspace*{3em}Example Analysis:\\
\hspace*{3em}1. Furniture: Vibanzi vya uwanja wa ndege\\
\hspace*{3em}2. Electronics: Kamera, Simu mahiri\\
\hspace*{3em}3. Clothing: Jaketi, Kofia\\
\hspace*{3em}4. Misc: Mizigo, Mabegi )\\
\hspace*{1em}\},\\
\hspace*{1em}2: \{\\
\hspace*{2em}``question'': ``Eleza mandhari yote ya picha hii. Ni wapi inafanyika na ni shughuli gani au matukio gani yanayoendelea?'',\\
\hspace*{2em}``example\_response'': (\\
\hspace*{3em}Example Analysis:\\
\hspace*{3em}``Jamii inayoonyesha mabango ya maandamano katika terminali ya uwanja wa ndege; baadhi wamekaa na wengine wamesimama.'' )\\
\hspace*{1em}\},\\
\hspace*{1em}3: \{\\
\hspace*{2em}``question'': ``Tambua mwingiliano au uhusiano wowote kati ya vitu au entiti katika picha hii. Eleza muunganisho wao.'',\\
\hspace*{2em}``example\_response'': (\\
\hspace*{3em}Example Analysis:\\
\hspace*{3em}``Kikundi cha waandamanaji waliozunguka mabango kinaonyesha maandamano yaliyopangwa.'' )\\
\hspace*{1em}\},\\
\hspace*{1em}4: \{\\
\hspace*{2em}``question'': ``Gawanya picha hii katika maeneo ya maana (semantic regions). Wewekee lebo kila eneo na elezea yaliyomo ndani.'',\\
\hspace*{2em}``example\_response'': (\\
\hspace*{3em}Example Analysis:\\
\hspace*{3em}- Mbele: Waandamanaji na mabango\\
\hspace*{3em}- Kati: Umati uliojaa watu\\
\hspace*{3em}- Nyuma: Ndani ya uwanja wa ndege )\\
\hspace*{1em}\},\\
\hspace*{1em}5: \{\\
\hspace*{2em}``question'': ``Toa maelezo ya kina kwa lugha ya asili ya kile kinachoendelea katika picha hii.'',\\
\hspace*{2em}``example\_response'': (\\
\hspace*{3em}Example Analysis:\\
\hspace*{3em}``Maandamano yanafanyika kwenye terminali ya uwanja wa ndege; washiriki wanashikilia mabango na kukusanyika kwa makundi.'' )\\
\hspace*{1em}\},\\
\hspace*{1em}6: \{\\
\hspace*{2em}``question'': ``Tenganisha na orodhesha sehemu maalum za maandishi zinazorejelea entiti, vitu, au mandhari yaliyoonyeshwa kwenye picha.'',\\
\hspace*{2em}``example\_response'': (\\
\hspace*{3em}Example Analysis:\\
\hspace*{3em}``Maandishi yanasema 'mabango ya maandamano' ambayo inaonekana wazi kwenye umati.'' )\\
\hspace*{1em}\},\\
\hspace*{1em}7: \{\\
\hspace*{2em}``question'': ``Tambua ni sehemu gani za maandishi ambazo hazionekani kwenye picha.'',\\
\hspace*{2em}``example\_response'': (\\
\hspace*{3em}Example Analysis:\\
\hspace*{3em}``Marejeo ya maagizo ya mahakama hayajaonyeshwa kwenye picha.'' )\\
\hspace*{1em}\},\\
\hspace*{1em}8: \{\\
\hspace*{2em}``question'': ``Orodhesha maeneo yaliyotajwa kwenye maandishi au kuonyeshwa kwenye picha.'',\\
\hspace*{2em}``example\_response'': (\\
\hspace*{3em}Example Analysis:\\
\hspace*{3em}``Uwanja wa ndege wa JFK unaashiriwa na mazingira ya terminali.'' )\\
\hspace*{1em}\}\\
\}\\

LANGUAGE\_CONFIGS = \{\\
\hspace*{1em}``Swahili'': \{\\
\hspace*{2em}``code'': ``Sw'',\\

\hspace*{2em}``system\_prompt'': ``Wewe ni msaidizi wa AI anayechambua picha na maandishi.'',\\

\hspace*{2em}``prompt\_template\_image\_only'': (\\
\hspace*{3em}``Example:'',\\
\hspace*{3em}``Question: \{example\_question\}'',\\
\hspace*{3em}``Response: \{example\_response\}'',\\
\hspace*{3em}``Now, analyze the following task:'',\\
\hspace*{3em}``\{task\_description\}'',\\
\hspace*{3em}``Provide your analysis.''\\
\hspace*{2em}),\\

\hspace*{2em}``prompt\_template\_image\_text'': (\\
\hspace*{3em}``Example:'',\\
\hspace*{3em}``Question: \{example\_question\}'',\\
\hspace*{3em}``Response: \{example\_response\}'',\\
\hspace*{3em}``Text associated with the image:'',\\
\hspace*{3em}``\{text\_content\}'',\\
\hspace*{3em}``Task:'',\\
\hspace*{3em}``\{task\_description\}'',\\
\hspace*{3em}``Provide your analysis based on both the image and the text.''\\
\hspace*{2em}),\\

\hspace*{2em}``tasks'': \{\\
\hspace*{3em}1: ``Chambua picha hii na orodhesha vitu vyote vilivyopo. Vikundi vipange kama samani, vifaa vya kielektroniki, nguo, n.k.'',\\
\hspace*{3em}2: ``Eleza mandhari yote ya picha hii. Ni wapi inafanyika na ni shughuli gani au matukio gani yanayoendelea?'',\\
\hspace*{3em}3: ``Tambua mwingiliano au uhusiano wowote kati ya entiti katika picha hii na eleza muunganisho wao.'',\\
\hspace*{3em}4: ``Gawanya picha hii katika maeneo ya maana (semantic regions). Wewekee lebo kila eneo na elezea yaliyomo ndani.'',\\
\hspace*{3em}5: ``Toa maelezo ya kina kwa lugha ya asili ya kile kinachoendelea katika picha hii.'',\\
\hspace*{3em}6: ``Tenganisha na orodhesha sehemu za maandishi zinazorejelea entiti, vitu, au mandhari yaliyoonyeshwa kwenye picha na eleza ushahidi wa kuona.'',\\
\hspace*{3em}7: ``Tambua ni sehemu gani za maandishi ambazo hazionekani kwenye picha na eleza kwanini.'',\\
\hspace*{3em}8: ``Orodhesha maeneo yaliyotajwa kwenye maandishi au kuonyeshwa kwenye picha; kwa kila moja, ongea kama inaonekana katika maandishi, picha, au zote mbili na elezea umuhimu wake.''\\
\hspace*{2em}\},\\

\hspace*{2em}``examples'': EXAMPLES\\
\hspace*{1em}\}\\
\}
\end{tcolorbox}

\clearpage
\begin{tcolorbox}[
  enhanced,
  breakable,
  colback=lightOrange,
  colframe=darkOrange,
  coltitle=white,
  fonttitle=\bfseries,
  title={Urdu: One-shot, No Rationales},
  boxrule=1pt,
  width=\linewidth,
  arc=2mm,
  halign title=center,
]
\tiny

EXAMPLES = \{\\
\hspace*{1em}1: \{ ``question'': \begin{UrduText}اس تصویر کا تجزیہ کریں اور موجود تمام اشیاء کی فہرست بنائیں۔ ہر شے کو فرنیچر، الیکٹرانک آلات، پوشاک وغیرہ کے گروپ میں تقسیم کریں۔\end{UrduText},\\
\hspace*{2em} ``example\_response'': ( \begin{UrduText}مثالی تجزیہ: 1) فرنیچر: ہوائی اڈے کی بینچیں؛ 2) الیکٹرانکس: کیمرے، اسمارٹ فونز؛ 3) پوشاک: جیکٹس، ٹوپیاں؛ 4) دیگر: سامان، بیگز۔\end{UrduText} ) \},\\
\hspace*{1em}2: \{ ``question'': \begin{UrduText}اس تصویر کے مجموعی منظر کی وضاحت کریں۔ یہ کہاں ہے اور کون سی سرگرمیاں یا واقعات ہو رہے ہیں؟\end{UrduText},\\
\hspace*{2em} ``example\_response'': ( \begin{UrduText}مثالی تجزیہ: ہوائی اڈے کے ٹرمینل میں احتجاجی پلے کارڈز اٹھائے ہجوم موجود ہے؛ کچھ لوگ بیٹھے ہیں اور کچھ کھڑے ہیں۔\end{UrduText} ) \},\\
\hspace*{1em}3: \{ ``question'': \begin{UrduText}اس تصویر میں موجود اشیاء یا اکائیوں کے درمیان تعاملات یا تعلقات کی نشاندہی کریں اور ان کے روابط بیان کریں۔\end{UrduText},\\
\hspace*{2em} ``example\_response'': ( \begin{UrduText}مثالی تجزیہ: پلے کارڈز کے گرد جمع افراد منظم احتجاج کی نشاندہی کرتے ہیں؛ گروہی قربت سماجی ربط دکھاتی ہے۔\end{UrduText} ) \},\\
\hspace*{1em}4: \{ ``question'': \begin{UrduText}اس تصویر کو معنوی (semantic) خطوں میں تقسیم کریں۔ ہر خطے کو لیبل کریں اور اس کے مواد کی وضاحت کریں۔\end{UrduText},\\
\hspace*{2em} ``example\_response'': ( \begin{UrduText}مثالی تجزیہ: — پیش منظر: احتجاج کرنے والے اور پلے کارڈز؛ — درمیانی منظر: گنجان ہجوم؛ — پس منظر: ہوائی اڈے کا اندرونی حصہ۔\end{UrduText} ) \},\\
\hspace*{1em}5: \{ ``question'': \begin{UrduText}اس تصویر میں کیا ہو رہا ہے، اس کا تفصیلی بیان قدرتی زبان میں فراہم کریں۔\end{UrduText},\\
\hspace*{2em} ``example\_response'': ( \begin{UrduText}مثالی تجزیہ: ہوائی اڈے کے ٹرمینل میں احتجاج جاری ہے؛ شرکاء پلے کارڈز تھامے چھوٹے گروہوں کی صورت میں جمع ہیں۔\end{UrduText} ) \},\\
\hspace*{1em}6: \{ ``question'': \begin{UrduText}متن کے ان مخصوص حصوں کو نکالیں اور فہرست بنائیں جو تصویر میں دکھائی گئی اکائیوں، اشیاء یا مناظر کا حوالہ دیتے ہیں۔\end{UrduText},\\
\hspace*{2em} ``example\_response'': ( \begin{UrduText}مثالی تجزیہ: متن میں ‘‘احتجاجی پلے کارڈز’’ کا ذکر ہے جو ہجوم میں واضح دکھائی دیتے ہیں۔\end{UrduText} ) \},\\
\hspace*{1em}7: \{ ``question'': \begin{UrduText}متن کے وہ حصے کون سے ہیں جو تصویر میں ظاہر نہیں کیے گئے؟\end{UrduText},\\
\hspace*{2em} ``example\_response'': ( \begin{UrduText}مثالی تجزیہ: عدالتی فیصلوں کا ذکر تصویر میں موجود نہیں۔\end{UrduText} ) \},\\
\hspace*{1em}8: \{ ``question'': \begin{UrduText}متن میں یا تصویر میں مذکور مقامات کی فہرست بنائیں۔\end{UrduText},\\
\hspace*{2em} ``example\_response'': ( \begin{UrduText}مثالی تجزیہ: ٹرمینل کے پس منظر سے JFK ہوائی اڈے کا اندازہ ہوتا ہے۔\end{UrduText} ) \} \}\\

LANGUAGE\_CONFIGS = \{\\
\hspace*{1em} ``Urdu'': \{\\
\hspace*{2em} ``code'': ``Ur'',\\
\hspace*{2em} ``system\_prompt'': \begin{UrduText}آپ ایک AI معاون ہیں جو تصاویر اور متن کا تجزیہ کرتا ہے۔\end{UrduText},\\

\hspace*{2em} ``prompt\_template\_image\_only'': (\\
\hspace*{3em} ``Example:'',\\
\hspace*{3em} ``\begin{UrduText}سوال:\end{UrduText} \{example\_question\}'',\\
\hspace*{3em} ``\begin{UrduText}جواب:\end{UrduText} \{example\_response\}'',\\
\hspace*{3em} ``\begin{UrduText}اب درج ذیل کام کا تجزیہ کریں:\end{UrduText}'',\\
\hspace*{3em} ``\{task\_description\}'',\\
\hspace*{3em} ``\begin{UrduText}اپنا تجزیہ فراہم کریں۔\end{UrduText}'' ),\\

\hspace*{2em} ``prompt\_template\_image\_text'': (\\
\hspace*{3em} ``Example:'',\\
\hspace*{3em} ``\begin{UrduText}سوال:\end{UrduText} \{example\_question\}'',\\
\hspace*{3em} ``\begin{UrduText}جواب:\end{UrduText} \{example\_response\}'',\\
\hspace*{3em} ``\begin{UrduText}تصویر سے متعلق متن:\end{UrduText}'',\\
\hspace*{3em} ``\{text\_content\}'',\\
\hspace*{3em} ``\begin{UrduText}کام:\end{UrduText}'',\\
\hspace*{3em} ``\{task\_description\}'',\\
\hspace*{3em} ``\begin{UrduText}تصویر اور متن دونوں کی بنیاد پر اپنا تجزیہ فراہم کریں۔\end{UrduText}'' ),\\

\hspace*{2em} ``tasks'': \{\\
\hspace*{3em} 1: \begin{UrduText}اس تصویر کا تجزیہ کریں اور موجود تمام اشیاء کی فہرست بنائیں۔ ہر شے کو فرنیچر، الیکٹرانک آلات، پوشاک وغیرہ کے گروپ میں تقسیم کریں۔\end{UrduText},\\
\hspace*{3em} 2: \begin{UrduText}اس تصویر کے مجموعی منظر کی وضاحت کریں، جس میں مقام اور نظر آنے والی سرگرمیاں بھی شامل ہوں۔\end{UrduText},\\
\hspace*{3em} 3: \begin{UrduText}اس تصویر میں موجود اکائیوں کے درمیان تعاملات یا تعلقات کی نشاندہی کریں اور ان کے روابط بیان کریں۔\end{UrduText},\\
\hspace*{3em} 4: \begin{UrduText}اس تصویر کو معنوی خطوں میں تقسیم کریں۔ ہر خطے کو لیبل کریں اور اس کے مواد کی وضاحت کریں۔\end{UrduText},\\
\hspace*{3em} 5: \begin{UrduText}اس تصویر میں کیا ہو رہا ہے، اس کا تفصیلی بیان قدرتی زبان میں فراہم کریں۔\end{UrduText},\\
\hspace*{3em} 6: \begin{UrduText}متن سے ان حصوں کو نکالیں جو تصویر میں دکھائی گئی اکائیوں، اشیاء یا مناظر کا حوالہ دیتے ہیں، اور بصری شواہد بیان کریں۔\end{UrduText},\\
\hspace*{3em} 7: \begin{UrduText}متن کے وہ حصے شناخت کریں جو تصویر میں ظاہر نہیں کیے گئے، اور وجہ بھی بیان کریں۔\end{UrduText},\\
\hspace*{3em} 8: \begin{UrduText}متن یا تصویر میں مذکور مقامات کی فہرست بنائیں؛ ہر مقام کے لیے بتائیں کہ آیا یہ متن، تصویر یا دونوں میں ہے، اور اس کی اہمیت بیان کریں۔\end{UrduText}\\
\hspace*{2em} \},\\
\hspace*{2em} ``examples'': EXAMPLES\\
\hspace*{1em} \} \}
\end{tcolorbox}

\clearpage
\subsubsection{Prompt Setting 4: One-shot, With rationales}
\begin{tcolorbox}[
  enhanced,
  breakable,
  bicolor,
  colback=lightOlive,
  colupper=darkOlive,
  colframe=darkOlive,
  coltitle=white,
  fonttitle=\bfseries,
  title={English: One-shot, With Rationales},
  boxrule=1pt,
  width=\linewidth,
  arc=2mm,
  halign title=center,
  boxsep=1mm
]
\tiny
EXAMPLES = \{\\
\hspace*{1em}1: \{\\
\hspace*{2em}``question'': ``Analyze this image and list all objects present. Categorize each object into groups such as furniture, electronic devices, clothing, etc.'',\\
\hspace*{2em}``example\_response'': (\\
\hspace*{3em}``Example Analysis: Objects present include airport seating, electronic devices such as cameras and smartphones, and various articles of clothing.''\\
\hspace*{3em}``Rationale: The image was scanned systematically; objects were identified based on clear visual boundaries and grouped according to their common functions and appearances.'' )\\
\hspace*{1em}\},\\
\hspace*{1em}2: \{\\
\hspace*{2em}``question'': ``Describe the overall scene in this image. What is the setting, and what activities or events are taking place?'',\\
\hspace*{2em}``example\_response'': (\\
\hspace*{3em}``Example Analysis: The scene depicts an airport terminal with a large group of people, likely engaged in a protest. Key elements include seating areas and visible signage.''\\
\hspace*{3em}``Rationale: Visual cues such as the arrangement of people, signage, and environmental details indicate a public gathering in an airport.'' )\\
\hspace*{1em}\},\\
\hspace*{1em}3: \{\\
\hspace*{2em}``question'': ``Identify any interactions or relationships between objects or entities in this image. Explain any spatial, functional, or social connections.'',\\
\hspace*{2em}``example\_response'': (\\
\hspace*{3em}``Example Analysis: Individuals holding similar protest signs are clustered together, indicating both a social bond and a unified demonstration.''\\
\hspace*{3em}``Rationale: The proximity and matching visuals (like similar signs) support the inference of coordinated activity.'' )\\
\hspace*{1em}\},\\
\hspace*{1em}4: \{\\
\hspace*{2em}``question'': ``Divide this image into different semantic regions and describe each region's contents.'',\\
\hspace*{2em}``example\_response'': (\\
\hspace*{3em}``Example Analysis: The image is segmented into foreground (people and protest signs), middle ground (a dense crowd), and background (airport terminal features).''\\
\hspace*{3em}``Rationale: Clear visual separations allow the division into regions, helping structure the overall scene analysis.'' )\\
\hspace*{1em}\},\\
\hspace*{1em}5: \{\\
\hspace*{2em}``question'': ``Provide a detailed, natural language description of what is happening in this image.'',\\
\hspace*{2em}``example\_response'': (\\
\hspace*{3em}``Example Analysis: The image shows a protest at an airport terminal with diverse participants; some are seated while others stand, holding signs that express protest messages.''\\
\hspace*{3em}``Rationale: Observing the various elements (signs, group formations, terminal features) supports a comprehensive narrative of the event.'' )\\
\hspace*{1em}\},\\
\hspace*{1em}6: \{\\
\hspace*{2em}``question'': ``Extract and list the specific parts of the text that reference entities, objects, or scenes depicted in the image. Explain the visual evidence.'',\\
\hspace*{2em}``example\_response'': (\\
\hspace*{3em}``Example Analysis: The text mentions 'protest signs' which are clearly visible in the image among the crowd.''\\
\hspace*{3em}``Rationale: The correspondence between the text and the visible signs confirms the match.'' )\\
\hspace*{1em}\},\\
\hspace*{1em}7: \{\\
\hspace*{2em}``question'': ``Identify which parts of the text are not represented in the image. Explain what is missing.'',\\
\hspace*{2em}``example\_response'': (\\
\hspace*{3em}``Example Analysis: Details about judicial rulings are mentioned in the text but are not depicted in the image, which shows only protest activity.''\\
\hspace*{3em}``Rationale: The absence of legal or courtroom elements in the image indicates no visual support for those text parts.'' )\\
\hspace*{1em}\},\\
\hspace*{1em}8: \{\\
\hspace*{2em}``question'': ``What places are mentioned in the text or shown in the image? Indicate whether they appear in the text, the image, or both; and explain their significance.'',\\
\hspace*{2em}``example\_response'': (\\
\hspace*{3em}``Example Analysis: JFK Airport is inferred from the airport scene in the image and is mentioned in the text.''\\
\hspace*{3em}``Rationale: The combined visual and textual context highlights the significance of the location as a major travel hub.'' )\\
\hspace*{1em}\}\\
\}

LANGUAGE\_CONFIGS = \{\\
\hspace*{1em}``English'': \{\\
\hspace*{2em}``code'': ``En'',\\

\hspace*{2em}``system\_prompt'': ``You are an AI assistant that analyzes images and text.'',\\

\hspace*{2em}``prompt\_template\_image\_only'': (\\
\hspace*{3em}``Example:'',\\
\hspace*{3em}``Question: \{example\_question\}'',\\
\hspace*{3em}``Response: \{example\_response\}'',\\
\hspace*{3em}``Now, analyze this task:'',\\
\hspace*{3em}``\{task\_description\}'',\\
\hspace*{3em}``Provide your analysis.''\\
\hspace*{2em}),\\

\hspace*{2em}``prompt\_template\_image\_text'': (\\
\hspace*{3em}``Example:'',\\
\hspace*{3em}``Question: \{example\_question\}'',\\
\hspace*{3em}``Response: \{example\_response\}'',\\
\hspace*{3em}``Text associated with the image:'',\\
\hspace*{3em}``\{text\_content\}'',\\
\hspace*{3em}``Task:'',\\
\hspace*{3em}``\{task\_description\}'',\\
\hspace*{3em}``Provide your analysis based on both the image and the text.''\\
\hspace*{2em}),\\

\hspace*{2em}``tasks'': \{\\
\hspace*{3em}1: ``Analyze this image and list all objects present. Categorize the objects into groups such as furniture, electronic devices, and clothing.'',\\
\hspace*{3em}2: ``Describe the overall scene in this image, including the setting and visible activities.'',\\
\hspace*{3em}3: ``Identify any interactions or relationships between entities in this image and explain their connections.'',\\
\hspace*{3em}4: ``Divide this image into semantic regions. Label each region and briefly describe its contents.'',\\
\hspace*{3em}5: ``Provide a detailed natural language description of what is happening in this image.'',\\
\hspace*{3em}6: ``Extract from the text the parts that reference entities, objects, or scenes depicted in the image and explain the visual evidence.'',\\
\hspace*{3em}7: ``Identify the parts of the text that are not represented in the image and explain why.'',\\
\hspace*{3em}8: ``List the places mentioned in the text or shown in the image; for each, indicate if it appears in the text, image, or both and explain its significance.''\\
\hspace*{2em}\},\\

\hspace*{2em}``examples'': EXAMPLES\\
\hspace*{1em}\}\\
\}
\end{tcolorbox}

\clearpage
\begin{tcolorbox}[
  enhanced,
  breakable,
  bicolor,
  colback=lightCyan,
  colupper=darkCyan,
  colframe=darkCyan,
  coltitle=white,
  fonttitle=\bfseries,
  title={Japanese: One-shot, With Rationales},
  boxrule=1pt,
  width=\linewidth,
  arc=2mm,
  halign title=center
]
\tiny
\begin{CJK*}{UTF8}{min}

EXAMPLES = \{\\
\hspace*{1em}1: \{\\
\hspace*{2em}``question'': ``この画像を分析し、存在するすべてのオブジェクトをリストアップしてください。家具、電子機器、衣類などのグループに分類してください。'',\\
\hspace*{2em}``example\_response'': (\\
\hspace*{3em}``Example Analysis: 空港の座席、カメラやスマートフォンなどの電子機器、さまざまな衣類が含まれています。'',\\
\hspace*{3em}``Rationale: 画像を体系的にスキャンし、視覚的境界に基づいてオブジェクトを識別し、機能や外観に応じて分類しました。''\\
\hspace*{2em})\\
\hspace*{1em}\},\\
\hspace*{1em}2: \{\\
\hspace*{2em}``question'': ``この画像の全体的なシーンを説明してください。設定は何で、どのような活動やイベントが行われていますか？'',\\
\hspace*{2em}``example\_response'': (\\
\hspace*{3em}``Example Analysis: 空港ターミナルで大勢の人々が抗議を行っているように見えます。座席エリアや目立つサインが含まれます。'',\\
\hspace*{3em}``Rationale: 人々の配置、サイン、環境の手掛かりから公共の集会であることが示唆されます。''\\
\hspace*{2em})\\
\hspace*{1em}\},\\
\hspace*{1em}3: \{\\
\hspace*{2em}``question'': ``この画像内のオブジェクトやエンティティ間の相互作用や関係を特定し、それらのつながりを説明してください。'',\\
\hspace*{2em}``example\_response'': (\\
\hspace*{3em}``Example Analysis: 同じ抗議用サインを持つ人々が集まっており、社会的なつながりと統一されたデモを示しています。'',\\
\hspace*{3em}``Rationale: 近接性と共通のビジュアル（同様のサイン）から共同活動が推測されます。''\\
\hspace*{2em})\\
\hspace*{1em}\},\\
\hspace*{1em}4: \{\\
\hspace*{2em}``question'': ``この画像を異なるセマンティック領域に分割し、各領域の内容を説明してください。'',\\
\hspace*{2em}``example\_response'': (\\
\hspace*{3em}``Example Analysis: 前景（人々と抗議用サイン）、中景（密集した群衆）、背景（空港ターミナルの構造）に分割できます。'',\\
\hspace*{3em}``Rationale: 明確な視覚的分離により領域化が可能となり、全体のシーン分析が構造化されます。''\\
\hspace*{2em})\\
\hspace*{1em}\},\\
\hspace*{1em}5: \{\\
\hspace*{2em}``question'': ``この画像で何が起こっているのか、詳細な自然言語で説明してください。'',\\
\hspace*{2em}``example\_response'': (\\
\hspace*{3em}``Example Analysis: 空港ターミナルで抗議が行われ、参加者は座ったり立ったりしながらサインを掲げています。'',\\
\hspace*{3em}``Rationale: サイン、集団の配置、ターミナルの特徴といった要素が物語全体を裏付けます。''\\
\hspace*{2em})\\
\hspace*{1em}\},\\
\hspace*{1em}6: \{\\
\hspace*{2em}``question'': ``テキストの特定の部分を抽出し、画像に描かれたエンティティやオブジェクト、シーンを参照している箇所を挙げ、視覚的証拠を説明してください。'',\\
\hspace*{2em}``example\_response'': (\\
\hspace*{3em}``Example Analysis: テキストには『抗議用サイン』が言及され、群衆の中で明確に確認できます。'',\\
\hspace*{3em}``Rationale: テキストと視覚要素の対応づけにより整合性が検証されます。''\\
\hspace*{2em})\\
\hspace*{1em}\},\\
\hspace*{1em}7: \{\\
\hspace*{2em}``question'': ``テキストのどの部分が画像に表現されていないかを特定し、何が欠けているか説明してください。'',\\
\hspace*{2em}``example\_response'': (\\
\hspace*{3em}``Example Analysis: テキストには司法判断の詳細がありますが、画像には抗議活動のみが描かれ、法的・法廷的要素は見当たりません。'',\\
\hspace*{3em}``Rationale: 視覚フレームに法的要素がないため、該当テキストは画像に支えられていません。''\\
\hspace*{2em})\\
\hspace*{1em}\},\\
\hspace*{1em}8: \{\\
\hspace*{2em}``question'': ``テキストに記載された、または画像に示された場所を挙げ、それぞれがテキスト、画像、または両方に現れるかを示し、その重要性を説明してください。'',\\
\hspace*{2em}``example\_response'': (\\
\hspace*{3em}``Example Analysis: シーンからJFK空港が示唆され、テキストにも言及があります。'',\\
\hspace*{3em}``Rationale: 視覚文脈とテキスト文脈が一致し、主要交通拠点としての重要性が強調されます。''\\
\hspace*{2em})\\
\hspace*{1em}\}\\
\}\\[0.5em]

LANGUAGE\_CONFIGS = \{\\
\hspace*{1em}``Japanese'': \{\\
\hspace*{2em}``code'': ``Jp'',\\
\hspace*{2em}``system\_prompt'': ``あなたは画像とテキストを分析するAIアシスタントです。'',\\
\hspace*{2em}``prompt\_template\_image\_only'': (\\
\hspace*{3em}``Example:\\\\
Question: \{example\_question\} \\\\
Response: \{example\_response\} \\\\
Now, analyze this task:\\\\
\{task\_description\} \\\\
Provide your analysis with rationale.''\\
\hspace*{2em}),\\
\hspace*{2em}``prompt\_template\_image\_text'': (\\
\hspace*{3em}``Example:\\\\
Question: \{example\_question\} \\\\
Response: \{example\_response\} \\\\
Text associated with the image:\\\\
\{text\_content\} \\\\
Task:\\\\
\{task\_description\} \\\\
Provide your analysis based on both the image and the text, including your rationale.''\\
\hspace*{2em}),\\
\hspace*{2em}``tasks'': \{\\
\hspace*{3em}1: ``この画像を分析し、存在するすべてのオブジェクトをリストアップしてください。家具、電子機器、衣類などのグループに分類してください。'',\\
\hspace*{3em}2: ``この画像の全体的なシーンを説明してください。設定は何で、どのような活動やイベントが行われていますか？'',\\
\hspace*{3em}3: ``この画像内のエンティティ間の相互作用や関係を特定し、それらのつながりを説明してください。'',\\
\hspace*{3em}4: ``この画像をセマンティック領域に分割し、各領域の内容を簡潔に説明してください。'',\\
\hspace*{3em}5: ``この画像で何が起こっているのか、詳細な自然言語で説明してください。'',\\
\hspace*{3em}6: ``テキストから、画像に描かれたエンティティ／オブジェクト／シーンを参照する部分を抽出し、視覚的証拠を説明してください。'',\\
\hspace*{3em}7: ``テキストのどの部分が画像に表現されていないかを特定し、その理由を説明してください。'',\\
\hspace*{3em}8: ``テキストに記載された、または画像に示された場所をリストアップし、それぞれがテキスト・画像・両方のどれに現れるかを示し、その重要性を説明してください。''\\
\hspace*{2em}\},\\
\hspace*{2em}``examples'': EXAMPLES\\
\hspace*{1em}\}\\
\}\\

\end{CJK*}
\end{tcolorbox}

\clearpage
\begin{tcolorbox}[
  enhanced,
  breakable,
  bicolor,
  colback=Honeydew,
  colupper=darkLime,
  colframe=darkLime,
  coltitle=white,
  fonttitle=\bfseries,
  title={Swahili: One-shot, With Rationales},
  boxrule=1pt,
  width=\linewidth,
  arc=2mm,
  halign title=center
]
\tiny

EXAMPLES = \{\\
    1: \{\\
        ``question'': ``Chambua picha hii na orodhesha vitu vyote vilivyopo. Vikundi vipange kama samani, vifaa vya kielektroniki, nguo, n.k.'',\\
        ``example\_response'': (
            ``Uchambuzi wa Mfano: Vitu vilivyopo ni viti vya uwanja wa ndege, vifaa vya kielektroniki kama kamera na simu mahiri, na makundi mbalimbali ya mavazi. ''
            ``Misingi: Picha ilichunguzwa kwa mpangilio; vitu vilitambuliwa kulingana na mipaka ya kuona na kuwekwa vikundi kulingana na kazi zao na muonekano.''
        )
    \},\\
    2: \{\\
        ``question'': ``Eleza mandhari ya jumla ya picha hii. Je, setting ni ipi, na ni shughuli gani au matukio gani yanayoendelea?'',\\
        ``example\_response'': (
            ``Uchambuzi wa Mfano: Mandhari inaonyesha terminali ya uwanja wa ndege ambapo kundi kubwa la watu linafanya maandamano. Sehemu muhimu ni maeneo ya kukaa na alama zinazoonekana. ''
            ``Misingi: Ishara za kuona kama upangaji wa watu, alama, na maelezo ya mazingira zinaonyesha mkusanyiko wa umma katika uwanja wa ndege.''
        )
    \},\\
    3: \{\\
        ``question'': ``Tambua mwingiliano au uhusiano wowote kati ya vitu au entiti katika picha hii na eleza uhusiano wao.'',\\
        ``example\_response'': (
            ``Uchambuzi wa Mfano: Watu wanaoshikilia alama zinazofanana wamekusanyika pamoja, ikionyesha muungano wa kijamii na maandamano yaliyopangwa. ''
            ``Misingi: Ukaribu na muonekano unaolingana (alama zinazofanana) unaunga mkono hitimisho la shughuli iliyoratibiwa.''
        )
    \},\\
    4: \{\\
        ``question'': ``Gawanya picha hii katika maeneo tofauti ya semantiki na elezea yaliyomo katika kila eneo.'',\\
        ``example\_response'': (
            ``Uchambuzi wa Mfano: Picha imegawanywa katika sehemu za mbele (watu na alama za maandamano), sehemu ya kati (umati mnene), na nyuma (vipengele vya terminali ya uwanja wa ndege). ''
            ``Misingi: Tofauti za wazi za kuona zinawezesha mgawanyiko wa maeneo, kusaidia muundo wa uchambuzi wa jumla wa mandhari.''
        )
    \},\\
    5: \{\\
        ``question'': ``Toa maelezo ya kina kwa lugha ya asili ya kile kinachoendelea katika picha hii.'',\\
        ``example\_response'': (
            ``Uchambuzi wa Mfano: Picha inaonyesha maandamano katika terminali ya uwanja wa ndege; washiriki wamesimama au wamekaa, wakishikilia alama za maandamano. ''
            ``Misingi: Kuchunguza vipengele mbalimbali (alama, muundo wa vikundi, muundo wa terminali) kunasaidia hadithi kamili ya tukio.''
        )
    \},\\
    6: \{\\
        ``question'': ``Tenganisha na orodhesha sehemu maalum za maandishi zinazorejelea entiti, vitu, au mandhari yaliyoonyeshwa kwenye picha. Elezea ushahidi wa kuona.'',\\
        ``example\_response'': (
            ``Uchambuzi wa Mfano: Maandishi yanataja 'alama za maandamano' ambazo zinaonekana wazi katika picha kati ya umati. ''
            ``Misingi: Ulinganifu kati ya maandishi na alama zinazoonekana unatilia nguvu mathaliko.''
        )
    \},\\
    7: \{\\
        ``question'': ``Tambua ni sehemu gani za maandishi ambazo hazionekani kwenye picha. Elezea kinacho kukosekana.'',\\
        ``example\_response'': (
            ``Uchambuzi wa Mfano: Maelezo kuhusu uamuzi wa mahakama yanatajwa kwenye maandishi lakini hayajaonyeshwa kwenye picha, ambayo inaonyesha tu shughuli za maandamano. ''
            ``Misingi: Ukosefu wa vipengele vya kisheria au mahakamani kwenye picha unaonyesha hakuna msaada wa kuona kwa sehemu hizo za maandishi.''
        )
    \},\\
    8: \{\\
        ``question'': ``Ni maeneo gani yaliyotajwa kwenye maandishi au kuonyeshwa kwenye picha? Onyesha kama yanatokea kwenye maandishi, picha, au zote mbili; na elezea umuhimu wake.'',\\
        ``example\_response'': (
            ``Uchambuzi wa Mfano: Uwanja wa ndege wa JFK unaashiriwa na mandhari ya terminali na umetajwa kwenye maandishi. ''
            ``Misingi: Muktadha wa kuona na maandishi unaangazia umuhimu wa eneo hilo kama kitovu kikuu cha usafiri.''
        )
    \}\\
\}

LANGUAGE\_CONFIGS = \{\\
    ``Swahili'': \{\\
        ``code'': ``Sw'',\\
        ``system\_prompt'': ``Wewe ni msaidizi wa AI anayechambua picha na maandishi.'',\\
        ``prompt\_template\_image\_only'': (
            ``Mfano:\\\\
            Swali: \{example\_question\} \\\\
            Jibu: \{example\_response\} \\\\
            Sasa, chambua kazi ifuatayo:\\\\
            \{task\_description\} \\\\
            Toa uchambuzi wako.''
        ),\\
        ``prompt\_template\_image\_text'': (
            ``Mfano:\\\\
            Swali: \{example\_question\} \\\\
            Jibu: \{example\_response\} \\\\
            Maandishi yanayohusiana na picha:\\\\
            \{text\_content\} \\\\
            Kazi:\\\\
            \{task\_description\} \\\\
            Toa uchambuzi wako ukitumia picha na maandishi.''
        ),\\
        ``tasks'': \{\\
            1: ``Chambua picha hii na orodhesha vitu vyote vilivyopo. Vikundi vipange kama samani, vifaa vya kielektroniki, nguo, n.k.'',\\
            2: ``Eleza mandhari ya jumla ya picha hii, ikijumuisha setting na shughuli zozote zinazoonekana.'',\\
            3: ``Tambua mwingiliano au uhusiano wowote kati ya entiti katika picha hii na eleza muunganisho wao.'',\\
            4: ``Gawanya picha hii katika maeneo ya semantiki. Wewekee lebo kila eneo na elezea yaliyomo ndani kwa ufupi.'',\\
            5: ``Toa maelezo ya kina kwa lugha asilia ya kile kinachoendelea katika picha hii.'',\\
            6: ``Tenganisha kutoka kwenye maandishi sehemu zinazorejelea entiti, vitu, au mandhari kwenye picha na elezea ushahidi wa kuona.'',\\
            7: ``Tambua sehemu za maandishi ambazo hazionekani kwenye picha na eleza kwa nini.'',\\
            8: ``Orodhesha maeneo yaliyotajwa kwenye maandishi au kuonyeshwa kwenye picha; kwa kila moja, onyesha kama inaonekana kwenye maandishi, picha, au zote mbili na elezea umuhimu wake.''\\
        \},\\
        ``examples'': EXAMPLES\\
    \}\\
\}
\end{tcolorbox}

\clearpage
\begin{tcolorbox}[
  enhanced,
  breakable,
  colback=lightOrange,
  colframe=darkOrange,
  coltitle=white,
  fonttitle=\bfseries,
  title={Urdu: One-shot, With Rationales},
  boxrule=1pt,
  width=\linewidth,
  arc=2mm,
  halign title=center,
]
\tiny

EXAMPLES = \{\\
\hspace*{1em}1: \{\\
\hspace*{2em}``question'': \begin{UrduText}اس تصویر کا تجزیہ کریں اور موجود تمام اشیاء کی فہرست بنائیں۔ ہر شے کو فرنیچر، الیکٹرانک آلات، لباس وغیرہ کے گروپ میں تقسیم کریں۔\end{UrduText},\\
\hspace*{2em}``example\_response'': (\\
\hspace*{3em}\begin{UrduText}مثالی تجزیہ: ہوائی اڈے کی نشستیں، کیمرے اور اسمارٹ فونز جیسے الیکٹرانک آلات، اور مختلف قسم کے لباس شامل ہیں۔\end{UrduText}\\
\hspace*{3em}\begin{UrduText}بنیاد: تصویر کو منظم انداز سے اسکین کر کے واضح بصری حدود کی بنیاد پر اشیاء کی شناخت کی گئی اور افعال و ظاہری شکل کے مطابق گروپ کیا گیا۔\end{UrduText}\\
\hspace*{2em})\\
\hspace*{1em}\},\\
\hspace*{1em}2: \{\\
\hspace*{2em}``question'': \begin{UrduText}اس تصویر کے مجموعی منظر کی وضاحت کریں۔ یہ کہاں ہے اور کون سی سرگرمیاں یا واقعات ہو رہے ہیں؟\end{UrduText},\\
\hspace*{2em}``example\_response'': (\\
\hspace*{3em}\begin{UrduText}مثالی تجزیہ: یہ منظر ہوائی اڈے کے ٹرمینل کا ہے، جہاں ایک بڑے گروہ کے لوگ احتجاج کر رہے ہیں۔ کلیدی عناصر میں نشست کا علاقہ اور نمایاں سائن بورڈ شامل ہیں۔\end{UrduText}\\
\hspace*{3em}\begin{UrduText}بنیاد: لوگوں کی ترتیب، سائن بورڈ اور ماحولیاتی تفصیلات عوامی اجتماع کی طرف اشارہ کرتی ہیں۔\end{UrduText}\\
\hspace*{2em})\\
\hspace*{1em}\},\\
\hspace*{1em}3: \{\\
\hspace*{2em}``question'': \begin{UrduText}اس تصویر میں موجود اشیاء یا عوامل کے درمیان کسی بھی تعامل یا تعلقات کی نشاندہی کریں اور وضاحت کریں۔\end{UrduText},\\
\hspace*{2em}``example\_response'': (\\
\hspace*{3em}\begin{UrduText}مثالی تجزیہ: یکساں احتجاجی پلے کارڈز اٹھائے افراد اکٹھے ہیں، جو سماجی ربط اور منظم مظاہرے کی علامت ہیں۔\end{UrduText}\\
\hspace*{3em}\begin{UrduText}بنیاد: نزدیکی اور مماثل بصری علامتیں (مثلاً یکساں پلے کارڈز) مشترکہ سرگرمی کے نتیجے کو مضبوط کرتی ہیں۔\end{UrduText}\\
\hspace*{2em})\\
\hspace*{1em}\},\\
\hspace*{1em}4: \{\\
\hspace*{2em}``question'': \begin{UrduText}اس تصویر کو مختلف معنوی (سیمانٹک) علاقوں میں تقسیم کریں اور ہر علاقے کے مواد کی وضاحت کریں۔\end{UrduText},\\
\hspace*{2em}``example\_response'': (\\
\hspace*{3em}\begin{UrduText}مثالی تجزیہ: تصویر کو پیش منظر (افراد اور احتجاجی پلے کارڈز)، درمیانی منظر (گنجان ہجوم)، اور پس منظر (ہوائی اڈے کے ٹرمینل کی ساخت) میں تقسیم کیا گیا ہے۔\end{UrduText}\\
\hspace*{3em}\begin{UrduText}بنیاد: واضح بصری تفریق نے علاقوں میں تقسیم ممکن بنائی، جس سے مجموعی منظر کا تجزیہ منظم ہو گیا۔\end{UrduText}\\
\hspace*{2em})\\
\hspace*{1em}\},\\
\hspace*{1em}5: \{\\
\hspace*{2em}``question'': \begin{UrduText}اس تصویر میں کیا ہو رہا ہے، اس کا تفصیلی بیان قدرتی زبان میں فراہم کریں۔\end{UrduText},\\
\hspace*{2em}``example\_response'': (\\
\hspace*{3em}\begin{UrduText}مثالی تجزیہ: ہوائی اڈے کے ٹرمینل میں احتجاج جاری ہے؛ شرکاء کچھ بیٹھے ہیں، کچھ کھڑے ہیں اور پلے کارڈز اٹھائے ہوئے ہیں۔\end{UrduText}\\
\hspace*{3em}\begin{UrduText}بنیاد: پلے کارڈز، گروہی ترتیب اور ٹرمینل کی خصوصیات جیسے عناصر جامع بیانیے کو تقویت دیتے ہیں۔\end{UrduText}\\
\hspace*{2em})\\
\hspace*{1em}\},\\
\hspace*{1em}6: \{\\
\hspace*{2em}``question'': \begin{UrduText}متن کے وہ مخصوص حصے نکالیں اور فہرست بنائیں جو تصویر میں دکھائے گئے عوامل، اشیاء یا مناظر کا حوالہ دیتے ہیں۔ بصری شواہد کی وضاحت کریں۔\end{UrduText},\\
\hspace*{2em}``example\_response'': (\\
\hspace*{3em}\begin{UrduText}مثالی تجزیہ: متن میں ‘‘احتجاجی پلے کارڈز’’ کا ذکر ہے جو ہجوم میں واضح طور پر نظر آ رہے ہیں۔\end{UrduText}\\
\hspace*{3em}\begin{UrduText}بنیاد: متن اور بصری عناصر کے مابین مطابقت نے اس مماثلت کی تصدیق کی۔\end{UrduText}\\
\hspace*{2em})\\
\hspace*{1em}\},\\
\hspace*{1em}7: \{\\
\hspace*{2em}``question'': \begin{UrduText}متن کے کون سے حصے تصویر میں موجود نہیں ہیں؟ واضح کریں کہ کیا غائب ہے۔\end{UrduText},\\
\hspace*{2em}``example\_response'': (\\
\hspace*{3em}\begin{UrduText}مثالی تجزیہ: متن میں عدالتی فیصلوں کا ذکر ہے لیکن تصویر صرف احتجاجی سرگرمی دکھاتی ہے—کوئی قانونی یا عدالت سے متعلق عنصر موجود نہیں۔\end{UrduText}\\
\hspace*{3em}\begin{UrduText}بنیاد: چونکہ تصویر میں قانونی/عدالتی عناصر موجود نہیں، اس لیے متعلقہ متن حصوں کے لیے بصری شہادت دستیاب نہیں۔\end{UrduText}\\
\hspace*{2em})\\
\hspace*{1em}\},\\
\hspace*{1em}8: \{\\
\hspace*{2em}``question'': \begin{UrduText}متن یا تصویر میں کن مقامات کا ذکر ہے؟ بتائیں کہ وہ متن، تصویر یا دونوں میں ہیں؛ اور ان کی اہمیت بیان کریں۔\end{UrduText},\\
\hspace*{2em}``example\_response'': (\\
\hspace*{3em}\begin{UrduText}مثالی تجزیہ: منظر سے JFK ہوائی اڈے کا عندیہ ملتا ہے اور متن میں بھی اس کا ذکر ہے۔\end{UrduText}\\
\hspace*{3em}\begin{UrduText}بنیاد: بصری اور متنی سیاق و سباق اس مقام کی ایک بڑے سفری مرکز کے طور پر اہمیت کو واضح کرتے ہیں۔\end{UrduText}\\
\hspace*{2em})\\
\hspace*{1em}\}\\
\}\\[0.5em]

LANGUAGE\_CONFIGS = \{\\
\hspace*{1em}``Urdu'': \{\\
\hspace*{2em}``code'': ``Ur'',\\
\hspace*{2em}``system\_prompt'': \begin{UrduText}آپ ایک AI معاون ہیں جو تصاویر اور متن کا تجزیہ کرتا ہے۔\end{UrduText},\\[0.2em]

\hspace*{2em}``prompt\_template\_image\_only'': (\\
\hspace*{3em}\begin{UrduText}مثال:\end{UrduText} ``\begin{UrduText}سوال:\end{UrduText} \{example\_question\}'',\\
\hspace*{3em}`` \begin{UrduText}جواب:\end{UrduText} \{example\_response\}'',\\
\hspace*{3em}\begin{UrduText}اب درج ذیل کام کا تجزیہ کریں:\end{UrduText} ``\{task\_description\}'',\\
\hspace*{3em}\begin{UrduText}براہِ کرم اپنا تجزیہ مع دلائل پیش کریں۔\end{UrduText}\\
\hspace*{2em}),\\[0.2em]

\hspace*{2em}``prompt\_template\_image\_text'': (\\
\hspace*{3em}\begin{UrduText}مثال:\end{UrduText} ``\begin{UrduText}سوال:\end{UrduText} \{example\_question\}'',\\
\hspace*{3em}`` \begin{UrduText}جواب:\end{UrduText} \{example\_response\}'',\\
\hspace*{3em}\begin{UrduText}تصویر سے متعلق متن:\end{UrduText} ``\{text\_content\}'',\\
\hspace*{3em}\begin{UrduText}کام:\end{UrduText} ``\{task\_description\}'',\\
\hspace*{3em}\begin{UrduText}تصویر اور متن دونوں کی بنیاد پر، معقول دلائل کے ساتھ اپنا تجزیہ فراہم کریں۔\end{UrduText}\\
\hspace*{2em}),\\[0.2em]

\hspace*{2em}``tasks'': \{\\
\hspace*{3em}1: \begin{UrduText}اس تصویر کا تجزیہ کریں اور موجود تمام اشیاء کی فہرست بنائیں؛ ہر شے کو فرنیچر، الیکٹرانک آلات، لباس وغیرہ کے گروپ میں تقسیم کریں۔\end{UrduText},\\
\hspace*{3em}2: \begin{UrduText}اس تصویر کے مجموعی منظر کی وضاحت کریں، جس میں مقام اور نظر آنے والی سرگرمیاں شامل ہوں۔\end{UrduText},\\
\hspace*{3em}3: \begin{UrduText}اس تصویر میں عوامل/اشیاء کے درمیان کسی بھی تعامل یا تعلقات کی نشاندہی کریں اور ان کے روابط بیان کریں۔\end{UrduText},\\
\hspace*{3em}4: \begin{UrduText}اس تصویر کو معنوی (سیمانٹک) علاقوں میں تقسیم کریں؛ ہر علاقے کو لیبل کریں اور اس کے مواد کی مختصر وضاحت کریں۔\end{UrduText},\\
\hspace*{3em}5: \begin{UrduText}اس تصویر میں کیا ہو رہا ہے—اس کا تفصیلی بیان قدرتی زبان میں فراہم کریں۔\end{UrduText},\\
\hspace*{3em}6: \begin{UrduText}متن سے وہ حصے نکالیں جو تصویر میں دکھائے گئے عوامل/اشیاء/مناظر کا حوالہ دیتے ہیں اور بصری شواہد کی وضاحت کریں۔\end{UrduText},\\
\hspace*{3em}7: \begin{UrduText}متن کے کون سے حصے تصویر میں موجود نہیں اور کیوں؟ واضح کریں۔\end{UrduText},\\
\hspace*{3em}8: \begin{UrduText}متن یا تصویر میں ذکر کردہ مقامات کی فہرست بنائیں؛ ہر مقام کے لیے بتائیں کہ آیا وہ متن، تصویر یا دونوں میں ہے، اور اس کی اہمیت بیان کریں۔\end{UrduText}\\
\hspace*{2em}\},\\
\hspace*{2em}``examples'': EXAMPLES\\
\hspace*{1em}\}\\
\}
\end{tcolorbox}

\end{document}